\documentclass[11pt]{article}
\usepackage{arxiv}
\usepackage{graphicx}
\usepackage{float} 
\usepackage{mathrsfs}
\usepackage{amsmath,amssymb,amsthm}
\usepackage{algorithm}
\usepackage{algpseudocode}
\usepackage{subcaption}
\usepackage{mathtools}
\DeclareMathOperator*{\argmin}{argmin} % thin space, limits underneath 
\usepackage{booktabs}
\newcommand\omicron{o}
\usepackage{xcolor}         % colors
\usepackage[T1]{fontenc}    % use 8-bit T1 fonts
\usepackage[colorlinks=true,linkcolor=blue,citecolor=blue]{hyperref}
\usepackage{cleveref} 
\usepackage{url}            % simple URL typesetting

\providecommand{\keywords}[1]{\textbf{\textit{Keywords:}} #1}

\def\bmu{\boldsymbol{\mu}}
\def\bsigma{\boldsymbol{\sigma}}

\def\msD{\mathscr{D}}
\def\msV{\mathscr{V}}
\newcommand{\nrm}[1]{\|#1\|}

\def\bF{{\mathbf{F}}}

\title{InVAErt networks: a data-driven framework for model synthesis and identifiability analysis}

\author{%
Guoxiang Grayson Tong$^{1}$, \quad Carlos A. Sing Long$^{2}$ , \quad Daniele E. Schiavazzi$^{1*}$ \\
$^1$Department of Applied and Computational Mathematics and Statistics, University of Notre Dame,\\ Notre Dame, 46656, IN, United States\\
$^2$Institute for Mathematical and Computational Engineering, Pontificia Universidad Católica de Chile, \\ Santiago, Chile\\
}

\begin{document}
\maketitle


%\tableofcontents

\begin{abstract}
% Use of Networks dominated by the construction of surrogates of models and operators
Use of generative models and deep learning for physics-based systems is currently dominated by the task of emulation.
% A few recent paper suggest to learn the forward and inverse map at the same time, but none of them discusses the approach for non-identifiable systems
However, the remarkable flexibility offered by data-driven architectures would suggest to extend this representation to other aspects of system synthesis including model inversion and identifiability. 
% What we propose
We introduce inVAErt (pronounced \emph{invert}) networks, a comprehensive framework for data-driven analysis and synthesis of parametric physical systems which uses a deterministic encoder and decoder to represent the forward and inverse solution maps, a normalizing flow to capture the probabilistic distribution of system outputs, and a variational encoder designed to learn a compact latent representation for the lack of bijectivity between inputs and outputs.
% We investigate the loss function and its derivatives
% We also investigate various approaches to sample from the latent space
We formally investigate the selection of penalty coefficients in the loss function and strategies for latent space sampling, since we find that these significantly affect both training and testing performance.
We validate our framework through extensive numerical examples, including simple linear, nonlinear, and periodic maps, dynamical systems, and spatio-temporal PDEs.
\end{abstract}

%======================================
\section{Introduction}\label{sec:intro}
%======================================
% Complex simulations are time expensive
In the simulation of physical systems, an increase in model complexity directly corresponds to an increase in the simulation time, posing substantial limitations to the use of such models for critical applications that depend on time-sensitive decisions.
% Solution: data-driven approach provide fast surrogates
Therefore, fast emulators learned by data-driven architectures and integrated in algorithms for the solution of forward and inverse problems are becoming increasingly successful.

% Examples of architectures that also use limited amounts of training data
On one hand, several contributions in the literature have proposed architectures for physics-based emulators designed to limit the number of model evaluations during training. These include, for example, physics-informed neural networks (PINN)~\cite{karniadakis2021physics}, deep operator networks (DeepONet)~\cite{lu2021learning}, and transformers-based architectures~\cite{geneva2022transformers}.
% Examples of generative models
On the other hand, generative approaches have been the subject of significant recent research due to their flexibility to quantify uncertainty in the predicted outputs. Unlike traditional deep learning tasks, generative models focus on capturing a distributional characterization of the latent variables, providing an improved understanding, and a superior way to interact with a given system.
Examples in this context include Gaussian Processes~\cite{parussini2017multi}, Bayesian networks~\cite{yang2021b}, generative adversarial networks (GAN)~\cite{yang2020physics}, diffusion models~\cite{wang2023generative}, optimal transport~\cite{marzouk2016sampling}, normalizing flow~\cite{kobyzev2020normalizing,papamakarios2021normalizing} and Variational Auto-Encoders (VAE)~\cite{zhong2023pi}.

% Inverse problems: why they are difficult
When using data-driven emulators in the context of inverse problems, other difficulties arise. Inverse problem are often ill-posed as a result of non-uniqueness of solutions, or of ill-conditioning due to high-dimensionality, data-sparsity, noise-corruption, and nonlinear response of the physical systems~\cite{kaipio2006statistical,benning_burger_2018,stuart_2010,arridge_maass_oktem_schonlieb_2019, ghattas_willcox_2021}.
%Detailed reviews discussing a wide range of solution strategies for inverse problems are available in the literature, for example, in the context of image reconstruction~\cite{arridge_maass_öktem_schönlieb_2019} or with emphasis in large-scale inverse problem and model reduction for physics-based systems~\cite{ghattas_willcox_2021}.
% Use of regularization
Thus, robust solutions heavily rely on regularization of the Tikhonov-Phillips type~\cite{kirsch2011introduction, kaipio2006statistical,benning_burger_2018}, on prior specification in Bayesian inversion~\cite{stuart_2010,cotter2010approximation,Knapik_2011} or, more recently, on learning data-driven regularizers (see, e.g., the Network Tikhonov approach~\cite{li2020nett}). 
% 
%Incremental improvements of data-driven emulators during inversion is also explored in~\cite{WANG2022111454} in the context of variational inference and~\cite{cao2023residual} for residual-based error correction of neural operators.
%
%Adaptive annealing for the Bayesian inversion of physics-based systems is finally discussed in~\cite{cobian2023adaann}.
% Example: Image recontruction 
%Finally, image reconstruction and de-noising, which constitute a significant class of inverse problems, have also benefited greatly from the the data-driven approaches. For instance, using diffusion models~\cite{chung2022diffusion}, recurrent variational networks~\cite{yiasemis2022recurrent}, normalizing flows~\cite{wei2022deep} etc.

% FIRST POINT: FORWARD AND INVERSE PROBLEMS ARE CLOSELY RELATED AND NEED TO BE SOLVED TOGETHER
First, we would like to emphasize that, even if forward and inverse problems are generally treated separately in the literature, they are strongly related. 
% Think about where an emulator should be accurate
An uniform emulator accuracy in the entire input space, for example, is not required for the accurate solution of inverse problems, and accuracy only around regions of high posterior density might suffice. In this context, incremental improvements of data-driven emulators during inversion is explored in~\cite{WANG2022111454} in the context of variational inference, \cite{cao2023residual} for residual-based error correction of neural operators, and~\cite{cobian2023adaann} for adaptive annealing.
% Studies combining both aspects
In addition, an increasing number of studies in the literature are looking at data-driven architectures that can jointly learn the forward and inverse solution map~\cite{goh2019solving,vadeboncoeur2022deep,tait2020variational}.
% Tan Bui-Thanh - solving inverse problems with VAE
%The use of VAEs is discussed in~\cite{goh2019solving} which provides a global Gaussian posterior approximation from the minimization of a loss function combining the KL and Jensen-Shannon divergence. 
% Girolami - deep probabilistic models for forward and inverse problems in parametric PDEs
%The approach in~\cite{vadeboncoeur2022deep} combines residual-based weak-form formulation of PDE problems with VAE probabilistic models and Gaussian parameter distributions. 
% Variational Autoencoding of PDE Inverse Problems -- Tait et al.
%Another work related to the discretized PDE inverse problem can be found in~\cite{tait2020variational}, where the authors manage to embed discretization knowledge into the VAE-based generative framework to bypass the costly forward model evaluations.
% INVERSE ANALYSIS WITH VARIATIONAL
%AUTOENCODERS: A COMPARISON OF
%SHALLOW AND DEEP NETWORKS-- Vesselinov et al. and % VI-DGP-Jin
%The VAE-assisted heterogeneous hydrogeological inverse problem was investigated by Wu et al.~\cite{wu2022inverse} and Xia et al.~\cite{xia2023vidgp}, highlighting the effectiveness of VAE in dimensional reduction for efficient inverse modeling and calibration. 

% SECOND POINT: REGULARIZATION IS NOT THE RIGHT ANSWER
Second we remark that, in many of the previous contributions, regularization and strong assumptions on the distribution of model outputs or parameters are used to promote certain solutions in ill-posed inverse problems.
% Regularization provides only a partial answer
However, this may not be fully informative on the nature of such problems. In other words, there may be several possible inputs belonging to a {\em manifold} (or, more generally, a \emph{fiber} or a \emph{level set}) embedded in the ambient input space, that constitute the \emph{preimage} of a given observation.
In this case, instead of {\em selecting} a solution with a particular structure, we would like to characterize the \emph{entire} manifold of possible solutions to the problem. 
Discovering this manifold is analogous to study the identifiability of a physical system with a surjective input-to-output map. Thus, understanding and characterizing such manifold becomes essential for gaining insights into the system and generating accurate emulators efficiently.

% What we propose 
In this paper, we introduce inVAErt networks, a new approach for jointly learning the forward and inverse model maps, the distribution of the model outputs, and also discovering non-identifiable input manifolds. 
%
% List the unique contributions 
We also claim the following novel contributions
\begin{itemize}
\item An inVAErt network goes beyond emulation, learning all essential aspects of a physics-based model, or, in other words, performing model \emph{synthesis}. 

\item Our approach does not require strong assumptions about the distribution of inputs and outputs, and provides a comprehensive sampling-based (nonparametric) characterization of the set of possible solutions to an inverse problem.

\item We formally investigate penalty selection through stationarity conditions for the loss function gradient, and explore several latent space sampling strategies to improve performance during inference.
\end{itemize}
% When used for evaluation and inference, the approach allows to interact with the model opening new possibilities for probabilistic reasoning with physics-based systems. 
After an inVAErt network is trained, it greatly enhances the possibility of interacting in real-time with a given physical model, by providing input parameters, spatial locations and time corresponding to a single or multiple observations. 
% Use for Bayesian estimation
For the solution of Bayesian inverse problems, an inVAErt network can be seen as jointly learning both an emulator and a proposal distribution, hence it can be easily integrated, for example, in MCMC simulation.
% Code and examples can be found online: Outline/github
For the interested reader, the source code and examples can be found at \url{https://github.com/desResLab/InVAErt-networks}.

% Similar approach found in the literature
While finalizing this paper, we became aware of a similar network architecture studied in~\cite{almaeen2021variational} and later applied to the problem of quantum chromodynamics~\cite{10069126}. 
% Details are not always clear 
% Extensive testing
% focus on non idetifiable physics-based models
% formal analysis
Even through it is difficult to exactly determine architectural differences between the two approaches, we believe our exposition explores in greater detail a number of non-identifiable physics-based systems, and provides formal arguments to improve the selection of penalty coefficients and latent space sampling strategy. 
% overlapping with simulation based inference. 
In addition, we would like to point out the similarities between our approach and the emerging paradigm of simulation-based (or likelihood-free) inference (SBI, see, e.g.,~\cite{cranmer2020frontier}). 
SBI combines amortized inference with a neural approximation of the posterior distribution, likelihood function or likelihood ratio. Once trained, the conditional distribution corresponding to a new set of observations is estimated without requiring additional model solutions. 
The same can be achieved by training our inVAErt networks from deterministic or noisy inputs and outputs. In the first case, the analysis of the latent space is indicative of the \emph{structural} identifiability of the system, in the second structural and \emph{practical} identifiability coexists.

% Section layout
The paper is organized as follows. In Section~\ref{sec:inVAErt}, we provide an in-depth description of each component in the inVAErt network. Section~\ref{sec:analysis} presents a rigorous analysis of its properties, with a focus on assessing the stationarity of the loss function gradients and latent space sampling. A number of numerical examples, covering input-to-output maps and the solution of nonlinear ODEs and PDEs in both spatial and temporal domains, are discussed in Section~\ref{sec: experiments}.

%===============================================
\section{The inVAErt network}\label{sec:inVAErt}
%===============================================

The network has three main components, an architecture-agnostic neural emulator, a density estimator for the model outputs, and a decoder network equipped with a VAE for latent space discovery and model inversion. Each of these components is explained in detail in the following sections. 

%===============================================================
\subsection{Neural emulator}\label{sec:inVAErt_emulator}
%===============================================================
%
Consider a physics-based system with inputs, possibly including space and time, defined through an input-to-output map or the solution of an ordinary or partial differential equation
\begin{equation}\label{equ:emulator_general}  
\boldsymbol{y} = \mathcal{F}\left(\boldsymbol{x},t,\boldsymbol{\Phi}; \boldsymbol{\Psi} \right). 
\end{equation}
%
% The operator $\mathcal{F}$ can be described as a generic \emph{emulator} (sometimes referred to as \emph{observation operator}~\cite{stuart_2010}), with parameters $\boldsymbol{\Psi} = \{\psi_1, \psi_2, \cdots\}$, mapping the spatial coordinates $\boldsymbol{x} \in \mathbb{R}^D$, time $t$ and a set of additional variables $\boldsymbol{\boldsymbol{\Phi}} = \{ \phi_1, \phi_2, \cdots \}$ to the system output $\boldsymbol{y} \in \boldsymbol{\mathcal{Y}}$. 
%
The operator $\mathcal{F}$ can be described as a generic \emph{emulator} (sometimes referred to as \emph{observation operator}~\cite{stuart_2010}), with parameterization $\boldsymbol{\Psi} = \{\psi_1, \psi_2, \cdots\}$, mapping the spatial coordinates $\boldsymbol{x} \in \mathbb{R}^D$, time $t$ and a set of additional problem-dependent parameters $\boldsymbol{\boldsymbol{\Phi}} = \{ \phi_1, \phi_2, \cdots \}$ to the system output $\boldsymbol{y} \in \boldsymbol{\mathcal{Y}}$. 
To simplify the notation, we define the model \emph{inputs} as $\boldsymbol{v} = [\boldsymbol{x}, t, \boldsymbol{\Phi}]\in \boldsymbol{\mathcal{V}}$ such that this forward process can be rewritten as 
\begin{equation}
\boldsymbol{y} = \mathcal{F}_{\boldsymbol{\Psi}}\left(\boldsymbol{v}\right) = \mathscr{N}_e\left(\boldsymbol{v}, \mathcal{\mathcal{D}}_{\boldsymbol{v}}; \boldsymbol{\theta}_e \right),
\end{equation}
where a generic neural network \emph{encoder} $\mathscr{N}_e$ with trainable parameters $\boldsymbol{\theta}_e$ and auxiliary input data $\mathcal{D}_{\boldsymbol{v}}$ is used in place of the abstract map $\mathcal{F}_{\boldsymbol{\Psi}}\left(\cdot \right)$. In the following sections, we omit the notation $(\cdot)_{\boldsymbol{\Psi}}$ for brevity.

The \emph{auxiliary input data} $\mathcal{D}_{\boldsymbol{v}}$ are included as part of inputs to assist the neural emulator $\mathscr{N}_e$ to learn complex forward models. 
For example, recurrent or residual networks~\cite{qin2019data, fu2022modeling,tong2023data} are used to emulate the evolution operator (flow map) of dynamical systems; convolutional neural networks (CNN) or message-passing graph neural networks (GNN)~\cite{stevens2020finitenet,sanchez2020learning, hamilton2020graph} can also be used to gather neighbor information in structured and unstructured discretized domains, respectively.
In general, to approximate the system output $\boldsymbol{y}$ at time instance $t_n$ and location $\boldsymbol{x}$, the auxiliary set $\mathcal{D}_{\boldsymbol{v}}$ has the form:
\begin{equation*}
    \mathcal{D}_{\boldsymbol{v}} = \big\{ \cdots, \boldsymbol{y}\big(t_{n-2}, \mathcal{B}(\boldsymbol{x})\big), \boldsymbol{y}\big(t_{n-1}, \mathcal{B}(\boldsymbol{x})\big) \big\} \ ,
\end{equation*}
where $\mathcal{B}(\boldsymbol{x})$ represents the neighborhood of node $\boldsymbol{x}$ when dealing with a mesh (graph)-based discretization of a physical system that involves space, e.g. $k$-hop in GNNs~\cite{hamilton2020graph}. 
In the numerical examples of Section~\ref{sec: experiments} involving discretized systems in space and time, we build  ResNet~\cite{qin2019data} emulators which learn to update the value of the state $\boldsymbol{y}$ from two successive time steps as
\begin{equation}\label{equ: aux system-encoder-resnet}
    \boldsymbol{y}(t_n,\boldsymbol{x}) = \boldsymbol{y}(t_{n-1},\boldsymbol{x}) + \mathscr{N}_e\left(\boldsymbol{v}, \mathcal{\mathcal{D}}_{\boldsymbol{v}}; \boldsymbol{\theta}_e \right). 
\end{equation}

%=====================================================================
\subsection{Density estimator for model outputs}\label{sec:inVAErt_nf}
%=====================================================================

Once the inVAErt network is trained, we would like the ability to generate representative output samples $\boldsymbol{y} \in \boldsymbol{\mathcal{Y}}$. In other words, assuming a collection of inputs with known distribution $\boldsymbol{v} \sim p_{\boldsymbol{v}}(\boldsymbol{v})$, we would like to generate output samples $\boldsymbol{y} \sim p_{\boldsymbol{y}}(\boldsymbol{y})$. To do so, we train a Real-NVP normalizing flow density estimator~\cite{dinh2016density}
\begin{equation}\label{equ: NF-NN map}
\boldsymbol{y}\approx  \boldsymbol{z}_K = \mathscr{N}_{f}\left(\boldsymbol{z}_0; \boldsymbol{\theta}_f \right),\,\,\text{where}\,\,\boldsymbol{z}_0 \sim \mathcal{N}(\boldsymbol{0}, \mathbf{I}) \ .
\end{equation}

Normalizing flows~\cite{rezende2015variational} consist of a collection of $K$ bijective transformations $\mathscr{T} = \boldsymbol{f}_{1}\circ \boldsymbol{f}_{2} \circ \cdots \circ \boldsymbol{f}_{K}$ (parameterized using $\boldsymbol{\theta}_{f} = \boldsymbol{\theta}_{f,1}\cup\cdots\cup\boldsymbol{\theta}_{f,K}$) trained to learn the mapping between an easy-to-sample distribution to the density of the available samples.
Under this transformation, the underlying density is modified following the change of variable formula
\begin{equation}\label{equ:change_var_nf}
p_{\boldsymbol{y}}(\boldsymbol{y}) = p_{\boldsymbol{z}}(\boldsymbol{z})|\det{\mathbf{J}}|^{-1},\,\,\text{or, equivalently}\,\,\log p_{\boldsymbol{y}}(\boldsymbol{y}) = \log q_0 (\boldsymbol{z}_0) + \sum^K_{k=1} \log\left|\frac{\partial \boldsymbol{z}_{k}}{\partial \boldsymbol{z}_{k-1}}\right|^{-1},
\end{equation}
where $q_0(\cdot)$ is usually the multivariate standard normal $\mathcal{N}(\boldsymbol{0}, \mathbf{I})$ and $\boldsymbol{z}_{k} = \boldsymbol{f}_{k}(\boldsymbol{z}_{k-1}; \boldsymbol{\theta}_{f,k})$, $\boldsymbol{z}_K \approx \boldsymbol{y}$. 
The parameters $\boldsymbol{\theta}_f$ are determined by maximizing the log-likelihood of the available samples of size $N$, i.e. $\{ \boldsymbol{y}_i\}_{i=1}^N$

\begin{equation}\label{equ:nf_den_estimation}
\begin{split}
\sum_{i=1}^{N}\,\log p_{\boldsymbol{y}}(\boldsymbol{y}_{i}) & = \sum_{i=1}^{N}\,\log \left\{q_0 (\boldsymbol{z}_{i,0}) \prod_{k=1}^K \left|\frac{\partial \boldsymbol{z}_{i,k}}{\partial \boldsymbol{z}_{i,k-1}}\right|^{-1}\right\}  \ ,\\
& = \sum_{i=1}^{N}\,\log q_0 \left(\mathscr{N}_{f}^{-1}(\boldsymbol{y}_i; \boldsymbol{\theta}_{f})\right) -\sum_{k=1}^K\,\sum_{i=1}^{N} \log\Big|\frac{\partial \boldsymbol{z}_{i,k}}{\partial \boldsymbol{z}_{i,k-1}}\Big|^{-1},
\end{split}
\end{equation}
where we used~\eqref{equ: NF-NN map} to represent $\boldsymbol{z}_{i,0}$ in terms of $\boldsymbol{y}_{i}$.

% The ideal NF is easy to invert and 
Of the many possible choices for the transformation $\boldsymbol{f}_{k},\,k=1,\dots,K$, an ideal candidate should be easy to invert and the computational complexity of computing the Jacobian determinant should increase linearly with the input dimensionality. 
Dinh et al. propose a block triangular autoregressive transformation, introducing the widely used real-valued non-volume preserving transformations (Real-NVP~\cite{dinh2016density}, but several other types of flows are discussed in the literature, see, e.g.~\cite{kobyzev2020normalizing,papamakarios2021normalizing}).
Given $\dim(\boldsymbol{y})=m$, the method of Real-NVP considers a $m^{*} < m$ and defines the bijection $\boldsymbol{z}_k = \boldsymbol{f}_k(\boldsymbol{z}_{k-1}; \boldsymbol{\theta}_{f,k})$ through the easily invertible \emph{affine coupling}
\begin{equation}\label{equ:realnvp_coupling}
\begin{aligned}
[\boldsymbol{z}_k]_{1:m^{*}} &= [\boldsymbol{z}_{k-1}]_{1:m^{*}}\ , \\
[\boldsymbol{z}_k]_{m^{*}+1:m} &= [\boldsymbol{z}_{k-1}]_{m^{*}+1:m} \odot \exp[{\boldsymbol{s}_k([\boldsymbol{z}_{k-1}]_{1:m^{*}})}]+\boldsymbol{t}_k([\boldsymbol{z}_{k-1}]_{1:m^{*}})\ ,\\
\end{aligned}
\end{equation}
where $[\cdot]_{a:b}$ is used to denote the range of components from $a$ to $b$ included, $\boldsymbol{s}_k\in \mathbb{R}^{m^{*}}\to \mathbb{R}^{m-m^{*}}$ and $\boldsymbol{t}_k\in \mathbb{R}^{m^{*}}\to \mathbb{R}^{m-m^{*}}$ are the scaling and translation functions implemented through dense neural networks, and $\odot$ denotes the Hadamard product. 
We alternate the variables that remain unchanged in each layer (see Section 3.5 in~\cite{dinh2016density}) and use batch normalization as proposed in~\cite{dinh2016density}.
Finally, observe how the above coupling does not apply to one-dimensional transformations, i.e. $m=1, y\in \mathbb{R}$. For such cases, we concatenate to $y$ auxiliary independent standard Gaussian samples. 

%=======================================================
\subsection{Inverse modeling}\label{sec:inVAErt_inverse}
%=======================================================

\def\by{\boldsymbol{y}}
\def\bv{\boldsymbol{v}}
\def\bw{\boldsymbol{w}}
\def\mcbV{\boldsymbol{\mathcal{V}}}
\def\mcbY{\boldsymbol{\mathcal{Y}}}
\def\mcbM{\boldsymbol{\mathcal{M}}}
\def\mcbW{\boldsymbol{\mathcal{W}}}
\def\mF{\mathcal{F}}
\def\bPsi{\boldsymbol{\Psi}}

% Consider the situation where the number of outputs $\boldsymbol{y}$ for the model $ \mathcal{F}$ is smaller than the number of inputs $\boldsymbol{v}$ i.e. $\dim{(\boldsymbol{y})} < \dim{(\boldsymbol{v})}$. 
Consider the situation where the dimension of the output space $\mcbY$ is smaller than the dimension of the input space $\mcbV$, i.e., $\dim(\mcbY) < \dim(\mcbV)$.
%
% This implies the existence of a low-dimensional manifold $\boldsymbol{\mathcal{M}}$ embedded in $\boldsymbol{\mathcal{V}}$, with $\dim(\boldsymbol{\mathcal{M}}) \leq \dim(\boldsymbol{\mathcal{V}})$, with respect to which the outputs are invariant, i.e. 
% \[
% \mathcal{F}_{\boldsymbol{\Psi}}(\boldsymbol{v}^{*})=\boldsymbol{y}^{*} \in \boldsymbol{\mathcal{Y}},\,\forall\,\boldsymbol{v}^{*}\in\boldsymbol{\mathcal{M}}.
% \]

In this case, to every input $\bv$ we can associate a manifold\footnote{In our discussion we assume the preimage of a single output $\{\boldsymbol{y}\}$ to have a local Euclidean structure, consistent with the nature of the problems we analyze in Section~\ref{sec: experiments}. However, the proposed approach is based on sampling and therefore, in principle, we can deal with more general settings and therefore the term \emph{manifold}, with an abuse of terminology, is sometimes used as a synonym for \emph{fiber}, or \emph{level set} of $\mF$.} $\mcbM_{\bv}$ embedded in $\mcbV$ with $\dim(\mcbM_{\bv}) \leq \dim(\mcbV)$ containing all the inputs producing the same output as $\bv$, i.e., 
$$
\mF(\bv')=\mF(\bv),\quad\forall\,\bv'\in\mcbM_{\bv} \ .
$$
As a result, it will not be possible to distinguish between two inputs $\bv,\bv'\in\mcbM_{\bv}$ from their outputs, or, in other words, the inputs in $\mcbM_{\bv}$ are \emph{non-identifiable} from their common output $\by$. 
%
% In this context, a forward map $\mathcal{F}$ is identifiable at parameter $\boldsymbol{v}_1 \in \boldsymbol{\mathcal{V}}$ if $\mathcal{F}(\boldsymbol{v}_1) = \mathcal{F}(\boldsymbol{v}_2)$ implies $\boldsymbol{v}_1=\boldsymbol{v}_2$.
A parameter \(\bv\in \mcbV\) is identifiable by the map \(\mF\) if $\mF(\bv) = \mF(\bv')$ implies $\bv=\bv'$~\cite{smith2013uncertainty}, that is, when \(\mcbM_{\bv} = \{\bv\}\). In general, we either have that \(\mcbM_{\bv}\cap \mcbM_{\bv'} = \varnothing
\) or that \(\mcbM_{\bv} = \mcbM_{\bv'}\). The collection of distinct manifolds of the form $\mcbM_{\bv}$ forms a partition of $\mcbV$ and the dimensionality of \(\mcbM_{\bv}\) may change with \(\bv\), depending on the rank of the Jacobian $\nabla\mF$.

A dimensionality mismatch between $\bv$ and $\by$ precludes the existence of an inverse, and may pose difficulties when constructing a pseudo-inverse
\begin{equation}\label{equ:inverse_map}
 \mathcal{G}_{\boldsymbol{\Xi}}\left( \boldsymbol{y}\right) = \mathcal{G}\left( \boldsymbol{y}; \boldsymbol{\Xi}\right) ,    
\end{equation}
for which $\mF\big(\mathcal{G}(\by; \boldsymbol{\Xi})\big) = \by$ and ${\boldsymbol{\Xi}} = \{\xi_1, \xi_2, \cdots\}$ parameterizes the operator $\mathcal{G}$.
To overcome this problem and restore bijectivity, we introduce a latent space $\mcbW$ such that $\dim(\mcbV) \leq \dim(\widetilde{\mcbY})$ where $\widetilde{\mcbY} = \mcbY\times \mcbW$ and  $\tilde{\by}\in \widetilde{\mcbY}$ is obtained via concatenation as $\tilde{\by} = [\by, \bw]^T$ with $\bw\in\mcbW$. This is similar to the concept of invertible neural networks in~\cite{ardizzone2018analyzing} and can be understood as a $\dim(\mcbW)$-dimensional extension of $\mcbY$. 
When the dimension of $\mcbM_{\bv}$ is constant, each one of the manifolds can be identified with the latent space $\mcbW$. Otherwise, $\mcbW$ must have a sufficiently high dimensionality to include the latent space representation of each $\mcbM_{\bv}$. As a concrete example, when all the above spaces are vector spaces and the forward map \(\mF\) is linear, the manifolds have the explicit form $\mcbM_{\bv} = \bv + \operatorname{\bf null}(\mF)$. Therefore, each one of them can be identified with \(\operatorname{\bf null}(\mF)\).
%
% This can be understood as a \csl{$\dim(\mcbW)$-dimensional} extension of \csl{$\boldsymbol{\mathcal{Y}}$}, denoted as \csl{$\widetilde{\boldsymbol{\mathcal{Y}}}$}, such that $\mathcal{F}( \boldsymbol{\mathcal{M}})$ is no longer invariant in $\widetilde{\boldsymbol{\mathcal{Y}}}$. 
%
% In other words, we have $\widetilde{\boldsymbol{\mathcal{Y}}} = \mcbW \oplus \boldsymbol{\mathcal{Y}}$ with our proposed latent variable $\boldsymbol{w} \in \mcbW \subset \mathbb{R}^{d-m}$. 
%
% Under general settings, when all the above spaces are Hilbert spaces, $\mcbW$ is the orthogonal complement~\cite{smith2013uncertainty} of the space $\boldsymbol{\mathcal{Y}}$ \csl{in $\widetilde{\boldsymbol{\mathcal{Y}}}$}, and, for every \csl{$\by\in\boldsymbol{\mathcal{Y}}$}, induces a decomposition of the input space in the form \csl{$\mcbV = \boldsymbol{\mathcal{M}}_{\by} \oplus \boldsymbol{\mathcal{M}}_{\by}^{\perp}$}.
%

We also consider the possibility that $\dim(\widetilde{\mcbY}) > \dim(\mcbV)$. As a consequence of the Whitney embedding theorem~\cite{whitney1936differentiable}, any smooth manifold can be embedded into a higher dimensional Euclidean space. Intuitively, it is simpler to approximate a lower dimensional structure in a higher dimensional space. In fact, an embedding into a high-dimensional space may smooth some geometric singularities, i.e., self-intersections, cusps, etc., that appear in a low-dimensional space. Think, for example, about a curve that does not self-intersect if represented in $\mathbb{R}^3$, but it does when projected on $\mathbb{R}^{2}$. Note that better performance of utilizing high-dimensional latent space for inverse modeling was also shown in~\cite{almaeen2021variational}.

% Second, our approach consists on using normalizing flows, i.e., a continuous bijection, to transform a Gaussian probability distribution into a distribution concentrated around \(\mcbM_{\bv}\). However, this problem is ill-posed when the manifold \(\mcbM_{\bv}\) of possible solutions has several connected components. However, the ill-possedness can be mitigated if the components are lifted into a high-dimensional space. Note that better performance of utilizing high-dimensional latent space for inverse modeling was also shown in~\cite{almaeen2021variational}.
%

\begin{figure}[ht!]
\centering
\includegraphics[scale=0.4]{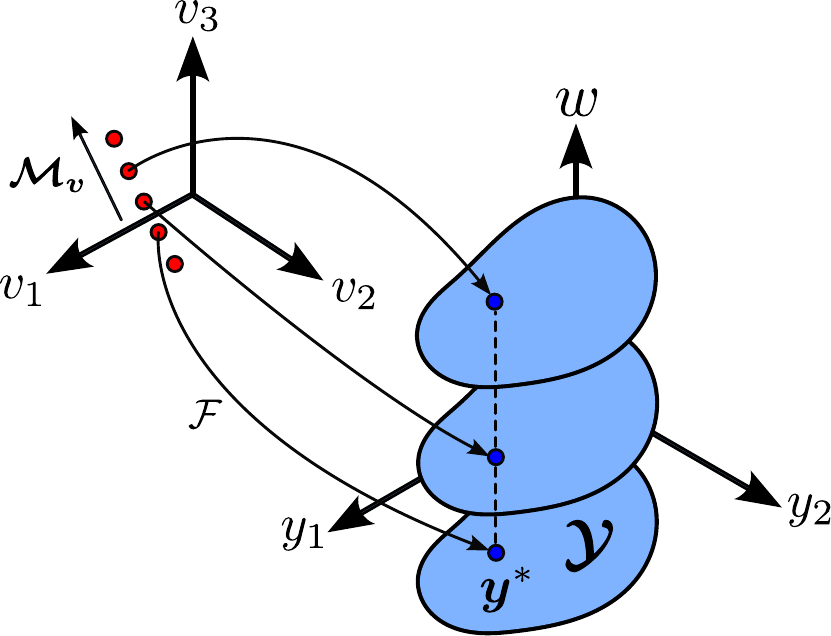}
\caption{A graphical illustration of a non-identifiable parameter embedding and how input-output bijectivity is restored by extending the output space.}
\label{fig:extension}
\end{figure}

% A graphical representation of this decomposition is illustrated in Figure~\ref{fig:extension}, where $\bv\in \mathbb{R}^3$, $\by\in \mathbb{R}^2$ and $\bw\in \mathbb{R}$, such that $\mcbM$ is an one-dimensional embedded manifold.
% %
% The space $\tilde{\mcbY}$ is also obtained by an \emph{extrusion} of $\mcbY$ along the $\bw$ direction.
% %
% As discussed, all $\bv\in\mcbM$ are mapped to a single $\by \in \mcbY$ through $\mF_{\bPsi}$, while $\mathcal{F}_{\boldsymbol{\Psi}}(\mcbM)$ varies in $\tilde{\mcbY}$.
% %
% We would also like to remark that under linear settings, i.e. when $\mathcal{F}\in \mathbb{R}^{m\times d}$, the above discussion is consistent with the rank–nullity theorem, stating that $\dim{(\mcbM)} + \dim{(\mcbY)} = \dim{(\boldsymbol{\mathcal{V}})}$, and $\mcbM$ is usually referred to as the null space or the kernel of $\boldsymbol{\mathcal{V}}$.
% %
A graphical representation of this decomposition is illustrated in Figure~\ref{fig:extension}, where $\bv\in \mathbb{R}^3$, $\by\in \mathbb{R}^2$ and $\bw\in \mathbb{R}$, such that $\mcbM_{\bv}$ is a one-dimensional fiber.
In this case, the space $\widetilde{\mcbY}$ is obtained by extending $\mcbY$ along the $\bw$ direction.
As discussed, all $\bv \in \mcbM_{\bv}$ are mapped to a single $\by^*\in \mcbY$ through $\mF$, while $\mathcal{F}(\mcbM_{\bv})$ varies in $\widetilde{\mcbY}$ when augmented with \(\bw\).
We would also like to remark that under linear settings, i.e. when $\mF \in \mathbb{R}^{\dim(\by)\times \dim(\bv)}$, the above discussion is consistent with the rank–nullity theorem, stating that $\dim{(\mcbM_{\bv})} + \dim{(\mcbY)} = \dim{(\boldsymbol{\mathcal{V}})}$, and $\mcbM_{\bv}$ can always be identified with the null space, or kernel, of $\mF$.

\def\vW{\boldsymbol{W}}
\def\bW{\boldsymbol{W}}
\def\bmu{\boldsymbol{\mu}}
\def\bsigma{\boldsymbol{\sigma}}
\def\beps{\boldsymbol{\epsilon}}
\def\btheta{\boldsymbol{\theta}}

% CSL: This is a suggestion how to write the following part.
In this paper, our goal is to learn the manifold \(\mcbM_{\bv}\) associated with every input \(\bv\) using a variational autoencoder. To do so, we consider a \(\mcbW\)-valued stochastic process \(\bW_{\bv}\) indexed by every \(\bv\in\mcbV\), such that the probability density of \(\bW_{\bv}\) is concentrated near the image of $\boldsymbol{\mathcal{M}}_{\bv}$ in $\boldsymbol{\mathcal{W}}$, which is responsible for the information gap between the spaces $\boldsymbol{\mathcal{V}}$ and $\boldsymbol{\mathcal{Y}}$.
Hence, we seek a neural network $\mathscr{N}_v (\cdot \ ; \boldsymbol{\theta}_v):  \boldsymbol{\mathcal{V}} \to \boldsymbol{\mathcal{W}}$ such that
\begin{align}
    \begin{split}
      \mathscr{N}_v(\bv; \boldsymbol{\theta}_{\bv}) &=  [\boldsymbol{\mu}_{\bv}, \boldsymbol{\sigma}_{\bv}]  \ , \ \\
        \vW_{\bv} &\sim \mathcal{N}\big(\boldsymbol{\mu}_{\bv}, \mathrm{\bf diag}(\boldsymbol{\sigma}^2_{\bv})\big),
    \end{split}
\end{align}
and another neural network $\mathscr{N}_d(\cdot, \cdot \ ; \btheta_d): \mcbW \times \mcbY \to \mcbV$ such that the parameters $(\btheta_v,\btheta_d)$ are optimally trained and
\begin{equation}
\mF \big(  \mathscr{N}_d ( \vW_{\bv}, \by ;\boldsymbol{\theta}_d) \big) \overset{w.h.p.}{\approx} \by \ . 
\label{equ: inverse id map-almost sure}
\end{equation}

Therefore, the stochastic process \(\vW_{\bv}\) represents the missing information about \(\bv\) with respect to the forward pass \(\by = \mF(\bv)\). 
The generation of $\bW_{\bv}$ during training, follows the well-known reparametrization trick in VAE~\cite{kingma2013auto}
\[
    \bW_{\bv} = \bmu_{\bv} + \bsigma_{\bv} \odot \beps\quad\mbox{with}\quad \beps \sim \mathcal{N}(\boldsymbol{0}, \mathbf{I}).
\]
We also impose that the samples of \(\bW_{\bv}\) provide unbiased information about \(\bv\) in the sense that
\begin{equation}
    \mathbb{E}_{\bw \sim \bW_{\bv}} \left[ \mathscr{N}_d ( \bw, \mF(\bv) ;\btheta_d ) \right] = \bv, \,\,  \forall \bv \in \mcbV \ .
    \label{equ: unbiased}
\end{equation}

\subsubsection{An introduction to variational autoencoders} \label{sec: vae}
%========================================================

The VAE uses deep neural network-based parametric models to handle the intractability in traditional inference tasks, as well as to learn a low-dimensional embedding of the input feature $\boldsymbol{v}$~\cite {kingma2013auto,kingma2019introduction,goodfellow2016deep,doersch2016tutorial}. Given a distribution $\boldsymbol{v} \sim p_{\boldsymbol{v}}(\boldsymbol{v})$ to be learned, we consider an unobserved latent variable $\boldsymbol{w} \sim p_{\boldsymbol{w}}(\boldsymbol{w})$, that correlates to $\boldsymbol{v}$ via the law of total probability
\begin{equation}
    p_{\boldsymbol{v}}(\boldsymbol{v}) = \mathbb{E}_{\boldsymbol{w}\sim p_{\boldsymbol{w}}(\boldsymbol{w})} \big[p(\boldsymbol{v} |\boldsymbol{w})\big] \ .
    \label{equ: total prob}
\end{equation}
The conditional probability $p(\boldsymbol{v} |\boldsymbol{w})$ quantifies the relation between a latent variable $\boldsymbol{w}$ (from an easy-to-sample prior distribution $p_{\boldsymbol{w}}(\boldsymbol{w})$, usually a standard Gaussian) and a specific input feature $\boldsymbol{v} \sim p_{\boldsymbol{v}}(\boldsymbol{v})$. 
However, computing $p(\boldsymbol{v} |\boldsymbol{w})$ is usually intractable, so one applies Bayes theorem to get
\begin{equation}
    p_{\boldsymbol{v}}(\boldsymbol{v}) = \mathbb{E}_{\boldsymbol{w}\sim p_{\boldsymbol{w}}(\boldsymbol{w})} \left[  \frac{ p_{\boldsymbol{v}}(\boldsymbol{v}) p(\boldsymbol{w} |\boldsymbol{v})}{p_{\boldsymbol{w}}(\boldsymbol{w})}  \right] = \mathbb{E}_{\boldsymbol{w}\sim p(\boldsymbol{w}|\boldsymbol{v})} \left[ \frac{p(\boldsymbol{v},\boldsymbol{w})}{p(\boldsymbol{w}|\boldsymbol{v})}  \right] \ .
    \label{equ: elbo-derivation}
\end{equation}
Drawing $\boldsymbol{w}$ samples from the exact posterior distribution $p(\boldsymbol{w} |\boldsymbol{v})$ is also intractable but a variational approximation $q(\boldsymbol{w} |\boldsymbol{v}) \approx p(\boldsymbol{w} |\boldsymbol{v})$ can be used instead, and the Jensen's inequality used to obtain
\begin{align}
    \log p_{\boldsymbol{v}}(\boldsymbol{v}) &\geq  
    -\mathcal{KL}\big[ q(\boldsymbol{w}|\boldsymbol{v}) \| p_{\boldsymbol{w}}(\boldsymbol{w}) \big] + \mathbb{E}_{\boldsymbol{w} \sim q(\boldsymbol{w}|\boldsymbol{v})} \big[ \log  p(\boldsymbol{v}|\boldsymbol{w}) \big],
    \label{equ: elbo des} 
\end{align}
where the term $\mathcal{KL}$ denotes the Kullback–Leibler (KL) divergence between two distributions $p_{1}$ and $p_{2}$, defined as
\begin{equation}
    \mathcal{KL}\big[p_1(\boldsymbol{x})\|p_2(\boldsymbol{x})\big] = \mathbb{E}_{\boldsymbol{x} \sim p_1(\boldsymbol{x})} \big[ \log{p_1(\boldsymbol{x})}-\log{p_2(\boldsymbol{x})}  \big],
    \label{equ: kl}
\end{equation}
and the right hand side of~\eqref{equ: elbo des} is also referred to as the \emph{evidence lower bound} (ELBO). 
Then maximizing $\log p_{\boldsymbol{v}}(\boldsymbol{v})$ is equivalent to minimize the negative ELBO, i.e. leading to an optimization problem of the form
\begin{equation}
   q(\boldsymbol{w}|\boldsymbol{v}), p(\boldsymbol{v}|\boldsymbol{w})   =  \argmin_{\tilde{q}(\boldsymbol{v}; \boldsymbol{\delta}), \tilde{p}(\boldsymbol{w}; \boldsymbol{\tau})} \big\{\mathcal{KL}\big[\tilde{q}(\boldsymbol{v}; \boldsymbol{\delta})\| p_{\boldsymbol{w}}(\boldsymbol{w})\big] - \mathbb{E}_{\boldsymbol{w} \sim \tilde{q}(\boldsymbol{v}; \boldsymbol{\delta})} \big[ \log  \tilde{p}(\boldsymbol{w}; \boldsymbol{\tau}) \big] \big\} \ .
    \label{equ: obj elbo}
\end{equation}
The candidate distributions, $\tilde{q}(\boldsymbol{v}; \boldsymbol{\delta})$ and $\tilde{p}(\boldsymbol{w}; \boldsymbol{\tau})$, are parameterized by variational and generative parameters, $\boldsymbol{\delta}$ and $\boldsymbol{\tau}$, respectively \cite{kingma2013auto}. The first objective of equation \eqref{equ: obj elbo} aims to align $\tilde{q}(\boldsymbol{v}; \boldsymbol{\delta})$ with the prior distribution of $\boldsymbol{w}$, while the second objective focuses on minimizing the reconstruction error. 
As previously mentioned, VAE models both $\tilde{q}$ and $\tilde{p}$ via deep neural networks and $\tilde{q}$ is usually chosen as an uncorrelated multivariate Gaussian distribution~\cite{kingma2019introduction}, with mean and variance equal to
\begin{equation}
\begin{aligned}
&\boldsymbol{\delta} = [\boldsymbol{\mu}_{\boldsymbol{v}}, \boldsymbol{\sigma}_{\boldsymbol{v}}]  = \mathscr{N}_v(\boldsymbol{v}; \boldsymbol{\theta}_v)  \ ,\\
& \tilde{q}(\boldsymbol{v}; \boldsymbol{\delta}) = \mathcal{N}\big(\boldsymbol{\mu}_{\boldsymbol{v}}, \mathrm{\bf diag}(\boldsymbol{\sigma}^2_{\bv})\big) .
\end{aligned}
\label{equ: VAE encoder}
\end{equation}
The network $\mathscr{N}_v$ is referred to as the variational encoder, characterized by the trainable parameter set $\boldsymbol{\theta}_v$.
Finally, \eqref{equ: obj elbo} is solved by stochastic gradient descent using the reparameterization trick~\cite{kingma2013auto}. 
This ensures the gradient and expectation operator can be exchanged, while also requiring minimal modifications to the deterministic back-propagation algorithm.

\subsubsection{The phenomenon of posterior collapsing}\label{sec: pos-colla}
% ===========================================================================

Posterior collapsing is a well-known phenomenon that can arise during the training of VAE networks It describes a scenario in which the learned posterior distribution $q(\bw|\bv)$ becomes remarkably similar to the prior distribution $p_{\bw}(\bw)$, resulting in a latent space independent on the input $\boldsymbol{v}$.
This collapse in the posterior undermines the fundamental nature of a generative model by preventing it from capturing inherent diversity between samples in $\boldsymbol{\mathcal{V}}$.

Posterior collapsing is often attributed to the variational inference objective, e.g. ELBO~\eqref{equ: elbo des} where the posterior distribution is drawn towards the prior, leading to the collapsing phenomenon. To address this issue, one can either reduce decoder flexibility or decrease the impact of the KL loss~\cite{zhao2018infovae,chen2017variational,havrylov2020preventing,fu2019cyclical}.
However, Lucas et al.~\cite{lucas2019don} challenge this perspective by comparing the ELBO objective to the MLE approach in linear VAEs. They suggest that the ELBO objective may not be the only reason for posterior collapse.
Razavi et al.~\cite{razavi2019preventing} propose an alternative approach that avoids modifying the ELBO objective. They introduce $\delta$-VAEs, which are specifically designed to constrain the statistical distance between the learned posterior and the prior.
Furthermore, Wang et al.~\cite{wang2021posterior} have shown that posterior collapses if and only the latent variable is not identifiable in the generative model. To mitigate this issue, they develop LIDVAE (Latent-IDentifiable VAE).

Posterior collapsing can also impact the proposed inVAErt network, making our variational decoder useless due to a trivial latent space.
Through our experiments, we have identified several factors that can readily induce posterior collapse, including network over-parameterization, inadequate penalization of loss function components related to $\mathscr{N}_v$ and $\mathscr{N}_d$, sub-optimal $\ell^2$ regularization and inappropriate data normalization.
Fortunately, during the early phase of training, we can detect a notable decline in the KL divergence loss, which clearly indicates posterior collapse. 
This early detection enables timely intervention, allowing for necessary adjustments of the hyperparameters.

%================================================================
\subsection{inVAErt network training}\label{sec:inVAErt_training}
%================================================================

\begin{figure}[ht!]
\centering
\includegraphics[width=0.8\textwidth]{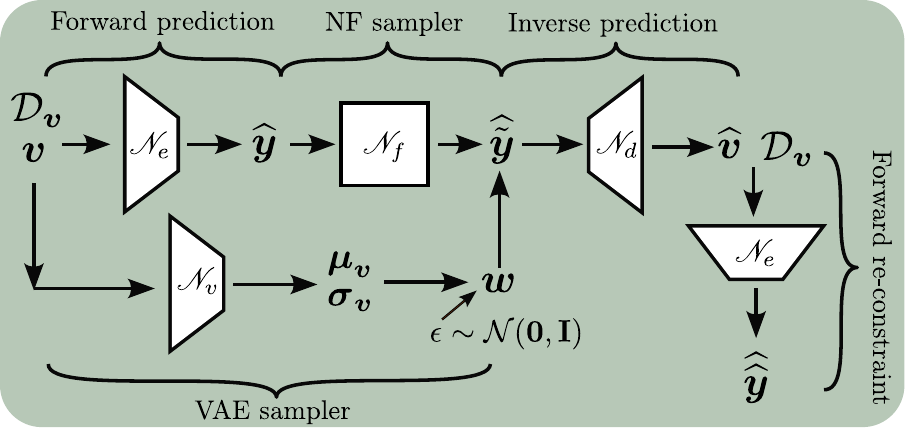}
\caption{Training diagram for the inVAErt network.}
\label{fig:inVAErt_training}
\end{figure}

A complete representation of the proposed inVAErt network is illustrated in Figure~\ref{fig:inVAErt_training}. Forward model evaluations are learned using the deterministic emulator $\mathscr{N}_{e}$. The network learns to generate model outputs from their density thanks to the normalizing flow density estimator $\mathscr{N}_{f}$. The non-identifiable manifold $\boldsymbol{\mathcal{M}}_{\bv}$ and the inverse model map are learned through the VAE and decoder $(\mathscr{N}_{v},\mathscr{N}_{d})$.
% These three components can be trained at the same time or separately. 
Note that, in the current implementation, the emulator, density estimator and VAE are independent components since the output \emph{labels} can be used directly for VAE training. Therefore, these three components can be trained separately. 
However, joint training might be required for extensions to stochastic models, which are not discussed in this paper.

The training and validation procedures of our inVAErt network model are illustrated in Figure~\ref{fig:inVAErt_training}. 
Given a set of $N$ independent samples uniformly distributed in $\boldsymbol{\mathcal{V}}$ and the corresponding outputs $\{(\boldsymbol{v}_i,\boldsymbol{y}_i)\}_{i=1}^N$, an optimal set of parameters $\boldsymbol{\theta} = \boldsymbol{\theta}_e \cup \boldsymbol{\theta}_f \cup \boldsymbol{\theta}_v \cup \boldsymbol{\theta}_d$, is obtained by minimizing the following loss
\begin{equation}\label{equ:loss_obj}
\mathcal{L} = \lambda_e \mathcal{L}_e + \lambda_v \mathcal{L}_v + \lambda_f \mathcal{L}_f + \lambda_d \mathcal{L}_d + \lambda_r \mathcal{L}_r \ ,
\end{equation}
where
\begin{align}
    \mathcal{L}_e(\boldsymbol{y}, \widehat{\boldsymbol{y}}) &= \frac{1}{N} \sum_{i=1}^N \|\boldsymbol{y}_i - \widehat{\boldsymbol{y}}_i \|_{2}^2 \ , \nonumber \\ 
    \mathcal{L}_v &= \frac{1}{2\cdot N} \sum_{i=1}^N \sum_{k=1}^{\dim(\boldsymbol{w})} \Big( \mu_{i,k}^2 + \sigma_{i,k}^2 -\log(\sigma_{i,k}^2)-1 \Big) \ , \nonumber \\ 
    \mathcal{L}_f &= -\frac{1}{N} \sum_{i=1}^N \Big( \log q_0 (\boldsymbol{z}_{i,0}) +  \sum^K_{k=1} \log \Big|\frac{\partial \boldsymbol{z}_{i,k}}{\partial \boldsymbol{z}_{i,k-1}}\Big|^{-1} \Big)  \ , \nonumber \\ 
    \mathcal{L}_d(\boldsymbol{v}, \widehat{\boldsymbol{v}}) &= \frac{1}{N} \sum_{i=1}^N \|\boldsymbol{v}_i -\widehat{\boldsymbol{v}}_i \|_{2}^2  \ , \nonumber \\ 
    \mathcal{L}_r(\boldsymbol{y}, \widehat{\widehat{\boldsymbol{y}}}) &= \frac{1}{N} \sum_{i=1}^N \|\boldsymbol{y}_i -\widehat{\widehat{\boldsymbol{y}}}_i \|_{2}^2  \ ,
    \label{equ: loss functions}
\end{align}
denote the forward MSE loss for the emulator ($\mathscr{N}_e$), KL divergence loss for the VAE encoder ($\mathscr{N}_v$), MLE loss for the Real-NVP density estimator ($\mathscr{N}_f$), reconstruction MSE loss for the decoder model ($\mathscr{N}_d$) and another forward MSE loss to constrain the inverse modeling, respectively. 
Besides, $\lambda_e$, $\lambda_v$, $\lambda_f$, $\lambda_d$, $\lambda_r$ are penalty coefficients associated with each loss function component.

In practice, the loss $\mathcal{L}_r$ is enforced via the trained emulator $\mathscr{N}_e$, computing an approximate output $\widehat{\widehat{\boldsymbol{y}}}_i$ for each inverse prediction $\widehat{\bv}_i$ as $\widehat{\widehat{\boldsymbol{y}}}_i = \mathscr{N}_e(\widehat{\bv}_i)$. 
This loss is consistent with~\eqref{equ: inverse id map-almost sure}, where the forward operator $\mF$ is approximated by $\mathscr{N}_e$.
However, for systems with less complex inverse processes, imposing $\mathcal{L}_r$ is usually not necessary for achieving good performance. Therefore, we do not enforce $\mathcal{L}_r$ in our first three numerical experiments (e.g. Sec~\ref{example: linear}, non-periodic part of Sec~\ref{example: non-linear}, Sec~\ref{example:rcr}). 
%
% \begin{remark}\label{rmk: training sep}
%     In the context of this paper, the forward encoder $\mathscr{N}_e$ and the NF model $\mathscr{N}_f$ can be separately trained with respect to other components of the gVAEs network. This allows the decoder model $\mathscr{N}_d$ always receives the exact label $\boldsymbol{y}$ instead of a prediction $\widehat{\boldsymbol{y}}$. \textbf{We may need to state a case that we could not do separate training.}
% \end{remark}
%
% \begin{remark}\label{rmk: L1L2}
%     For achieving model parsimonicity, we implement the Lasso ($\ell^1$) or Ridge ($\ell^2$) regularization~\cite{goodfellow2016deep} if needed.
% \end{remark}

%=======================================================================
\subsection{inVAErt network interrogation}\label{sec:inVAErt_validation}
%=======================================================================

\begin{figure}[ht!]
\centering
\includegraphics[width=0.8\textwidth]{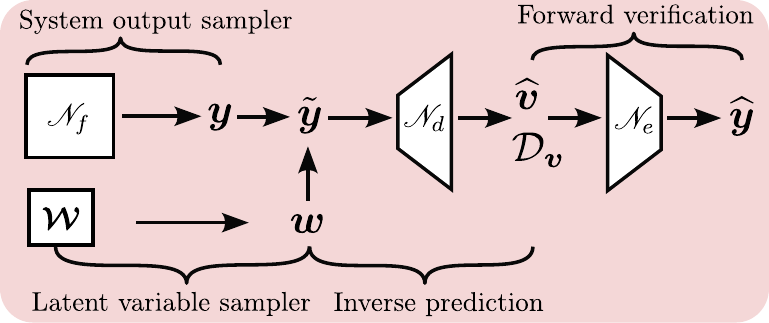}
\caption{Sampling and inference diagram for the inVAErt network.}\label{fig:invaert_inference}
\end{figure}

% instead of inference we use "interrogation" since, once trained, the network can be used in multiple modes
Once trained, an inVAErt network can be used to answer a number of interesting questions about the physical system it synthesizes. A diagram of how the network is used during inference is shown in Figure~\ref{fig:invaert_inference}.

% Fix one output, interrogate for multiple samples in the latent space
We might be interested, for example, in sampling from $\boldsymbol{\mathcal{M}}_{\bv}$, the manifold of input parameters corresponding to the specific model output $\boldsymbol{y}^{*}$. To do so, we sample a number of latent space realizations $\boldsymbol{w}$ from a multivariate standard normal or, alternatively, we use one of the approaches discussed in Section~\ref{sec:analysis_sampling}. We then concatenate these realizations with the observation $\boldsymbol{y}^{*}$ and feed the resulting vector to the decoder. 
% This provides all the solutions instead of regularization
Note that this approach produces multiple possible input parameters and therefore characterizes the inverse properties of the system in a way that is superior to most regularization approaches, where a single solution is promoted, instead of the $\boldsymbol{\mathcal{M}}_{\bv}$ manifold.
%
% The parameters resulting from the decoder can also be filtered
Note also that the inputs $\widehat{\boldsymbol{v}}$ obtained from the decoder can include time and spatial variables (see examples in Section~\ref{example: lorenz} and~\ref{example:pde}).

% Fix one value of the latent space and interrogate for multiple outputs
Alternatively, we can sample from the density of the model outputs $\boldsymbol{y}$ using $\mathscr{N}_{f}$ and set a constant value for the latent variable $\boldsymbol{w}$. 
Ideally, decoding under this process, as opposed to sampling from $\boldsymbol{\mathcal{M}}_{\bv}$, allows one to approximate the manifold transverse to $\boldsymbol{\mathcal{M}}_{\bv}$ (which, in the linear case, coincides with the orthogonal complement $\boldsymbol{\mathcal{M}}_{\bv}^{\perp} = \boldsymbol{\mathcal{V}} \setminus \boldsymbol{\mathcal{M}}_{\bv}$).
This transverse manifold is embedded in $\boldsymbol{\mathcal{V}}$, and is associated with the highest sensitivity in the model outputs.

% local non-identifiable parameters corresponding to a single v
In addition, we can also determine a local collection of non-identifiable neighbors of a given input $\boldsymbol{v}^{*}$. To do so, we decode the concatenation of $\mathcal{F}(\boldsymbol{v}^{*})$ and samples of $\boldsymbol{w}$ from the multivariate normal distribution $\mathcal{N}\big(\boldsymbol{\mu}_{\bv^*}, \mathrm{\bf diag}(\boldsymbol{\sigma}^2_{\bv^*})\big)$, as obtained from the variational encoder $\mathscr{N}_v$ for the input $\boldsymbol{v}^{*}$. 
This process should help in parameter searches to identify directions along which the cost function, formulated in terms of the system outputs, remains constant.

Additional examples related to the practical interrogation of an inVAErt network are provided in Section~\ref{sec: experiments}.

%===================================================
\section{Analysis of the method}\label{sec:analysis}
%===================================================

%======================================================================
\subsection{Gradient of the loss function}\label{sec:analysis_gradient}
%======================================================================

To successfully train the proposed inVAErt network, we empirically find that selection of the penalty coefficients in~\eqref{equ:loss_obj} is of paramount importance. Thus we formally investigate strategies for penalty selection that are informed by first-order stationarity conditions. 
Additionally, since in this work the three components of an inVAErt network (i.e., $\mathscr{N}_e, \mathscr{N}_f, (\mathscr{N}_v, \mathscr{N}_d)$) are trained separately (see the Appendix), we focus on $\lambda_v, \lambda_d, \lambda_r$, and assume the exact forward model $\mathcal{F}$ is available.

First, note that the relations~\eqref{equ: inverse id map-almost sure} and~\eqref{equ: unbiased} lead us to an optimization problem of the form
\begin{align}
   \mathscr{N}_d,\mathscr{N}_v &= \argmin_{\mathscr{D},\mathscr{V} \in \mathfrak{D}, \mathfrak{V}} T(\mathscr{D},\mathscr{V}) \ , \nonumber \\ 
   &= \argmin_{\mathscr{D},\mathscr{V} \in \mathfrak{D}, \mathfrak{V}} \big[ \lambda_r J (\mathscr{D} ,\mathscr{V}) + \lambda_d H (\mathscr{D},\mathscr{V} ) + \lambda_v K( \mathscr{V}) \big] \ ,
    \label{equ: argmin problem}
\end{align}
with the smooth candidate functions $\mathscr{D}$, $\mathscr{V}$ belonging to some general vector spaces $\mathfrak{D}$ and $\mathfrak{V}$, respectively. The target functional $T: \mathfrak{D} \times \mathfrak{V} \to \mathbb{R}$ is composed of three objective functionals $J$, $H$, $K$ corresponding to our loss function components $\mathcal{L}_r, \mathcal{L}_d$ and $\mathcal{L}_v$ in equations~\eqref{equ: loss functions}.

Let $\{(\boldsymbol{v}_i,\boldsymbol{y}_i)\}_{i=1}^N$ (disregarding, for simplicity, the auxiliary data $\mathcal{D}_{\boldsymbol{v}}$) be a set of \emph{i.i.d} training samples, and we associate each sample with a set of standard Gaussian random variables $\{\beps_{ij}\}_{i,j=1}^{N,M}$.
We then define $\boldsymbol{w}_{ij}= \mathscr{V}(\boldsymbol{v}_i)_{\boldsymbol{\epsilon}_{ij}}=\boldsymbol{\mu}_i + \boldsymbol{\epsilon}_{ij}\odot \boldsymbol{\sigma}_i$ to indicate the latent realization obtained from the $i$-th training sample by adding noise $\beps_{ij}$ (i.e. the reparametrization trick).

The first functional $J: \mathfrak{D} \times \mathfrak{V} \to \mathbb{R}$, as our main objective, consists of a discrete version of equation~\eqref{equ: inverse id map-almost sure},
and enforces consistency between the forward and inverse problems
\begin{align}
    J(\mathscr{D},\mathscr{V}) &\coloneqq 
      \frac{1}{2\cdot N\cdot M}  \sum_{i,j = 1}^{N,M} \|\mathcal{F} \big( \mathscr{D} ( \boldsymbol{w}_{ij}, \boldsymbol{y}_i ) \big) - \boldsymbol{y}_i \|^2_2 \,
    \label{equ: J functional}
\end{align}
where we use $\sum_{i,j = 1}^{N,M}$ in place of the nested sum $\sum_{i= 1}^{N}\sum_{j = 1}^{M}$. 
The first constraint $H: \mathfrak{D} \times \mathfrak{V} \to \mathbb{R}$ focuses on minimizing the reconstruction loss of the system input:
\begin{align}
    H(\mathscr{D},\mathscr{V}) &\coloneqq   \frac{1}{2\cdot N\cdot M}  \sum_{i,j = 1}^{N,M} \| \mathscr{D} ( \boldsymbol{w}_{ij}, \boldsymbol{y}_i  ) - \boldsymbol{v}_i  \|^2_2  \ .
    \label{equ: H functional}
\end{align}
Finally, the second constraint $K: \mathfrak{V} \to \mathbb{R}$ promotes compactness in the support of $\boldsymbol{w}$ by minimizing the distance between the posterior distribution of the latent variables and a standard normal prior. 
While any statistical measure of discrepancy between distributions can be used, here we adopt the KL divergence~\eqref{equ: kl}, resulting in
\begin{equation}
     K\big(\mathscr{V}\big) \coloneqq   \frac{1}{2\cdot N}  \sum_{i=1}^{N} \sum_{k=1}^{\dim(\boldsymbol{w})}  \big( \mu_{i,k}^2 + \sigma_{i,k}^2 -\log(\sigma_{i,k}^2)-1 \big) \ .
     \label{equ: K functional}
\end{equation}
In practice, a local minimizer of the full objective may not attain $J(\mathscr{D},\mathscr{V}) = 0$ nor $H(\mathscr{D},\mathscr{V}) = 0$. However, we can still affirm that the local minimizer must satisfy the stationarity conditions for the full objective. Hence, to better understand and interpret the results obtained using our approach, we determine the first-order conditions for a local minimizer of $T$.
Assuming the first functional $J$ is G\^ateaux-differentiable, we can express its directional derivative in terms of its first-order expansion with respect to $\alpha\in \mathbb{R}$ along an arbitrary perturbation function $\Delta \mathscr{D} \in \mathfrak{D}$ as
\begin{equation}
     J(\mathscr{D} + \alpha \Delta \mathscr{D},\mathscr{V}) =  J(\mathscr{D},\mathscr{V}) + \alpha \delta J(\mathscr{D}, \mathscr{V}; \Delta \mathscr{D}) + \omicron(\alpha) \ ,
    \label{equ: J exp}
\end{equation}
where the term $\omicron(\alpha)$ represents any function such that $\lim_{\alpha \to 0} \alpha^{-1}\omicron(\alpha) = 0$. It follows that
\begin{equation}
    \delta J(\mathscr{D}, \mathscr{V}; \Delta \mathscr{D}) \coloneqq \lim_{\alpha \to 0}\frac{ J(\mathscr{D} + \alpha \Delta \mathscr{D},\mathscr{V}) -  J(\mathscr{D},\mathscr{V}) }{\alpha} \ ,
    \label{equ: dir-J}
\end{equation}
is the G\^ateaux derivative, or \emph{first variation}, of the functional $J$ at $\mathscr{D}$ in the direction of $\Delta \mathscr{D}$. In order to simplify our analysis, we denote with the symbol $\omicron(\alpha)$ to any term where $\omicron(\alpha)/\alpha\to 0$ holds as $\alpha\to 0$.

Assuming the forward model $\mathcal{F}$ is Fr\'echet differentiable, we have its first-order Taylor expansion as:
\begin{equation}
    \mathcal{F}( \mathscr{D} + \alpha \Delta \mathscr{D}) = \mathcal{F} ( \mathscr{D} )+ \alpha \nabla \mathcal{F}(\mathscr{D}) \Delta \mathscr{D} + \omicron(\alpha) \ ,
    \label{equ: F exp}
\end{equation}
where $\nabla \mathcal{F}$ denotes the Jacobian matrix of map $\mathcal{F}$ with respect to the map $\mathscr{D}$.
We then merge the expansion~\eqref{equ: F exp} with the definition~\eqref{equ: J functional}, leading to
\begin{align}
     J(\mathscr{D} + \alpha \Delta \mathscr{D},\mathscr{V}) &= \frac{1}{2\cdot N\cdot M}  \sum_{i,j = 1}^{N,M} \Big( \| \mathcal{F}( \mathscr{D} + \alpha \Delta \mathscr{D})( \boldsymbol{w}_{ij}, \boldsymbol{y}_i) - \boldsymbol{y}_i \|^2_2 \Big) \ , \nonumber \\ 
     & =  \frac{1}{2\cdot N\cdot M}  \sum_{i,j = 1}^{N,M} \Big( \| \mathcal{F} \big( \mathscr{D} ( \boldsymbol{w}_{ij}, \boldsymbol{y}_i) \big) -\boldsymbol{y}_i \|^2_2  \ , \nonumber \\ 
      & \quad + 2 \alpha \big \langle \mathcal{F} \big( \mathscr{D} ( \boldsymbol{w}_{ij}, \boldsymbol{y}_i) \big) -\boldsymbol{y}_i ,  \nabla \mathcal{F} \big( \mathscr{D} ( \boldsymbol{w}_{ij}, \boldsymbol{y}_i) \big) \Delta \mathscr{D}( \boldsymbol{w}_{ij}, \boldsymbol{y}_i) \big \rangle  \ , \nonumber \\
     & \quad+ \alpha^2 \| \nabla \mathcal{F} \big( \mathscr{D} ( \boldsymbol{w}_{ij}, \boldsymbol{y}_i) \big) \Delta \mathscr{D} ( \boldsymbol{w}_{ij}, \boldsymbol{y}_i) \|^2_2 \ , \nonumber \\ 
     & \quad +  2 \alpha \big \langle \nabla \mathcal{F} \big( \mathscr{D} ( \boldsymbol{w}_{ij}, \boldsymbol{y}_i) \big) \Delta \mathscr{D} ( \boldsymbol{w}_{ij}, \boldsymbol{y}_i) , \omicron(\alpha)\big \rangle \ , \nonumber \\ 
     & \quad + 2 \big \langle \mathcal{F} \big( \mathscr{D} ( \boldsymbol{w}_{ij}, \boldsymbol{y}_i) \big) -\boldsymbol{y}_i, \omicron(\alpha)\big \rangle + \|\omicron(\alpha)\|^2_2 \Big) \ ,
     \label{equ: J-expanded}
\end{align}
which should match the expansion~\eqref{equ: J exp} above, after neglecting the high order terms related to $\alpha^2$ and $\omicron(\alpha)$. Therefore, we have
\begin{align}
    \delta J(\mathscr{D}, \mathscr{V}; \Delta \mathscr{D}) &= \frac{1}{N\cdot M}  \sum_{i,j=1}^{N,M} \big \langle \mathcal{F} \big( \mathscr{D} ( \boldsymbol{w}_{ij}, \boldsymbol{y}_i) \big) -\boldsymbol{y}_i,  \nabla \mathcal{F} \big( \mathscr{D} ( \boldsymbol{w}_{ij}, \boldsymbol{y}_i) \big) \Delta \mathscr{D}( \boldsymbol{w}_{ij}, \boldsymbol{y}_i) \big \rangle \ , \nonumber \\ 
    & = \frac{1}{N\cdot M}  \sum_{i,j=1}^{N,M} \big \langle \nabla \mathcal{F} \big( \mathscr{D} ( \boldsymbol{w}_{ij}, \boldsymbol{y}_i) \big)^T \Big( \mathcal{F} \big( \mathscr{D} ( \boldsymbol{w}_{ij}, \boldsymbol{y}_i) \big) -\boldsymbol{y}_i \Big), \Delta \mathscr{D}( \boldsymbol{w}_{ij}, \boldsymbol{y}_i) \big \rangle \ .
    \label{equ: 1st-order J}
\end{align}
If we were minimizing only this term, then by the first-order necessary condition for optimality~\cite{liberzon2011calculus} (i.e. $\delta J = 0$), plus the arbitrariness of $\Delta \mathscr{D} \in \mathfrak{D}$, we must have
\begin{equation}
   \nabla \mathcal{F} \big( \mathscr{D} ( \boldsymbol{w}_{ij}, \boldsymbol{y}_i) \big)^T  \Big( \mathcal{F} \big( \mathscr{D} ( \boldsymbol{w}_{ij}, \boldsymbol{y}_i) \big) -\boldsymbol{y}_i \Big) = 0 , \ \forall \ i,j \ ,
\end{equation}
from where it follows that $\mathcal{F} \big( \mathscr{D} ( \boldsymbol{w}_{ij}, \boldsymbol{y}_i) \big)  = \boldsymbol{y}_i$ whenever the Jacobian is full-rank. This matches our expected condition~\eqref{equ: inverse id map-almost sure}.

Following the same procedure, we have that
\begin{align}
    \delta H(\mathscr{D}, \mathscr{V}; \Delta \mathscr{D}) &= \frac{1}{N\cdot M}  \sum_{i,j=1}^{N,M} \big \langle \mathscr{D} ( \boldsymbol{w}_{ij}, \boldsymbol{y}_i) - \boldsymbol{v}_i, \Delta \mathscr{D}( \boldsymbol{w}_{ij}, \boldsymbol{y}_i) \big \rangle \ .
    \label{equ: 1st-order H}
\end{align}
Again, suppose we were minimizing only this term, then the first-order optimality condition applied to the functional $H$ with respect to its first argument would imply that $\mathscr{D} ( \boldsymbol{w}_{ij}, \boldsymbol{y}_i) = \boldsymbol{v}_i$, has to be satisfied for all $i,j$. 
When minimizing the full-objective, i.e. $T$, the first-order conditions with respect to $\mathscr{D}$ become
\begin{equation}
\label{equ: T-diff D}
    0 = \nabla \mathcal{F} \big( \mathscr{D} ( \boldsymbol{w}_{ij}, \boldsymbol{y}_i) \big)^T  \Big( \mathcal{F} \big( \mathscr{D} ( \boldsymbol{w}_{ij}, \boldsymbol{y}_i) \big) -\boldsymbol{y}_i \Big) + \frac{\lambda_d}{\lambda_r} \big(\mathscr{D} ( \boldsymbol{w}_{ij}, \boldsymbol{y}_i) - \boldsymbol{v}_i\big), \qquad\forall\, i,j \ .
\end{equation}
From this, we deduce the following. On one hand, 
\begin{equation}
    \mathscr{D} ( \boldsymbol{w}_{ij}, \boldsymbol{y}_i) = \boldsymbol{v}_i - \frac{\lambda_r}{\lambda_d}\nabla \mathcal{F} \big( \mathscr{D} ( \boldsymbol{w}_{ij}, \boldsymbol{y}_i) \big)^T  \Big( \mathcal{F} \big( \mathscr{D} ( \boldsymbol{w}_{ij}, \boldsymbol{y}_i) \big) -\boldsymbol{y}_i \Big) , \qquad\forall\, i,j \ .
\end{equation}
Therefore, recall in practice, $\mathscr{N}_e \approx \mathcal{F}$ is utilized during the training of $(\mathscr{N}_v, \mathscr{N}_d)$, hence any error in the emulator might negatively affect the decoder. This suggests, for example, increasing the value of \(\lambda_d\) while letting $\lambda_r$ fixed can reduce bias in $\mathscr{D}$. 
On the other hand, if the Jacobian of \(\mathcal{F}\) is full-rank, we then have
\begin{equation}
    \mathcal{F} \big( \mathscr{D} ( \boldsymbol{w}_{ij}, \boldsymbol{y}_i) \big) = \boldsymbol{y}_i - \frac{\lambda_d}{\lambda_r} \Big(\nabla \mathcal{F}\big(\mathscr{D}(\boldsymbol{w}_{ij},\boldsymbol{y}_i)\big)^T\Big)^{+}\big(\mathscr{D} ( \boldsymbol{w}_{ij}, \boldsymbol{y}_i) - \boldsymbol{v}_i\big), \qquad\forall i,j \ ,
    \label{equ: on the other hand}
\end{equation}
where \({(\cdot)}^+\) denotes the Moore-Penrose pseudo-inverse. Therefore, any error in the decoder becomes a prediction error. In fact, the effect that the error in the decoder has on the prediction error is scaled by the pseudo-inverse of the Jacobian and thus depends on the local conditioning of the problem at \(\mathscr{D}(\boldsymbol{w}_{ij}, \boldsymbol{y}_i)\). Furthermore, this effect is proportional to $\frac{\lambda_d}{\lambda_r}$ , showing that a balance must be reached when modulating the combination of $\lambda_d, \lambda_r$, based on the expected training complexity of the decoder and the accuracy of the trained emulator, respectively. 
For instance, when dealing with difficult-to-learn forward maps and so the expected emulator accuracy is limited, it becomes crucial to reduce $\lambda_r$ to prevent the training of $\mathscr{N}_d$ being misled. 
%
%As a result, as suggested by equation~\eqref{equ: on the other hand}, an increase in $\lambda_d$ becomes necessary for proper compensation.

% %
% For challenging inverse problems, an increase in $\lambda_1$ can be complemented with an increase in the dimensionality of $\boldsymbol{\mathcal{W}}$.

We then proceed to compute the derivatives of $\mathscr{V} \in \mathfrak{V}$. 
This process is equivalent to deriving first-order conditions with respect to $\bmu_i$ and $\bsigma_i$ through the VAE architecture. First, consider the following expansion of $J$
\begin{align}
     J(\msD, \msV + \alpha \Delta &\msV) = \frac{1}{2\cdot N\cdot M}  \sum_{i,j=1}^{N,M} \nrm{\mF\big(\msD(\bmu_i + \alpha\Delta\bmu_i + (\bsigma_i + \alpha\Delta\bsigma_i)\odot \beps_{ij}, \by_i)\big) - \by_i}^2_2 \ , \nonumber\\ 
     &=J(\msD,\msV) \ , \nonumber \\
     &+ \frac{\alpha}{N\cdot M}  \sum_{i,j=1}^{N,M} \big \langle \mF\big(\msD(\bw_{ij}, \by_i)\big) - \by_i, \nabla \mF \big(\msD(\bw_{ij},\by_i)\big) \nabla \msD(\bw_{ij},\by_i)\Delta\bmu_i \big \rangle \ , \nonumber\\
     &+ \frac{\alpha}{N\cdot M}  \sum_{i,j=1}^{N,M} \big \langle \mF \big(\msD(\bw_{ij}, \by_i)\big) - \by_i, \nabla \mF \big(\msD(\bw_{ij},\by_i)\big) \nabla \msD(\bw_{ij},\by_i)\Delta\bsigma_i \odot \beps_{ij} \big \rangle \ , \nonumber\\
     &+ \omicron(\alpha) \ ,
\end{align}
where $\nabla \mathscr{D}$ denotes the Jacobian matrix of $\mathscr{D}$ with respect to $\mathscr{V}$. Following similar procedure as~\eqref{equ: 1st-order J}, the G\^ateaux derivative of \(J\) is then
\begin{align}
\begin{split}
    \delta J(\msD,&\msV;\Delta \msV) = \frac{1}{N\cdot M}  \sum_{i,j=1}^{N,M} \big \langle \nabla \msD(\bw_{ij},\by_i)^T \nabla \mF \big( \msD(\bw_{ij},\by_i)\big)^T \Big(\mF \big( \msD(\bw_{ij}, \by_i)\big) - \by_i\Big), \Delta\bmu_i \big \rangle \ , \\
    &+ \frac{1}{N\cdot M}  \sum_{i,j=1}^{N,M} \big \langle \beps_{ij}\odot \nabla \msD(\bw_{ij},\by_i)^T \nabla \mF \big( \msD(\bw_{ij},\by_i)\big)^T \Big(\mF \big( \msD(\bw_{ij}, \by_i)\big) - \by_i\Big), \Delta\bsigma_i\big \rangle \ .
\end{split}
\label{eq: J-diff V}
\end{align}
The same expansion for \(H\) and its G\^ateaux derivative results in
\begin{align}
     H(\msD, \msV + \alpha \Delta \msV) &= \frac{1}{2\cdot N\cdot M}  \sum_{i,j=1}^{N,M} \nrm{\msD \big(\bmu_i + \alpha\Delta\bmu_i + (\bsigma_i + \alpha\Delta\bsigma_i)\odot \beps_{ij}, \by_i \big) - \bv_i}^2_2 \ , \nonumber \\ 
     &=H(\msD,\msV) \ , \nonumber \\
     &+ \frac{\alpha}{N\cdot M}  \sum_{i,j=1}^{N,M} \big \langle \msD(\bw_{ij}, \by_i) - \bv_i, \nabla \msD(\bw_{ij},\by_i)\Delta\bmu_i \big \rangle \ , \nonumber \\
     &+ \frac{\alpha}{N\cdot M}  \sum_{i,j=1}^{N,M} \big \langle  \msD(\bw_{ij}, \by_i) - \bv_i,  \nabla \msD(\bw_{ij},\by_i) \Delta\bsigma_i \odot \beps_{ij} \big \rangle \ , \nonumber \\
     &+ \omicron(\alpha) \ .
\end{align}
\begin{align}
\begin{split}
    \delta H(\msD,\msV;\Delta\msV) &= \frac{1}{N\cdot M} \sum_{i,j=1}^{N,M} \big \langle \nabla \msD(\bw_{ij},\by_i)^T \big( \msD(\bw_{ij}, \by_i) - \bv_i\big) , \Delta\bmu_i \big \rangle \ , \\
    &+ \frac{1}{N\cdot M}  \sum_{i,j=1}^{N,M} \big \langle \beps_{ij}\odot \nabla \msD(\bw_{ij},\by_i)^T \big( \msD(\bw_{ij}, \by_i) - \bv_i\big) , \Delta\bsigma_i \big \rangle \ .
\end{split}
\label{eq: H-diff V}
\end{align}

Finally, we compute the derivative of \(K\) in~\eqref{equ: K functional}. Its variation along an arbitrary direction $\Delta \msV \in \mathfrak{V}$ is
\begin{align}
    K(\msV + \alpha\Delta\msV) &= K(\bmu_i + \alpha \Delta \bmu_i, \bsigma_i + \alpha \Delta \bsigma_i) \ , \nonumber \\ 
    & = K(\msV) + \frac{\partial K}{\partial \bmu_i} \Delta \bmu_i + \frac{\partial K}{\partial \bsigma_i} \Delta \bsigma_i +\omicron(\alpha)  \ ,  \nonumber \\
    & = K(\msV) + \frac{\alpha}{N} \sum_{i=1}^N \big( \bmu_i \Delta \bmu_i + (\bsigma_i - \bsigma_i^{-1})\Delta \bsigma_i  \big ) + \omicron(\alpha) \ ,
    \label{equ: K directional derivative}
\end{align}
where \(\bsigma_i^{-1}\) is to be interpreted componentwise. Hence, the G\^ateaux derivative $\delta K$ can be computed explicitly as:
\begin{equation}
    \delta K(\msV; \Delta\msV) = \frac{1}{N}\sum_{i=1}^N \left( \bmu_i \Delta \bmu_i + (\bsigma_i - \bsigma_i^{-1})\Delta \bsigma_i  \right).
    \label{equ: K with mu sigma} 
\end{equation}
Collecting terms, the first-order optimality condition with respect to the mean \(\bmu_i\) implies that 
\begin{align}
\begin{split}
0 &= \frac{\lambda_r}{M}\sum_{j=1}^{M} \nabla \msD(\bw_{ij},\by_i)^T \nabla \mF \big(\msD(\bw_{ij},\by_i)\big)^T \Big( \mF \big(\msD(\bw_{ij}, \by_i)\big) - \by_i\Big) \ , \\
    &+ \frac{\lambda_d}{M}  \sum_{j=1}^{M} \nabla \msD(\bw_{ij},\by_i)^T \big( \msD(\bw_{ij}, \by_i) - \bv_i \big) \ , \\
    &+ \lambda_v \bmu_i, \,\,\forall\, i \ .
\end{split}
\end{align}
In particular, this yields an implicit relation for the mean
\begin{align}
\begin{split}
\bmu_i &= -\frac{\lambda_r}{\lambda_v} \frac{1}{M} \sum_{j=1}^{M} \nabla\msD(\bw_{ij},\by_i)^T \nabla\mF \big(\msD(\bw_{ij},\by_i)\big)^T \Big(\mF \big(\msD(\bw_{ij}, \by_i) \big) - \by_i\Big ) \ , \\
    &- \frac{\lambda_d}{\lambda_v} \frac{1}{M}  \sum_{j=1}^{M} \nabla \msD(\bw_{ij},\by_i)^T \big(\msD(\bw_{ij}, \by_i) - \bv_i \big) \ , \qquad\forall\, i \ .
\end{split}
\label{equ: optimal mu}
\end{align}
The first-order optimality with respect to the standard deviation \(\bsigma_i\) yields
\begin{align}
\begin{split}
0 &= \frac{\lambda_r}{M}\sum_{j=1}^{M} \beps_{ij}\odot \nabla \msD(\bw_{ij},\by_i)^T \nabla \mF \big(\msD(\bw_{ij},\by_i) \big)^T \Big(\mF \big(\msD(\bw_{ij}, \by_i) \big) - \by_i \Big) \ , \\
    &+ \frac{\lambda_d}{M}  \sum_{j=1}^{M} \beps_{ij}\odot \nabla \msD(\bw_{ij},\by_i)^T \big( \msD(\bw_{ij}, \by_i) - \bv_i \big) \ , \\
    &+ \lambda_v (\bsigma_i - \bsigma_i^{-1}) \ ,   \qquad\forall\, i \ .
\end{split}
\end{align}
This expression also leads to an implicit relation for the variance. Since \(\bw_{ij} - \bmu_i = \bsigma_i \odot \beps_{ij}\), we have that
\begin{align}
\bsigma_i^2 &= -\frac{\lambda_r}{\lambda_v} \frac{1}{M} \sum_{j=1}^{M} (\bw_{ij} - \bmu_i)\odot \nabla \msD(\bw_{ij},\by_i)^T \nabla\mF \big(\msD(\bw_{ij},\by_i)\big)^T \Big(\mF \big( \msD(\bw_{ij}, \by_i) \big) - \by_i \Big) \ ,\nonumber\\
    &- \frac{\lambda_d}{\lambda_v} \frac{1}{M}  \sum_{j=1}^{M} (\bw_{ij} - \bmu_i)\odot \nabla \msD(\bw_{ij},\by_i)^T \big( \msD(\bw_{ij}, \by_i) - \bv_i \big)  + \boldsymbol{1} \ ,    \qquad\forall\, i \ ,
\label{equ: optimal sigma}
\end{align}
where $\boldsymbol{1}$ denotes a vector of ones with size $\dim(\boldsymbol{w})$.

It is illustrative to evaluate these conditions when \(\mF\) is a full-rank linear map \(\bF\). In this case, the condition~\eqref{equ: T-diff D} becomes
\begin{equation}
    0 = \bF^T \big( \bF \msD(\bw_{ij}, \by_i) - \by_i \big) + \frac{\lambda_d}{\lambda_r} \big(\msD(\bw_{ij}, \by_i) - \bv_i \big) \ , \qquad\forall i,j \ ,
\end{equation}
whence
\begin{align}
    \msD(\bw_{ij}, \by_i) &= (\bF^T \bF + \frac{\lambda_d}{\lambda_r} \mathbf{I})^{-1}(\frac{\lambda_d}{\lambda_r} \bv_i + \bF^\top \by_i) \ , \nonumber \\
    &= (\bF^T\bF + \frac{\lambda_d}{\lambda_r} \mathbf{I})^{-1}(\frac{\lambda_d}{\lambda_r} \mathbf{I} + \bF^T \bF)\bv_i \ ,\\
    &= \bv_i \ , \qquad\forall i,j \ .
\end{align}
Hence, the decoder in this case is exact and independent of \(\bw\), which then turns the conditions~\eqref{equ: optimal mu} and~\eqref{equ: optimal sigma} into
\[
\bmu_i = \boldsymbol{0}\quad\mbox{and}\quad \bsigma^2_i = \boldsymbol{1} \ .
\]
In other words, we have {\em posterior collapse} (see Section~\ref{sec: pos-colla}). In this example, this seems to be the consequence of {\em exact decoding}. In fact, when we have
\[
    \msD(\bw_{ij},\by_i) \equiv \bv_i \ ,
\]
then~\eqref{equ: optimal mu} and~\eqref{equ: optimal sigma} show that we must have posterior collapse. Although surprising, this is intuitive. 
When the decoding is exact, the decoder becomes independent of \(\bw\), which implies that optimal training is achieved at the global optimum of \(K\).

%==============================================================
\subsection{Latent space sampling}\label{sec:analysis_sampling}
%==============================================================

As discussed in Section~\ref{sec: vae}, a VAE forces the posterior distribution of $\boldsymbol{w}$ to be close to a standard normal, promoting latent spaces with concentrated support in $\boldsymbol{\mathcal{W}}$.
%
%Consequently, during the evaluation phase, we draw samples from this distribution to conduct inverse prediction. 
%
%Geometrically, the goal is for the standard Gaussian distribution to (\textbf{\color{green}compactly?}) cover all data-dependent posterior distributions of $\boldsymbol{w}$.
%
However, in practice, the process $\boldsymbol{W}_{\boldsymbol{v}}$ is composed by the mixture of multiple uncorrelated multivariate Gaussian densities (one for each training example) and might contain regions with probability much lower than the standard normal. 
As we will discuss in Section~\ref{sec: experiments}, this may lead the decoder to generate incorrect inverse predictions when $\boldsymbol{w}$ comes from low-density posterior regions.
We refer to these spurious predictions as \emph{outliers} and dedicate the next sections to compare approaches to effectively sample from the latent space posterior.

% \begin{remark} \label{rmk: outliers}
% The structure of the non-identifiable manifold $\boldsymbol{\mathcal{M}}\subset \boldsymbol{\mathcal{V}}$ is learned via the stochastic component of the VAE. In other words, our inVAErt network learns to transform a standard Gaussian distribution into a distribution that resembles $\boldsymbol{\mathcal{M}}$. 
% %
% However, learning this deformation becomes challenging, especially when $\boldsymbol{\mathcal{M}}$ exhibits complex features such as discontinuity.
% %
% Therefore, as demonstrated in Section~\ref{sec: experiments}, we hypothesize that a primary contributor to the outliers is the transformation from a continuous Gaussian distribution into a disconnected one that mimics $\boldsymbol{\mathcal{M}}$.
% \end{remark}

%==========================================================================
\subsubsection{Direct sampling from the prior} \label{sec: direct sampling}
%==========================================================================

A straightforward approach to generate samples from the entire $\boldsymbol{\mathcal{W}}$ is to sample from the standard normal prior. This approach performs well for simple inverse problems, as shown in a few numerical experiments of Section~\ref{sec: experiments}.
Nevertheless, as the complexity of the inverse problem increases, this straightforward approach tends to produce sub-optimal results. 
To address this limitation, we introduce alternative strategies in the following sections, which have proven to outperform the basic sampling method.  

%===============================================================================
\subsubsection{Predictor-Corrector (PC) sampling}\label{sec: pc sampling}
%===============================================================================
An alternative approach to reduce the number of outliers is the \emph{predictor-corrector} method (see Algorithm~\ref{alg: pc sampling}).
\begin{algorithm*}[ht!]
\caption{Predictor-corrector (PC) sampling.}
\begin{algorithmic}[1]
\State For a given $\boldsymbol{y}^*$, sample $\boldsymbol{w}_{[0]}$ from $\mathcal{N}(\boldsymbol{0}, \mathbf{I})$, or use a given set by other sampling method
\State Concatenate and decode to produce $\widehat{\boldsymbol{v}}_{[0]} = \mathscr{N}_d (\boldsymbol{w}_{[0]}, \boldsymbol{y}^*)$ \Comment{Predictor step}
\For{$r= 1, 2, \cdots, R$} \Comment{Correction loop}
\State VAE encode: $[\boldsymbol{\mu}_{[r]}, \boldsymbol{\sigma}_{[r]}] = \mathscr{N}_v (\widehat{\boldsymbol{v}}_{[r-1]})$
\State Assign mean: $\boldsymbol{w}_{[r]} = \boldsymbol{\mu}_{[r]} $  \Comment{Denoted as the operator: $\mathscr{N}_{v, \mu} (\cdot )$}
\State Concatenate and decode: $\widehat{\boldsymbol{v}}_{[r]}  = \mathscr{N}_d (\boldsymbol{w}_{[r]}, \boldsymbol{y}^*)$
\EndFor
\State Output $\widehat{\boldsymbol{v}}_{[R]}$
\end{algorithmic}
\label{alg: pc sampling}
\end{algorithm*}

It consists of an iterative approach expressed as
\[
\boldsymbol{\beta}_{[r]} + \widehat{\boldsymbol{v}}_{[r]} = \mathscr{N}_d(\boldsymbol{w}_{[r]}, \boldsymbol{y}^*) = \mathscr{N}_d(\mathscr{N}_{v,\mu} (\widehat{\boldsymbol{v}}_{[r-1]}), \boldsymbol{y}^*),\,\,r=1,\dots,R \ ,
\]
where $\boldsymbol{\beta}_{[r]}$ is the prediction error.
%, and $\mathring{\boldsymbol{v}}$ denotes the exact inverse solution with latent space encoding $[\mathring{\boldsymbol{\mu}}, \mathring{\boldsymbol{\sigma}}]^T = \mathscr{N}_v(\mathring{\boldsymbol{v}})$. 
%
Now assuming the term $H$ is sufficiently small in~\eqref{equ: argmin problem} as a result of gradient-based optimization, then $\widehat{\boldsymbol{v}}_{[r]} \approx\widehat{\boldsymbol{v}}_{[r-1]}$, , at least $\forall\,\widehat{\boldsymbol{v}}_{[r]}$ in the training set. Therefore, the operator $\mathscr{N}_d\big(\mathscr{N}_{v,\mu}(\cdot)\big)$ behaves as a \emph{contraction}.
In other words, a deterministic inVAErt network acts as a denoising autoencoder, but it learns to reduce the effects of latent space noise rather than noise in the input space.
%
% Since $\boldsymbol{w}\sim \mathcal{N}\big(\mathring{\boldsymbol{\mu}}, \textbf{diag}(\mathring{\boldsymbol{\sigma}})\big)$, so if we are most likely to obtain $\mathring{\boldsymbol{v}}$ during inverse prediction if $\boldsymbol{w} = \mathring{\boldsymbol{\mu}}$.
% %
% During the evaluation phase, the reparametrization process is turned off, thus if we plug $\widehat{\boldsymbol{v}}$ back to the trained VAE encoder, we will have $\widehat{\boldsymbol{\mu}} = \mathscr{N}_v(\widehat{\boldsymbol{v}})$. Now provided $|\boldsymbol{\gamma}|$ is small enough such that $\widehat{\boldsymbol{v}} \approx \mathring{\boldsymbol{v}}$ and $\widehat{\boldsymbol{\mu}} \approx \mathring{\boldsymbol{\mu}}$, then we are more like to get closer to $\mathring{\boldsymbol{v}}$ if we re-assign $\widehat{\boldsymbol{\mu}}$ to $\boldsymbol{w}$ and re-apply the trained decoder $\mathscr{N}_d$.
% %
% Through the above illustration, it becomes apparent that PC-sampling functions as a mechanism for \emph{de-noising}, as it relies on the quality of the initial set of standard normal samples of $\boldsymbol{w}$ that keeps the prediction error $\boldsymbol{\gamma}$ at a low level. 

In addition, the presence of potential overlap among latent density components $\{\boldsymbol{W}_{\boldsymbol{v}_{i}}\}_{i=1}^{N}$ can lead our iterative correction process to prioritize distributions with smaller support, and therefore to promote contractions towards a small subset of the entire training set.
A latent space sample $\boldsymbol{w}$ belonging to the support of multiple density components from $\{\boldsymbol{W}_{\boldsymbol{v}_{i}}\}_{i=1}^{N}$ will tend to move towards the mean of the component with the highest density at $\boldsymbol{w}$, since such samples have been observed more often during training.
This poses limitation on the flexibility of PC sampling to explore the global structure of the non-identifiable manifold (see, e.g., Figure~\ref{fig: r50-traj} in Section~\ref{example: non-linear}).
Thus, in practical applications, the trade-off between the number of outliers and maintaining an interpretable manifold structure can be achieved by varying the number $R$ of iterations.

%================================================================
\subsubsection{High-Density (HD) sampling} \label{sec: sampling rank pdf}
%================================================================
A second sampling method (see Algorithm~\ref{alg: pdf sampling}) leverages the training data to draw more relevant latent samples, then ranks the samples based on the corresponding posterior density.
\begin{algorithm*}[ht!]
\caption{High-Density (HD) sampling.}
\begin{algorithmic}[1]
\State Take a random subset of the training data input $\mathbf{V}^S \subset \mathbf{V} = \{\boldsymbol{v}_i\}_{i=1}^N  $ of size $S$
\For{$\boldsymbol{v}_i$ in $\mathbf{V}^S$}
\State VAE encode: $[\boldsymbol{\mu}_i, \boldsymbol{\sigma}_i] = \mathscr{N}_v (\boldsymbol{v}_{i})$
\For{$j=1:Q$} \Comment{$Q$: sub-sampling size}
\State Sample latent variable from the data-dependent distribution: $\boldsymbol{w}_{ij} \sim \mathcal{N}\big(\boldsymbol{\mu}_i, \textbf{diag}(\boldsymbol{\sigma}_i^2)\big)$
\State Record $[\boldsymbol{w}_{ij}, \boldsymbol{\mu}_i, \boldsymbol{\sigma}_i]$
\EndFor
\EndFor
\State Initialize the PDF matrix as zeros: $\boldsymbol{p} \in \mathbb{R}^{S\times Q}$
\For{$i=1:S$}
\State Evaluate PDF and consider the overlapping: $\boldsymbol{p} \mathrel{+}= \mathcal{N}\big(\boldsymbol{w}; \boldsymbol{\mu}_i, \textbf{diag}(\boldsymbol{\sigma}_i^2)\big)$
\EndFor
\State Rank all $S\cdot Q$ samples based on their stacked PDF values and take samples from the top
\end{algorithmic}
\label{alg: pdf sampling}
\end{algorithm*}

It is important to note that the selection of the subset size $S$ and the sub-sampling size $Q$ requires careful consideration to prevent the loss of important latent space features. 
Similar to PC sampling, HD sampling also tends to contract samples towards the means of highly-concentrated latent distributions. In practice, relative to the size of the entire dataset, we typically keep both $S$ and $Q$ small to attain more informative samples.

%======================================================================
\subsubsection{Sampling with normalizing flow} \label{sec: nf sampling}
%======================================================================

An additional sampling method considered in this paper uses normalizing flows to map the VAE latent space to a standard normal (we refer this approach as \emph{NF sampling}). Similar approaches have been suggested in~\cite{brehmer2020flows,morrow2020variational}, for improving VAE and manifold learning.

\begin{figure}[ht!]
    \centering
    \includegraphics[scale=0.25]{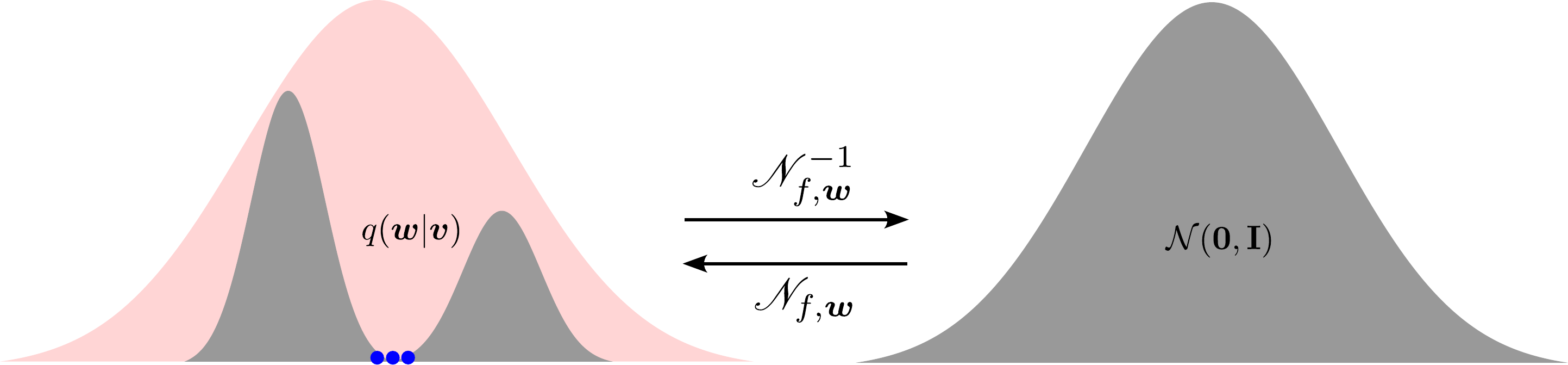}
    \caption{Diagram of the normalizing flow model built on posterior distribution $q(\boldsymbol{w}|\boldsymbol{v})$ of the latent variable $\boldsymbol{w}$.}
    \label{fig: nf-w}
\end{figure}

The learned Real-NVP based NF model, denoted as $\mathscr{N}_{f,\boldsymbol{w}}$, is conditioned on the learned posterior distribution $q(\bw|\bv)$, which is a mixture of multivariate Gaussian densities. 
We show the benefit of NF sampling by an exaggerated case in Figure~\ref{fig: nf-w}, where the transformation $\mathscr{N}_{f,\boldsymbol{w}}$ can effectively reduce the probability of selecting samples with high prior but low posterior density (see the blue dots in Figure~\ref{fig: nf-w}).

Indeed, the NF sampling produces similar results to HD sampling (see Algorithm~\ref{alg: pdf sampling}), has minimal reliance on hyperparameters (as opposed to $S$ and $Q$ in HD sampling), and improves with an increased amount of training data.
In practice, like Algorithm~\ref{alg: pdf sampling}, we first feed the trained VAE encoder $\mathscr{N}_v$ with the input set $\mathbf{V}^S$ to obtain samples of $\boldsymbol{w}$ which correspond to inputs in the training set, then estimate their density with the new NF model $\mathscr{N}_{f,\boldsymbol{w}}$.
%===========================================
\section{Experiments}\label{sec: experiments}
%===========================================

%===========================================
\subsection{Underdetermined linear system} \label{example: linear}
%===========================================
As a first example, we consider an under-determined linear system from $ \mathbb{R}^3 \to \mathbb{R}^2$ defined as
\begin{equation}
    \boldsymbol{y} = 
    \begin{bmatrix}
        y_1 \\ y_2 
    \end{bmatrix}
    =
    \begin{bmatrix}
        \pi & e & 0\\
          0 & e & \pi
    \end{bmatrix}
    \cdot 
    \begin{bmatrix}
        v_1 \\ v_2 \\ v_3 
    \end{bmatrix}
    = \mathbf{F}\,\boldsymbol{v}.
    \label{equ: linear system}
\end{equation}

The linear map represented by the matrix $\mathbf{F}$ is surjective (on-to) but non-injective (one-to-one) such that we have a non-trivial one-dimensional kernel of the form
\begin{equation}
\textrm{Ker}(\mathbf{F}) \subset \mathbb{R} \approx c^*[0.5475278,-0.6327928, 0.5475278]^T,
\label{equ: kernel of linear sys}
\end{equation}
where $c^* \in \mathbb{R}$ is an arbitrary constant and the kernel direction is normalized. As a result, any fixed $\boldsymbol{v}$ in $\mathbb{R}^3$ translating in the direction of $\textrm{Ker}(\mathbf{F})$ will leave the output $\boldsymbol{y}$ invariant. In another words, this system is non-identifiable along $\textrm{Ker}(\mathbf{F})$. This line represents the manifold $\boldsymbol{\mathcal{M}}_{\bv}$ of non-identifiable parameters for this linear map $\mathbf{F}$, as discussed above.

Due to the simplicity of this example, we omit showing the performance of the emulator $\mathscr{N}_e$ and density estimator $\mathscr{N}_f$, and focus on the VAE and decoder components $\mathscr{N}_e$ and $\mathscr{N}_d$.
We generate the dataset by letting: $\boldsymbol{v} \sim [\mathcal{U}(0,5)]^3$
and forward the exact model $10^4$ times to gather the ground truth. 
In addition, a recommended set of hyperparameters for this example is reported in the Appendix.

\begin{figure}[ht!]
     \centering
     \begin{subfigure}[b]{0.4\textwidth}
         \centering
         \includegraphics[scale=0.23]{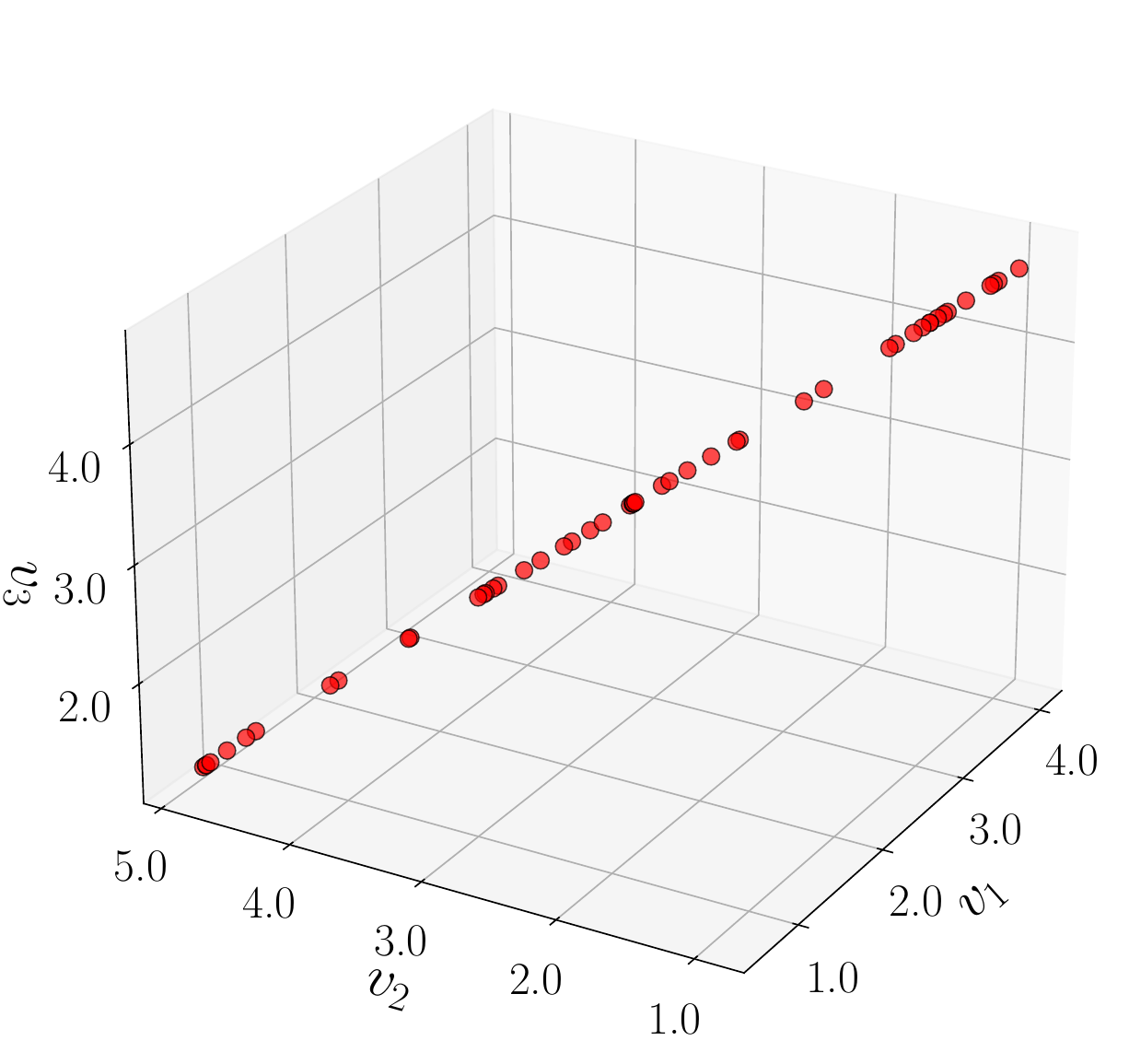}
         \caption{Decoded samples $\widehat{\boldsymbol{v}}$ for a fixed output $\boldsymbol{y}^*$.}
     \end{subfigure}
     \hfill
     \begin{subfigure}[b]{0.48\textwidth}
         \centering
         \includegraphics[scale=0.23]{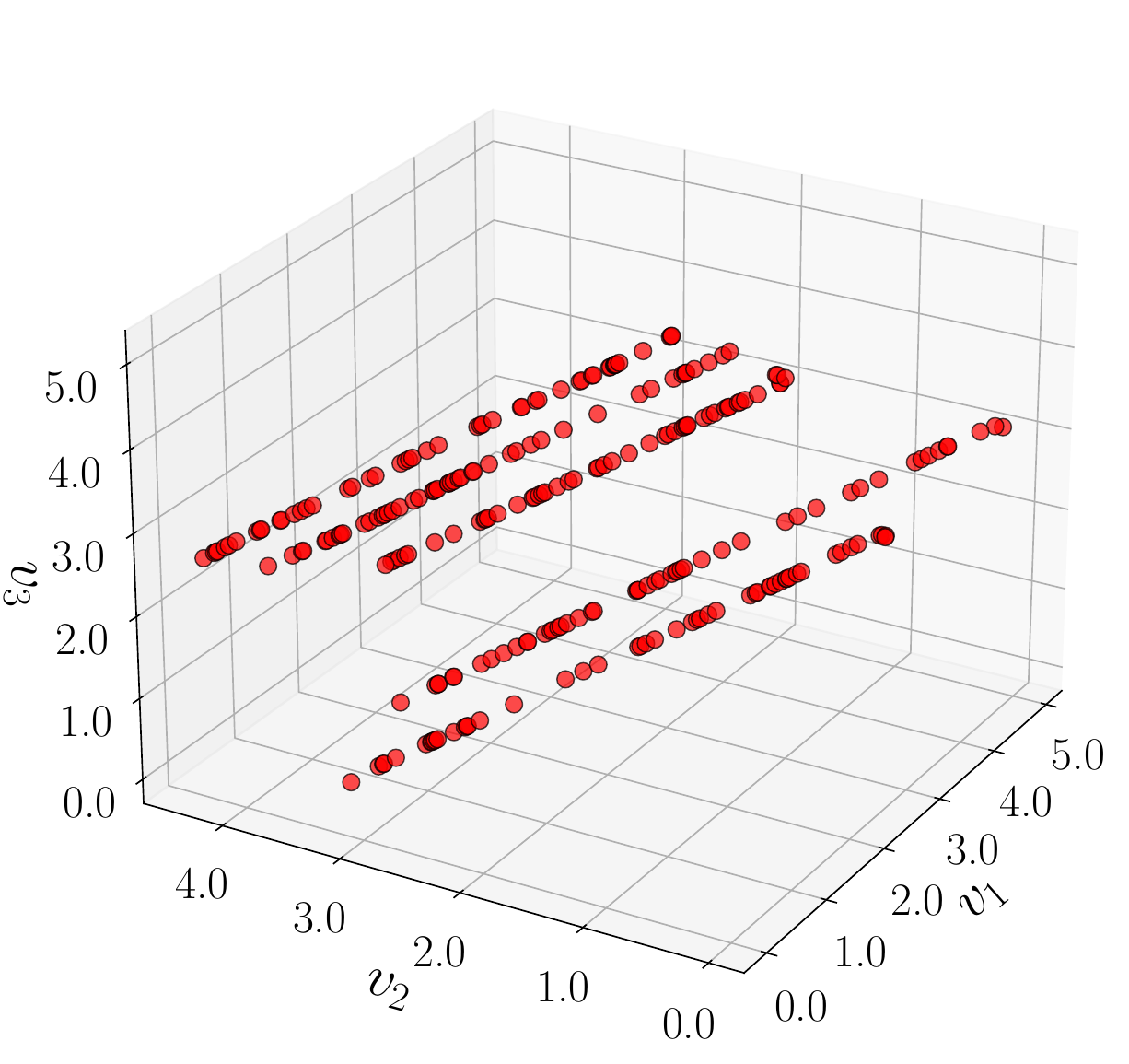}
         \caption{Decoded samples $\widehat{\boldsymbol{v}}$ when considering five different outputs $\boldsymbol{y}^*$.}
     \end{subfigure}
    \caption{Model inversion results for the underdetermined linear system. We fix one or multiple outputs $\boldsymbol{y}^*$ and concatenate it with scalar samples $w$ from the one-dimensional latent space.}
    \label{fig: simple-linear fixy-sampleW}
\end{figure}

% Fix output and sample from the latent space
First, we select a fixed $\boldsymbol{y}^*$ and reconstruct inputs in $\mathbb{R}^{3}$ by sampling a scalar $w$ from the latent space. 
Due to a non-trivial null space, if $\boldsymbol{y}^* = \mathbf{F}\boldsymbol{v}^*$, any input of the form $\boldsymbol{v}^*+\textrm{Ker}(\mathbf{F})$ will map to the same $\boldsymbol{y}^*$.
Therefore the reconstructed inputs are expected to lay on a straight line in $\mathbb{R}^{3}$ aligned with the direction of $\textrm{Ker}(\mathbf{F})$~\eqref{equ: kernel of linear sys}, passing through $\boldsymbol{v}^*$. 
This behaviour is correctly reproduced as shown in Figure~\ref{fig: simple-linear fixy-sampleW}, where we either fix $\boldsymbol{y}^*$ at 1 value or 5 different values, sampled from $\mathscr{N}_f$. For each fixed $\boldsymbol{y}^*$, we draw 50 samples of the latent variable $w$ from $\mathcal{N}(0,1)$ to formulate $\tilde{\boldsymbol{y}}^*$ and then pass it through the trained decoder ($\mathscr{N}_d$), resulting a series of points in $\boldsymbol{\mathcal{V}}$. 
A simple linear regression through these locations provides a normalized direction vector (averaged of the 5 sets shown in Figure~\ref{fig: simple-linear fixy-sampleW}) equal to
\[
[-0.549, 0.632, -0.547]^T
\]
which is close to $\textrm{Ker}(\mathbf{F})$~\eqref{equ: kernel of linear sys}. 
%
% Note that this phenomenon also matches with our illustration in diagram~\ref{fig: extension}, where space $\boldsymbol{\mathcal{W}}$ is assumed to be the orthogonal complement of space $\boldsymbol{\mathcal{Y}}$.

\begin{figure}[ht!]
     \centering
     \begin{subfigure}[b]{0.22\textwidth}
         \centering
         \includegraphics[scale=0.165]{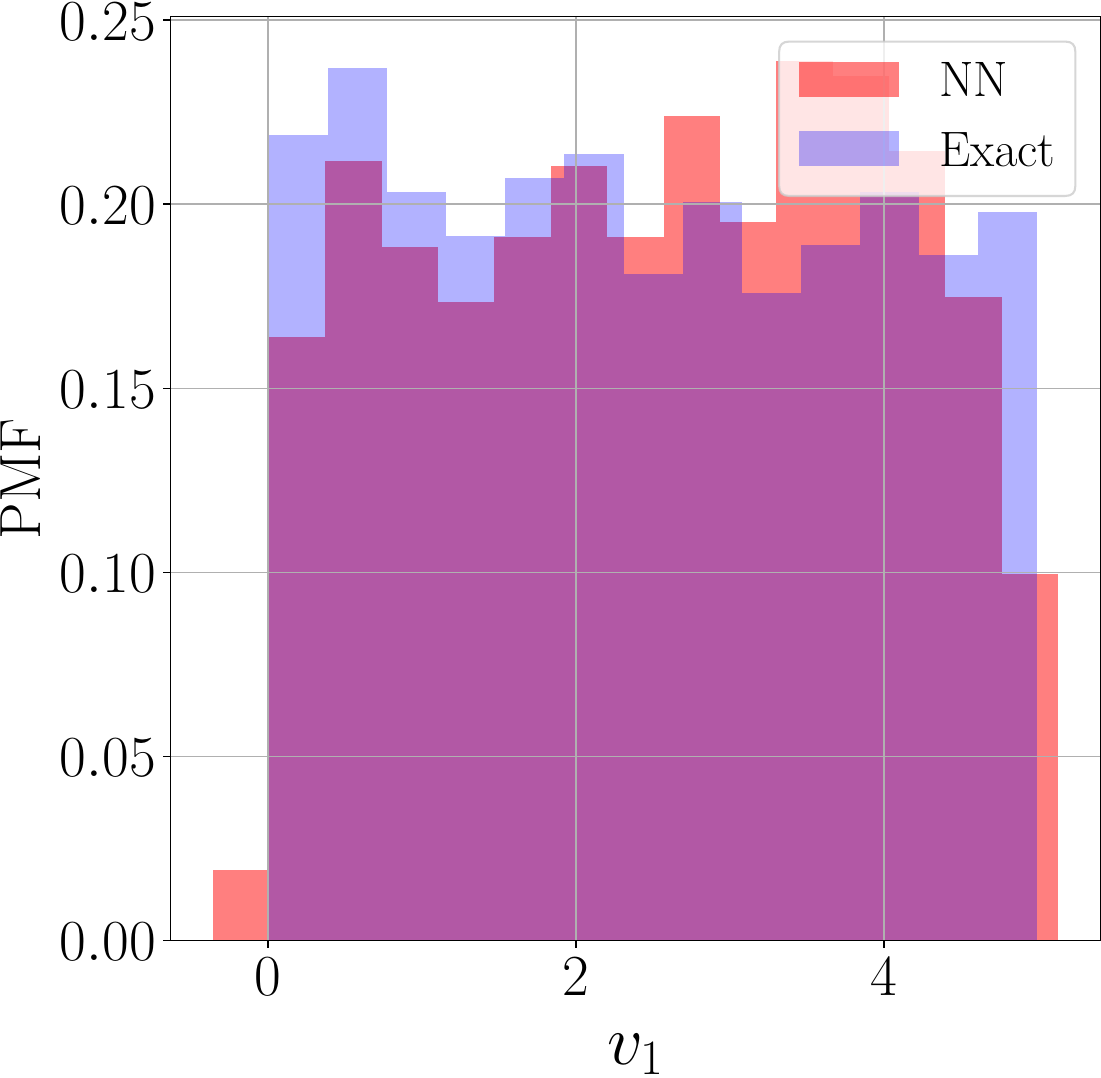}
         \caption{Histogram of $v_1$.}
     \end{subfigure}
     \hfill
    \begin{subfigure}[b]{0.22\textwidth}
         \centering
         \includegraphics[scale=0.165]{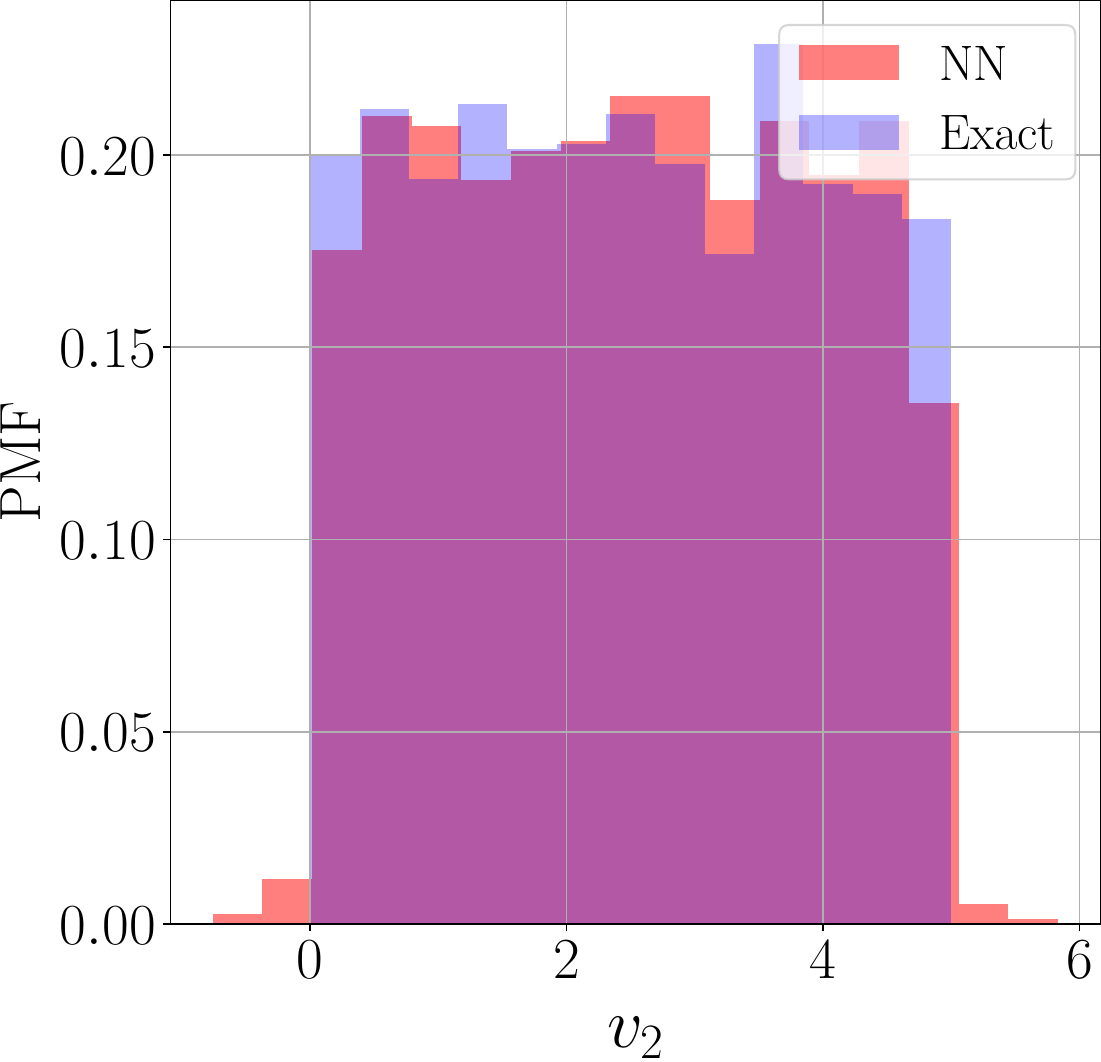}
         \caption{Histogram of $v_2$.}
     \end{subfigure}
     \hfill
    \begin{subfigure}[b]{0.22\textwidth}
         \centering
         \includegraphics[scale=0.165]{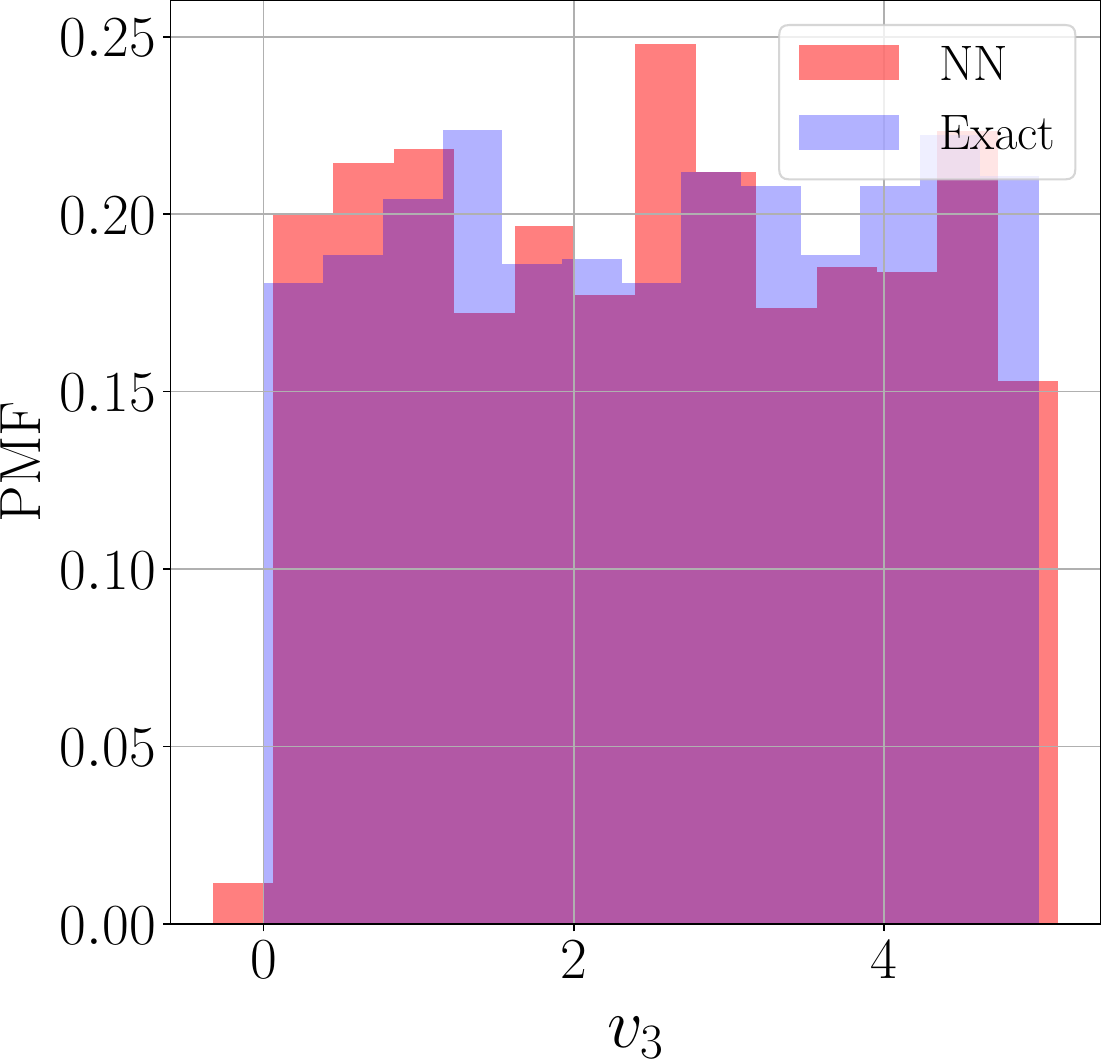}
         \caption{Histogram of $v_3$.}
     \end{subfigure}
          \hfill
    \begin{subfigure}[b]{0.28\textwidth}
         \centering
         \includegraphics[scale=0.18]{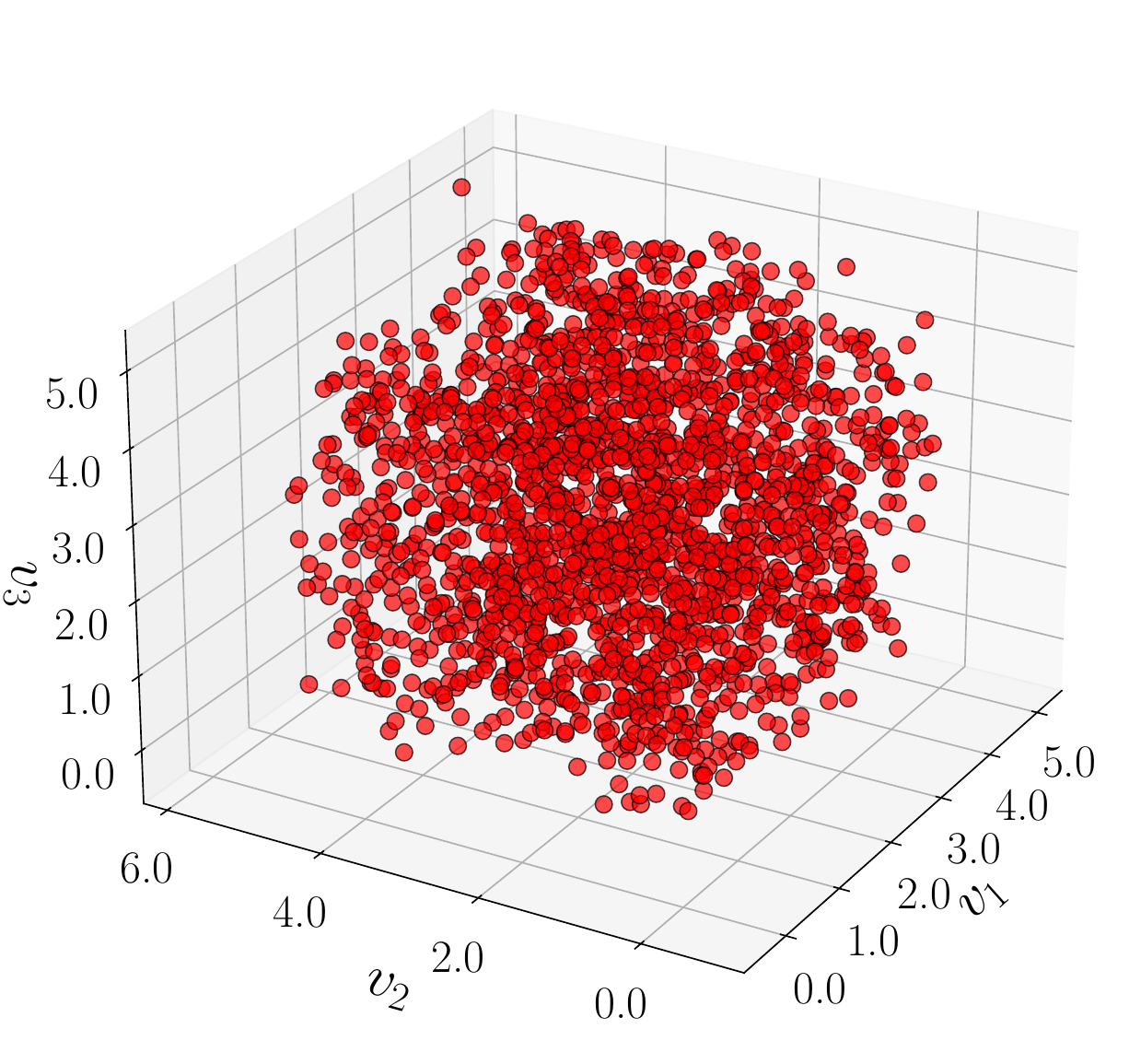}
         \caption{Predicted samples of $\widehat{\boldsymbol{v}}$.}
     \end{subfigure}
    \caption{Model inversion results for the underdetermined linear system, when jointly sampling form $w$ and $\boldsymbol{y}$. The initial uniform distribution of $\boldsymbol{v}$ is correctly recovered.}
    \label{fig: simple linear fix-nothing}
\end{figure}

% Sampling from both y and w
Next we jointly sample from $w$ and $\boldsymbol{y}$. 
This should pose no restrictions to the ability to recover all possible realizations in the input space $\boldsymbol{\mathcal{V}}$, since $\textrm{Ker}(\mathbf{F})$ as well as its orthogonal complement can be both reached. 
To test this, we draw 2000 random samples of $\boldsymbol{y}$ from the trained $\mathscr{N}_f$ and $w$ from $\mathcal{N}(0,1)$. We then use the trained decoder $\mathscr{N}_d$ to determine 2000 values of the corresponding $\widehat{\boldsymbol{v}}$. 
From Figure~\ref{fig: simple linear fix-nothing}, it can be observed that most of the predicted samples reside uniformly in the $[0,5]^3$ cube, despite the existence of a few outliers. 

%=============================================================
\subsection{Nonlinear periodic map}\label{example: non-linear}
%=============================================================
We move to a nonlinear map $\mathcal{F}: \mathbb{R}^2 \to \mathbb{R}$:
\begin{equation}\label{equ: nonlinear system}
y = \sin{k x}.
\end{equation}
Initially, we consider $k \in [1,3]$ and $x\in [-\frac{\pi}{6}, \frac{\pi}{6}]$ to avoid a periodic response. In this case $[k,x]^T = \boldsymbol{v} \in \boldsymbol{\mathcal{V}}$, and the forward map $y = \mathcal{F}(\boldsymbol{v})$ is not identifiable on the subspace $\boldsymbol{\mathcal{M}}_{\bv} \subset \boldsymbol{\mathcal{V}}$, where the product $kx$ is constant.
We again focus only on the inverse analysis task. The inputs $k$ and $x$ are uniformly sampled from their prior ranges for $10^4$ times, and the optimal hyperparameters are reported in the Appendix.

\begin{figure}[ht!]
     \centering
     \begin{subfigure}[b]{0.24\textwidth}
         \centering
         \includegraphics[scale=0.18]{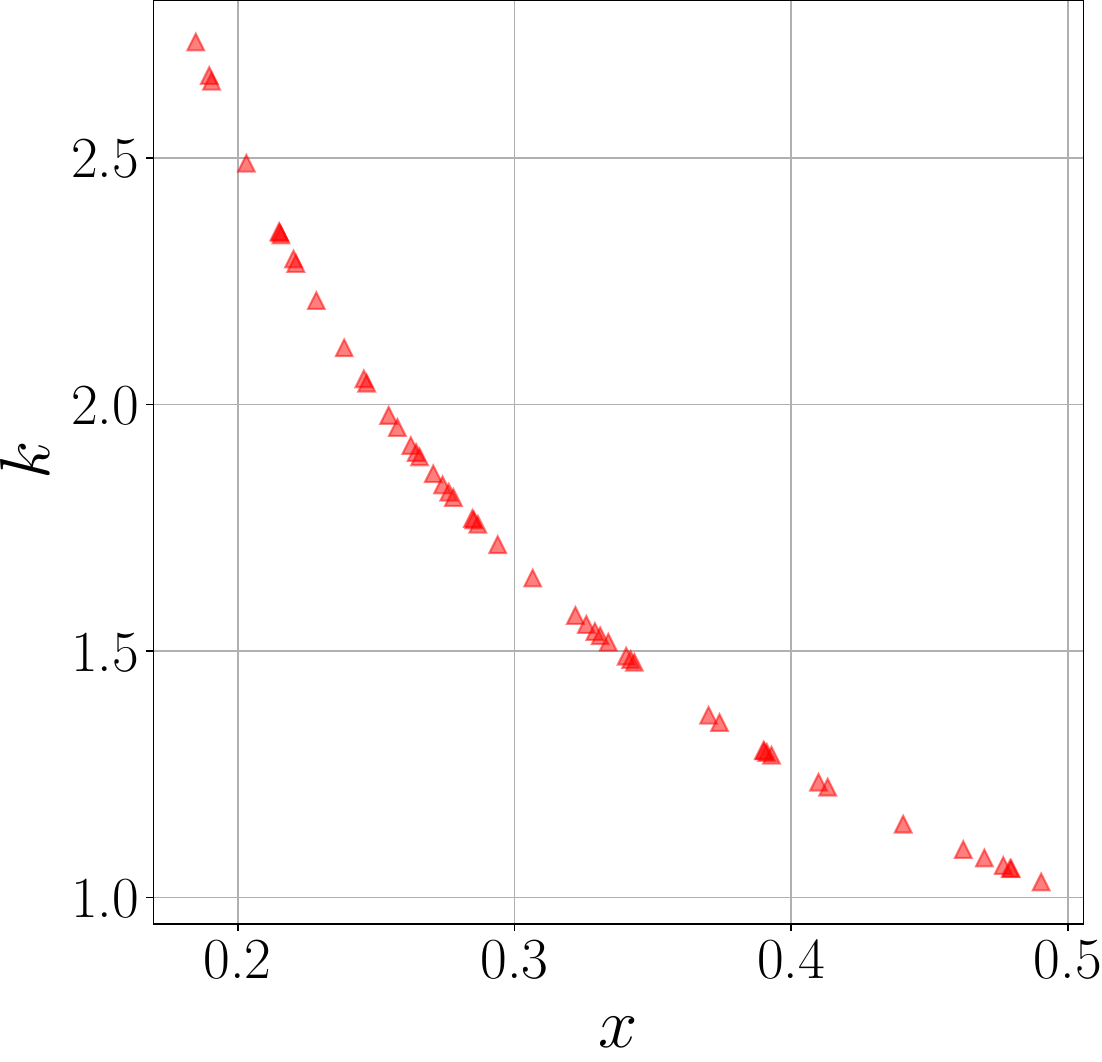}
         \caption{Trajectory of the inversion samples $\widehat{\boldsymbol{v}}$ when fixing a $y^* > 0$.}
        \label{fig: x_traj yp}
     \end{subfigure}
     \hfill
     \begin{subfigure}[b]{0.245\textwidth}
         \centering
         \includegraphics[scale=0.18]{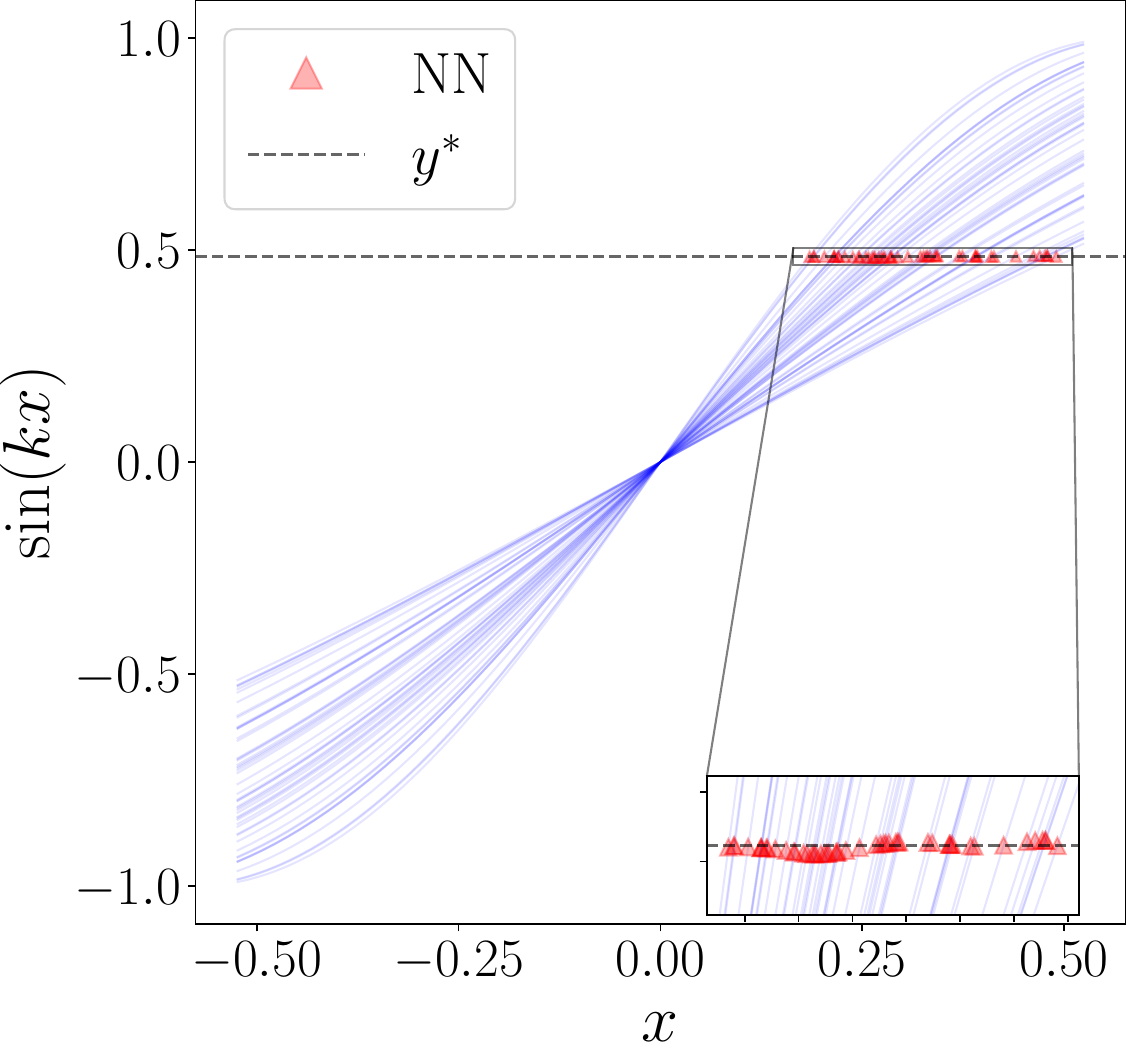}
         \caption{Predicted system output from $\widehat{\boldsymbol{v}}$ when fixing a $y^* > 0$.}
         \label{fig: yp}
     \end{subfigure} 
     \hfill
    \begin{subfigure}[b]{0.24\textwidth}
         \centering
         \includegraphics[scale=0.18]{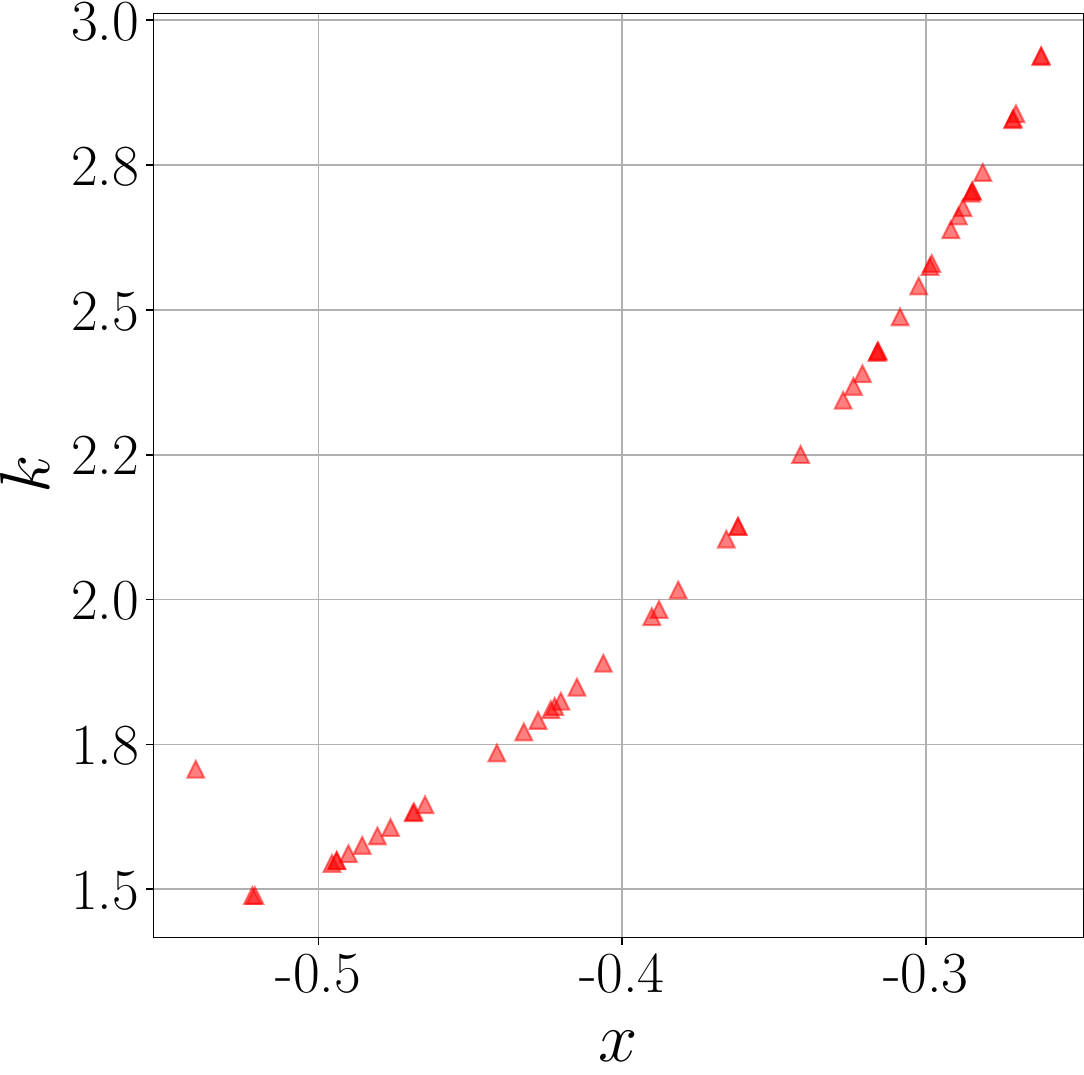}
         \caption{Trajectory of the inversion samples $\widehat{\boldsymbol{v}}$ when fixing a $y^* < 0$.}
         \label{fig: x_traj yn}
     \end{subfigure}
     \hfill
     \begin{subfigure}[b]{0.245\textwidth}
         \centering
         \includegraphics[scale=0.18]{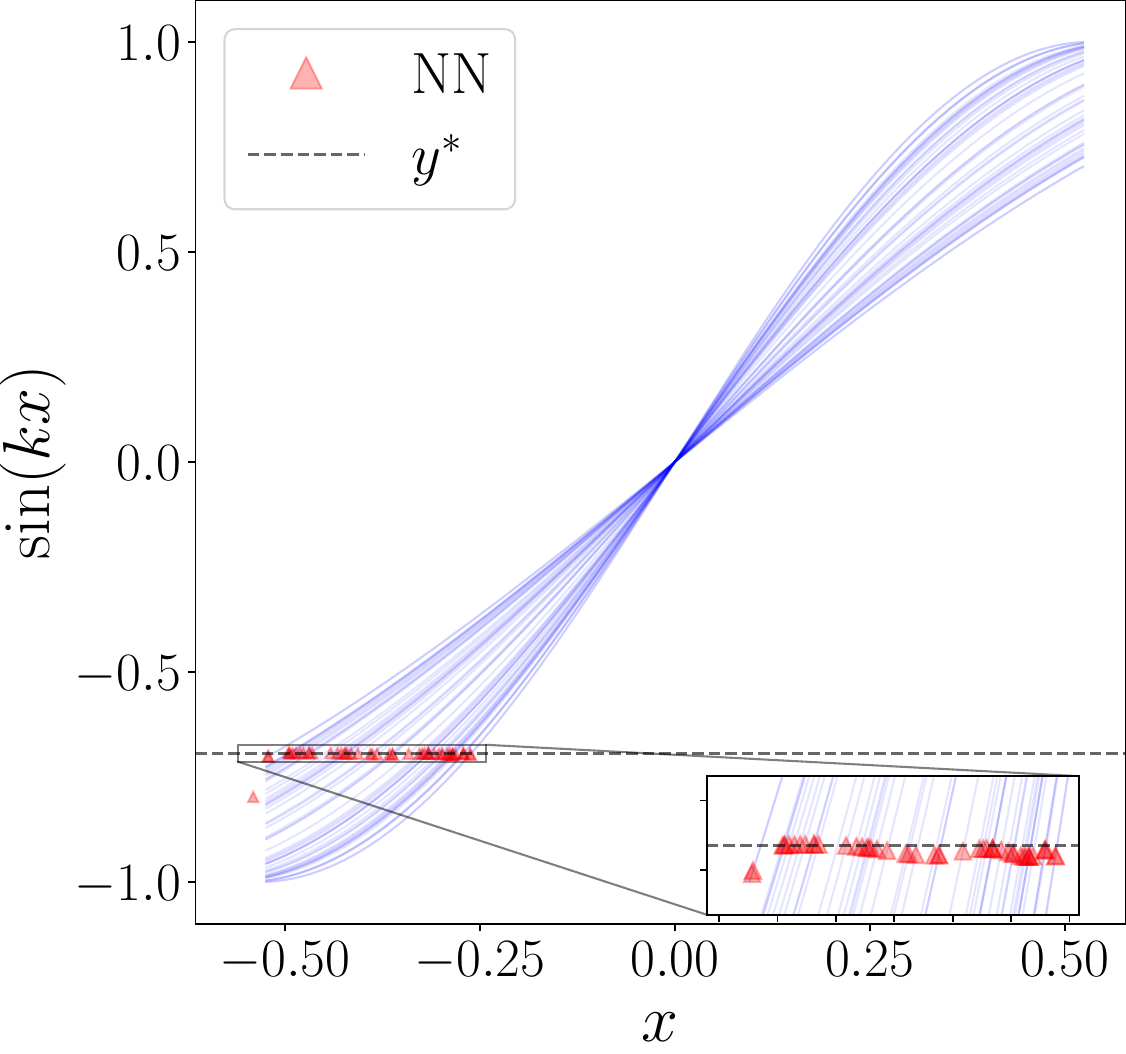}
         \caption{Predicted system output from $\widehat{\boldsymbol{v}}$ when fixing a $y^* < 0$.}
         \label{fig: yn}
     \end{subfigure} 
    \caption{Model inversion results for the sine wave model \textbf{without} periodicity. We fix $y^*$ at both positive and negative values, sample $w$ from $\mathcal{N}(0,1)$, concatenate and decode. The inverse prediction $\widehat{\boldsymbol{v}}$ leads to system output predictions (NN) close to $y^*$, as expected.}
    \label{fig: simple-nonlinear fixy}
\end{figure}

\begin{figure}[ht!]
     \centering
     \begin{subfigure}[b]{0.3\textwidth}
         \centering
         \includegraphics[scale=0.2]{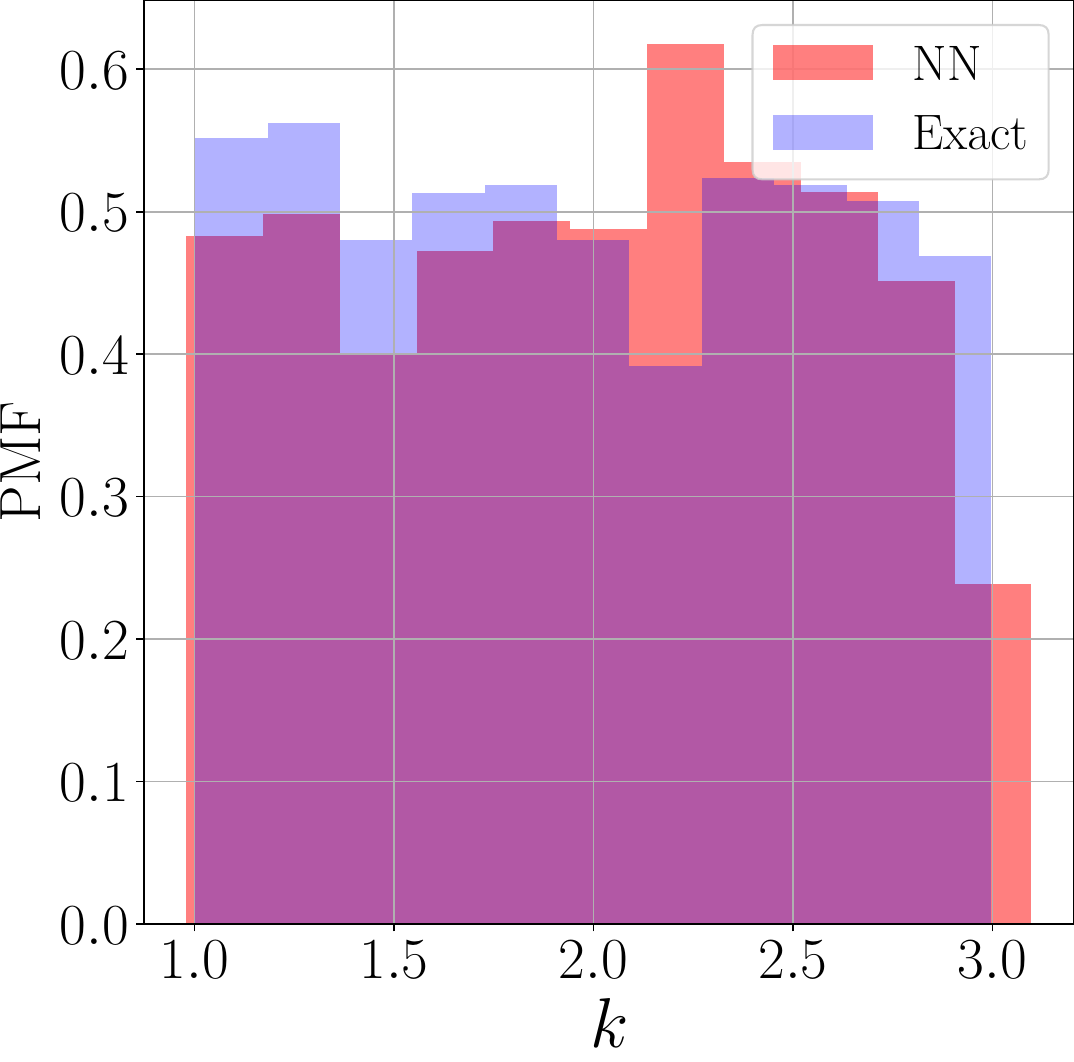}
         \caption{Histogram of $k$.}
     \end{subfigure}
     \hfill
    \begin{subfigure}[b]{0.3\textwidth}
         \centering
         \includegraphics[scale=0.2]{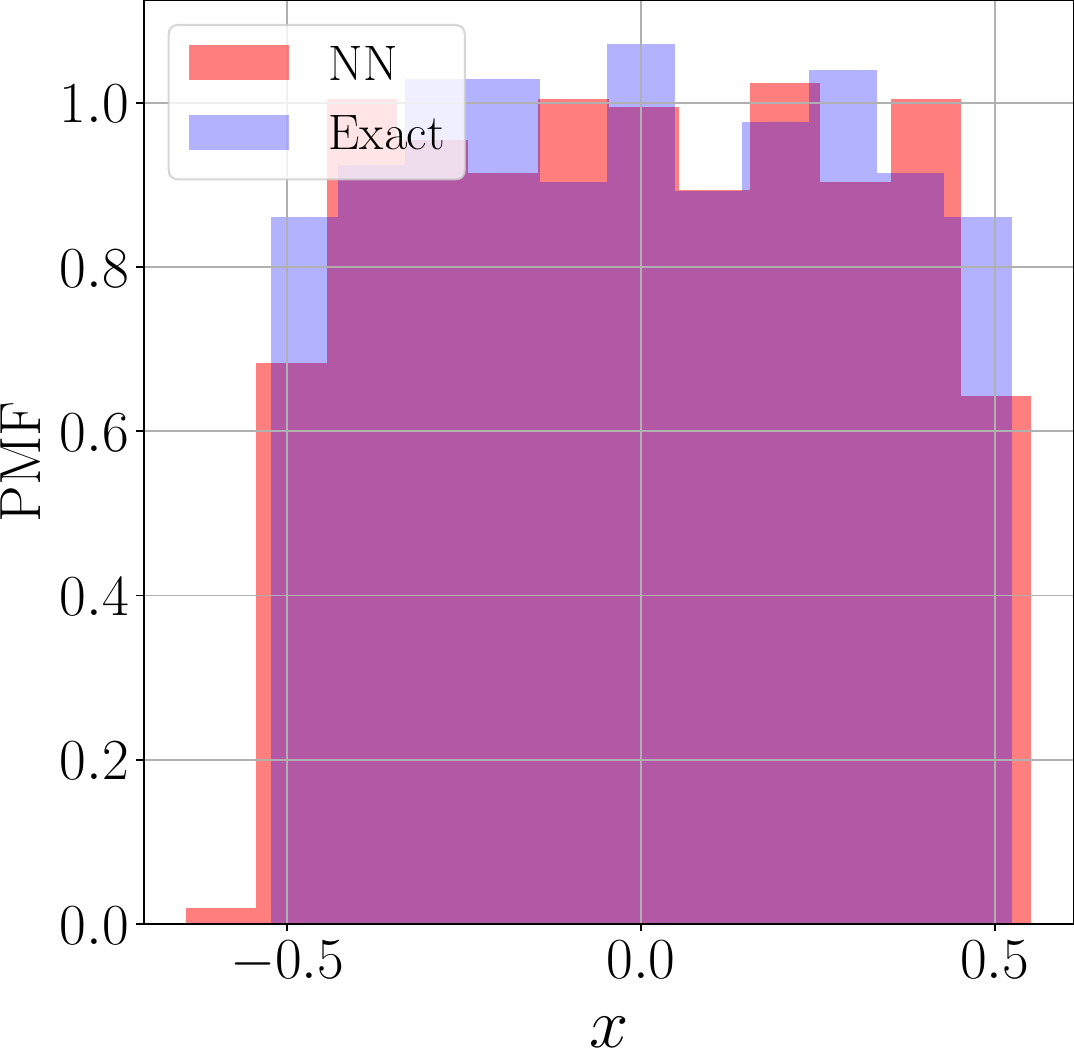}
         \caption{Histogram of $x$.}
     \end{subfigure}
     \hfill
    \begin{subfigure}[b]{0.3\textwidth}
         \centering
         \includegraphics[scale=0.2]{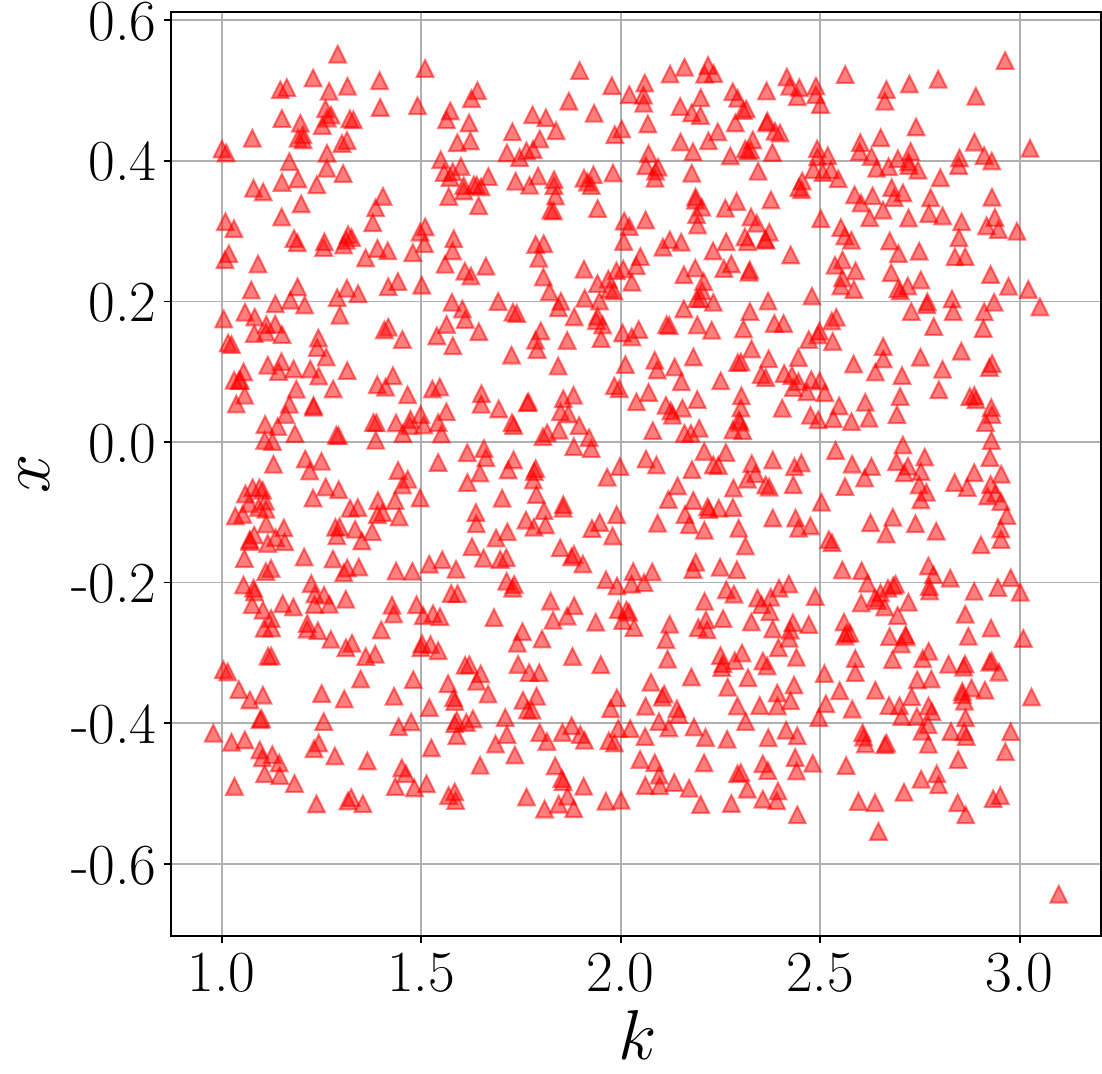}
         \caption{Predicted samples of $\widehat{\boldsymbol{v}}$.}
     \end{subfigure}
    \caption{Model inversion results for the sine wave model \textbf{without} periodicity: sample $w$ and $y$ together and the initial uniform distributions of $k$ and $x$ are recovered.}
    \label{fig: simple nonlinear fix-nothing}
\end{figure}

As discussed for the previous example, we freeze $y=y^*$ at both positive and negative values and, for each fixed $y^*$, we pick 50 samples from the latent variable $w\sim \mathcal{N}(0,1)$, concatenate and decode with $\mathscr{N}_{d}$.
This results in the sample trajectories of $\widehat{\boldsymbol{v}}$ shown in Figure~\ref{fig: x_traj yp} and Figure~\ref{fig: x_traj yn}, respectively. 
These one-dimensional manifolds accurately capture the correct correlation between $k$ and $x$, leading to the same $y=y^*$ as confirmed in Figure~\ref{fig: yp} and Figure~\ref{fig: yn} (the superimposed blue sine curves are based on the predicted $\widehat{k}$ values and a fixed sequence of $x$ values. The red triangle marks the predicted $\widehat{x}$ value that should lead to $\sin \widehat{k}\widehat{x} = \widehat{y} \approx y^* $).
We then sample together from $w$ and $y$, recovering the initial distributions utilized during training data preparation, for $k$ and $x$, as shown in Figure~\ref{fig: simple nonlinear fix-nothing}.

Next, we expand $x$ from $[-\pi/6, \pi/6]$ to $[-\pi, \pi]$ and keep $k \in [1,3]$. Consequently, the latent manifold is no longer $kx = \text{const}$ in this scenario due to periodicity. 
To prepare for training, we still generate $10^4$ uniform samples from the given ranges while we figure out a more complicated network structure is required for achieving robust performance (please refer to Table~\ref{table:nonlinear hyper-periodicity} for recommended hyperparameters), compared to the above monotonic case.
%

% By construction, the latent space $\boldsymbol{\mathcal{W}}$ built by the VAE network should be one-dimensional since $\dim{(\boldsymbol{v})}=2$ and $y\in \mathbb{R}$. However, for the current periodic system, we find the inverse prediction can be significantly improved if $\dim{(\boldsymbol{\mathcal{W}}})$ is increased beyond 1. 
% %
% This is not surprising considering any smooth (\textbf{\color{blue} our learned manifold is actually discrete, how do we explain it..}) lower-dimensional manifold can be embedded into a higher one, as a consequence of the Whitney embedding theorem~\cite{whitney1936differentiable}. 
% %
% Besides, it is intuitively simpler to approximate a lower dimensional structure in a higher dimensional space. Think, for example, about a curve that does not self-intersect if represented in $\mathbb{R}^3$, but it does when projected on $\mathbb{R}^{2}$. Note that better performance of utilizing high-dimensional latent space for inverse modeling was also shown in~\cite{almaeen2021variational}.
% %

\begin{figure}[ht!]
     \centering
     \begin{subfigure}[b]{0.26\textwidth}
         \centering
         \includegraphics[scale=0.2]{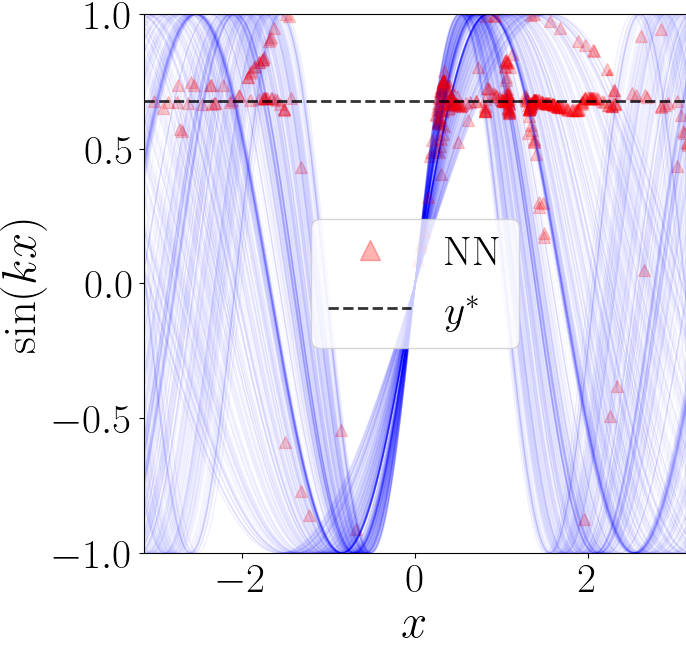}\caption{$\dim{(\boldsymbol{\mathcal{W}})} = 1$.}
     \end{subfigure}
     \hfill
    \begin{subfigure}[b]{0.24\textwidth}
         \centering
         \includegraphics[scale=0.2]{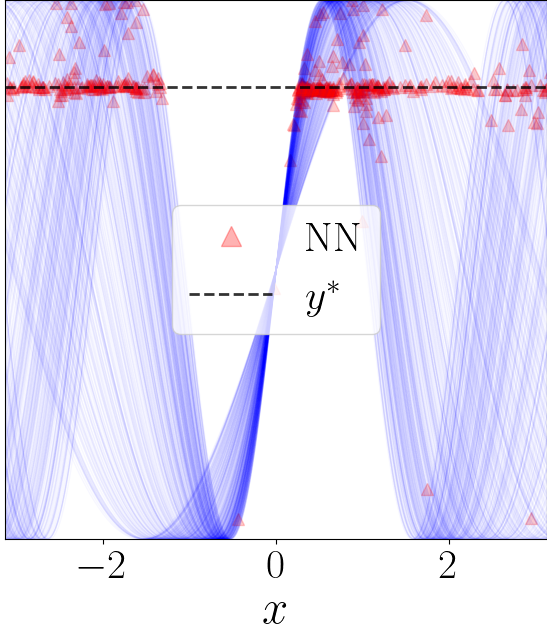}\caption{$\dim{(\boldsymbol{\mathcal{W}})} = 2$.}
     \end{subfigure}
     \hfill
    \begin{subfigure}[b]{0.24\textwidth}
         \centering
         \includegraphics[scale=0.2]{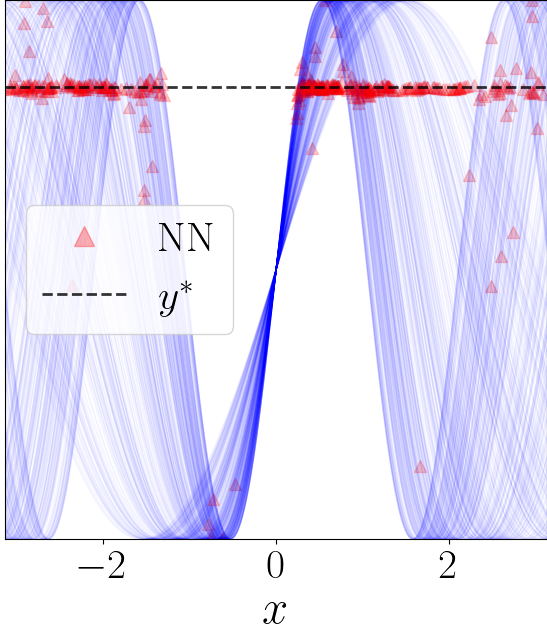}\caption{$\dim{(\boldsymbol{\mathcal{W}})} = 4$.}
     \end{subfigure}
     \hfill
    \begin{subfigure}[b]{0.24\textwidth}
         \centering
         \includegraphics[scale=0.2]{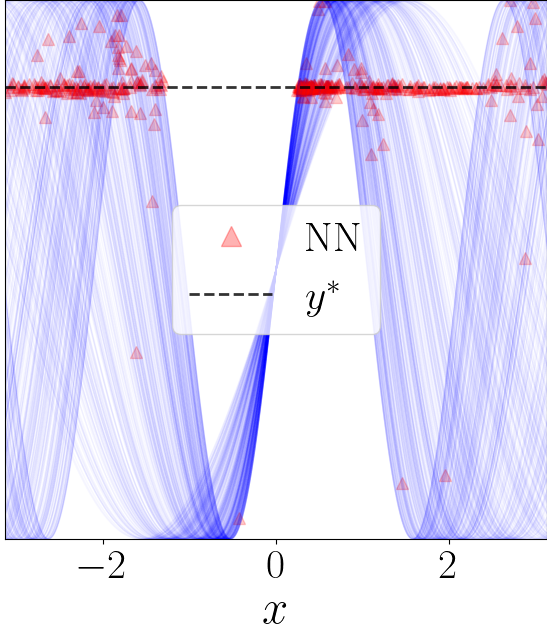}\caption{$\dim{(\boldsymbol{\mathcal{W}})} = 8$.}
         \label{fig: sine-8d N01}
     \end{subfigure}
     \caption{Model inversion results for the sine wave model \textbf{with} periodicity. We fix $y^* = 0.676$ and draw 400 latent variable samples from the standard Gaussian spaces of dimension $1,2,4,8.$ Each blue curve is based on a predicted $\widehat{k}$ value and a fixed sequence of $x$. The red triangle marks the predicted $\widehat{x}$ value that should lead to $y^* \approx \sin \widehat{k}\widehat{x}$. }
    \label{fig: peroidc-sine-dimW}
\end{figure}

By construction, the latent space $\boldsymbol{\mathcal{W}}$ built by the VAE network should be one-dimensional since $\dim{(\boldsymbol{\mathcal{V}})}=2$ and $y\in \mathbb{R}$. However, for the current periodic system, we find the inverse prediction can be significantly improved if $\dim{(\boldsymbol{\mathcal{W}}})$ is increased beyond 1. We first compare inverse prediction results under $\dim{(\boldsymbol{\mathcal{W}})} = 1,2,4,8$ with respect to the same fixed $y^* = 0.676$. 
In Figure~\ref{fig: peroidc-sine-dimW}, we find the original 1D latent space produces the poorest result, while if $\dim{(\boldsymbol{\mathcal{W}})}$ increases beyond 1, the performance remains relatively consistent.

Next, we demonstrate how the three sampling methods discussed in Section~\ref{sec:analysis_sampling} enhance our inverse predictions by removing the spurious outliers shown in Figure~\ref{fig: peroidc-sine-dimW}. We focus only on the case with $\dim{(\boldsymbol{\mathcal{W}})} = 8$ and keep the sampling size as 400. 
We hypothesize that a more robust sampling method requires fewer samples to unveil the underlying structure of the non-identifiable manifold $\boldsymbol{\mathcal{M}}_{\bv}$, though a larger sample size always leads to better visualizations.
\begin{figure}[ht!]
     \centering
     \begin{subfigure}[b]{0.245\textwidth}
         \centering
         \includegraphics[scale=0.195]{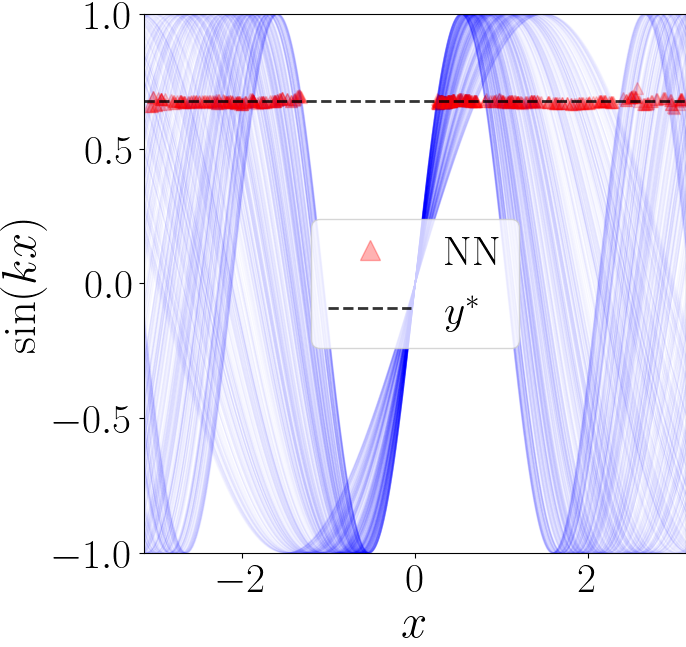}\caption{Predicted system output: $R=2$.}
     \end{subfigure}
     \hfill
     \begin{subfigure}[b]{0.245\textwidth}
         \centering
         \includegraphics[scale=0.175]{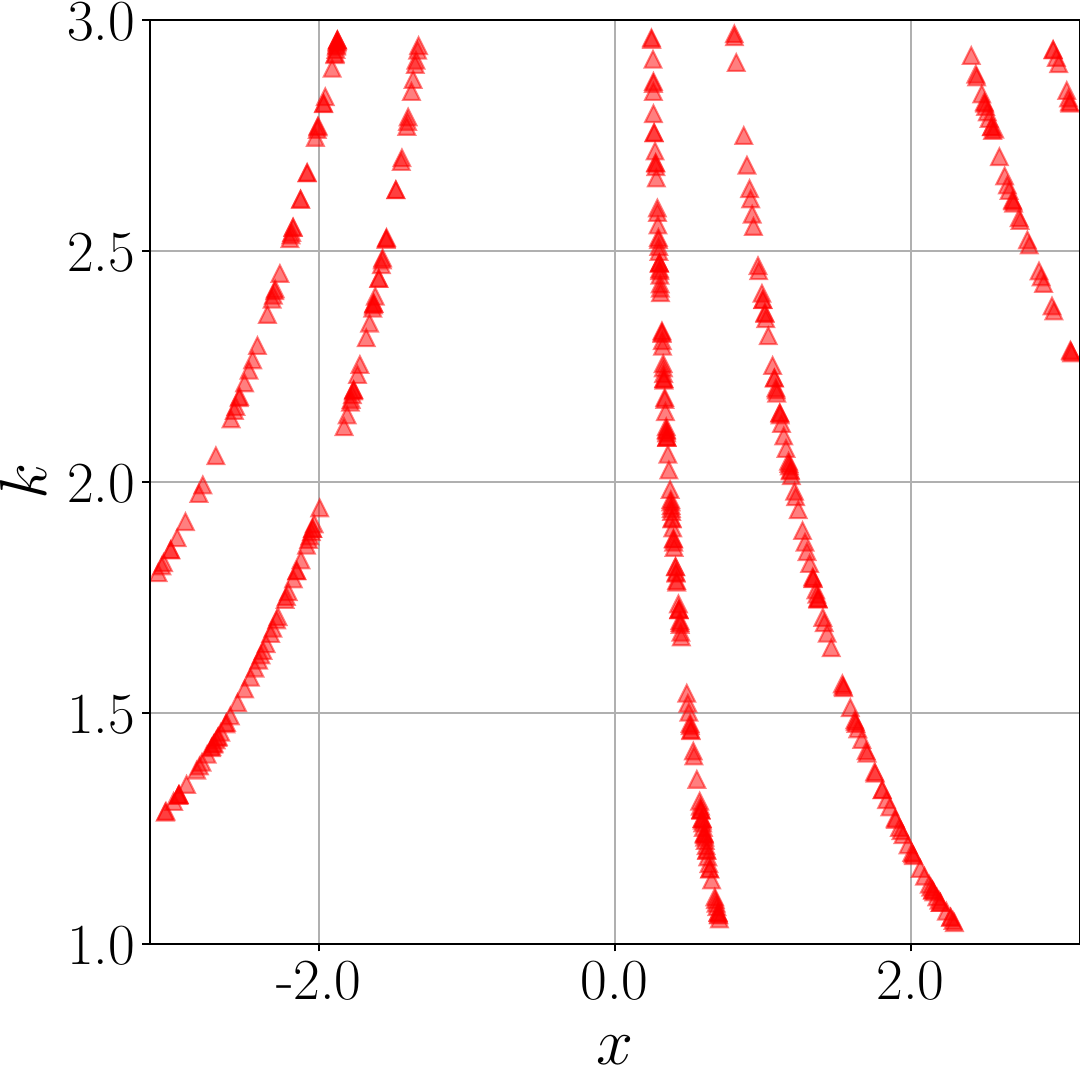}\caption{Inversion sample trajectory: $R=2$.}
         \label{fig: r2-traj}
     \end{subfigure}
     \hfill
      \begin{subfigure}[b]{0.245\textwidth}
         \centering
         \includegraphics[scale=0.195]{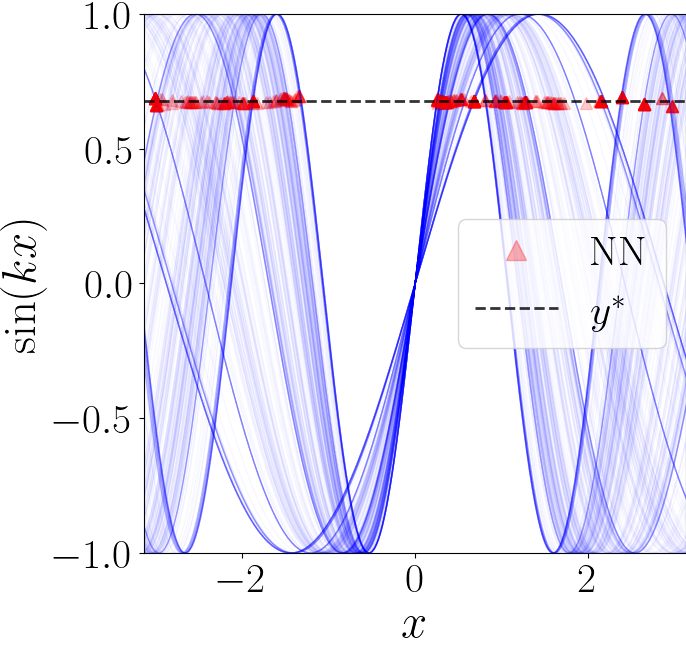}\caption{Predicted system output: $R=50$.}
     \end{subfigure}
     \hfill
     \begin{subfigure}[b]{0.245\textwidth}
         \centering
         \includegraphics[scale=0.175]{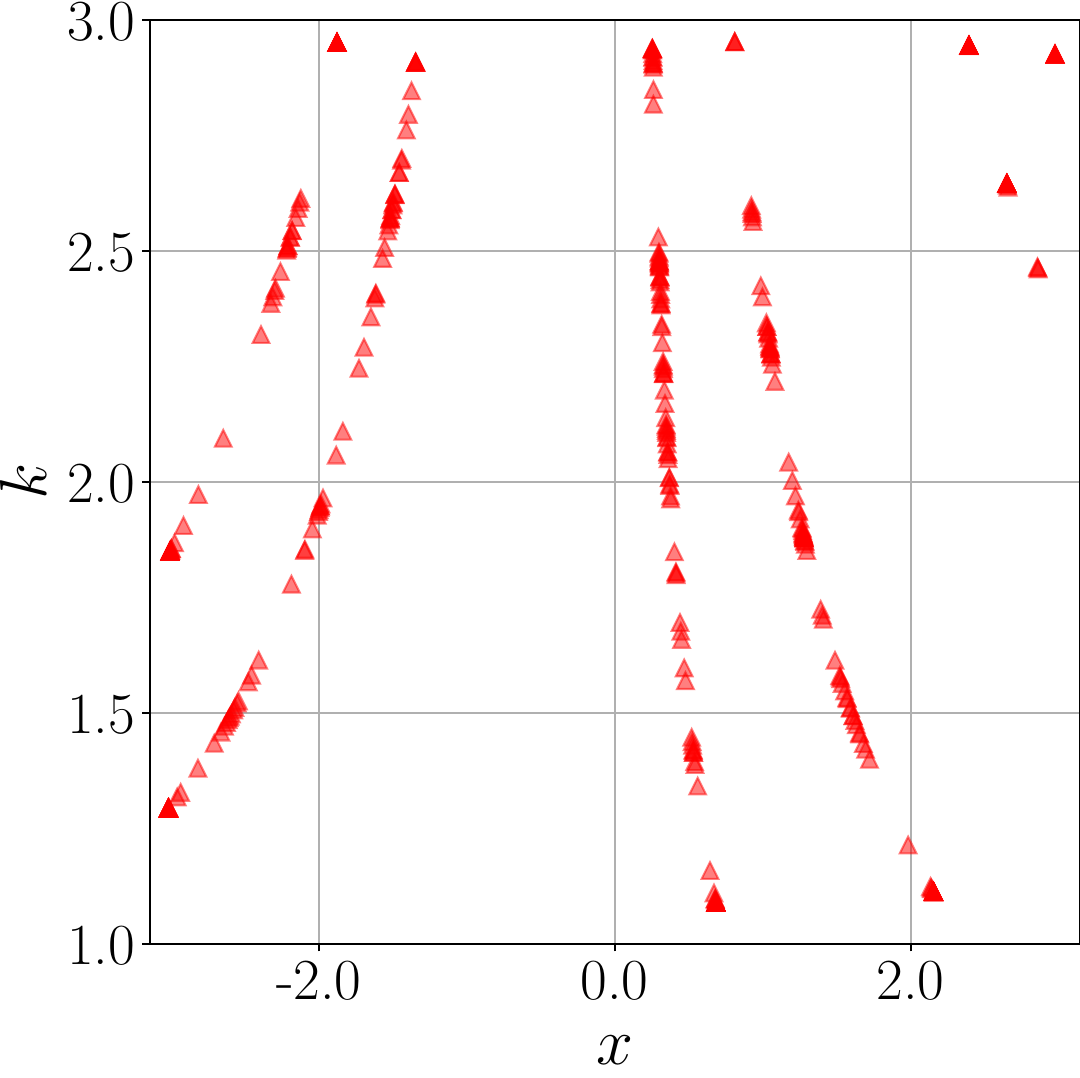}\caption{Inversion sample trajectory: $R=50$.}
         \label{fig: r50-traj}
     \end{subfigure}
     \caption{Model inversion results for the sine wave model \textbf{with} periodicity and using PC sampling. The model output $y^* = 0.676$ is fixed and $\dim{(\boldsymbol{\mathcal{W}}) = 8}$. The results are shown for iteration number $R=2$ and $R=50$. Sample size: 400.}
     \label{fig: pc-sampling sine wave}
\end{figure}

First, compared to Figure~\ref{fig: sine-8d N01}, the number of outliers is significantly smaller with PC sampling applied. 
Furthermore, as the number of iteration $R$ increases, details of the learned manifold gradually fade away, as the decoded samples are drawn towards the means of the posterior components with smaller support, as shown in \Cref{fig: r2-traj,fig: r50-traj}. This confirms our analysis in Section~\ref{sec: pc sampling}.

\begin{figure}[ht!]
     \centering
     \begin{subfigure}[b]{0.245\textwidth}
         \centering
         \includegraphics[scale=0.195]{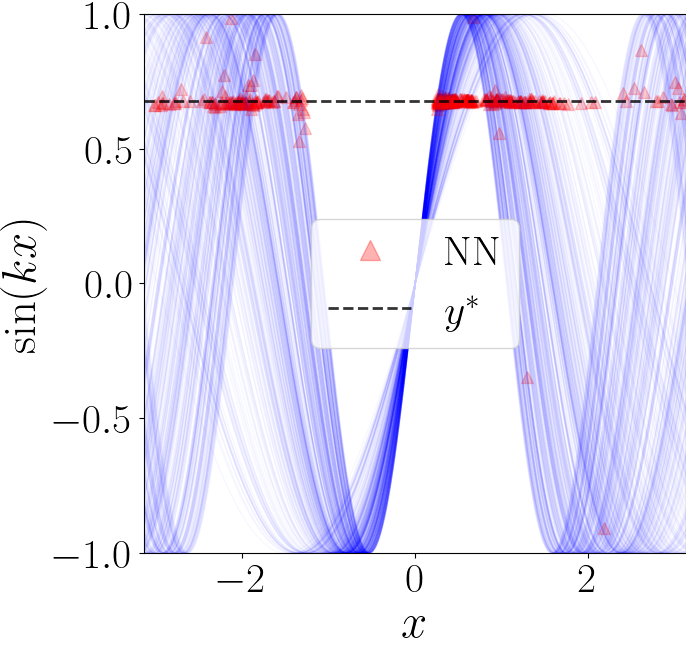}\caption{Predicted system output: $S=500, Q=3$. Sample size: 400.}
     \end{subfigure}
     \hfill
      \begin{subfigure}[b]{0.245\textwidth}
         \centering
         \includegraphics[scale=0.175]{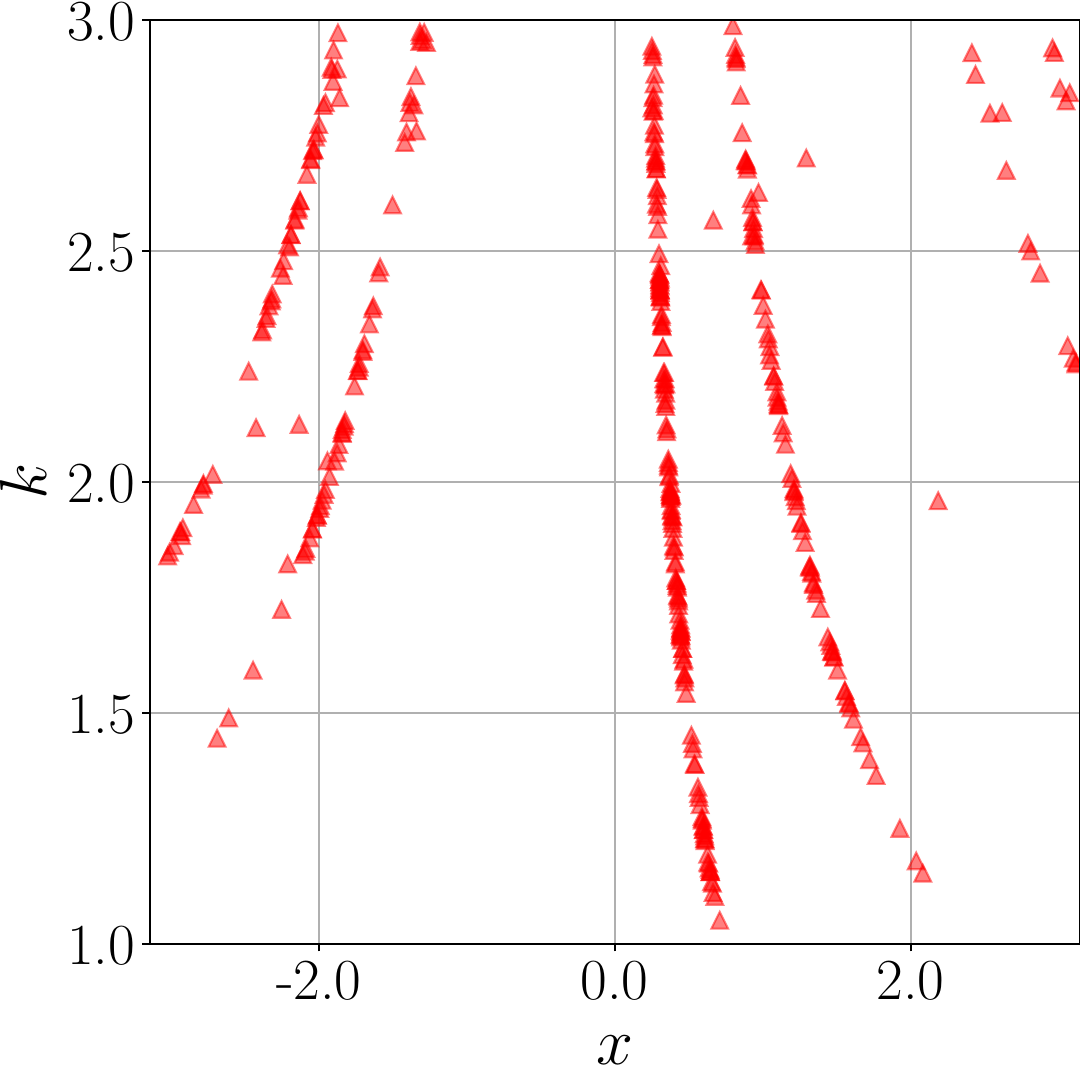}\caption{Inversion sample trajectory: $S=500, Q=3$. Sample size: 400.}
         \label{fig: HD-500-3}
     \end{subfigure}
     \hfill
     \begin{subfigure}[b]{0.245\textwidth}
         \centering
         \includegraphics[scale=0.175]{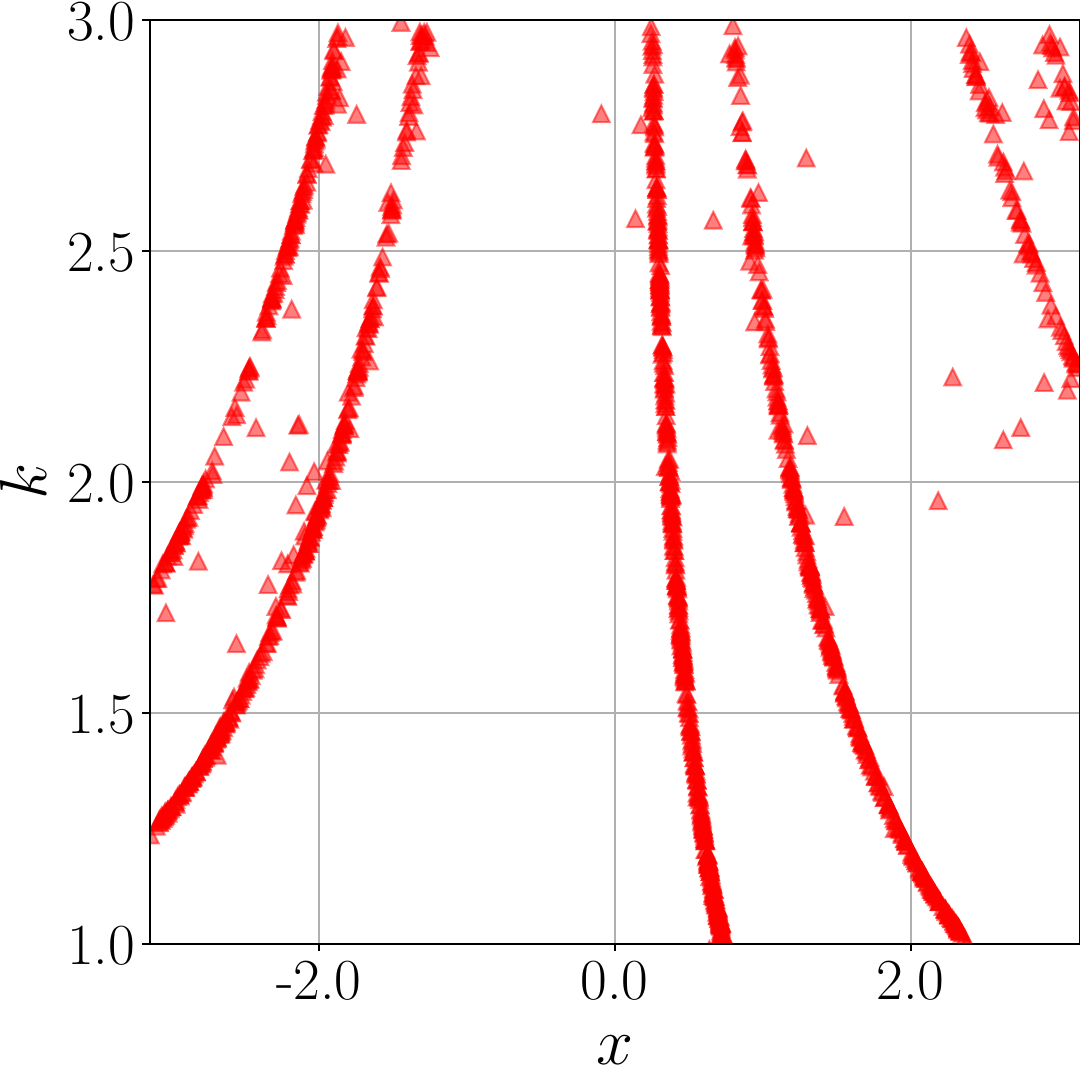}\caption{Inversion sample trajectory: $S=500, Q=3$. Sample size: 1500.}
         \label{fig: noise-1500}
     \end{subfigure}
     \hfill
      \begin{subfigure}[b]{0.245\textwidth}
         \centering
         \includegraphics[scale=0.175]{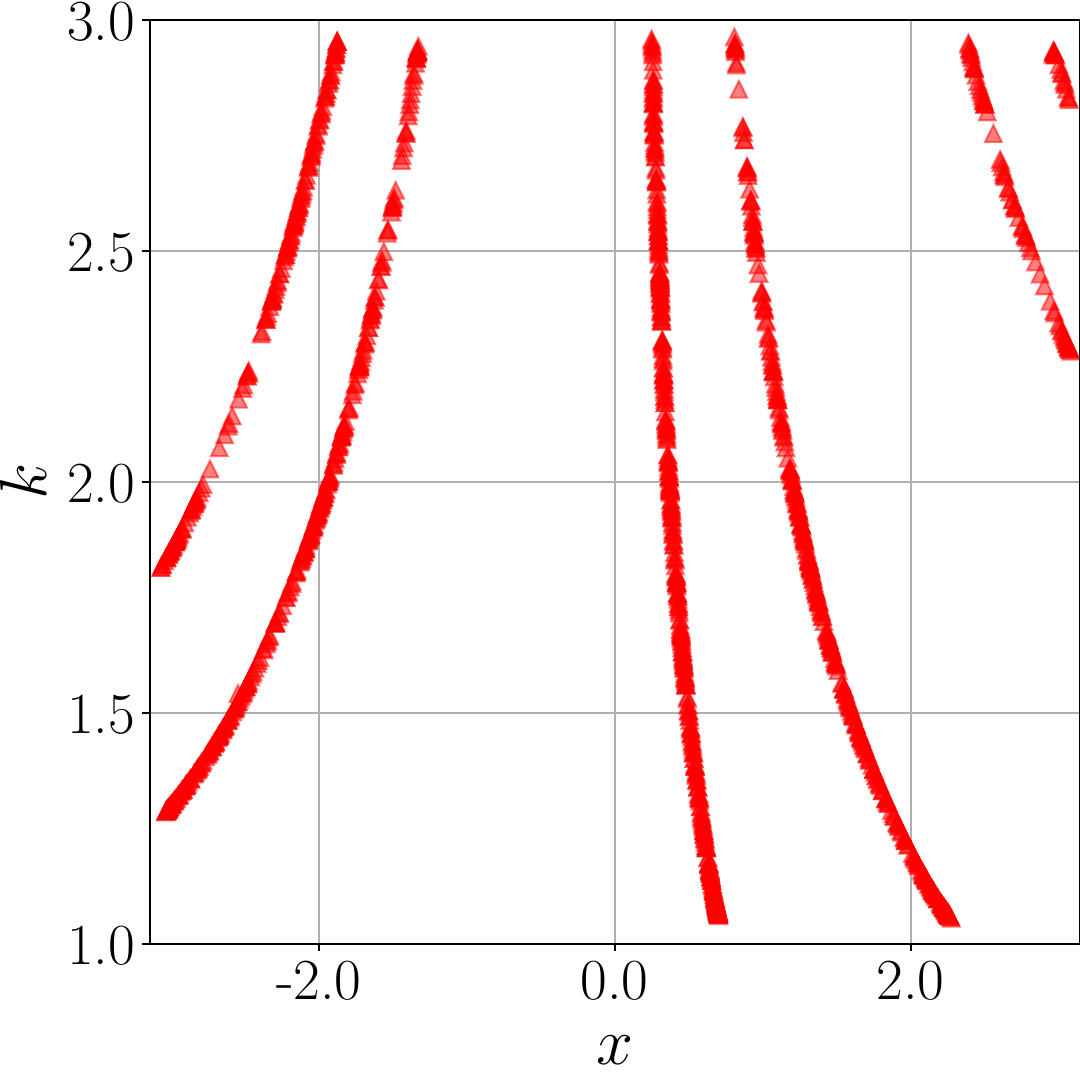}\caption{Apply PC-sampling to denoise the results shown in Figure~\ref{fig: noise-1500}, $R=3$.}
         \label{fig: noise-free-1500}
     \end{subfigure}
     \caption{Model inversion results for the sine wave model \textbf{with} periodicity, using HD sampling. The model output $y^* = 0.676$ is fixed and $\dim{(\boldsymbol{\mathcal{W}}) = 8}$. The results are shown for subset size $S = 500$ and sub-sampling size $Q=3$.}
     \label{fig: HP sampling-sine}
\end{figure}

Moving to HD sampling, use of $S=500, Q=3$ can approximate the global structure of the non-identifiable manifold with 400 samples, even through several outliers are still visible and some details of the learned trajectories are missing (e.g., compare Figure~\ref{fig: HD-500-3} with Figure~\ref{fig: r2-traj}).
The missing details attribute to the density ranking mechanism of the HD sampling scheme, consistent with our discussion in Section~\ref{sec: sampling rank pdf}. 
In this application, the performance of HD sampling can be improved just by utilizing more samples, as illustrated in Figure~\ref{fig: noise-1500}.
Finally, we show that the two sampling schemes (PC and HD) can be combined in Figure~\ref{fig: noise-free-1500}.

\begin{figure}[ht!]
     \centering
     \begin{subfigure}[b]{0.245\textwidth}
         \centering
         \includegraphics[scale=0.195]{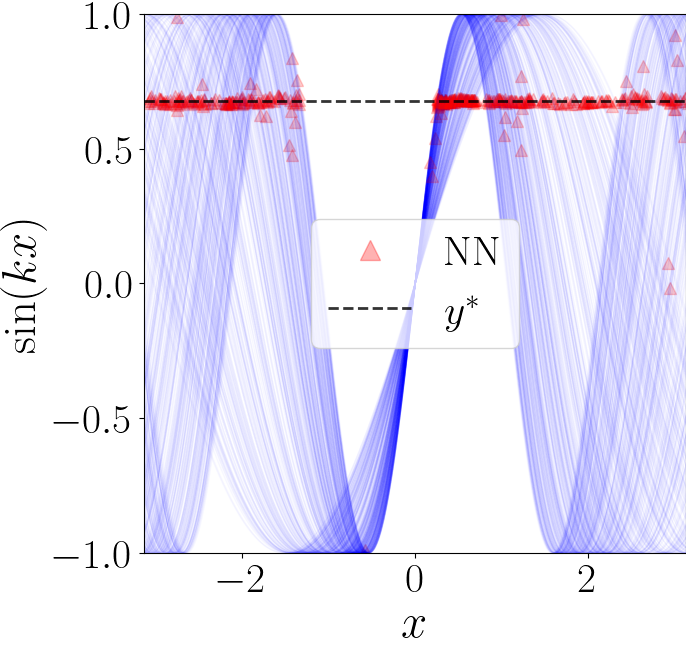}\caption{Predicted system output.}
         \label{fig: sine-nf-check}
     \end{subfigure}
     \hfill
      \begin{subfigure}[b]{0.245\textwidth}
         \centering
         \includegraphics[scale=0.175]{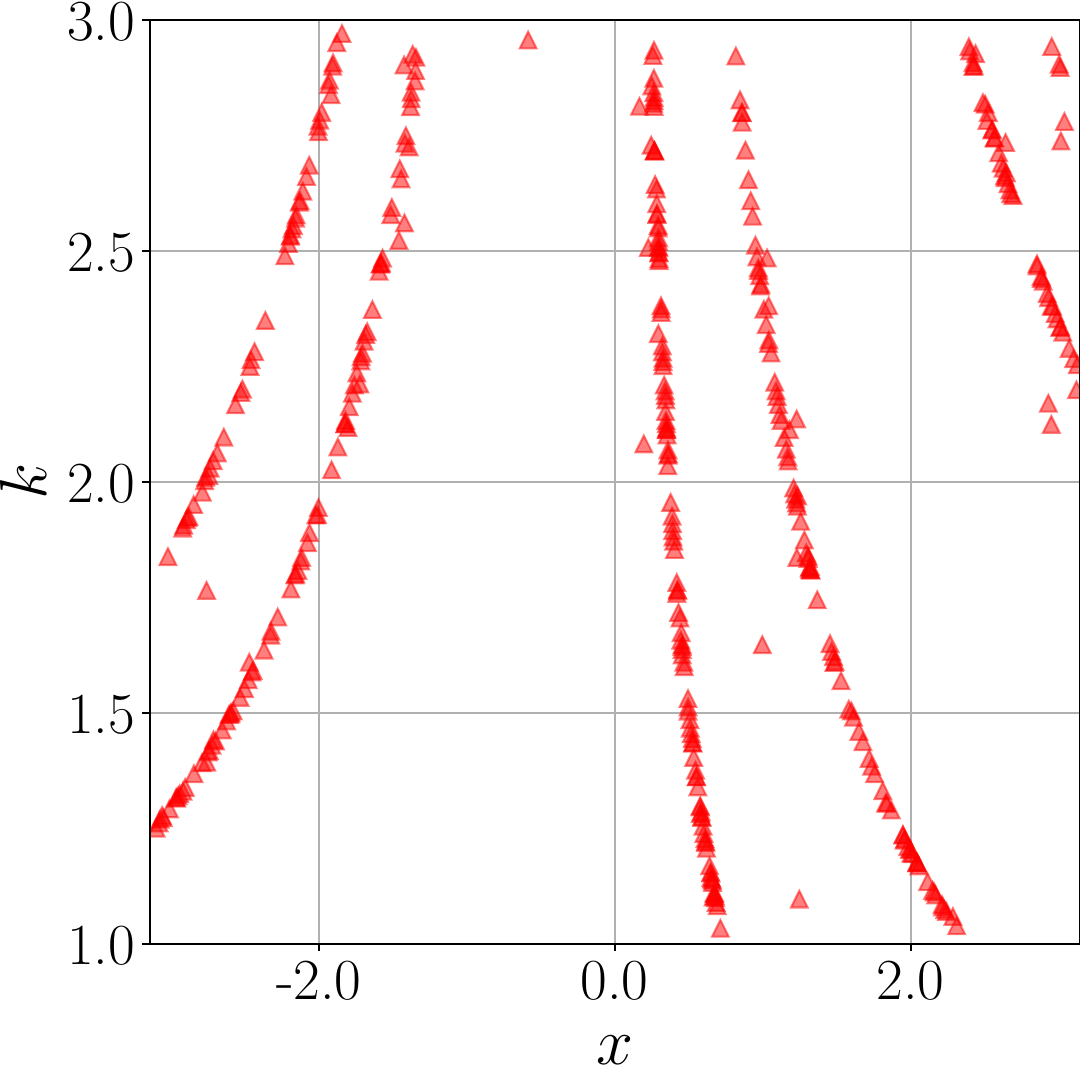}\caption{Inversion sample trajectory.}
     \label{fig: sine-nf-traj}
     \end{subfigure}
     \hfill
     \begin{subfigure}[b]{0.245\textwidth}
         \centering
         \includegraphics[scale=0.18]{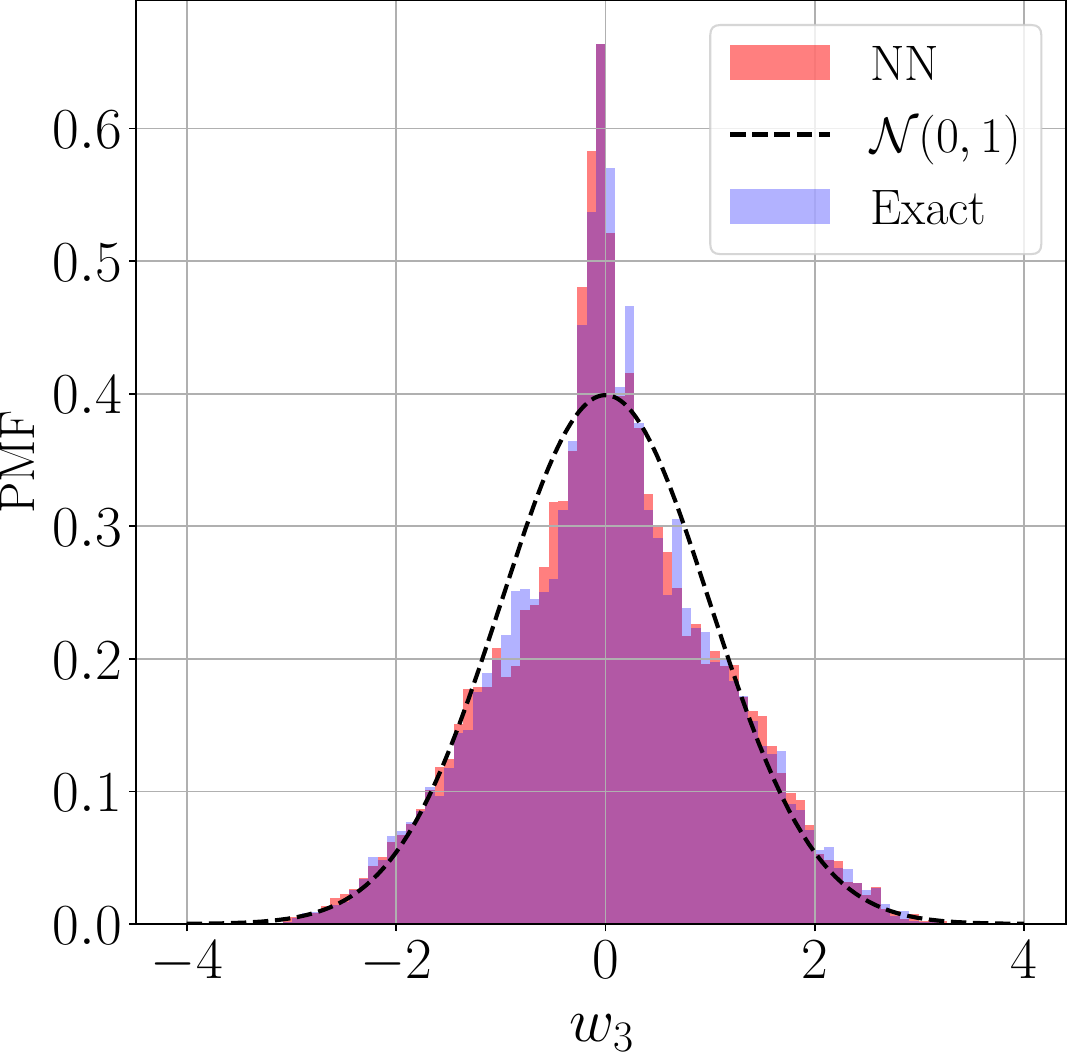}\caption{Learned distribution of $w_3$ by $\mathscr{N}_{f, \boldsymbol{w}}$.}
         \label{fig: sine-w3}
     \end{subfigure}
     \hfill
     \begin{subfigure}[b]{0.245\textwidth}
         \centering
         \includegraphics[scale=0.18]{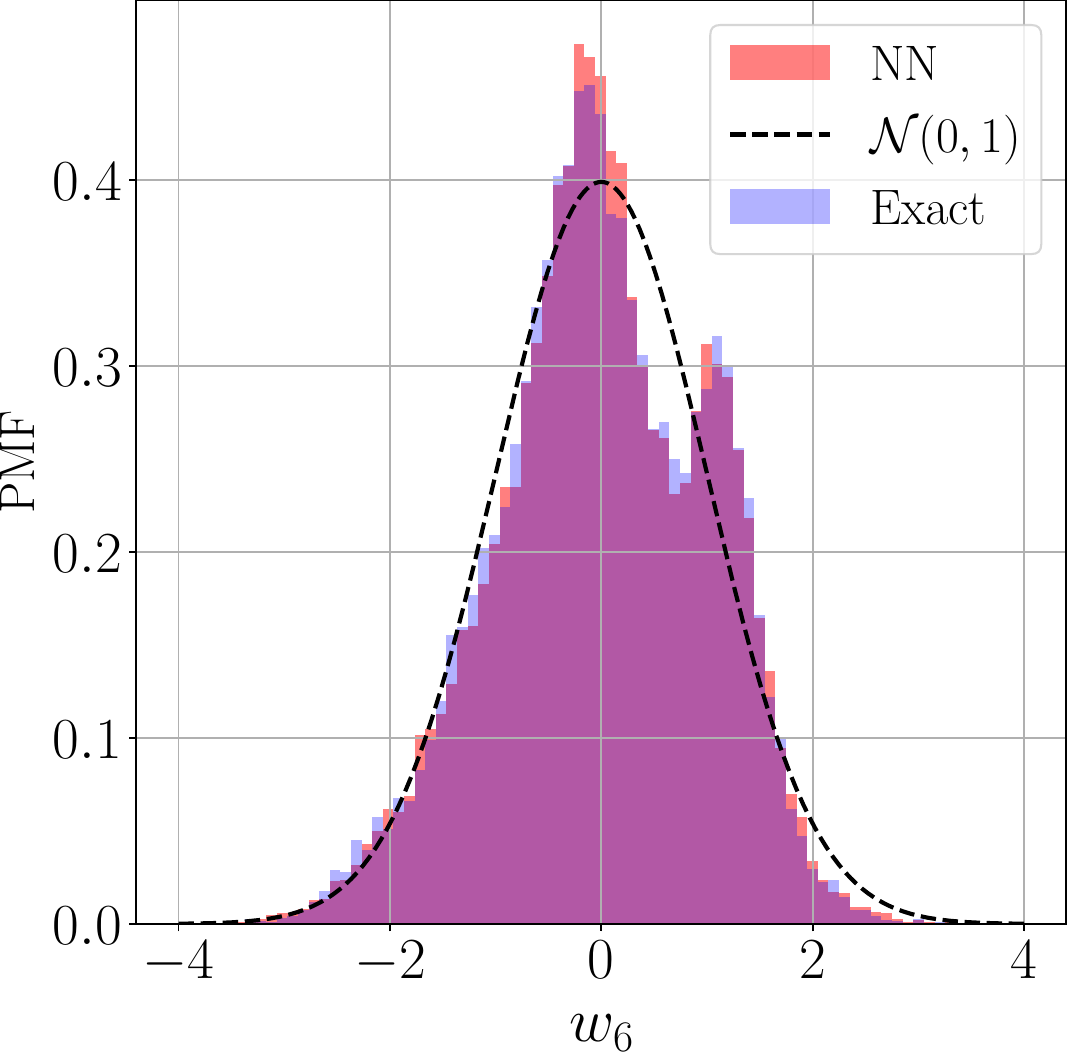}\caption{Learned distribution of $w_6$ by $\mathscr{N}_{f, \boldsymbol{w}}$.}
         \label{fig: sine-w6}
    \end{subfigure}
    \caption{Model inversion results for the sine wave model \textbf{with} periodicity, using NF sampling. The model output $y^* = 0.676$ is fixed and $\dim{(\boldsymbol{\mathcal{W}}) = 8}$. Sample size: 400. Quantities in the histogram: Exact: data distribution of the latent variable samples generated by the trained VAE encoder $\mathscr{N}_v$. NN: Learned distribution by the NF sampler $\mathscr{N}_{f, \boldsymbol{w}}$.}
     \label{fig: NF sampling-sine}
\end{figure}

We then consider NF sampling. To do so, we first train $\mathscr{N}_{f, \boldsymbol{w}}$ based on the latent variable samples generated by the trained VAE encoder $\mathscr{N}_v$ (Please refer to Table~\ref{table: nonlinear-nf} for hyperparameter choices). 
The transformation associated with this flow should be simple if not trivial when the posterior $q(\boldsymbol{w}|\boldsymbol{v})$ is close (in terms of an appropriate statistical distance or divergence) to a standard normal distribution.
However, this is not the case as shown in \Cref{fig: sine-w3,fig: sine-w6}. The learned latent variable $\boldsymbol{w}$ displays spurious structures in two of its components, which we suspect to be associated with low-density posterior regions that would be incorrectly sampled according to a standard normal (see Section~\ref{sec: nf sampling}).

In contrast to Figure~\ref{fig: sine-8d N01}, we notice a reduction of the outliers in Figure~\ref{fig: sine-nf-check} when using NF sampling. 
The remaining outliers, which may come from insufficient network training, or just due to the fact we are transforming a continuous latent space into disconnected branches, can be effectively removed if we further apply the PC sampling method to denoise (e.g. see~\Cref{fig: noise-free-1500,fig: noise-1500}, results not shown for brevity). 

Finally, the comparison of~\Cref{fig: sine-nf-traj,fig: HD-500-3,fig: r2-traj} indicates that NF sampling can adequately preserve the details of the non-identifiable manifold compared to HD sampling.
Also due to their similar nature of drawing high-density posterior samples, we omit presenting results from HD sampling in the next sections.
%=============================================================
\subsection{Three-element Windkessel model}\label{example:rcr}
%=============================================================

\begin{figure}[ht!]
     \centering
     \begin{subfigure}[b]{0.42\textwidth}
         \centering
         \includegraphics[scale=0.35]{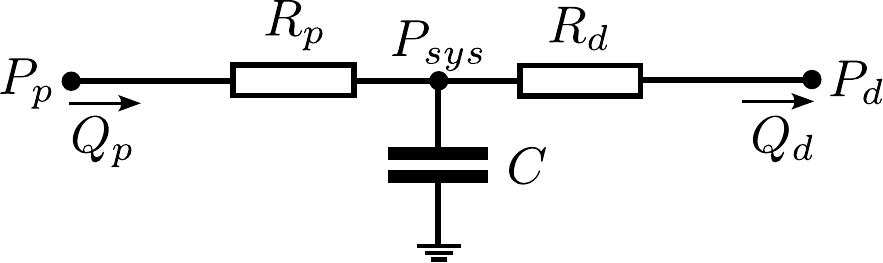}
         \caption{Schematic representation of an RCR circuit model.}
         \label{fig:RCR-diag}
     \end{subfigure}
     \hfill
     \begin{subfigure}[b]{0.55\textwidth}
         \centering
         \includegraphics[scale=0.18]{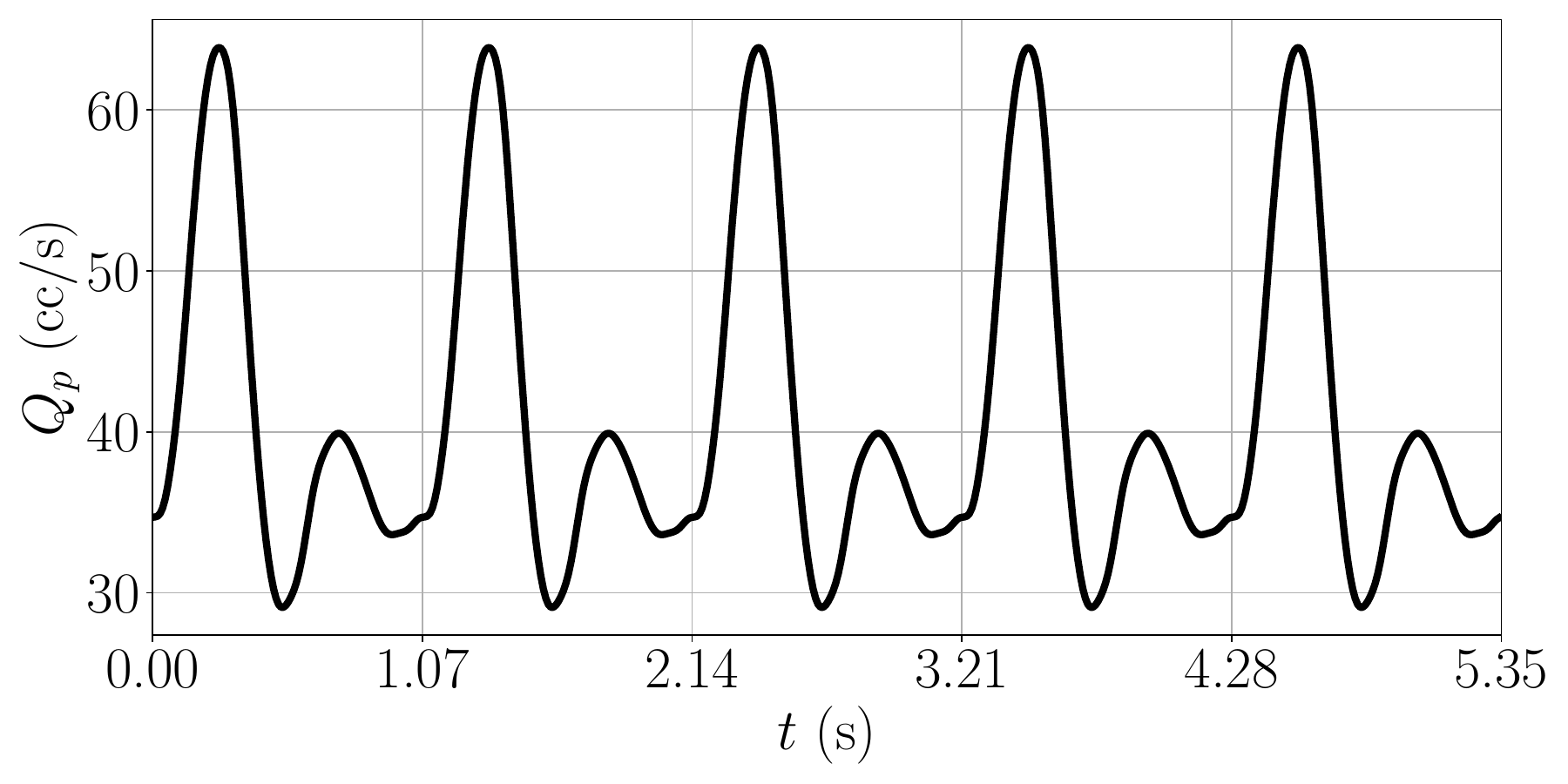}
            \caption{Proximal flow rate $Q_p$ over 5 cardiac cycles.}
    \label{fig:q_p history}
     \end{subfigure}
    \caption{Schematic representation of the RCR model and assigned inflow.}
\end{figure}

The three-element Windkessel model~\cite{shi2011review}, provides one of the simplest models for the human arterial circulation and can be formulated as a RCR circuit through the hydrodynamic analogy (see Figure~\ref{fig:RCR-diag}). 
Despite its simplicity, the RCR model has extensive applications in data-driven physiological studies and is widely used to provide boundary conditions for three-dimensional hemodynamic models (see, e.g.,~\cite{harrod2021predictive,WANG2022111454}).
It is formulated through the following coupled algebraic and ordinary differential system of equations
\begin{equation}
\begin{cases}
Q_p = \displaystyle{\frac{P_p-P_{sys}}{R_p}}, \\
Q_d = \displaystyle{\frac{P_{sys}-P_d}{R_d}}, \\
\dot{P}_{sys} = \displaystyle{\frac{Q_p-Q_d}{C}},
\end{cases}
\label{equ: RCR-ODE}
\end{equation}
where $C$ is the overall systemic capacitance which represents vascular compliance. Two resistors, $R_p$ and $R_d$ are used to model the viscous friction in vessels and $P_p$, $P_d$ and $P_{sys}$ stand for the aortic (proximal) pressure, fixed distal pressure (see Table~\ref{table:rcr paras}) and the systemic pressure, respectively. 
In addition, $Q_d$ represents the distal flow rate, whereas the proximal flow rate data $Q_p (t)$ is assigned~\cite{harrod2021predictive} (see Figure~\ref{fig:q_p history}).

This simple circuit model is non-identifiable. The average proximal pressure $\bar{P}_p = (P_{p,\text{max}} + P_{p,\text{min}})/2$ depends only on the total systemic resistance $R_p+R_d$, rather than on each of these individual parameters. 
Thus, to keep the same $\bar{P}_p$, an increment of the proximal resistance ${R}_p$ will cause a reduction of the distal resistance ${R}_d$, which also allows more flow ($Q_p-Q_d$) exiting the capacitor, resulting in an increasing capacitance $C$ to balance the system, which affects the \emph{pulse} pressure, i.e., the difference between maximum and minimum proximal pressure. 
A nonlinear correlation thus exists between the capacitance, proximal and distal resistances when maximum and minimum proximal pressures are provided as data.

Rearranging equations~\eqref{equ: RCR-ODE} with respect to the variable of interest, $P_p(t)$, the aortic pressure, resulting in the following first-order, linear ODE
\begin{equation}
\begin{cases}
\dot{P}_p = R_p \dot{Q}_p + \displaystyle{\frac{Q_p}{C}} - \displaystyle{\frac{P_p-Q_pR_p-P_d}{CR_d}}, \\
P_p(0) = 0.
\end{cases}
\label{equ: Pp ode}
\end{equation}
\begin{table}[ht!]
{\footnotesize
\begin{center}
\begin{tabular}{@{} l l @{}}
\toprule
Cardiac cycle ($t_c$) & $1.07$ (s) \\
Distal pressure ($P_d$) & $55$ (mmHg) \\ 
Proximal resistance ($R_p$) & $[500,1500]$ (Barye $\cdot$s/ml) \\
Distal resistance ($R_d$) & $[500,1500]$ (Barye $\cdot$s/ml) \\
Capacitance ($C$) & $[1\times 10^{-5}, 1\times 10^{-4}]$ (ml/Barye) \\
\bottomrule
\end{tabular}
\end{center}}
\caption{RCR model parameters and ranges.}
\label{table:rcr paras}
\end{table}

When tuning boundary conditions in numerical hemodynamics, the diastolic and systolic pressures $P_{p,\text{min}}$ and $P_{p,\text{max}}$ are usually available. As a result, we consider the following map: 
\begin{equation}\label{equ:rcr-map}
[P_{p,\text{max}} , P_{p,\text{min}}]^T = \mathcal{F}(\boldsymbol{v}),
\end{equation}
where $\boldsymbol{v} = [R_p, R_d, C]^T$ and by construction, the latent space is assumed to be one-dimensional, i.e. $\dim(\boldsymbol{\mathcal{W}})= 1$, enough for achieving robust inverse predictions.

To generate training data, we take uniform random samples of $R_p$, $R_d$ and $C$ from the ranges listed in Table~\ref{table:rcr paras} and solve the ODE $10^4$ times using the fourth order Runge-Kutta time integrator (RK4).
To achieve stable periodic solutions, we choose a time step size equal to $\Delta t = 0.01$ s and simulate the system up to 10 cardiac cycles (10.7 s), where $P_{p,\text{max}}$ and $P_{p,\text{min}}$ are extracted from the last three heart cycles.
We then train the inVAErt network with the hyperparameter combination listed in the Appendix.

\begin{figure}[ht!]
     \centering
     \begin{subfigure}[b]{0.3\textwidth}
         \centering
         \includegraphics[scale=0.2]{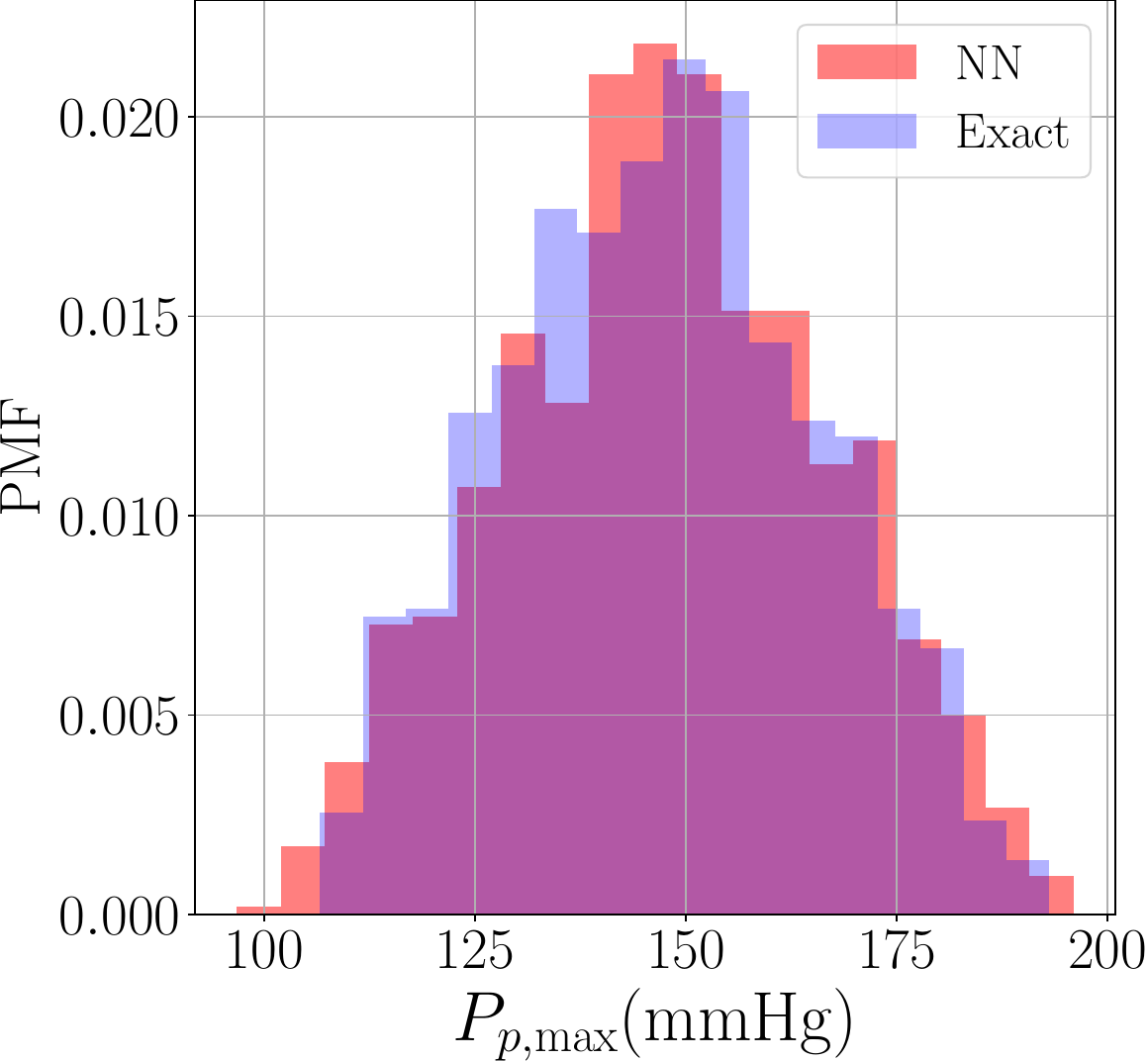}
         \caption{Histogram of $P_{p,\max}$.}
     \end{subfigure}
     \hfill
    \begin{subfigure}[b]{0.3\textwidth}
         \centering
         \includegraphics[scale=0.2]{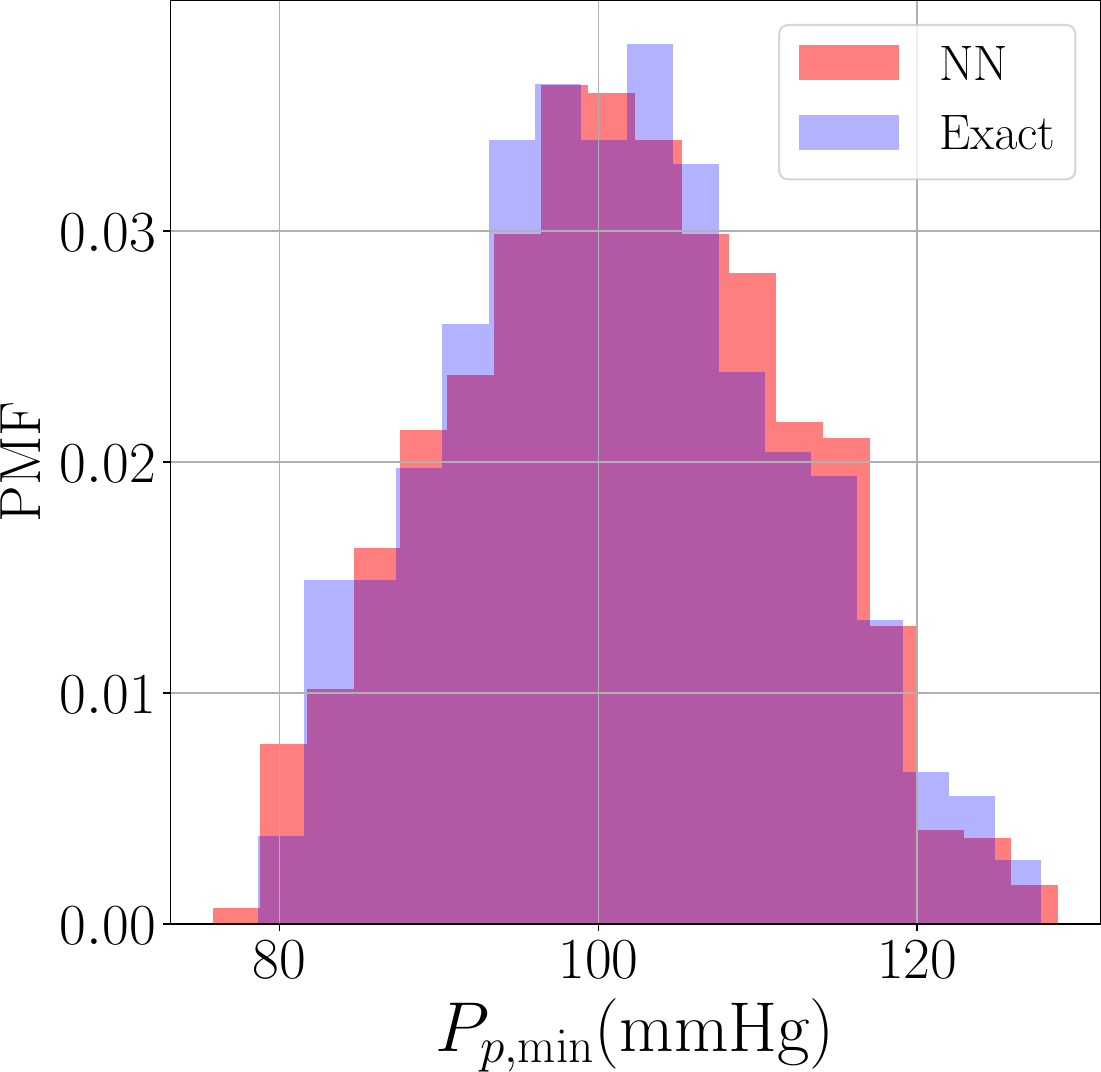}
         \caption{Histogram of $P_{p,\min}$.}
     \end{subfigure}
     \hfill
    \begin{subfigure}[b]{0.3\textwidth}
         \centering
         \includegraphics[scale=0.2]{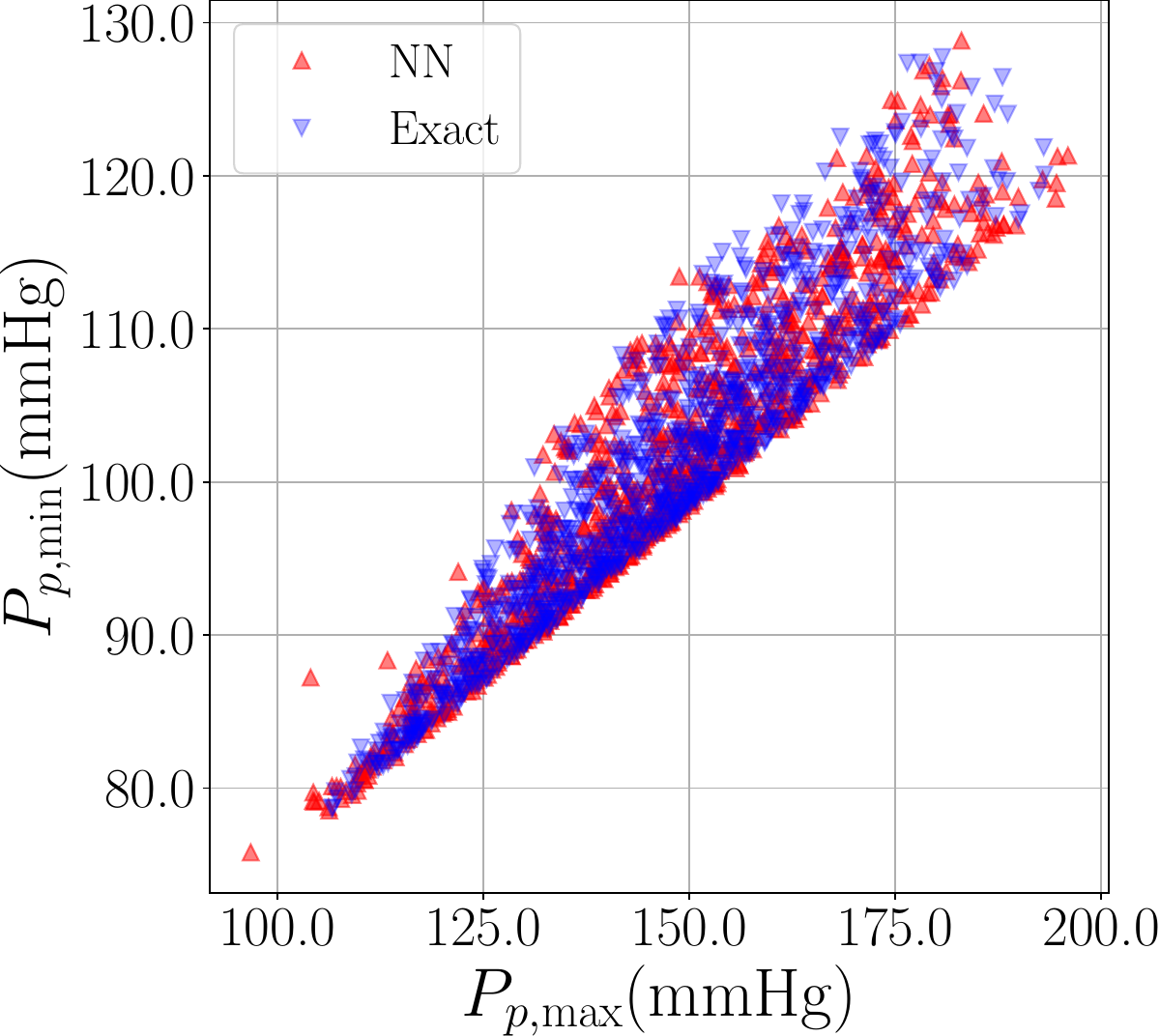}
         \caption{$P_{p,\max}$-$P_{p,\min}$ correlation.}
     \end{subfigure}
    \caption{Distributions of parametric RCR model outputs as learned by $\mathscr{N}_{f}$.}
    \label{fig: RCR-NF-results}
\end{figure}

First, we discuss the learned distributions of the system outputs $P_{p,\text{max}}$ and $P_{p,\text{min}}$. The density estimator correctly learns the parameter correlations and ranges, as shown in Figure~\ref{fig: RCR-NF-results}.

\begin{figure}[ht!]
    \begin{subfigure}[b]{0.39\textwidth}
         \centering
         \includegraphics[scale=0.25]{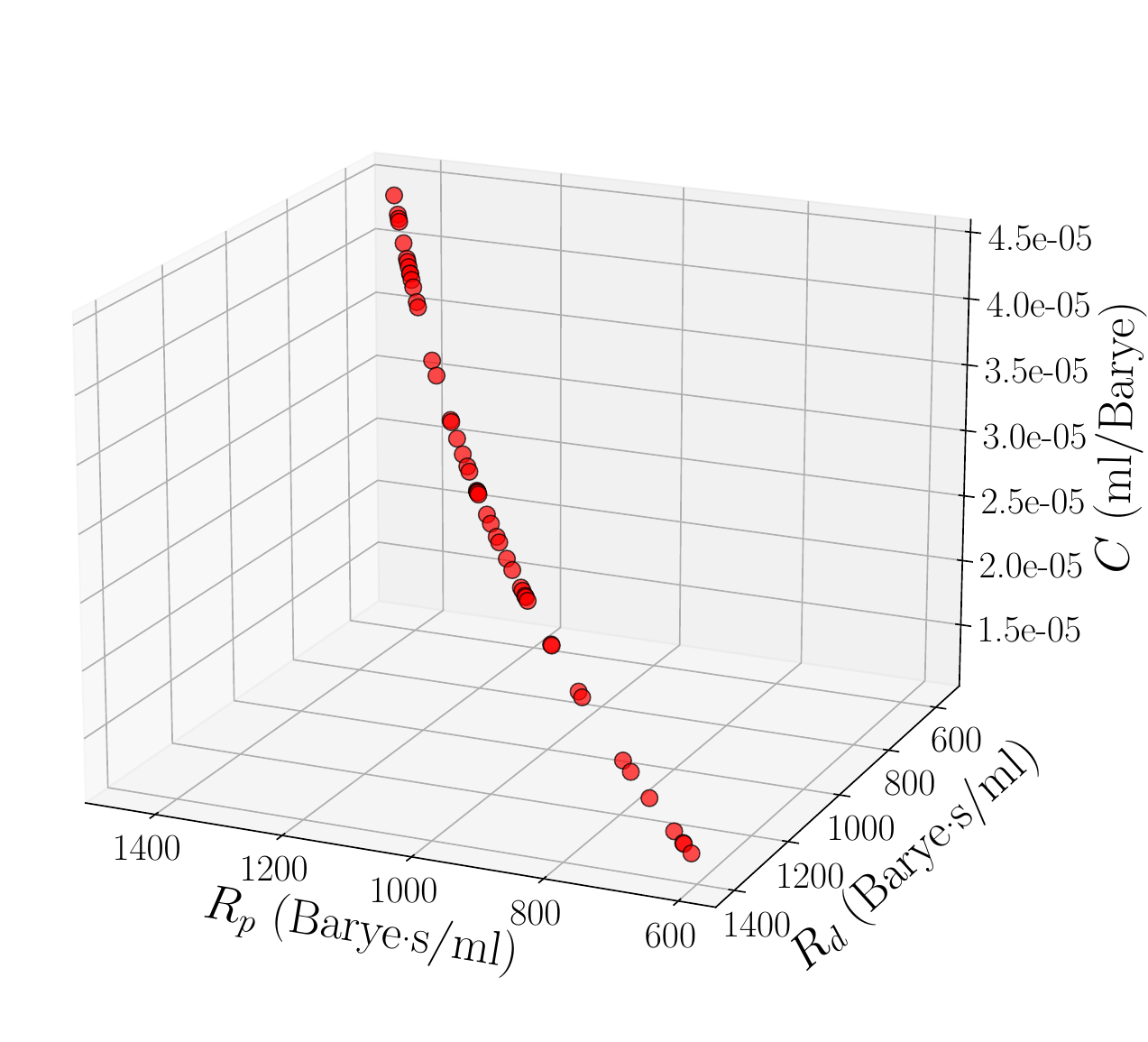}
         \caption{Predicted samples of $\widehat{\boldsymbol{v}}$.}
         \label{fig: rcr-traj-fixmax-min}
     \end{subfigure}
     \hfill
    \begin{subfigure}[b]{0.6\textwidth}
         \centering
         \includegraphics[scale=0.265]{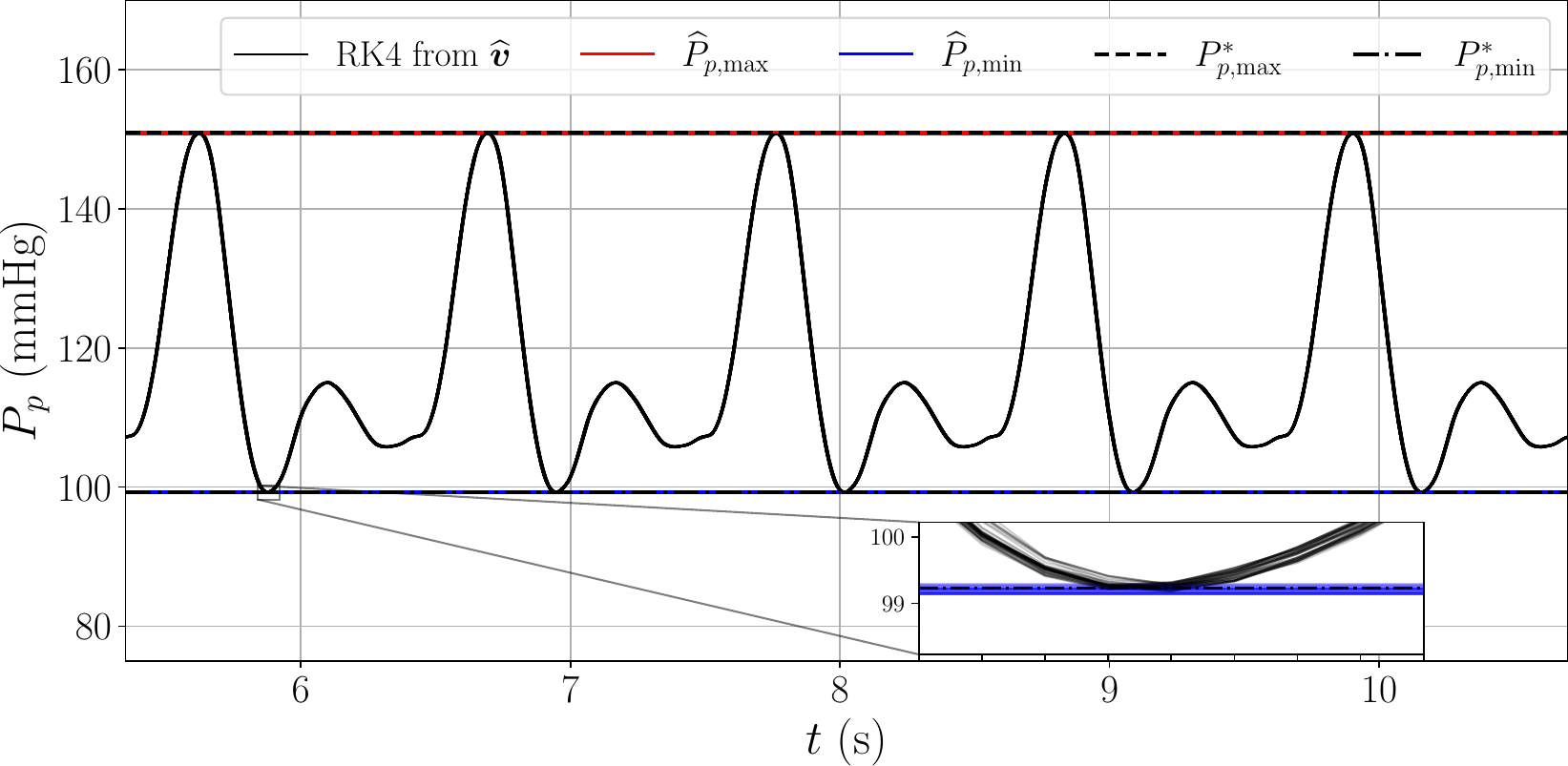}
         \caption{Predicted system output from $\widehat{\boldsymbol{v}}$.}
         \label{fig: rcr-plug-backin}
     \end{subfigure}
    \caption{Model inversion results for the parametric RCR system, setting $[P_{p,\text{max}}^* , P_{p,\text{min}}^*]$ and sampling from the latent space $\mathcal{W}$. The decoded parameters $\widehat{\boldsymbol{v}}$ leads to system output predictions (RK4) close to $[P_{p,\text{max}}^* , P_{p,\text{min}}^*]$. The trained forward model $\mathscr{N}_e$ also provides accurate predictions $[\widehat{P}_{p,\text{max}} , \widehat{P}_{p,\text{min}}]$.}
    \label{fig: RCR-fix y and sample w}
\end{figure}

\begin{figure}[ht!]
    \begin{subfigure}[b]{0.325\textwidth}
         \centering
         \includegraphics[scale=0.22]{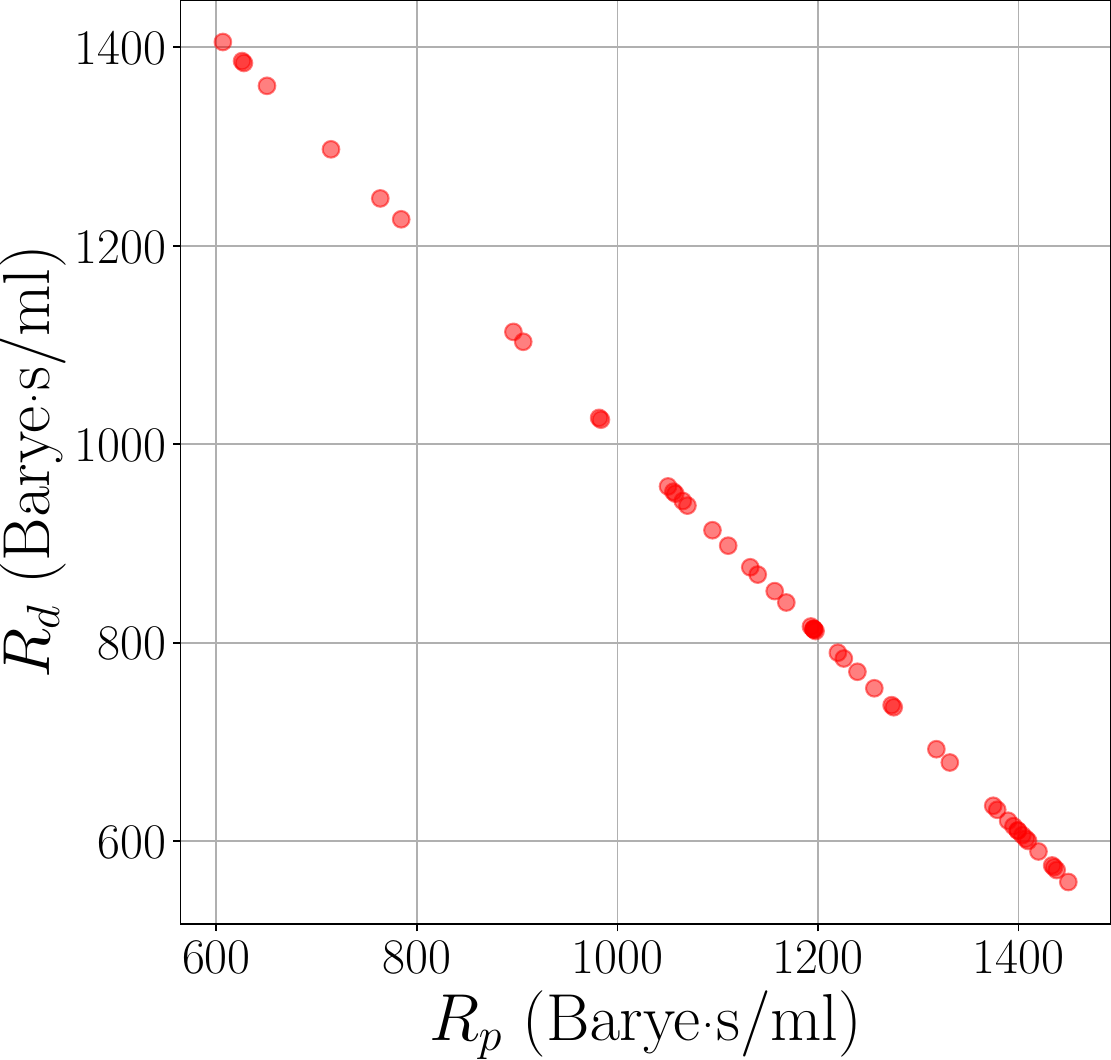}
         \caption{$R_p$-$R_d$ correlation.}
     \end{subfigure}
     \hfill
    \begin{subfigure}[b]{0.325\textwidth}
         \centering
         \includegraphics[scale=0.22]{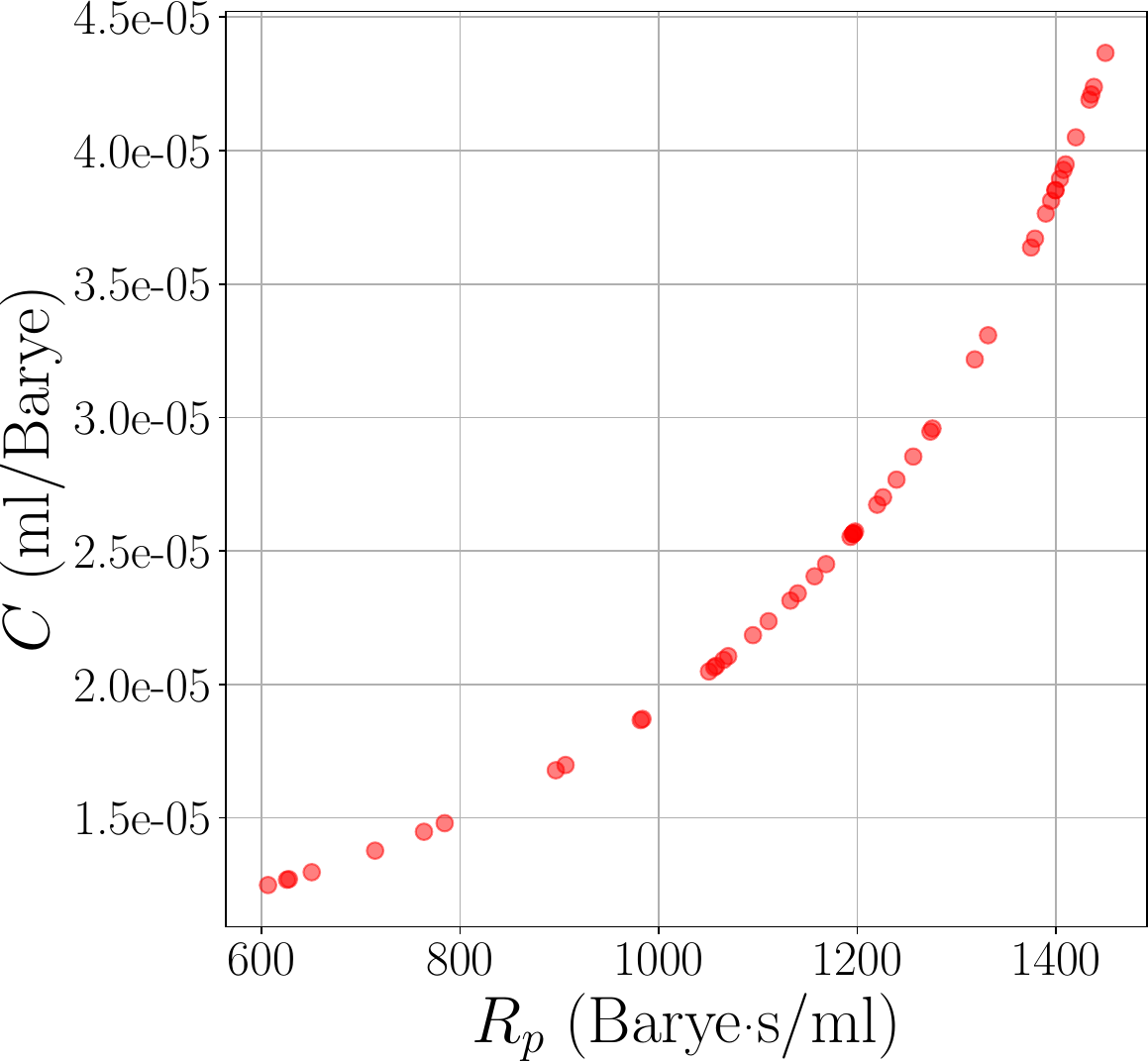}
         \caption{$R_p$-$C$ correlation.}
     \end{subfigure}
          \hfill
    \begin{subfigure}[b]{0.325\textwidth}
         \centering
         \includegraphics[scale=0.22]{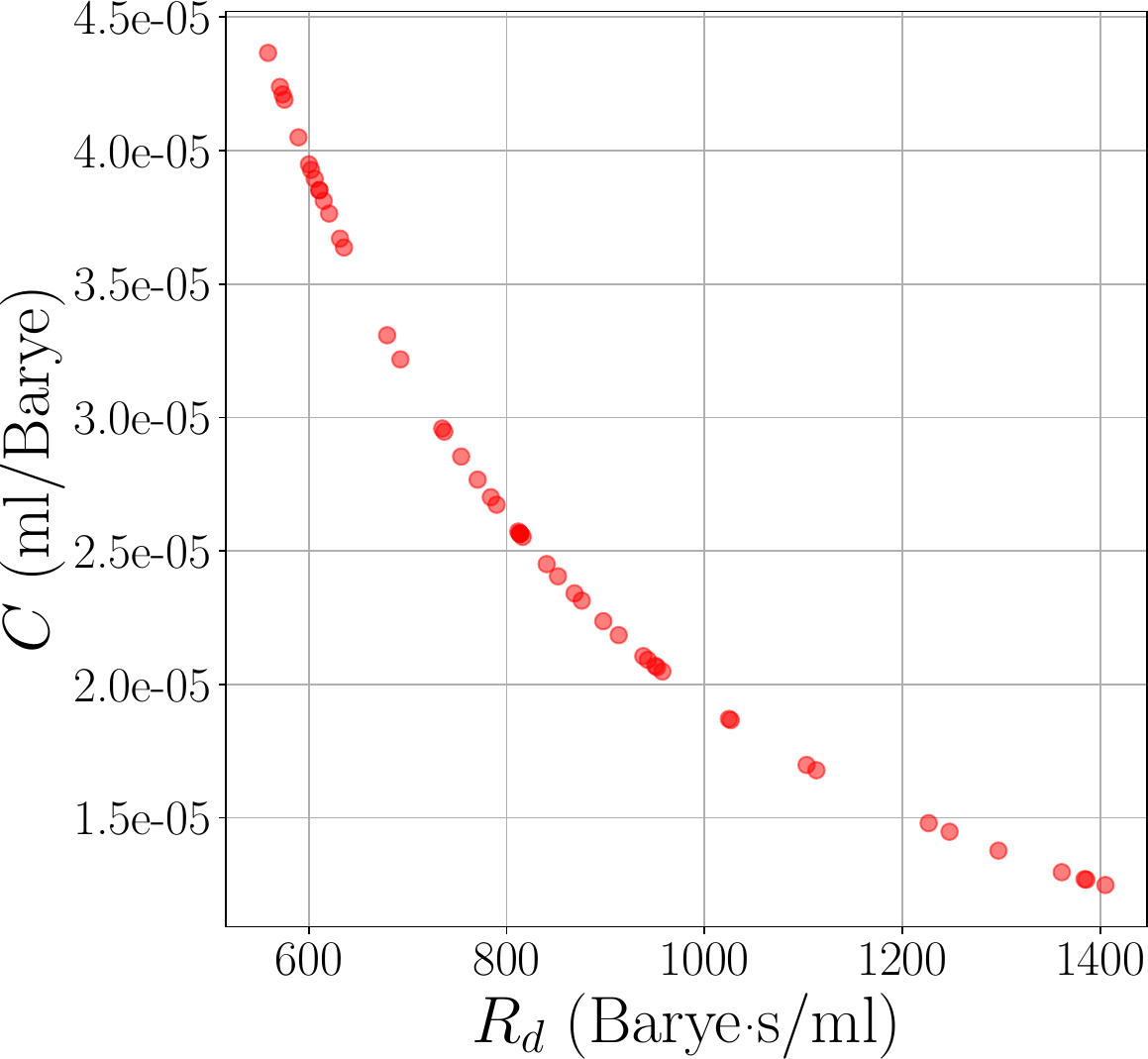}
         \caption{$R_d$-$C$ correlation.}
     \end{subfigure}
    \caption{Projected sample trajectories from Figure~\ref{fig: rcr-traj-fixmax-min} when fixing $[P_{p,\text{max}}^* , P_{p,\text{min}}^*]$ and sample from $\mathcal{W}$.}
    \label{fig: RCR-projected samples}
\end{figure}

Next, by inverse analysis, we would like to determine all combinations of $R_{p}$, $R_{d}$ and $C$ which correspond to given systolic and diastolic distal pressures. To do so, we feed the trained decoder $\mathscr{N}_d$ with a valid $[P^*_{p,\max}, P^*_{p,\min}]$ sampled from the trained $\mathscr{N}_f$ and 50 $w$ drawn from $\mathcal{N}(0,1)$, resulting in the trajectory displayed in Figure~\ref{fig: rcr-traj-fixmax-min}. Additionally, we plot the binary correlations of these predicted samples in Figure~\ref{fig: RCR-projected samples}.

To confirm that the RCR parameters computed by the decoder correspond to the expected maximum and minimum pressures, we integrate in time the 50 resulting parameter combinations with RK4.
The resulting periodic curves are displayed in Figure~\ref{fig: rcr-plug-backin}. They are found to almost perfectly overlap with one another, all oscillating between the correct systolic and diastolic pressures.
In addition, we can evaluate the performance of the trained emulator by examining the predicted $\widehat{P}_{p,\max}$ and $\widehat{P}_{p,\min}$ values and comparing them with those obtained from the exact RK4 integration in time. 
These predictions provide insight into the quality of the forward model evaluations.

Note that the process of obtaining correlated parameters from observations would normally require multiple optimization tasks, since each of these tasks would converge to a single parameter combination. Using the inVAErt network, this process is almost instantaneous and multiple parameter combinations are generated at the same time, providing a superior characterization of all possible solutions for this ill-posed inverse problem. Note also that any form of regularization would have only provided a partial characterization of the right answer, since there is no one right answer. 


%====================================================
\subsection{Lorenz oscillator}\label{example: lorenz}
%====================================================

The Lorenz system describes the dynamics of atmospheric convection, where $x(t)$ relates to the rate of convection and $y(t)$, $z(t)$ models the horizontal and vertical temperature variations, respectively~\cite{sparrow2012lorenz}. It is formulated as a system of ordinary differential equations of the form
\begin{equation}
    \begin{cases}
        \dot{x} = Pr (y - x) , \\ 
        \dot{y} = x (Ra - z) - y  ,\\ 
        \dot{z} = xy - b z \ . \\
    \end{cases}
    \label{equ: lorenz}
\end{equation}
The parameters $Pr, Ra$ are proportional to the Prandtl number and the Rayleigh number, respectively and $b$ is a dependent geometric factor. 

The forward map of this nonlinear ODE system is defined as
\begin{equation}\label{equ:lorenz-map}
\boldsymbol{y} = [x(t), y(t), z(t)]^T = \mathcal{F}(\boldsymbol{v}, \mathcal{D}_{\boldsymbol{v}}),
\end{equation}
where $\boldsymbol{v} = [Pr, Ra, b, t]^T$, resulting in a one dimensional latent space $\boldsymbol{\mathcal{W}}$. The auxiliary dataset $\mathcal{D}_{\boldsymbol{v}}$, as described above, contains the time delayed solutions
\[
\mathcal{D}_{\boldsymbol{v}} = \left\{\boldsymbol{y}(t-n_p\Delta t), \cdots, \boldsymbol{y}(t-2\Delta t), \boldsymbol{y}(t-\Delta t) \right\}.
\]
A training dataset is obtained by randomly sample the parameter vector $\boldsymbol{v}$ from the specified ranges listed in Table~\ref{table:lorenz paras}.
Unlike previous examples, it is worth noting that we now consider the simulation ending time $t$ as an additional random parameter, drawn from a discrete uniform distribution with an interval of $\Delta t = 5\times 10^{-4}$ s. 
This interval aligns with the time step size used in the RK4 time integration. We solve the Lorenz system using 5000 sets of inputs $\boldsymbol{v}$ and take 30 time points at random from each simulation, leading to $1.5 \times 10^5$ total training samples.
A suggested hyperparameter choice is listed in the Appendix and additionally, in our experiments, we have observed that increasing the number of lagged steps $n_p$ up to 10 leads to noticeable benefits for the emulator learning.
\begin{table}[ht!]
{\footnotesize
\begin{center}
\begin{tabular}{@{} l l @{}}
\toprule
Initial conditions ($x_0, y_0, z_0$) & 0,1,0 \\
Largest possible simulation time ($T_f$) & 4 (s) \\
Prandtl number  ($Pr$) &  [$8, 12$] \\ 
Rayleigh number ($Ra$) &  [$26, 30$] \\
Geometric factor ($b$) &  [8/3-1, 8/3+1] \\
\bottomrule
\end{tabular}
\end{center}}
\caption{Parameters of the Lorenz system.}
\label{table:lorenz paras}
\end{table}

\begin{figure}[ht!]
    \begin{subfigure}[b]{0.48\textwidth}
         \centering
         \includegraphics[scale=0.3]{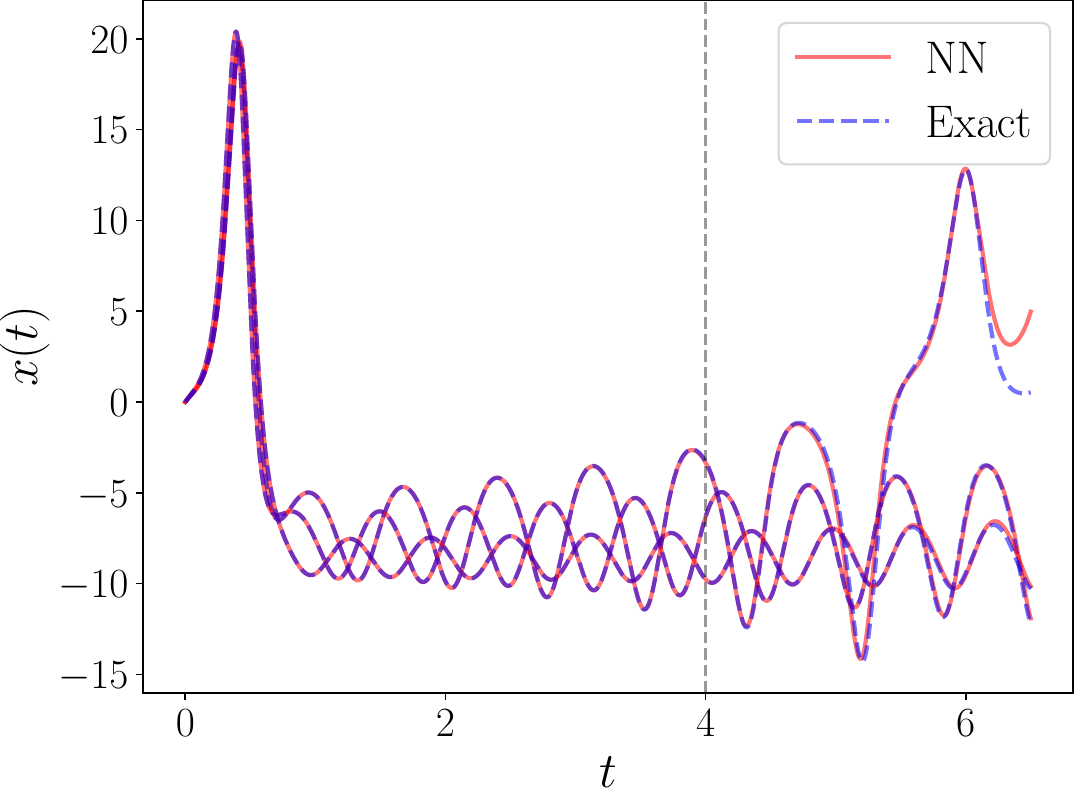}
         \caption{Three examples of learned dynamical response for $x(t)$, superimposed to the exact Lorenz trajectory.}
         \label{fig: lorenz x}
     \end{subfigure}
     \hfill
     \begin{subfigure}[b]{0.48\textwidth}
         \centering
         \includegraphics[scale=0.3]{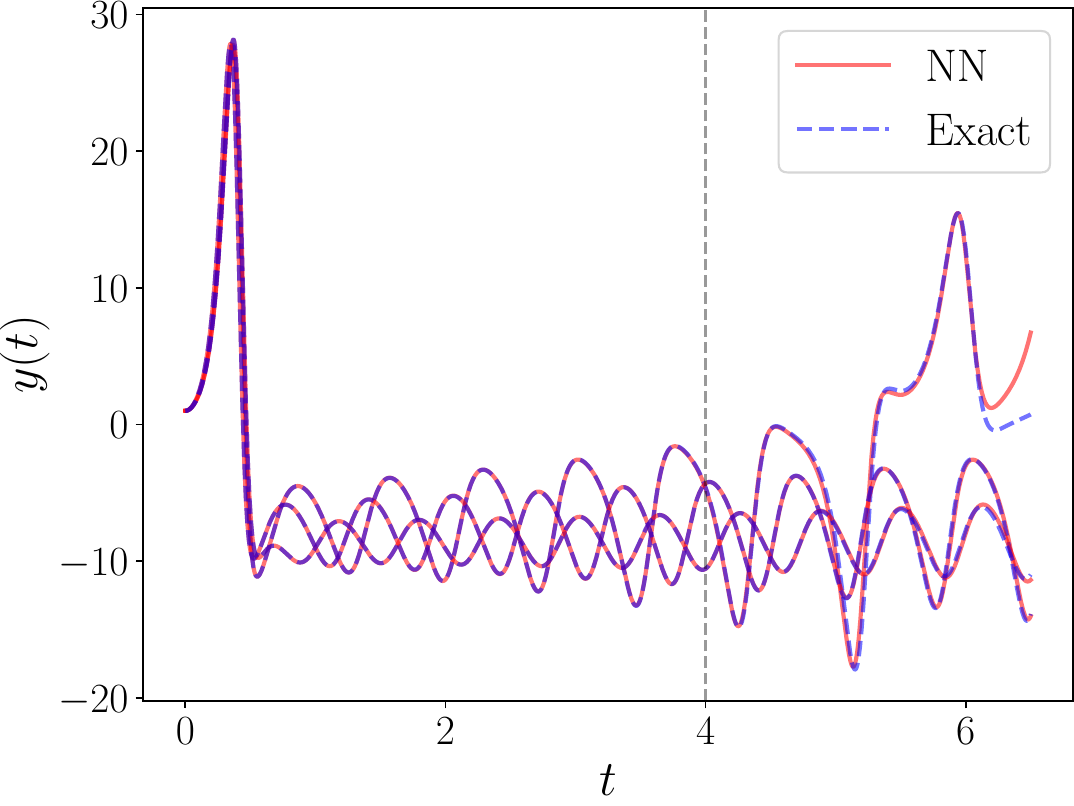}
         \caption{Three examples of learned dynamical response for $y(t)$, superimposed to the exact Lorenz trajectory.}
          \label{fig: lorenz y}
     \end{subfigure}\\
     \begin{subfigure}[b]{0.48\textwidth}
         \centering
         \includegraphics[scale=0.3]{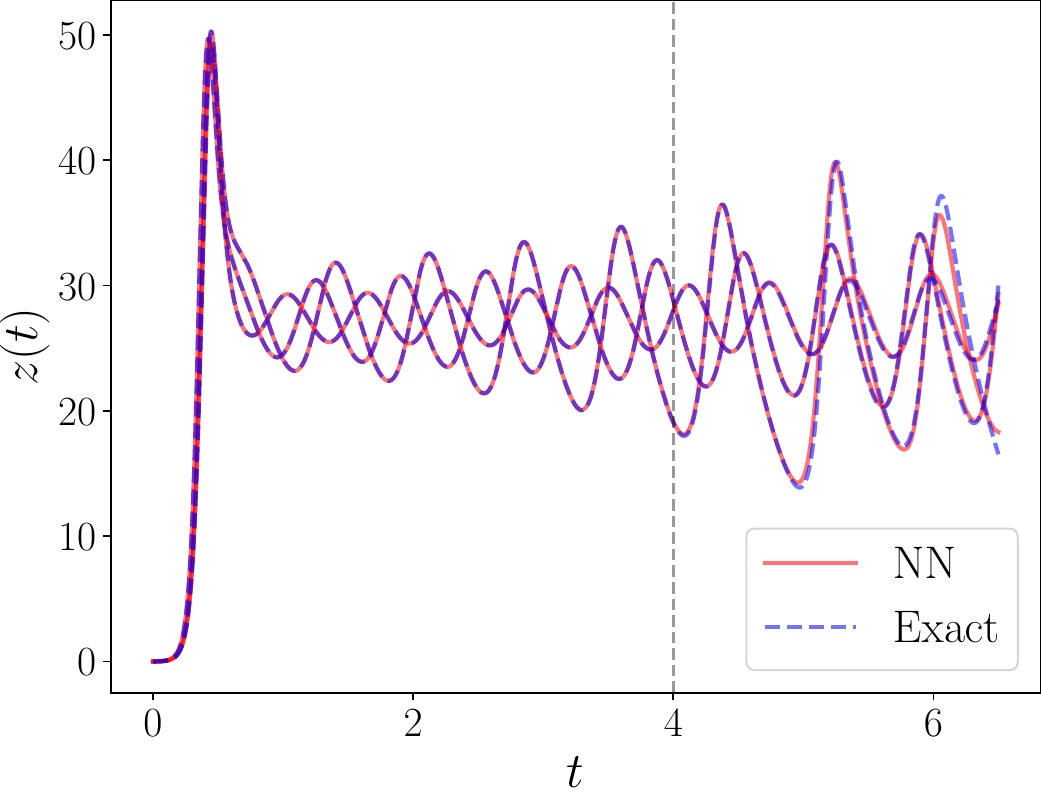}
         \caption{Three examples of learned dynamical response for $z(t)$, superimposed to the exact Lorenz trajectory.}
        \label{fig: lorenz z}
     \end{subfigure}
     \hfill
    \begin{subfigure}[b]{0.48\textwidth}
         \centering
         \includegraphics[scale=0.32]{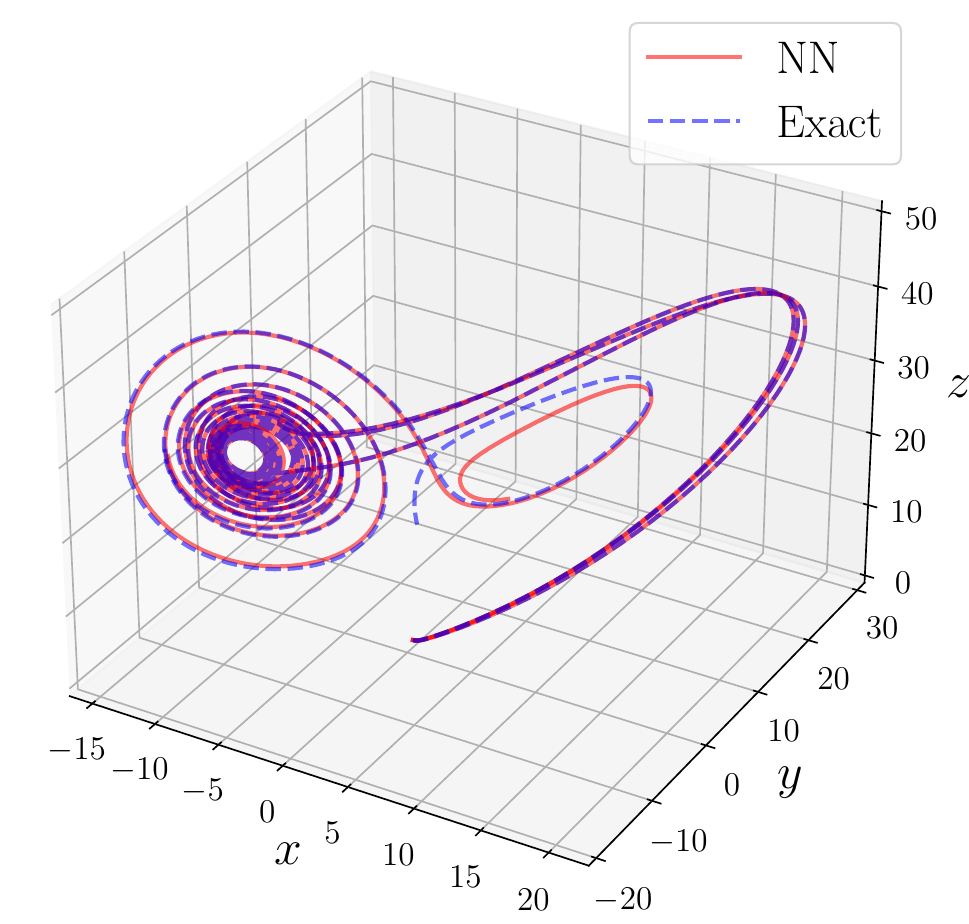}
         \caption{Phase plots for the learned dynamical response and exact Lorenz trajectory.}
         \label{fig: lorenz xyz}
     \end{subfigure}
    \caption{Comparison between the dynamic response learned by the emulator $\mathscr{N}_e$ and the exact solution computed with RK4 for the Lorenz system up to 6.5 seconds.}
    \label{fig: Lorenz-encoder}
\end{figure}

%===========================================
% \subsubsection{Results} \label{example: lorenz-results}
%===========================================
%
We first focus on the accuracy of the trained emulator.
We pick three sets of random samples of $Pr, Ra$ and $b$ unseen during training and forward the trained encoder model $\mathscr{N}_e$ up to 6.5 seconds. 
Note that the model only requires the first $n_p$ exact steps as inputs, after which, the emulator outputs from previous time steps are used as inputs for successive steps (since the emulator, in this example, is designed to learn the \emph{flow map}).
For the prior ranges of $Pr, Ra, b$ listed in Table~\ref{table:lorenz paras}, and up to 4 seconds simulation time, almost all solution trajectories converge towards one of the attractors with regular oscillatory patterns, although varying in frequency, magnitude and speed of convergence.
Nevertheless, due to a relatively large $Ra$, the system has the potential to bifurcate towards another attractor much earlier in time, even though this event is rare within the selected prior ranges. 
Thus, as shown in Figure~\ref{fig: Lorenz-encoder}, an insufficient amount of training samples may lead to a sub-optimal performance in predicting rare dynamical responses.
%, particularly when aperiodic solution behavior emerges.
%

\begin{figure}[ht!]
    \begin{subfigure}[b]{0.33\textwidth}
         \centering
         \includegraphics[scale=0.22]{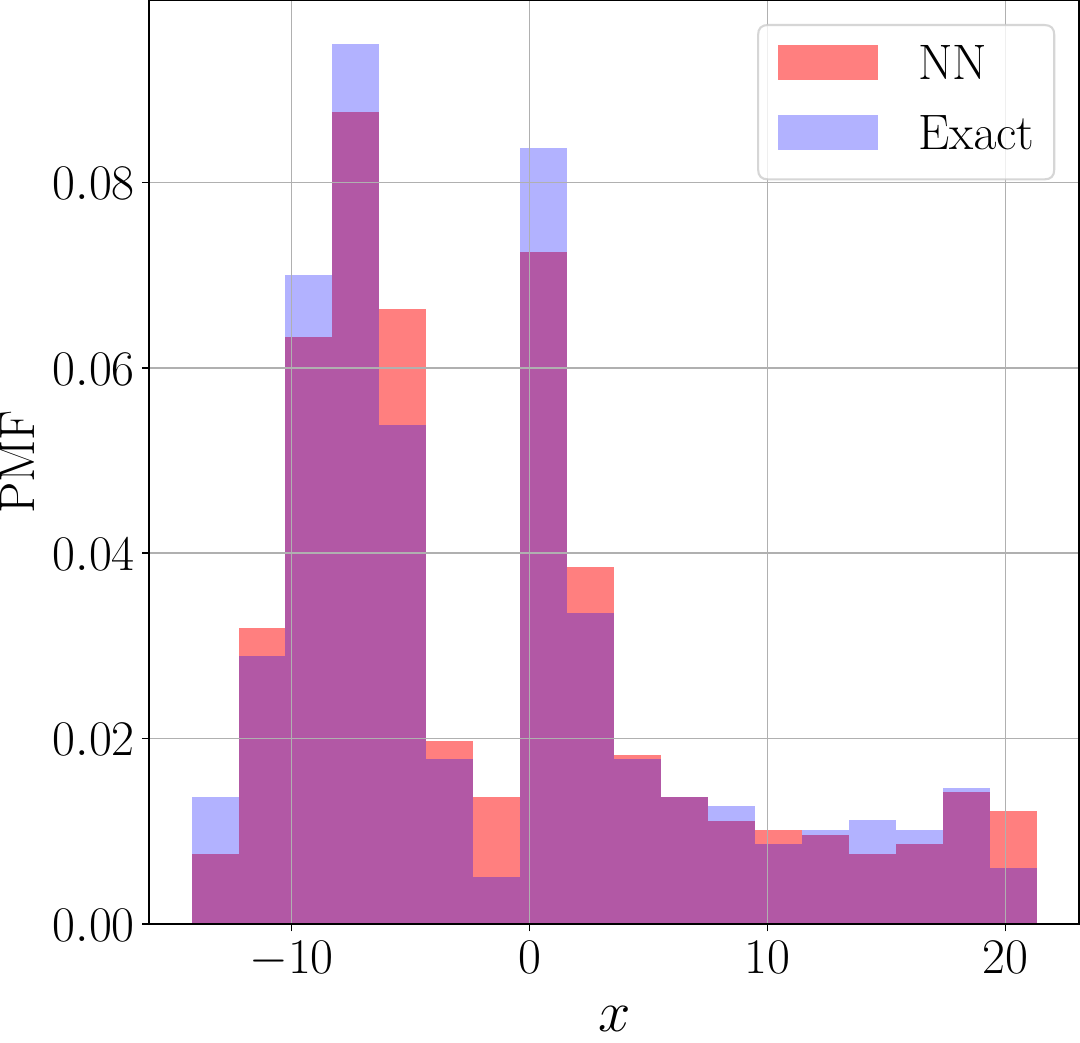}
         \caption{$x(t)$-histogram.}
     \end{subfigure}
     \hfill
    \begin{subfigure}[b]{0.33\textwidth}
         \centering
         \includegraphics[scale=0.22]{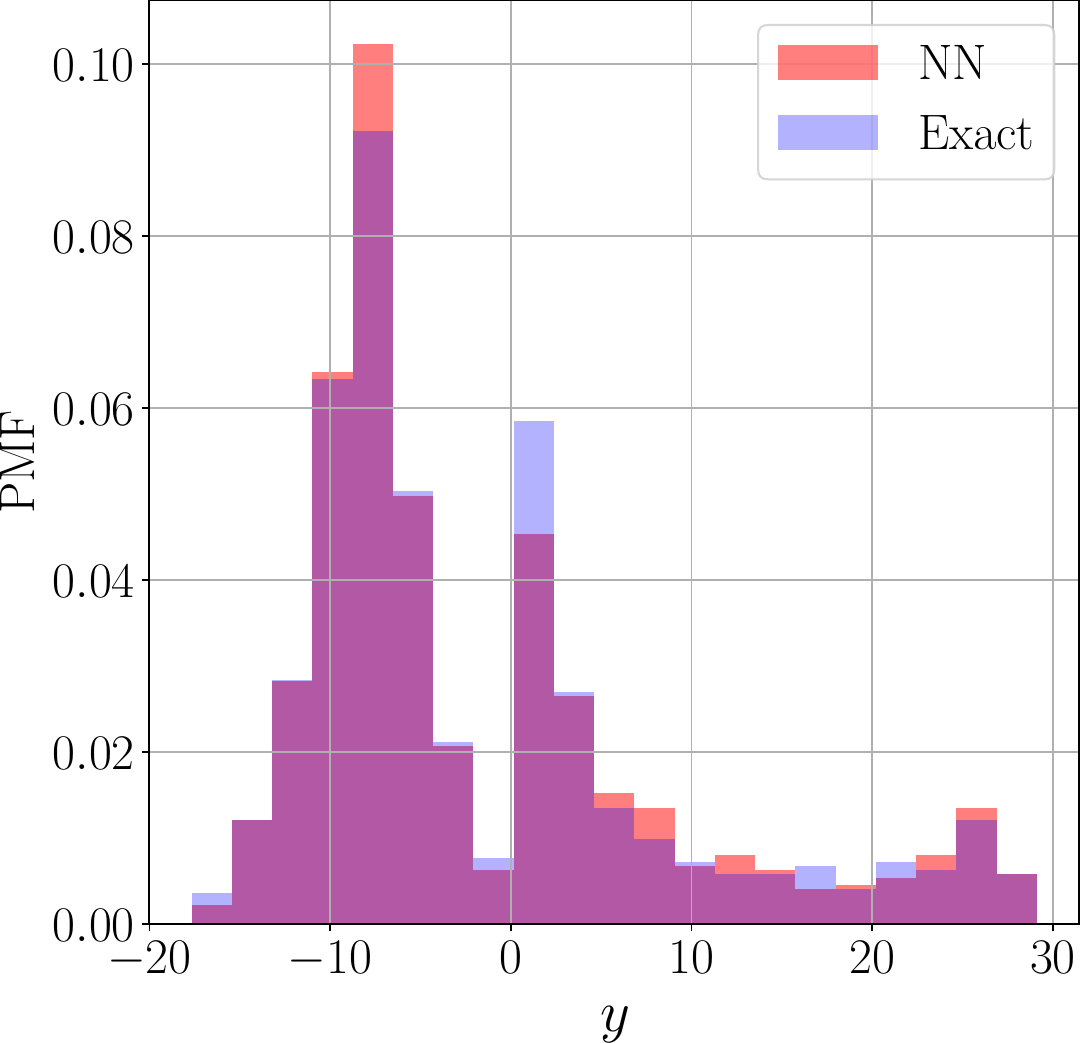}
         \caption{$y(t)$-histogram.}
     \end{subfigure}
    \hfill
     \begin{subfigure}[b]{0.33\textwidth}
         \centering
         \includegraphics[scale=0.22]{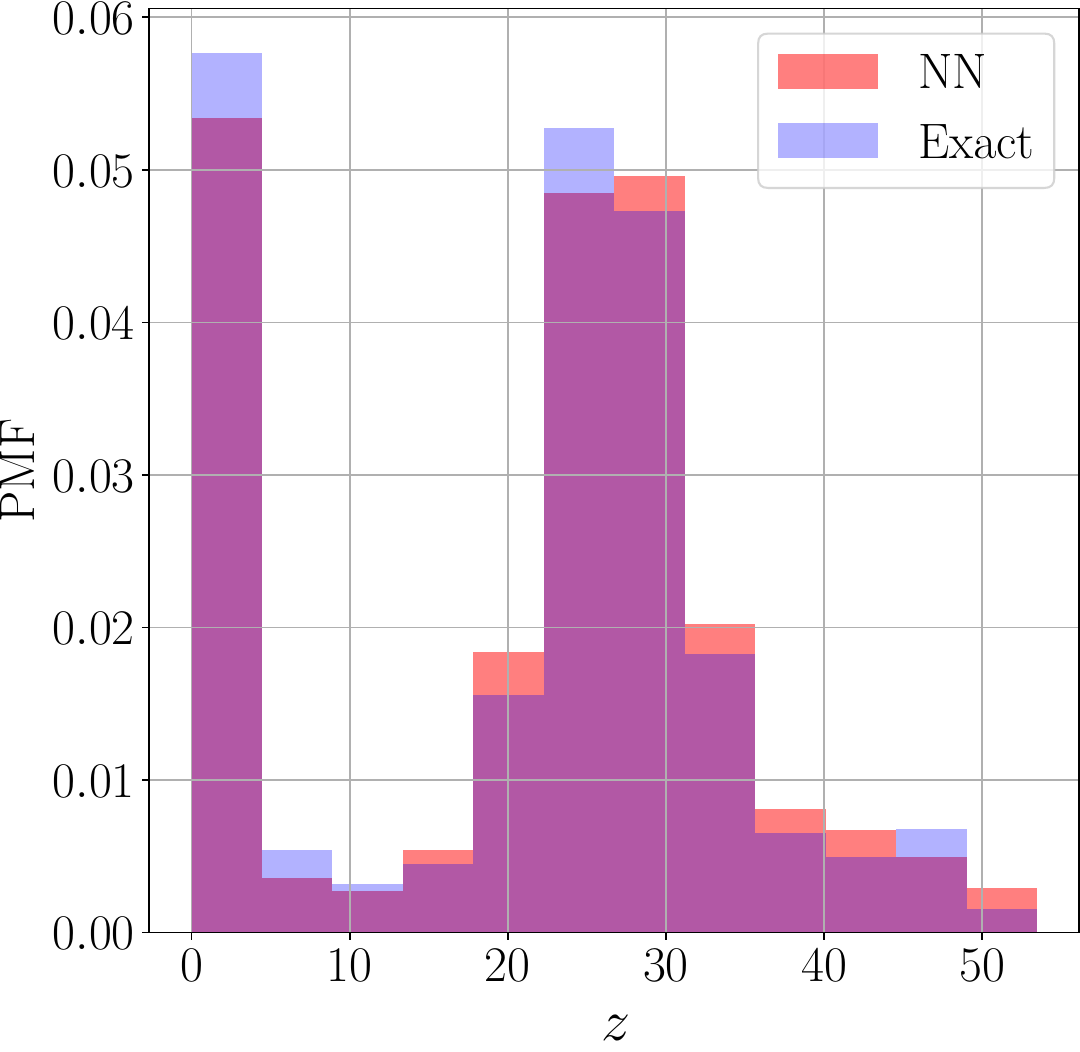}
         \caption{$z(t)$-histogram.}
     \end{subfigure} \\
     
        \begin{subfigure}[b]{0.32\textwidth}
         \centering
         \includegraphics[scale=0.22]{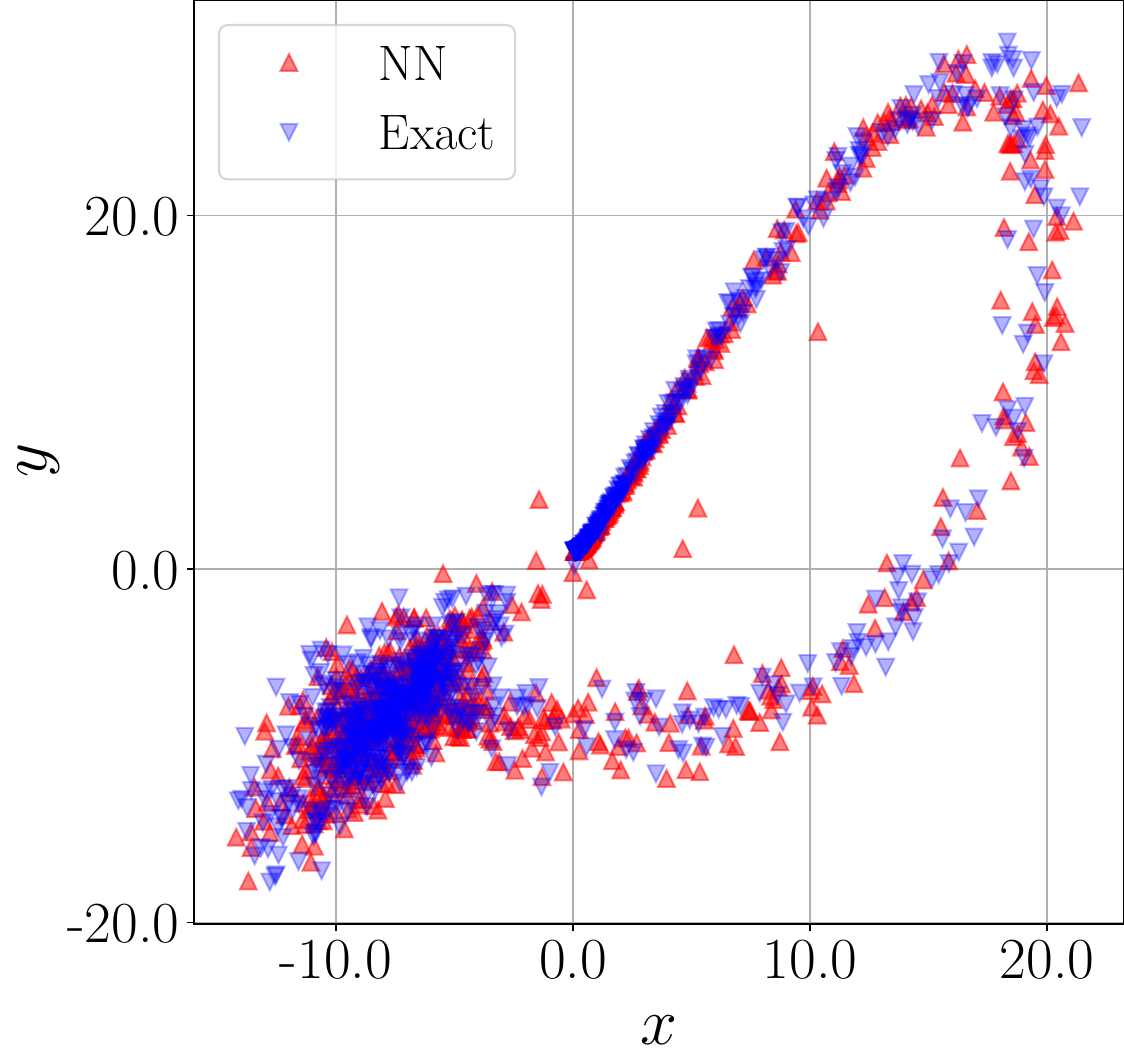}
         \caption{$x(t)$-$y(t)$ correlation.}
     \end{subfigure}
     \hfill
     \begin{subfigure}[b]{0.32\textwidth}
         \centering
         \includegraphics[scale=0.22]{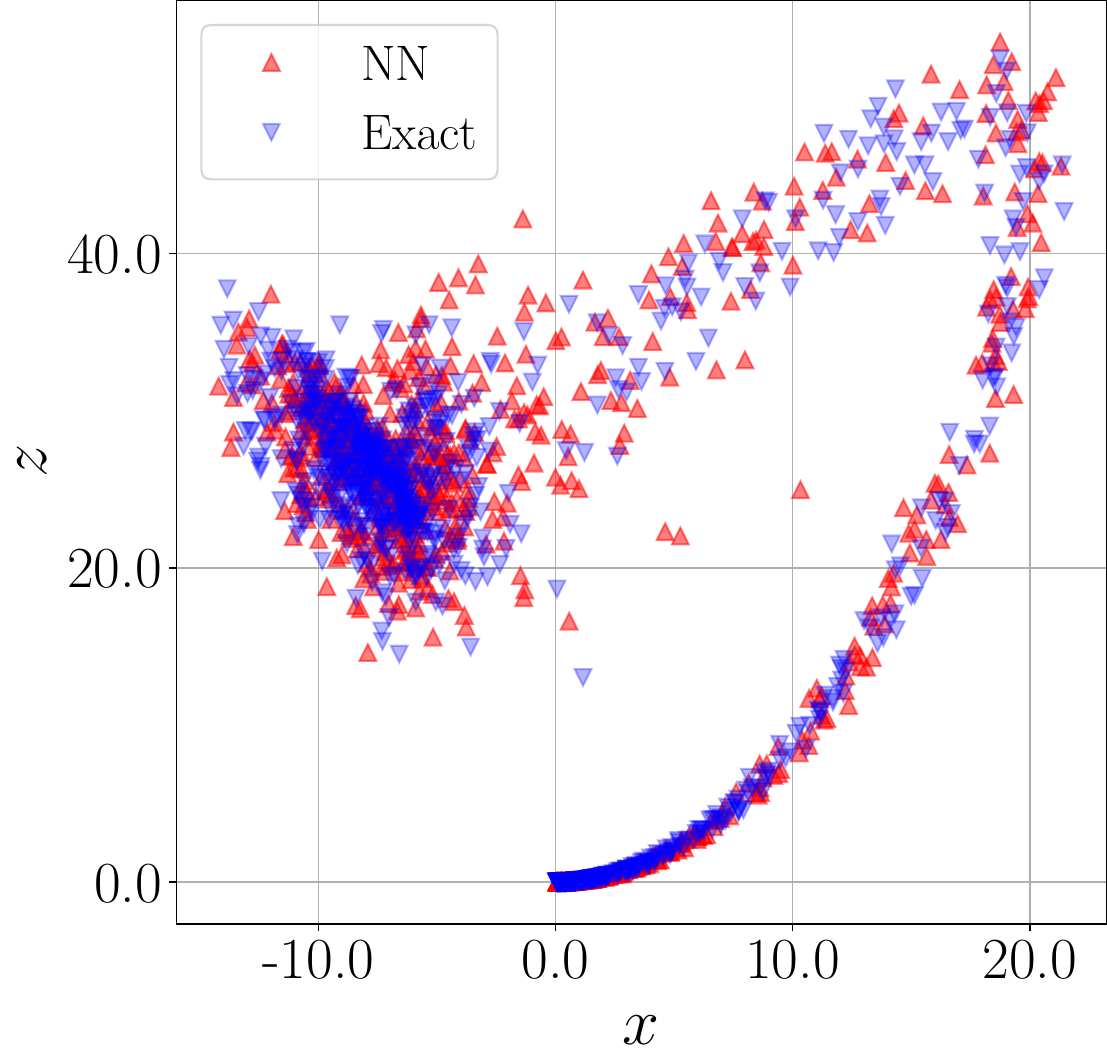}
         \caption{$x(t)$-$z(t)$ correlation.}
     \end{subfigure}
    \hfill
    \begin{subfigure}[b]{0.32\textwidth}
         \centering
         \includegraphics[scale=0.22]{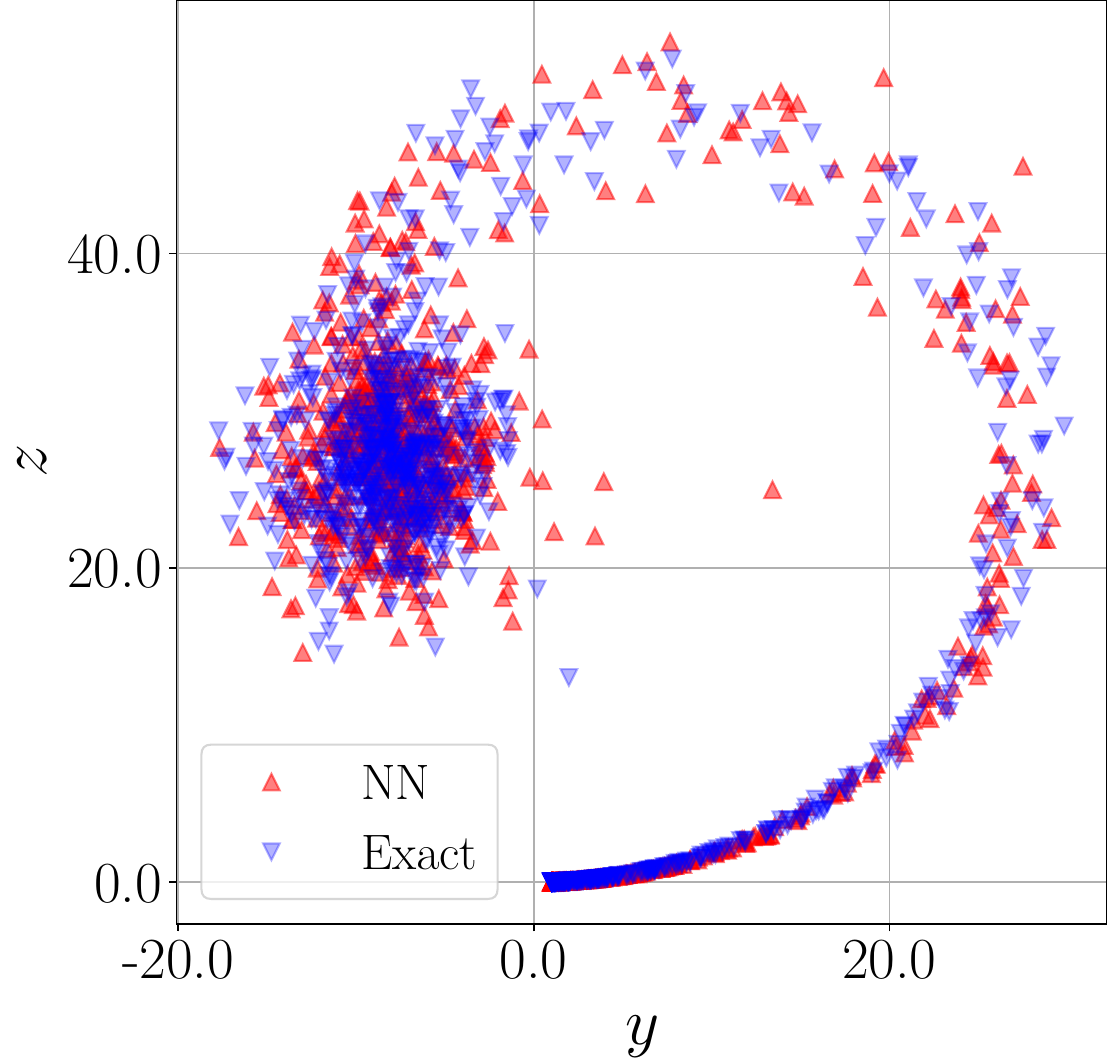}
         \caption{$y(t)$-$z(t)$ correlation.}
     \end{subfigure}
    \caption{Density estimation of parametric Lorenz system outputs.}
    \label{fig: Lorenz-NF}
\end{figure}

Next we find that the Real-NVP sampler $\mathscr{N}_f$ effectively identifies whether a spatial location resides within the high-density regions, such as those around one of the attractors or in proximity to the initial condition, as shown in Figure~\ref{fig: Lorenz-NF}.
However its accuracy is reduced at the tails, which corresponds to rare system trajectories.

We then investigate the ability of the inVAErt network to solve inverse problems for the given parametric Lorenz system.
From our definition of inputs and outputs in \eqref{equ:lorenz-map}, we provide a predetermined system output $\boldsymbol{y}^*$, obtained by sampling from the trained Real-NVP model, and ask the inVAErt network to provide all parameters $Pr, Ra, b$ and time $t$ where the resulting solution trajectory should reach $\boldsymbol{y}^*$ at time $t$.

To do so, we fix $\boldsymbol{y}^* = [-3.4723, -8.9758, 26.2026]^T$, initially focusing on a state that the parametric Lorenz system can reach at an early time. We then draw 100 standard Gaussian samples for $\boldsymbol{w}$ and apply $\mathscr{N}_d$ to the concatenated array $\widetilde{\boldsymbol{y}}^* = [\boldsymbol{y}^*, \boldsymbol{w}]$. 
To verify the accuracy, we forward the RK4 solver with each inverted sample $\widehat{\boldsymbol{v}} = [\widehat{Pr}, \widehat{Ra}, \widehat{b}, \widehat{t}]$ and terminate the simulation exactly at $t = \widehat{t}$.

Our definition of the forward model~\eqref{equ:lorenz-map} implies an one-dimensional latent space $\boldsymbol{\mathcal{W}}$, but in practice, improved results can be obtained by increasing the latent space dimensionality.
In this context, we investigated latent spaces $\boldsymbol{\mathcal{W}}$ with dimensions 2, 4, 6, and 8 and observed relatively consistent performances across these dimensions. Nevertheless, it is crucial to recognize that utilizing a lower-dimensional latent space can occasionally result in the loss of certain features when constructing non-identifiable input parameter manifolds.

\begin{figure}[ht!]
    \begin{subfigure}[b]{0.495\textwidth}
         \centering
        \includegraphics[scale=0.28]{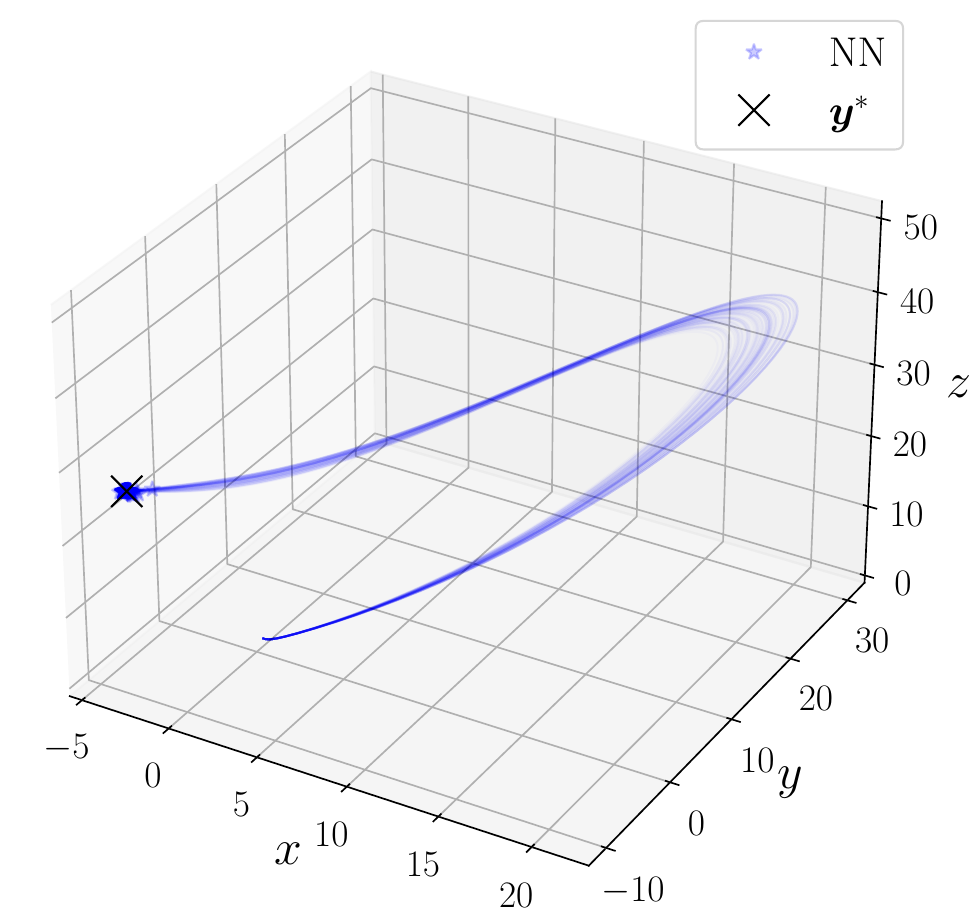}
         \caption{Phase plot trajectories generated from $\widehat{\boldsymbol{v}}$.}
     \end{subfigure}
     \hfill
     \begin{subfigure}[b]{0.495\textwidth}
         \centering
         \includegraphics[scale=0.22]{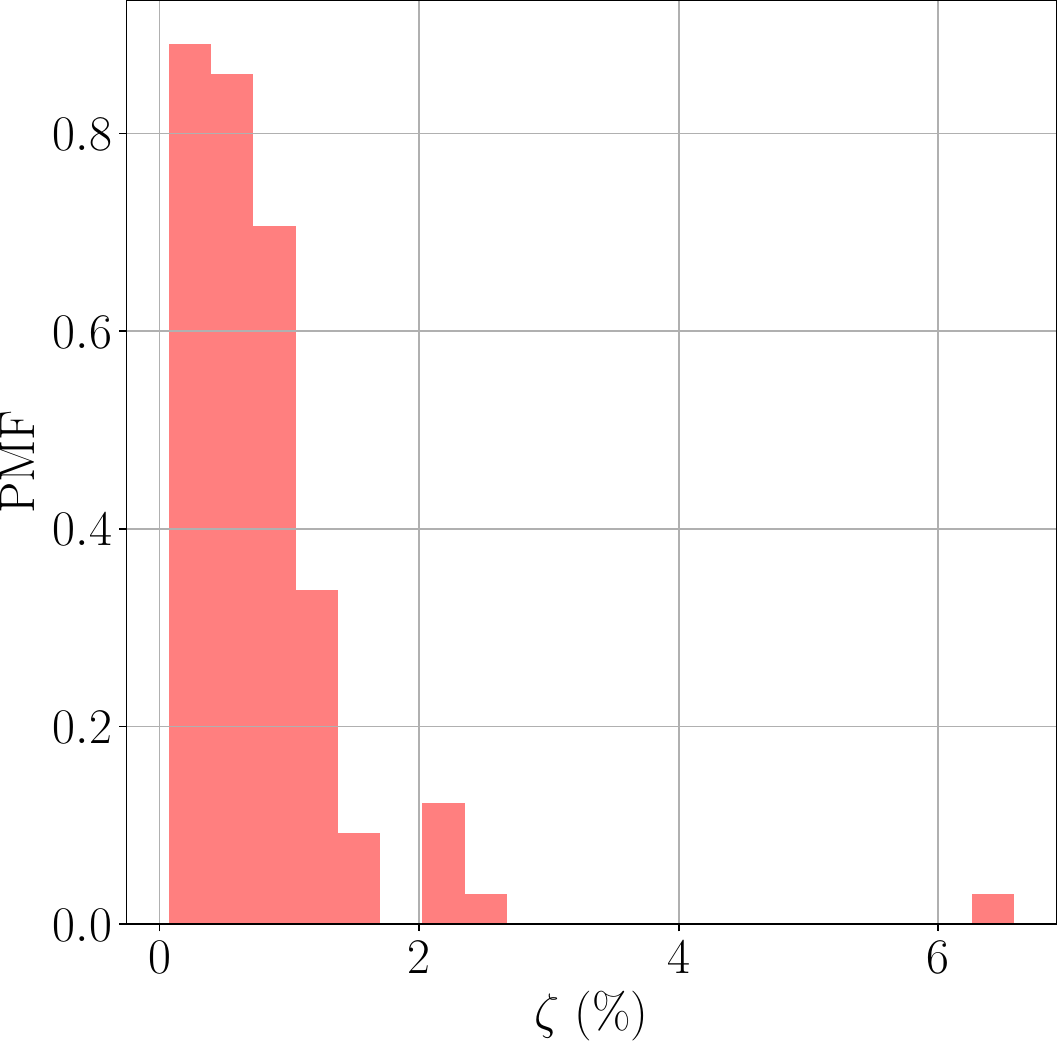}
         \caption{Histogram of $\zeta$.}
         \label{fig: lorenz-p1-error hist}
     \end{subfigure} \\
    
     \begin{subfigure}[b]{0.32\textwidth}
         \centering
         \includegraphics[scale=0.2]{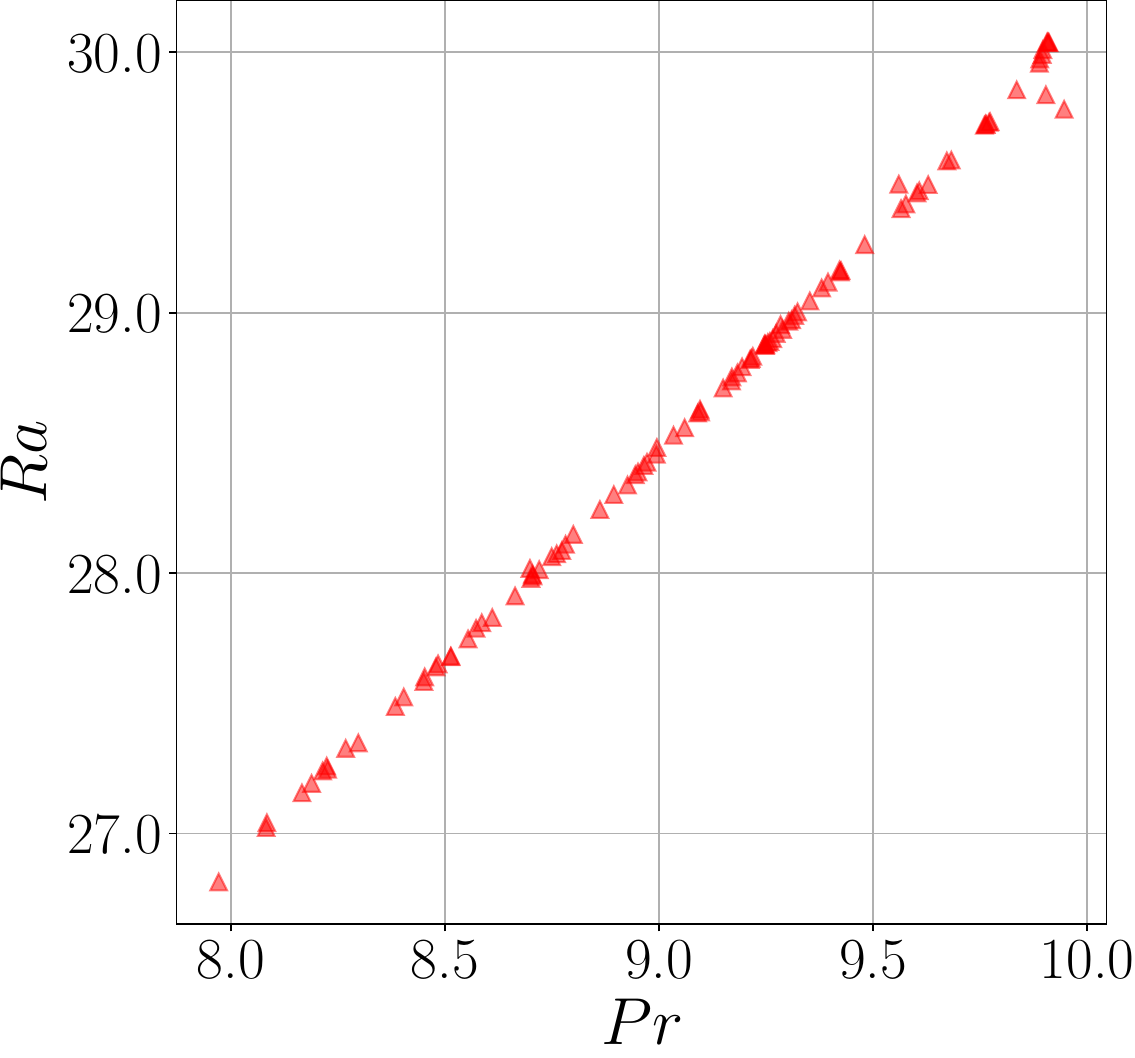}
         \caption{$Pr$-$Ra$ correlation.}
         \label{fig: lorenz-point1-cor1}
     \end{subfigure}
     \hfill
    \begin{subfigure}[b]{0.32\textwidth}
         \centering
        \includegraphics[scale=0.2]{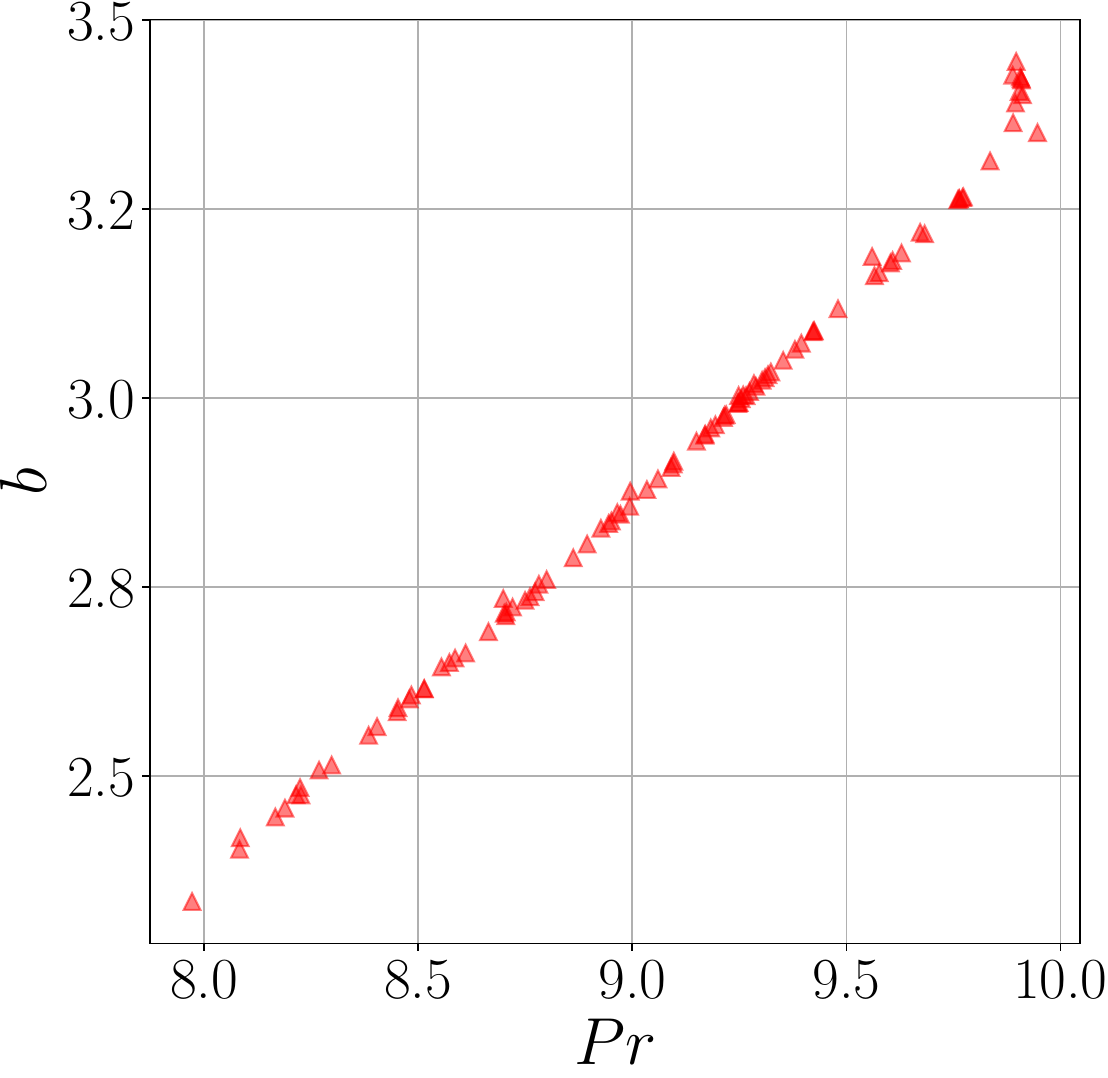}
         \caption{$Pr$-$b$ correlation.}
         \label{fig: lorenz-point1-cor2}
     \end{subfigure}
     \hfill
    \begin{subfigure}[b]{0.32\textwidth}
         \centering
        \includegraphics[scale=0.2]{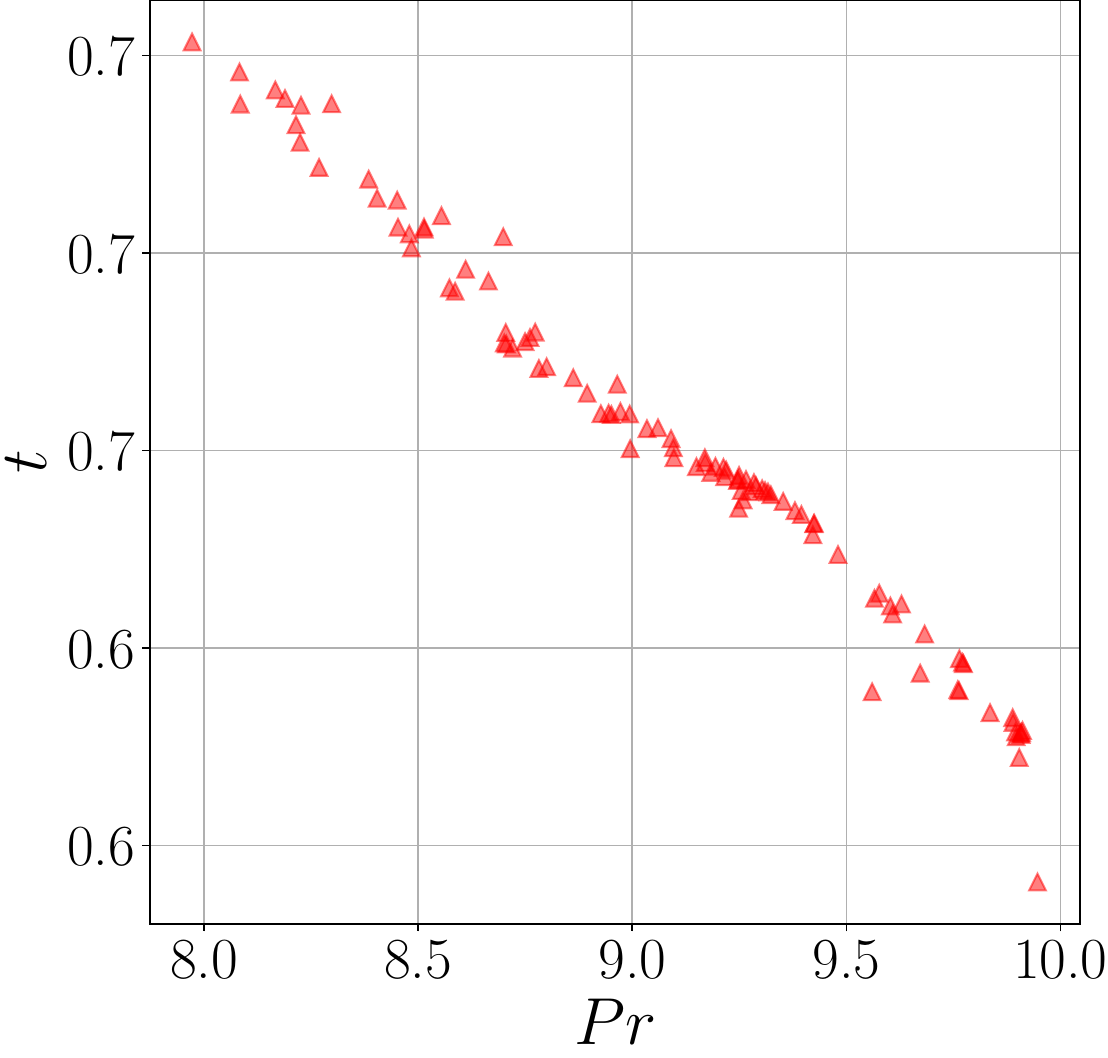}
         \caption{$Pr$-$t$ correlation.}
         \label{fig: lorenz-point1-cor3}
     \end{subfigure}
    \caption{Model inversion results for the parametric Lorenz system: fix $\boldsymbol{y}^* = [-3.4723, -8.9758, 26.2026]^T$ and sample $\boldsymbol{w}$ from a 6-dimensional standard Gaussian distribution. Phase plots, error measure and learned correlations between $Pr$ and the other three parameters.}
    \label{fig: Lorenz-inverse-phase-error-correlations-point1}
\end{figure}

For brevity, we only plot the best results in Figure~\ref{fig: lorenz inverse, point 1} and Figure~\ref{fig: Lorenz-inverse-phase-error-correlations-point1}, obtained using a six-dimensional latent space. To evaluate how close the decoded parameters led to trajectories ending at the desired coordinates, we introduce a relative error measure $\zeta$ defined as
\begin{equation}\label{equ: error measure}
\zeta = \frac{\|\boldsymbol{y}^* - \widehat{\boldsymbol{y}}^*\|_2}{\|\boldsymbol{y}^*\|_2},
\end{equation}
where $\widehat{\boldsymbol{y}}^*$ is the RK4 numerical solution based on the inverse prediction $\widehat{\boldsymbol{v}}$. 
A histogram of $\zeta$ generated from all 100 inverse predictions is presented in Figure~\ref{fig: lorenz-p1-error hist}, which reveals that almost all of these predictions yield a relative error less than 2\%. 

We also plot the learned correlations between the Prandtl number $Pr$ and the other three input parameters in~\Cref{fig: lorenz-point1-cor1,fig: lorenz-point1-cor2,fig: lorenz-point1-cor3}. 
The correlation plots presented here depict the projections from the learned 4D non-identifiable manifold $\boldsymbol{\mathcal{M}}_{\bv} \subset \boldsymbol{\mathcal{V}}$, associated with our fixed $\boldsymbol{y}^*$, onto 2D planes. 
From these plots, we notice a positive correlation between $Pr$-$Ra$ and $Pr$-$b$, and negative for $Pr$-$t$.
\begin{figure}[ht!]
    \centering
    \includegraphics[scale=0.22]{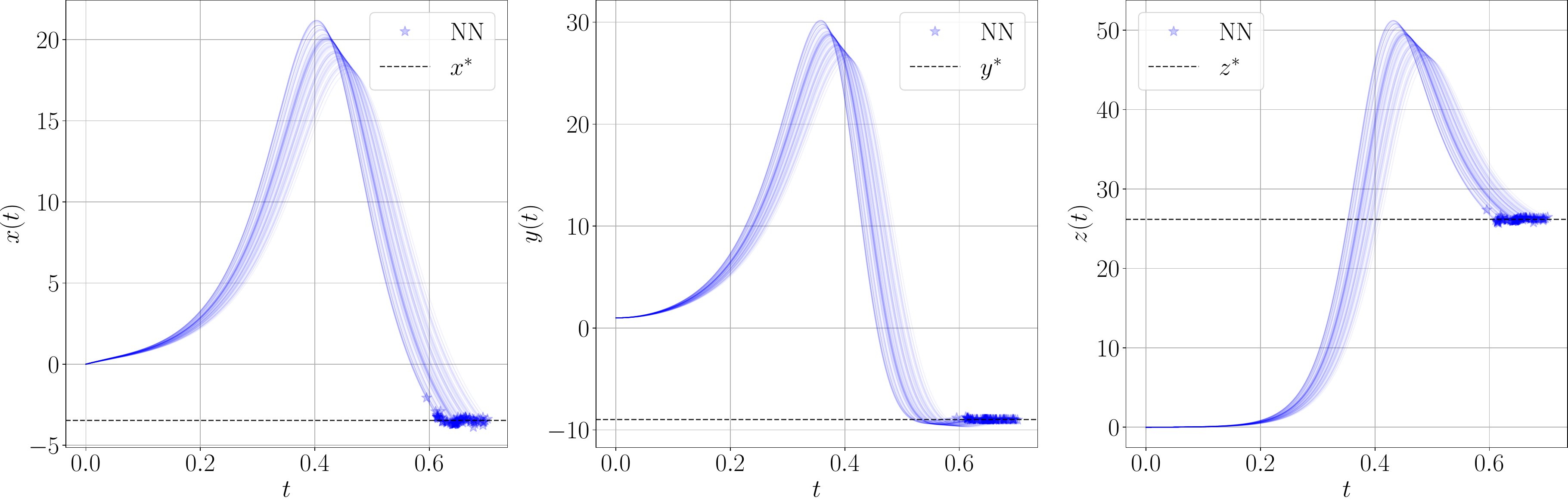}
    \caption{Model inversion results for the parametric Lorenz system. We fix $\boldsymbol{y}^* = [-3.4723, -8.9758, 26.2026]^T$ and sample $\boldsymbol{w}$ from a 6-dimensional standard Gaussian. Each RK4 solution trajectory generated from the decoded inputs $\widehat{\boldsymbol{v}}$ is plotted (NN) and a marker is added to the point computed at $t = \widehat{t}$.}
    \label{fig: lorenz inverse, point 1}
\end{figure}

For a fixed output $\boldsymbol{y}^*$, direct sampling of the latent variable $\boldsymbol{w}$ from a standard Gaussian performs relatively well. Therefore, we omit utilizing other sampling approaches mentioned in Section~\ref{sec:analysis_sampling} to refine the inverse prediction.
However, sampling from a standard normal behaves poorly if we significantly increase the complexity of the inverse problem by selecting $\boldsymbol{y}^* = [-11.5224, -10.3361, 30.7660]^T$, a state that can be revisited by a single system multiple times.

\begin{figure}[ht!]
     \centering
     \begin{subfigure}[b]{0.245\textwidth}
         \centering
         \includegraphics[scale=0.18]{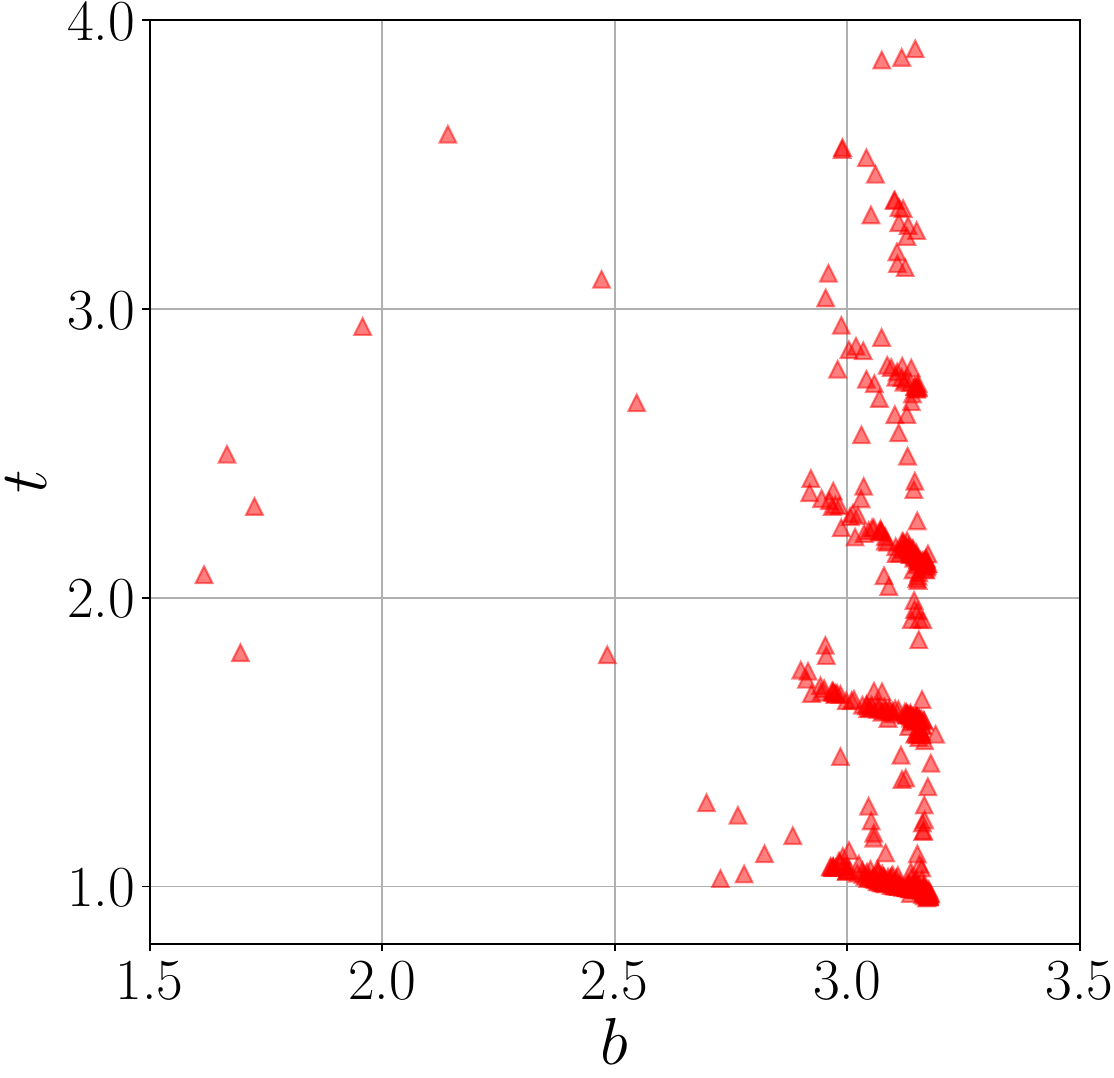}\caption{$\mathcal{N}(\boldsymbol{0}, \mathbf{I})$ sampling.}
         \label{fig: lorenz-bt-n01}
     \end{subfigure}
     \hfill
      \begin{subfigure}[b]{0.245\textwidth}
         \centering
         \includegraphics[scale=0.18]{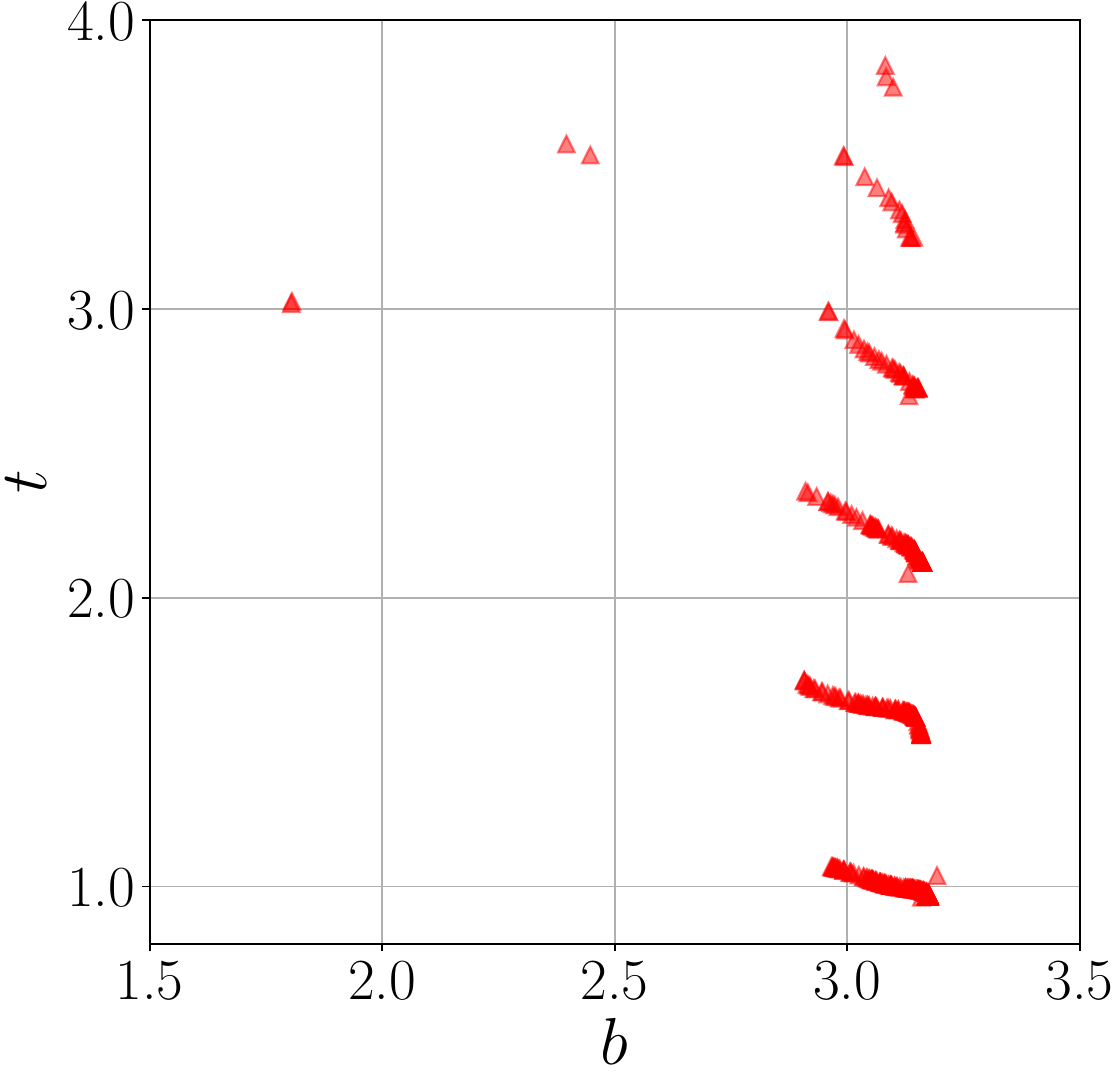}\caption{PC sampling.}
     \end{subfigure}
      \hfill
      \begin{subfigure}[b]{0.245\textwidth}
         \centering
         \includegraphics[scale=0.18]{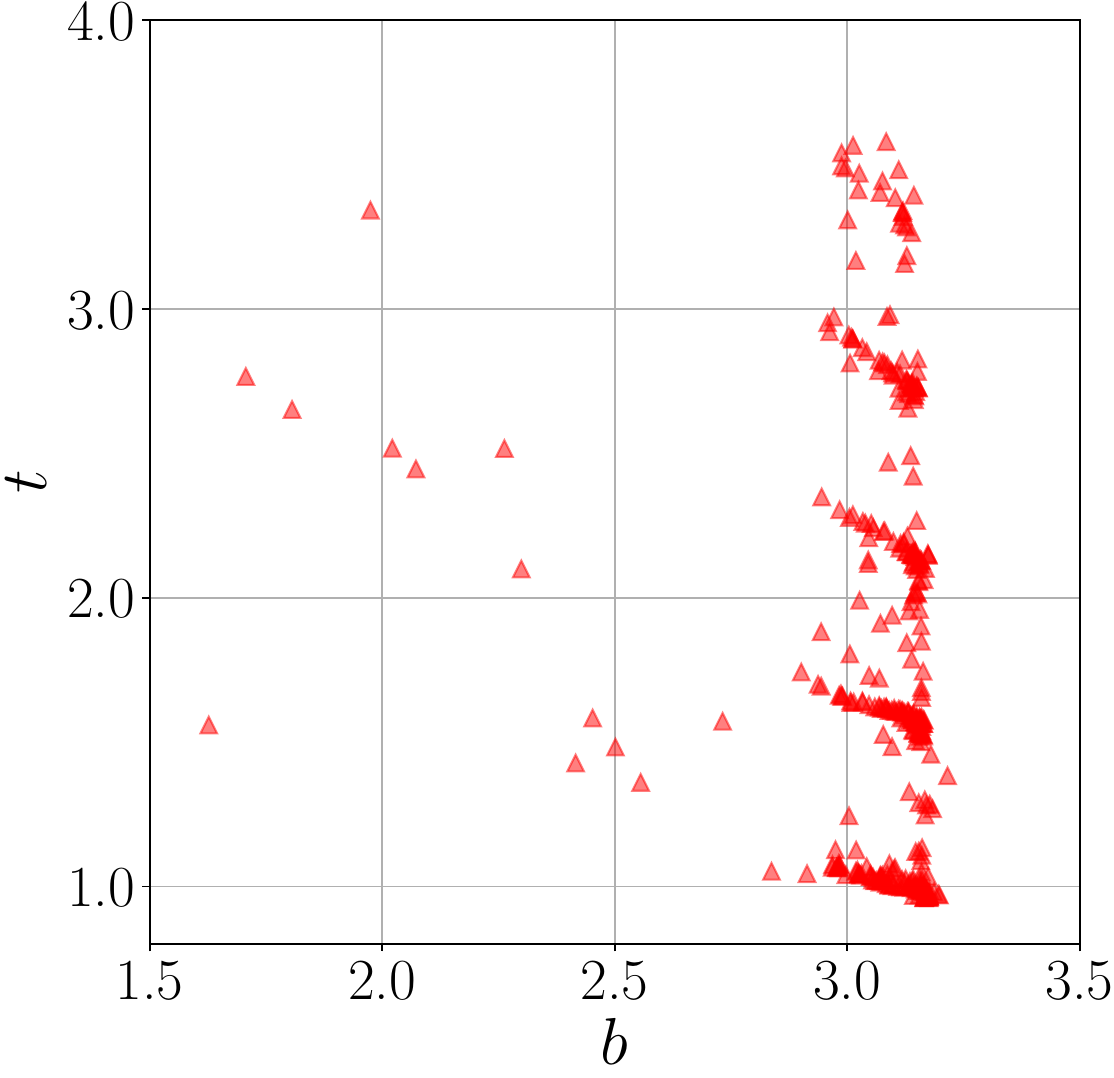}\caption{NF sampling.}
         \label{fig: bt-NF}
     \end{subfigure}
     \hfill
      \begin{subfigure}[b]{0.245\textwidth}
         \centering
         \includegraphics[scale=0.18]{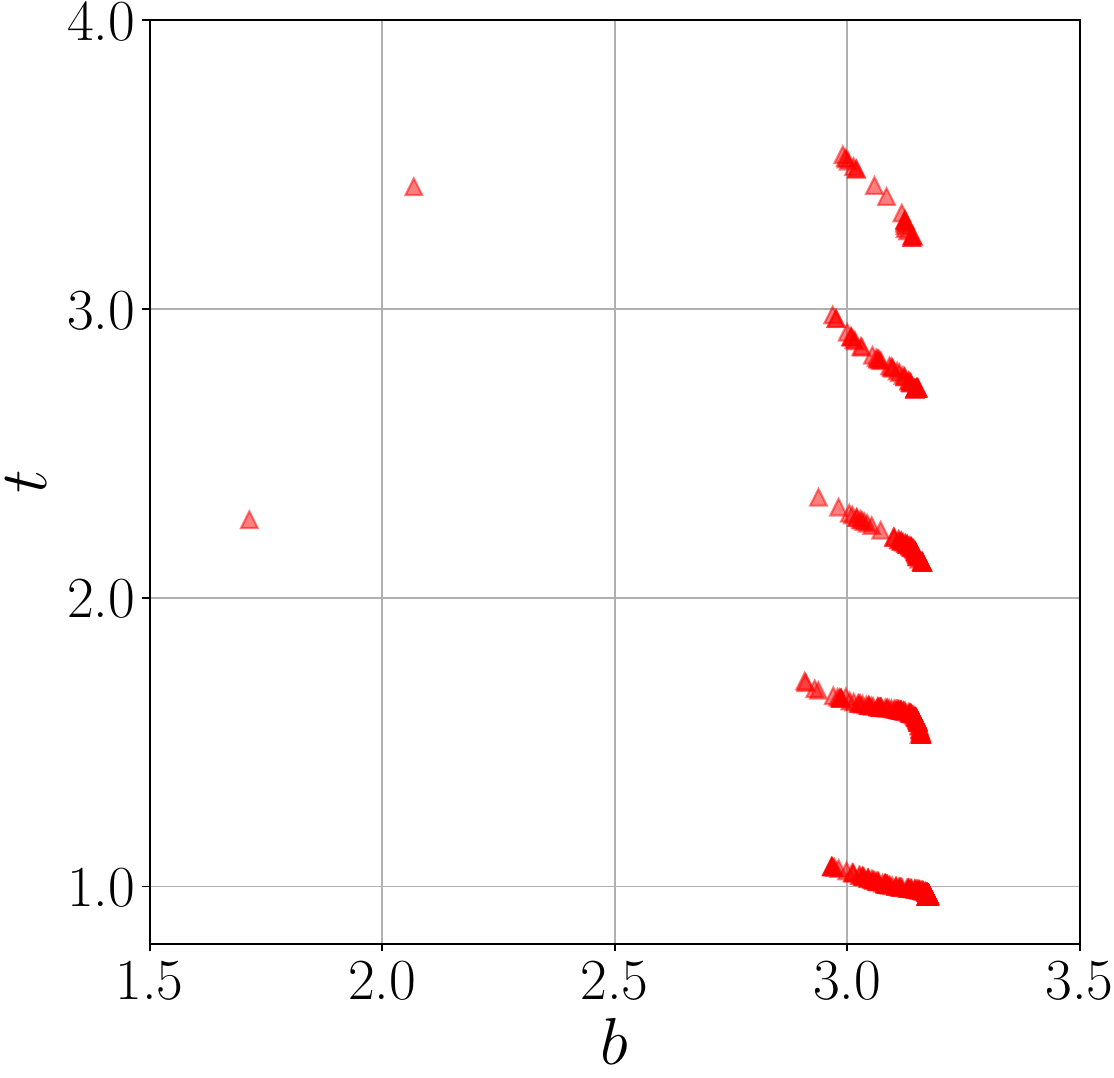}\caption{NF+PC.}
         \label{fig: bt-NFPC}
     \end{subfigure}
     \caption{Model inversion results for the parametric Lorenz system: fix $\boldsymbol{y}^* = [-11.5224, -10.3361, 30.7660]^T$. Comparing $b$-$t$ correlations using different latent variable sampling schemes. Sample size: 400, 6D latent space. Methods: sampling from $\mathcal{N}(\boldsymbol{0}, \mathbf{I})$, PC sampling ($R=4$), NF sampling (see Table~\ref{table: lorenz-nf}), combined NF+PC sampling (apply PC sampling to inverse predictions $\widehat{\boldsymbol{v}}$ associated with Figure~\ref{fig: bt-NF}, $R=4$). }
     \label{fig: lorenz-p2-bt}
\end{figure}

Like the periodic wave example studied in Section~\ref{example: non-linear}, the non-identifiable manifold induced by the new fixed $\boldsymbol{y}^*$ contains outliers. To illustrate this, we first show the learned correlations between the geometric factor $b$ and time $t$ in Figure~\ref{fig: lorenz-p2-bt}, where the latent variables are generated using the sampling methods discussed in Section~\ref{sec:analysis_sampling}.

The results in Figure~\ref{fig: lorenz-p2-bt} suggest that the values of $(b,t)$ for which the Lorenz systems pass through the selected spatial location consist of five disjoint regions. This is likely due to the presence of five instances in time where the given $\boldsymbol{y}^*$ can be reached within 4 seconds under a specific parameter combination, i.e. $[Pr, Ra, b]$. 
Again, as shown in Section~\ref{example: non-linear}, it is challenging for our inVAErt network to deform a continuous Gaussian latent space to disconnected sets that represent the non-identifiable manifold. 
Consequently, despite the concentration of samples around the correct locations, the classical VAE sampling scheme is still contaminated by numerous outliers (see Figure~\ref{fig: lorenz-bt-n01}). 
These outliers can be effectively reduced using PC sampling, or by combining PC and NF sampling (see Figure~\ref{fig: bt-NFPC}). Besides, it is not surprising that NF sampling remains susceptible to outliers, as we observed the posterior $q(\boldsymbol{w}|\boldsymbol{v})$ in this case is almost the standard normal distribution. In other words, the outliers may not come from the previously mentioned low posterior density regions.

\begin{figure}[ht!]
    \begin{subfigure}[b]{0.495\textwidth}
         \centering
        \includegraphics[scale=0.28]{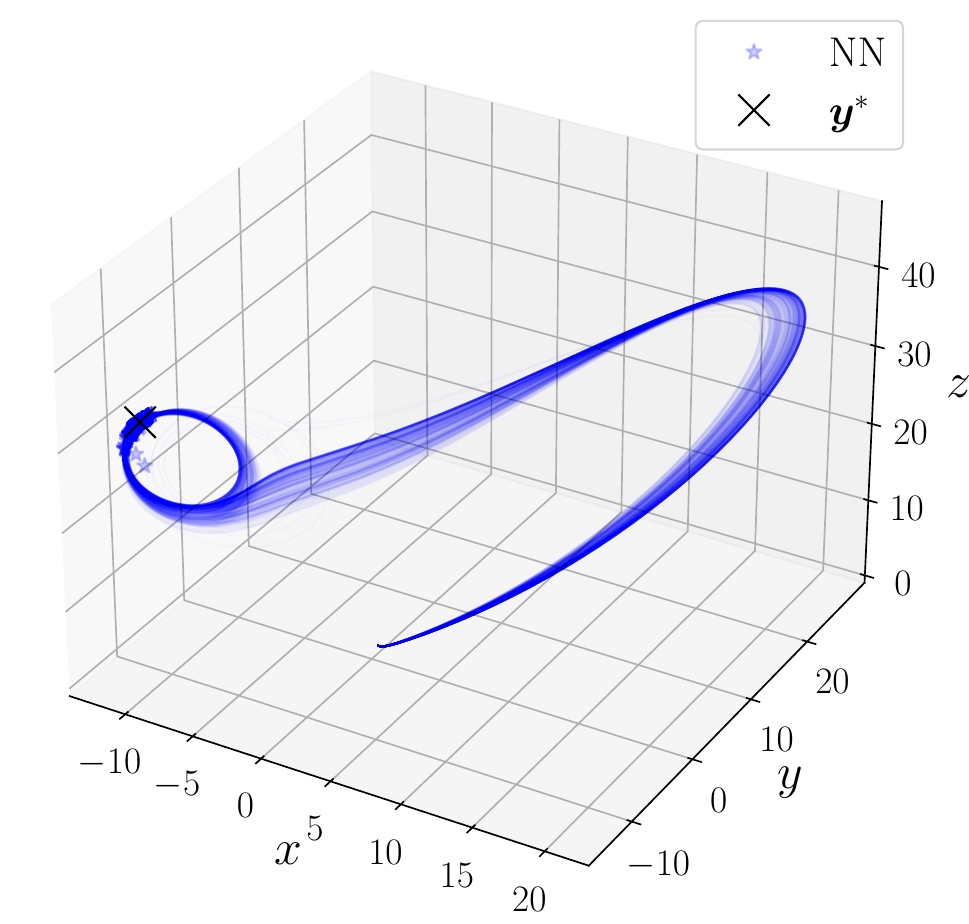}
         \caption{Phase plot trajectories generated from $\widehat{\boldsymbol{v}}$.}
     \end{subfigure}
     \hfill
     \begin{subfigure}[b]{0.495\textwidth}
         \centering
         \includegraphics[scale=0.22]{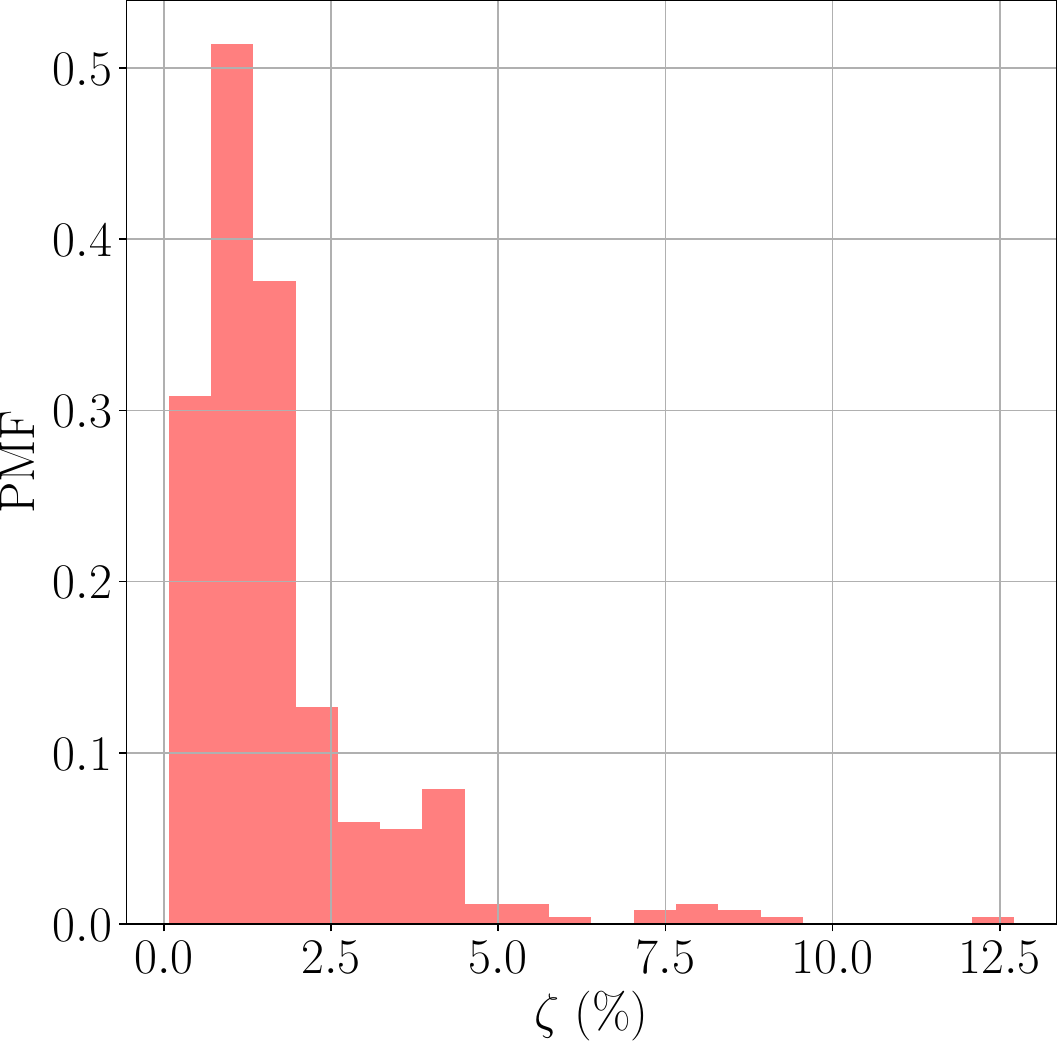}
         \caption{Histogram of $\zeta$.}
         \label{fig: lorenz-p2-error hist}
     \end{subfigure} \\
     
     \begin{subfigure}[b]{0.99\textwidth}
         \centering
         \includegraphics[scale=0.2]{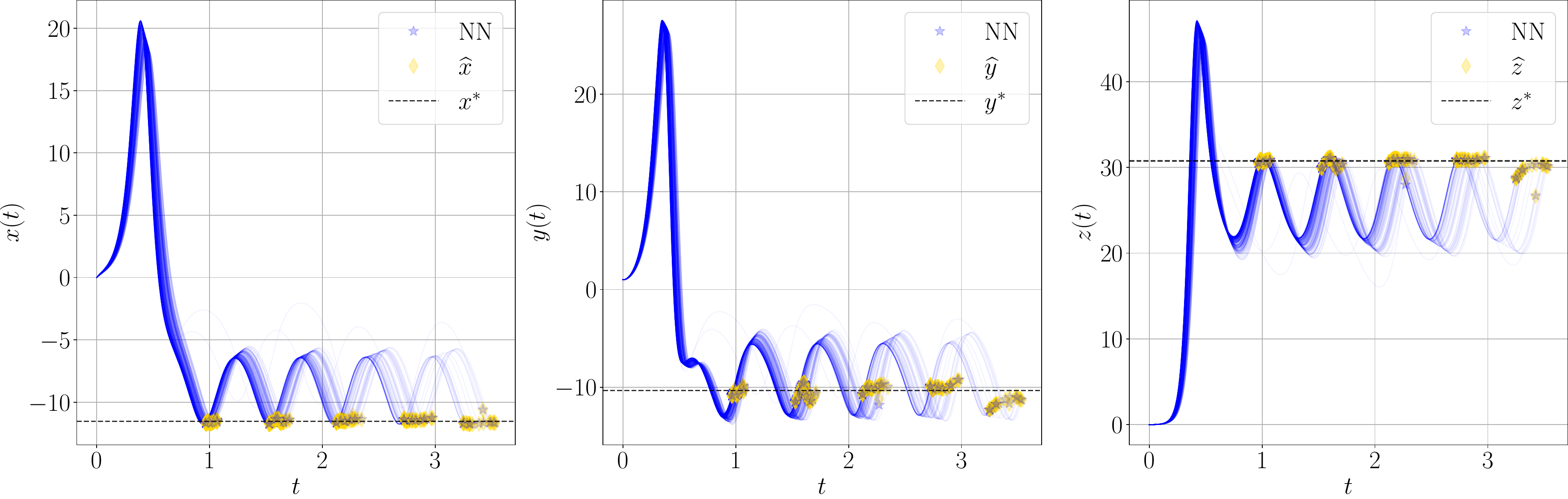}
     \end{subfigure}
     \caption{Model inversion results for the parametric Lorenz system: fix $\boldsymbol{y}^* = [-11.5224, -10.3361, 30.7660]^T$. NF sampling (see Table~\ref{table: lorenz-nf}) is applied to generate 400 samples of $\boldsymbol{w}$, followed by PC sampling ($R=4$) for denoising. Results corresponding to inversion samples of $\widehat{\boldsymbol{v}}$ associated with Figure~\ref{fig: bt-NFPC}. Each RK4 solution trajectory from $\widehat{\boldsymbol{v}}$ is plotted (NN) and the end point is highlighted that corresponds to $t = \widehat{t}$. Endpoint predictions from the trained emulator $\mathscr{N}_e$ is also plotted, denoted as $\widehat{\boldsymbol{y}} = [\widehat{x},\widehat{y},\widehat{z}]^T$. }
     \label{fig: lorenz-p2-verify}
\end{figure}

Next, we verify our inverse prediction by numerically integrating the Lorenz system using the samples of $\widehat{\boldsymbol{v}}$ associated with Figure~\ref{fig: bt-NFPC}, obtaining a relative error $\zeta$ smaller than 5\% in most cases (see Figure~\ref{fig: lorenz-p2-error hist}).
Finally, we would like to point out that we can also use the emulator $\mathscr{N}_e$ learnt by the inVAErt network to verify whether a prediction $\widehat{\boldsymbol{v}}$ suggested by the decoder results in an acceptable error $\zeta$, and deploy a rejection sampling approach to complement those discussed in Section~\ref{sec:analysis_sampling}. To verify the applicability of this approach, we forward $\mathscr{N}_e$ with all 400 samples of $\widehat{\boldsymbol{v}}$ produced by the decoder $\mathscr{N}_d$ and plot the endpoints, i.e. $t = \widehat{t}$, for each solution component in Figure~\ref{fig: lorenz-p2-verify}. 
These predictions from the emulator, denoted as $\widehat{\boldsymbol{y}}$, match well with the RK4 solutions, and thus may replace $\widehat{\boldsymbol{y}}^*$ in equation~\eqref{equ: error measure}, providing accurate approximations for $\zeta$.

%============================================================
\subsection{Reaction-diffusion PDE system}\label{example:pde}
%============================================================

Finally, we consider applications to space- and time-dependent PDEs. To prepare the dataset, we utilize the reaction-diffusion solver from the scientific machine learning benchmark repository PDEB{\scriptsize{ENCH}}~\cite{PDEBench2022}. 
The reaction-diffusion equations~\eqref{equ: pde-DR} model the evolution in space and time of two chemical concentrations $\boldsymbol{c}(\boldsymbol{x},t) = [c_1(\boldsymbol{x},t), c_2(\boldsymbol{x},t)]^T$ in the domain $\Omega=[-1,1]^2$, 
\begin{equation}
    \begin{cases}
       \partial \boldsymbol{c}/\partial t = \mathbf{D} \Delta \boldsymbol{c} + \mathbf{R}(\boldsymbol{c}) & \textrm{in} \ \Omega \times (0,T],\\
       \partial \boldsymbol{c} /\partial \boldsymbol{x} =\boldsymbol{0} & \textrm{on} \ \partial\Omega \times (0,T],\\
       \boldsymbol{c}(\boldsymbol{x}, 0) = \boldsymbol{c}_0(\boldsymbol{x}) & \textrm{in} \ \Omega \ , \textrm{at} \ t = 0,
    \end{cases}\,\,\text{where}\,\,
    \mathbf{D} = 
    \begin{bmatrix} 
    D_1 & 0 \\ 
    0 & D_2 \\ 
    \end{bmatrix},    
    \label{equ: pde-DR}
\end{equation}
where the nonlinear reaction function $\mathbf{R}(\boldsymbol{c})$ follows the Fitzhugh-Nagumo model~\cite{PDEBench2022} with reaction constant $\kappa$, and is expressed as
\begin{equation}
    \mathbf{R}(\boldsymbol{c}) = 
    \begin{bmatrix}
        c_1 - c_1^3 - \kappa - c_2 \\ 
        c_1 - c_2
    \end{bmatrix}.        
    \label{equ: pde-reaction-model}
\end{equation}

PDEB{\scriptsize{ENCH}}~\cite{PDEBench2022} utilizes first order finite volumes (FV) in space and a 4-th order Runge-Kutta integrator in time to solve the above reaction-diffusion system, and we set $\boldsymbol{c}_0(\boldsymbol{x}) \sim \mathcal{N}([2,2]^T, \mathbf{I})$, $\forall  \boldsymbol{x} \in \Omega$ for initializations. Additional details on the simulations considered in this section are reported in Table~\ref{table:dr paras}. 
\begin{table}[ht!]
{\footnotesize
\begin{center}
\begin{tabular}{@{} l l @{}}
\toprule
Spatial discretization & $32 \times 32$ \\
Time step size  ($\Delta t$) & 0.005\\
Largest possible simulation time ($T_f$) & 5 \\
Diffusivity for $c_1$ ($D_1$) &  [$2\times10^{-3}, 5\times10^{-3}$] \\ 
Diffusivity for $c_2$ ($D_2$) &  [$2\times10^{-3}, 5\times10^{-3}$] \\ 
Reaction constant ($\kappa$) &   [$2\times10^{-3}, 5\times10^{-3}$] \\
\bottomrule
\end{tabular}
\end{center}}
\caption{Simulation parameters of 2D reaction-diffusion system.}
\label{table:dr paras}
\end{table}

\begin{figure}[ht!]
    \centering
    \includegraphics[scale=0.2]{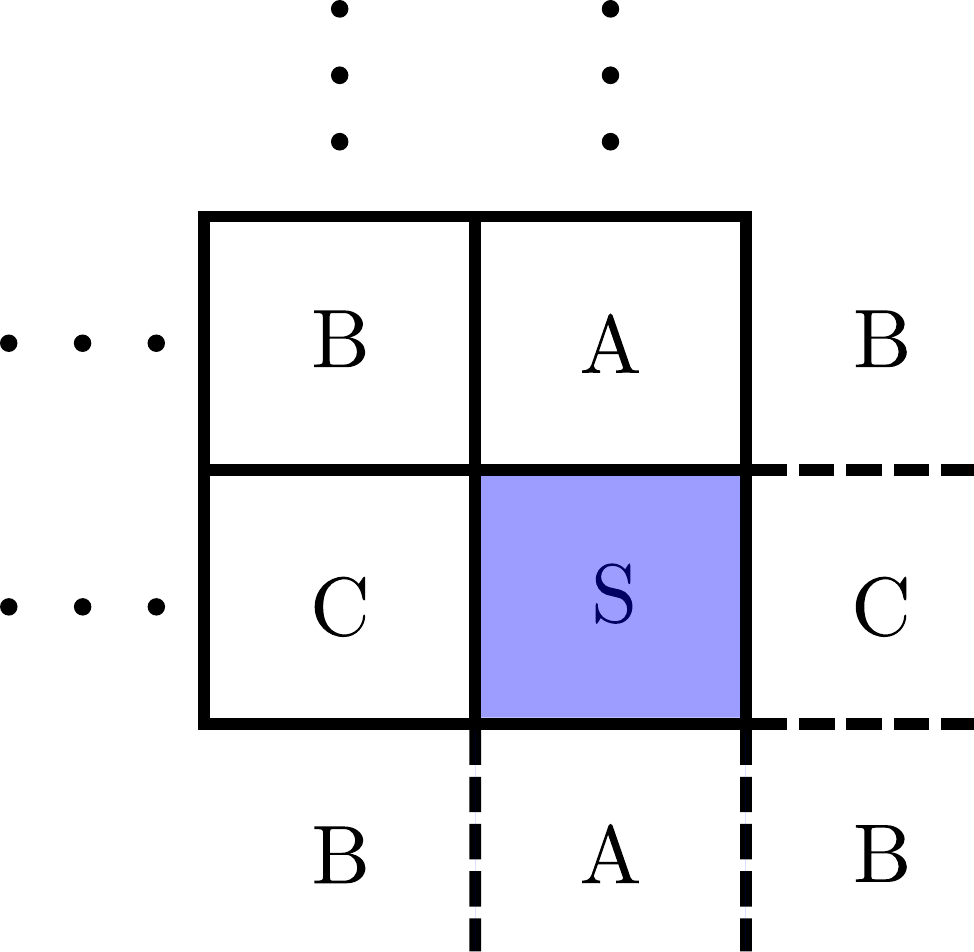}
    \caption{Symmetry condition used to gather auxiliary data $\mathcal{D}_{\boldsymbol{v}}$ at the boundary. Cell ``$\textrm{S}$" here is at the bottom right corner of a two-dimensional discretized domain.} 
    \label{fig: GNN symm}
\end{figure}

The first-order finite volume method assumes constant solution within each grid cell, with space location identified through cell-center coordinates $\boldsymbol{x} = (x,y)$.
We then consider a forward process $\mathcal{F}$ of the form
\begin{equation}
    [c_1(x,y,t), c_2(x,y,t)]^T = \mathcal{F}(\boldsymbol{v}, \mathcal{D}_{\boldsymbol{v}}) \ ,
    \label{equ:RD-map}
\end{equation}
with dependent parameters $\boldsymbol{v} = [D_1, D_2, \kappa, t, x, y]^T$, and auxiliary data defined as
\[
\mathcal{D}_{\boldsymbol{v}} = \{ \boldsymbol{c}(x \pm \Delta x,y \pm \Delta y , t - \Delta t),\dots, \boldsymbol{c}(x,y, t - n_p\Delta t),\dots, \boldsymbol{c}(x,y, t - \Delta t) \},
\]
containing solutions at a given cell ``S" (see diagram in Figure~\ref{fig: GNN symm}) from the previous $n_p$ time steps and the solutions at its neighbors {\color{black} from the last step}.
Eight neighbors are considered for all cells by assuming the solution outside $\Omega$ is obtained by mirroring the solution within the domain (see Figure~\ref{fig: GNN symm}).  
To gather training data, we simulate the system using PDEB{\scriptsize{ENCH}}~\cite{PDEBench2022} for 5000 times with uniform random samples of $D_1, D_2$ and $\kappa$. 
From each simulation, we randomly pick 10 cells at 5 different time instances to effectively manage the amount of training samples ($2.5\times 10^5$ data points in total) and to mitigate over-fitting.

The emulator $\mathscr{N}_e$ is trained by a residual network as discussed in Section~\ref{sec:inVAErt_emulator}, equation~\eqref{equ: aux system-encoder-resnet}. We apply logarithmic transformations to the inputs $D_1, D_2, \kappa$ since their magnitudes are much smaller compared to $x, y, t, c_1, c_2$. For a detailed list of hyperparameters, the interested reader is referred to the Appendix.

\begin{figure}[ht!]
    \begin{subfigure}[b]{0.245\textwidth}
         \centering
         \includegraphics[scale=0.14]{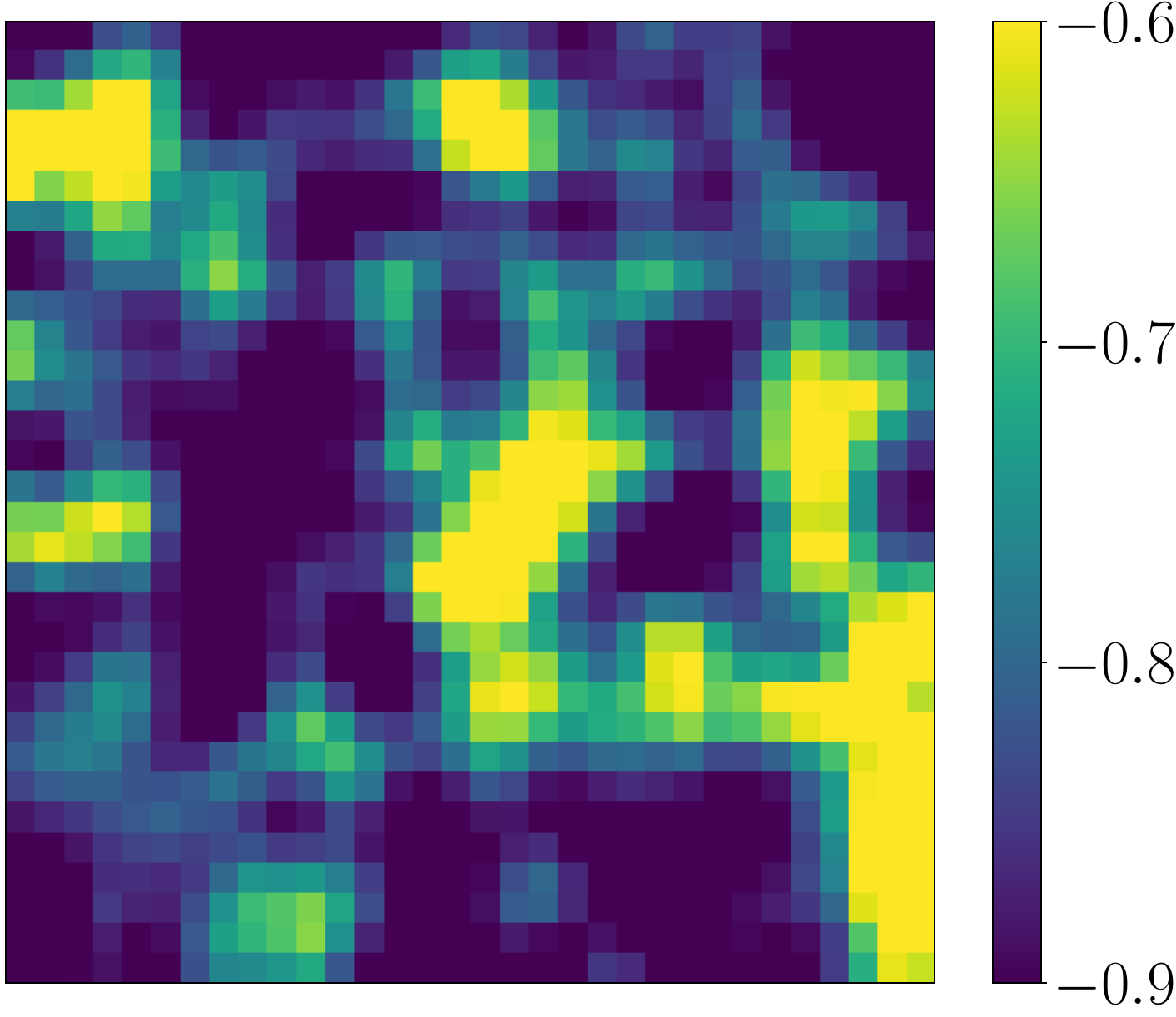}
         \caption{Exact: $c_1$.}
     \end{subfigure}
     \hfill
     \begin{subfigure}[b]{0.245\textwidth}
         \centering
         \includegraphics[scale=0.14]{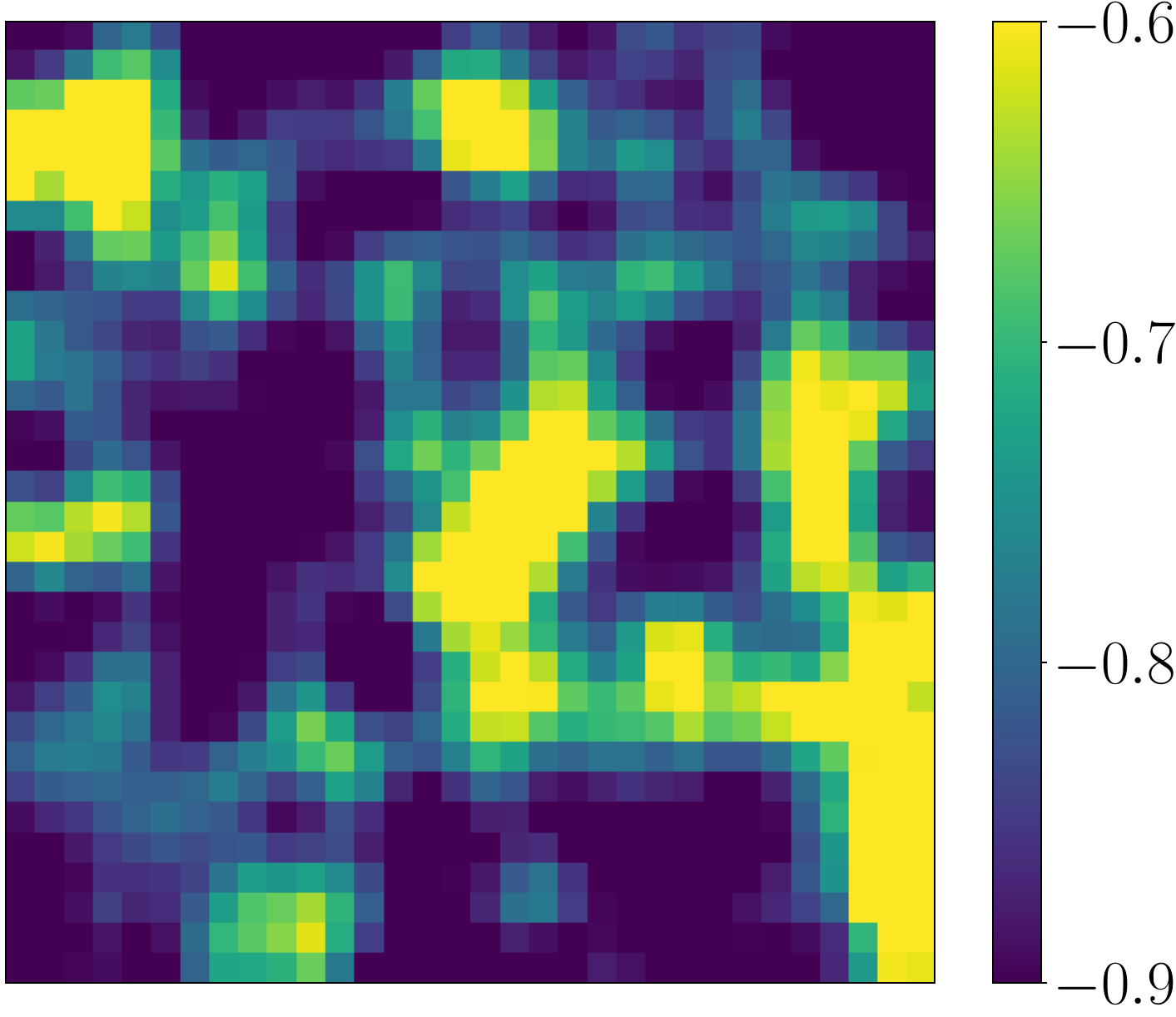}
         \caption{NN: $c_1$.}
     \end{subfigure}
     \hfill
        \begin{subfigure}[b]{0.245\textwidth}
         \centering
         \includegraphics[scale=0.14]{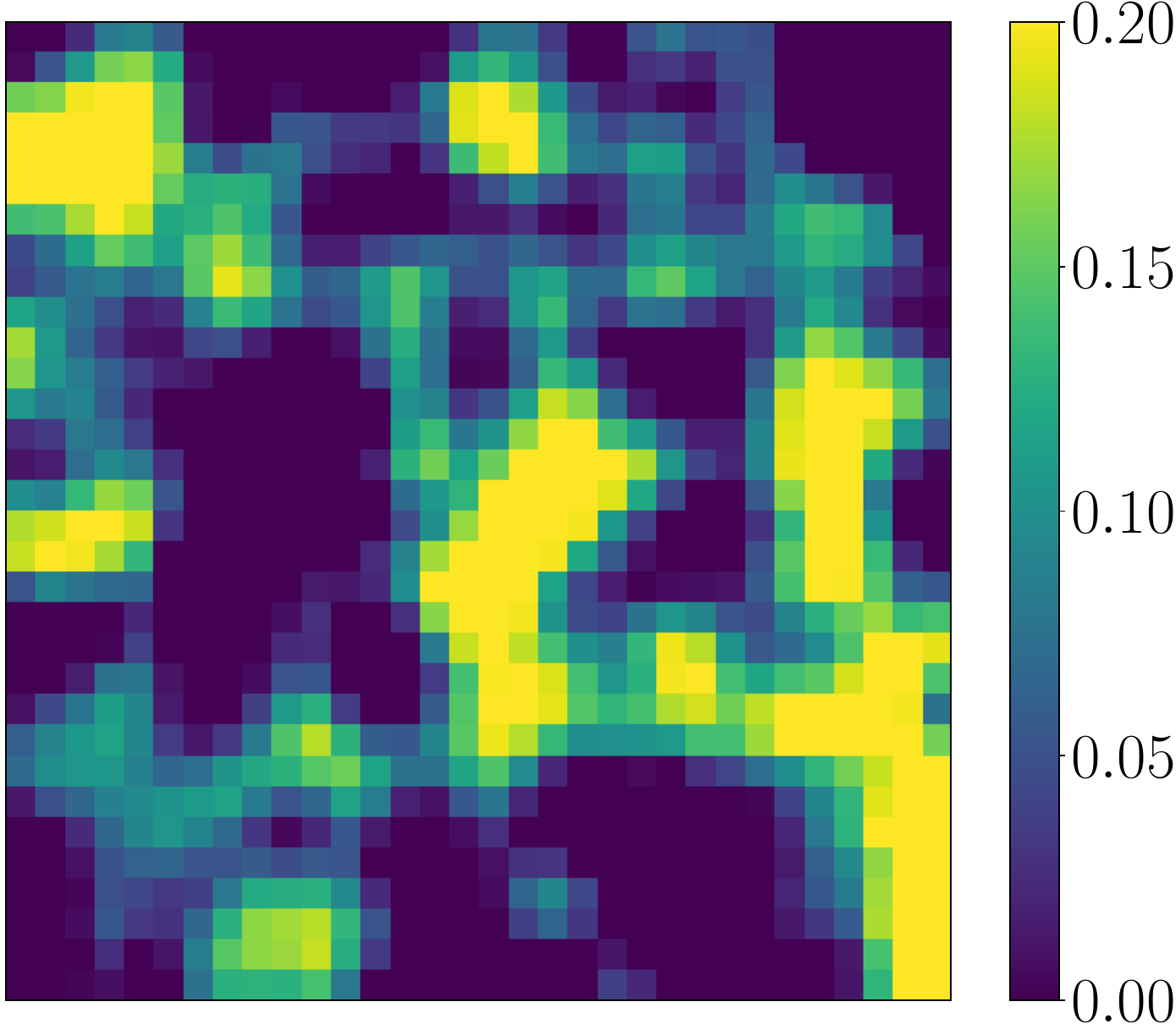}
         \caption{Exact: $c_2$.}
     \end{subfigure}
     \hfill
     \begin{subfigure}[b]{0.245\textwidth}
         \centering
         \includegraphics[scale=0.14]{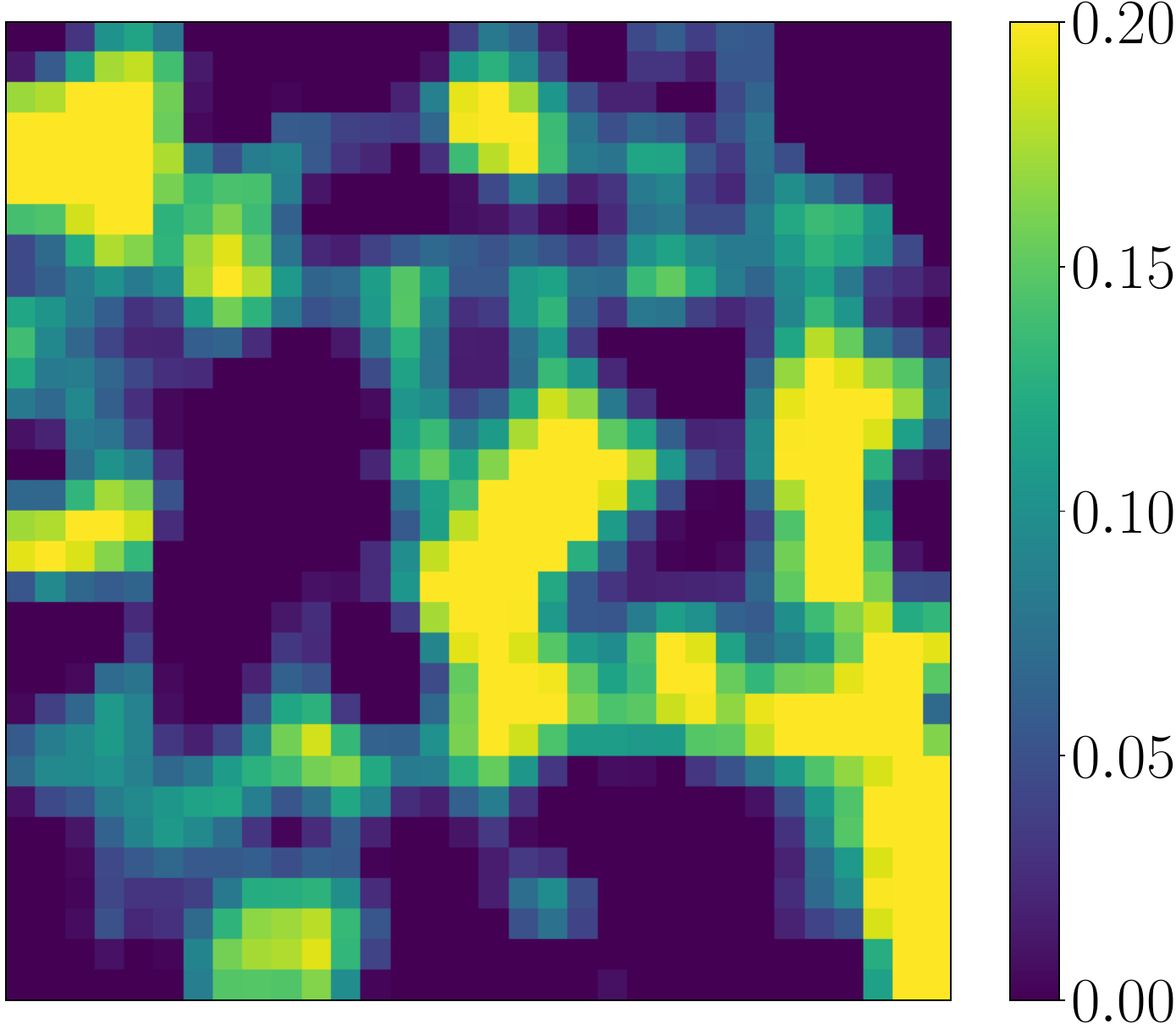}
         \caption{NN: $c_2$.}
     \end{subfigure}
    \\

     \begin{subfigure}[b]{0.245\textwidth}
         \centering
         \includegraphics[scale=0.14]{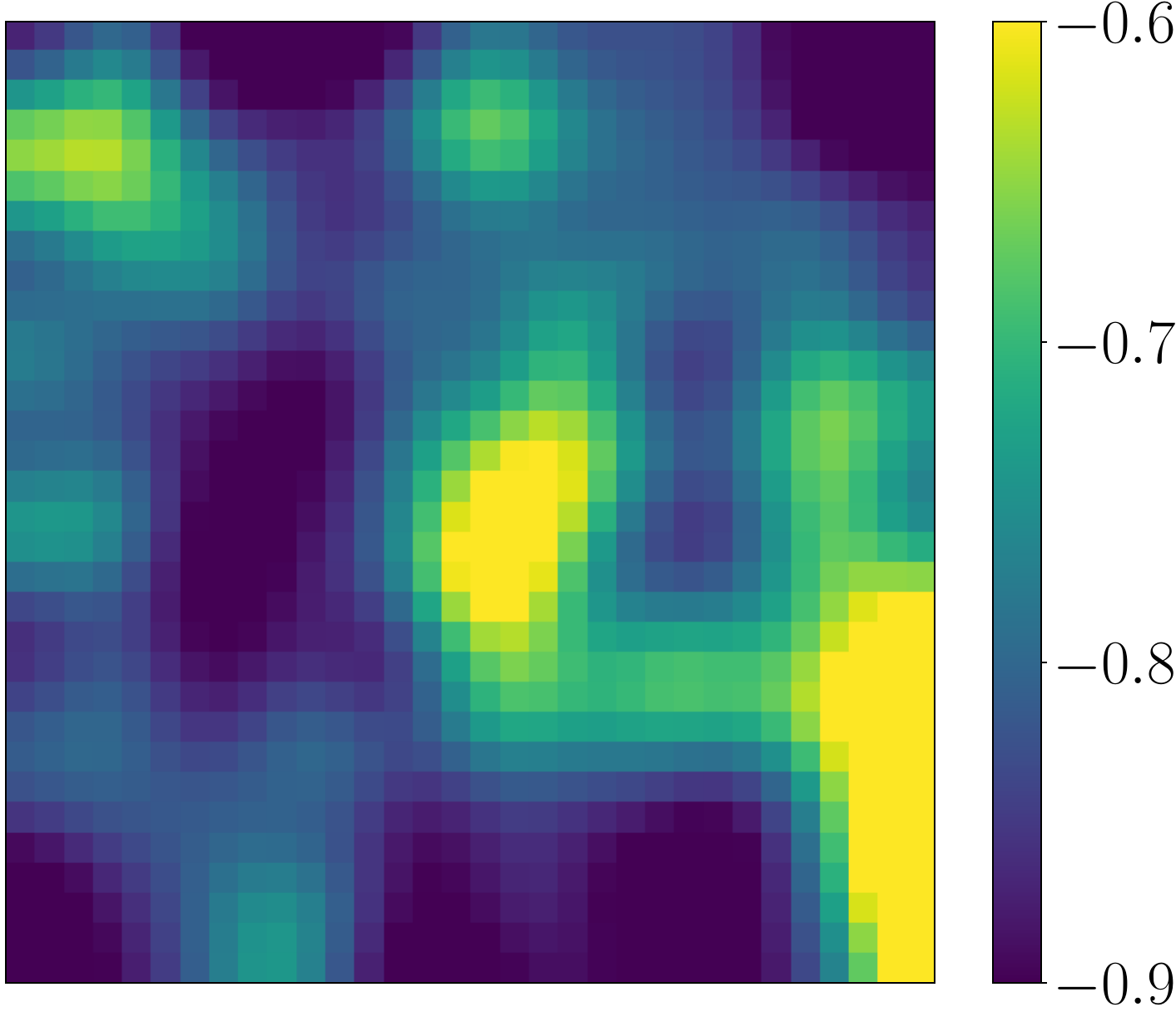}
         \caption{Exact: $c_1$.}
     \end{subfigure}
     \hfill
     \begin{subfigure}[b]{0.245\textwidth}
         \centering
         \includegraphics[scale=0.14]{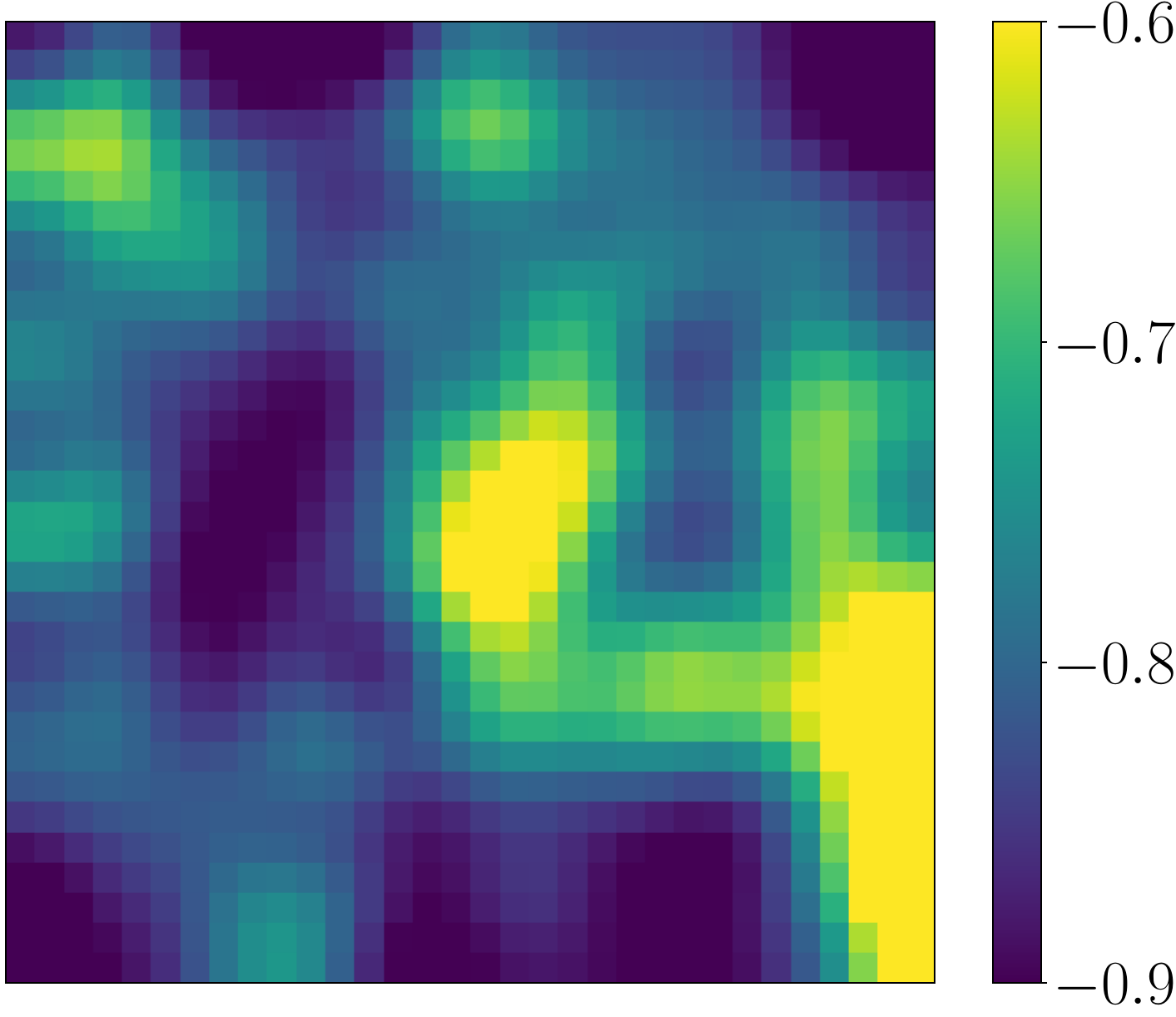}
         \caption{NN: $c_1$.}
     \end{subfigure}
     \hfill
        \begin{subfigure}[b]{0.245\textwidth}
         \centering
         \includegraphics[scale=0.14]{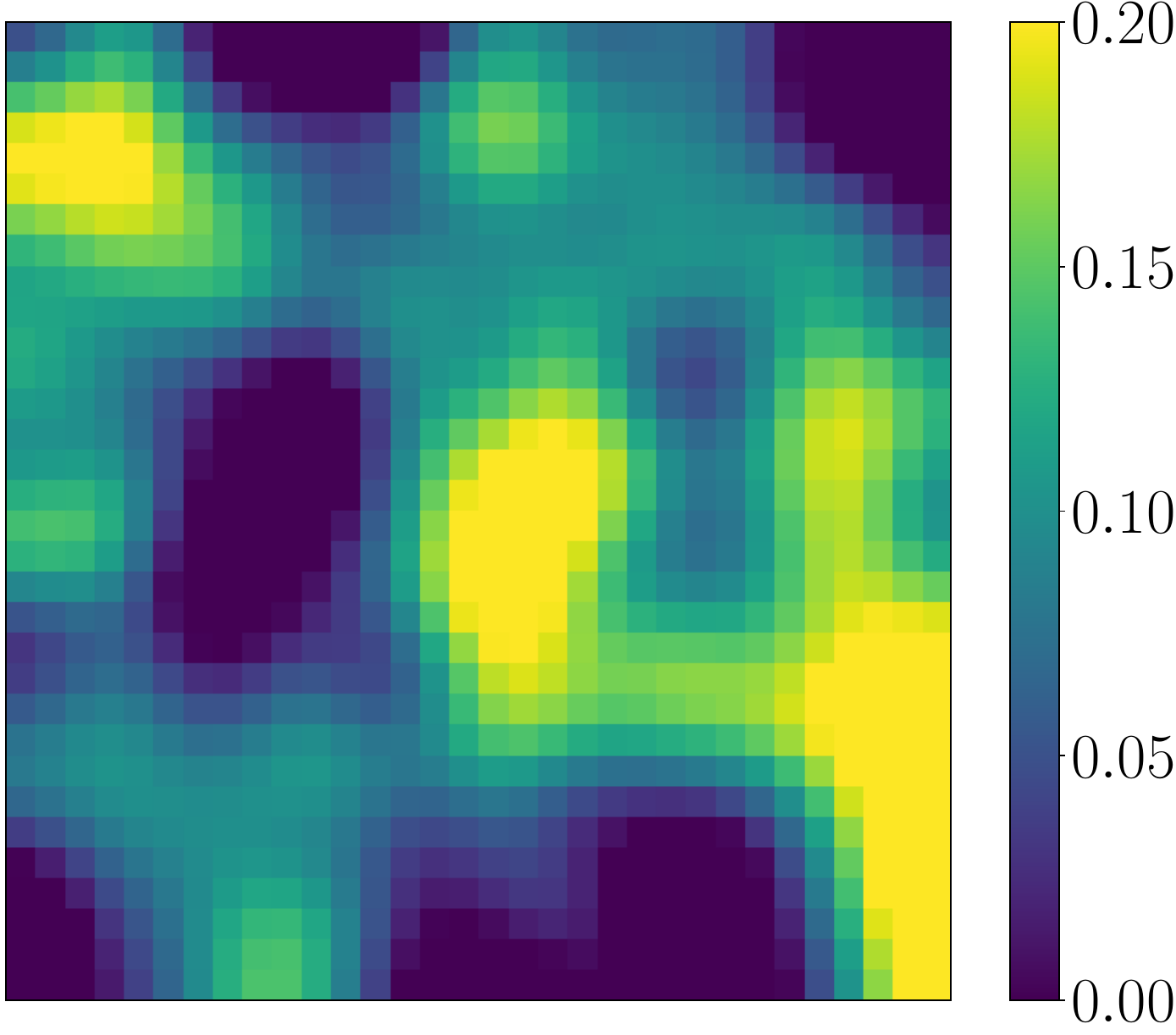}
         \caption{Exact: $c_2$.}
     \end{subfigure}
     \hfill
     \begin{subfigure}[b]{0.245\textwidth}
         \centering
         \includegraphics[scale=0.14]{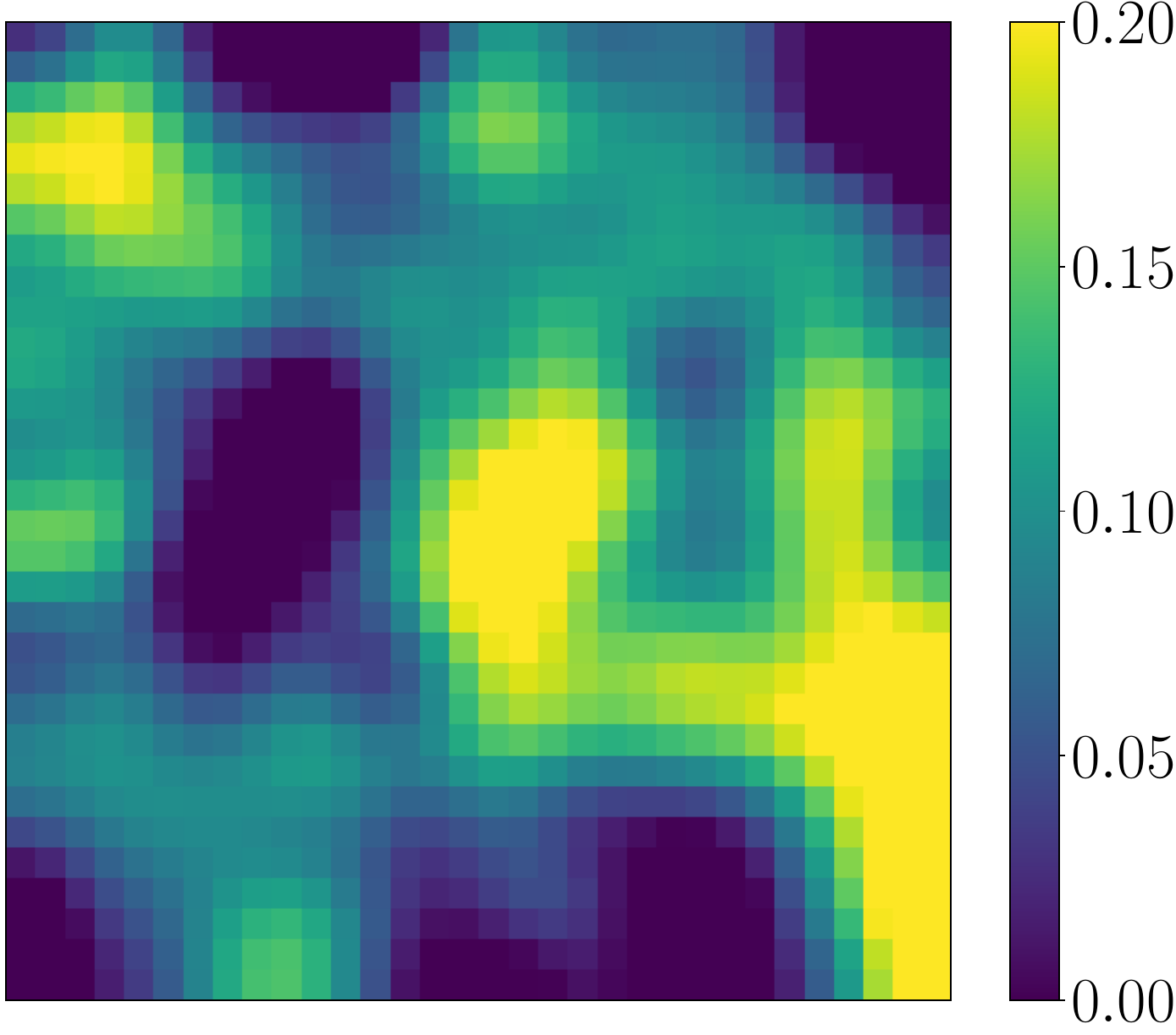}
         \caption{NN: $c_2$.}
     \end{subfigure}
     \caption{Comparison between the learned emulator dynamics (NN) at $T=2.0$ seconds and reference numerical solutions (from the PDEB{\scriptsize{ENCH}} package~\cite{PDEBench2022}) for the parametric reaction-diffusion system. Top: Low diffusion regime ($D_1=D_2 = 2\times 10^{-3}, \kappa = 3\times 10^{-3}$). Bottom: High diffusion regime ($D_1=D_2 = 5\times 10^{-3}, \kappa = 3\times 10^{-3}$). The coordinate of the left-bottom corner: (-1,-1).}
    \label{fig: RD-forward}
\end{figure}

\begin{figure}[ht!]
    \begin{subfigure}[b]{0.49\textwidth}
         \centering
         \includegraphics[scale=0.32]{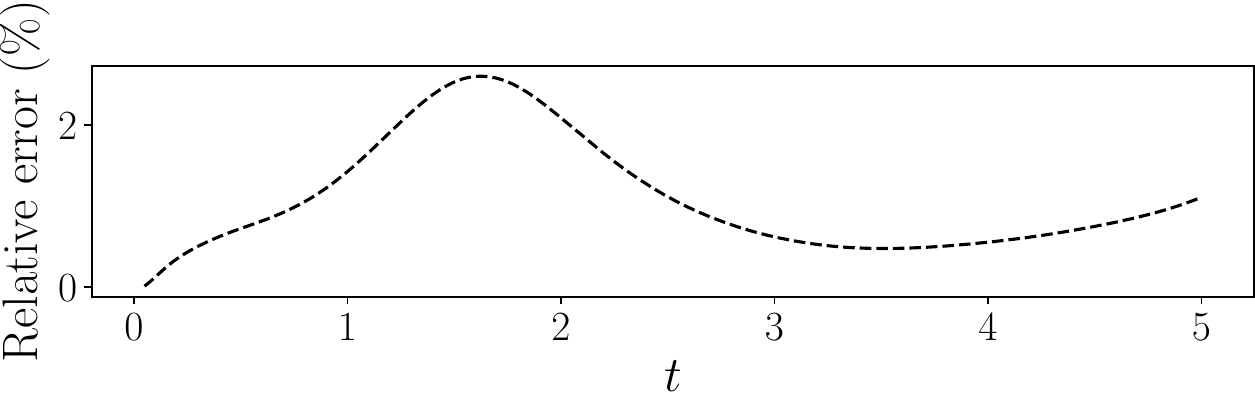}
         \caption{Error evolution of the low-diffusion system.}
     \end{subfigure}
     \hfill
     \begin{subfigure}[b]{0.49\textwidth}
         \centering
         \includegraphics[scale=0.32]{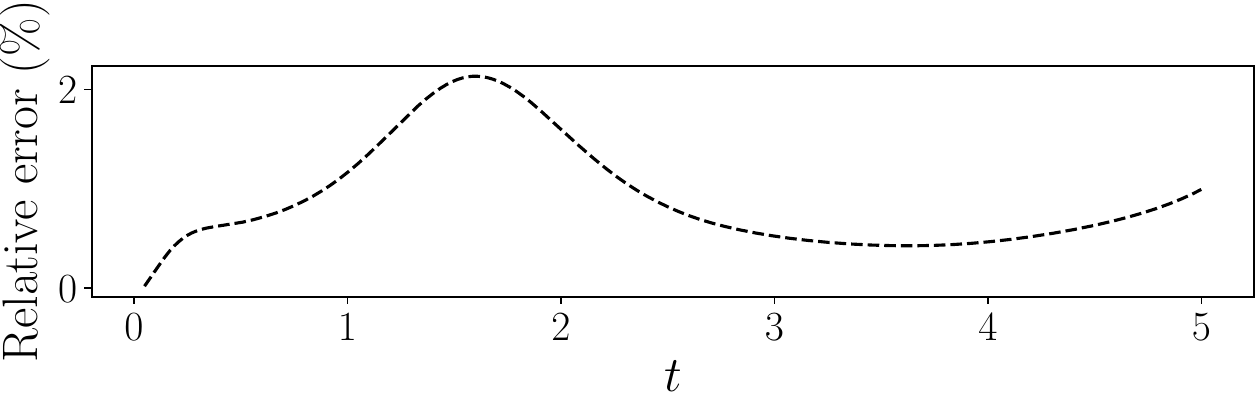}
         \caption{Error evolution of the high-diffusion system.}
     \end{subfigure}
     \caption{Spatial-averaged, relative $l^2$ prediction error of the two systems shown in Figure~\ref{fig: RD-forward}, up to $T=5.0$ seconds.}
     \label{fig: RD-error}
\end{figure}

% Emulator Accuracy
We first show the performance of the emulator $\mathscr{N}_e$. 
%
% To do so, we randomly choose $D_1, D_2, \kappa$ from the a-priori ranges listed in Table~\ref{table:dr paras} and predict the system dynamics at every cell in the discretized spatial domain. 
%
% Reference numerical solutions and emulator predictions at $T=1.0$ s and $3.0$ s are shown in Figure~\ref{fig: RD-forward}, suggesting satisfactory accuracy.
To do so, we pick two sets of parameters $D_1, D_2, \kappa$ that correspond to a low-diffusive and a high-diffusive regime and plot the contours at $t=2.0$ seconds in Figure~\ref{fig: RD-forward}. 
We also show the evolution of spatial-averaged, relative $l^2$-error of these two systems in Figure~\ref{fig: RD-error}, up to 5 seconds.

\begin{figure}[ht!]
    \begin{subfigure}[b]{0.32\textwidth}
         \centering
         \includegraphics[scale=0.21]{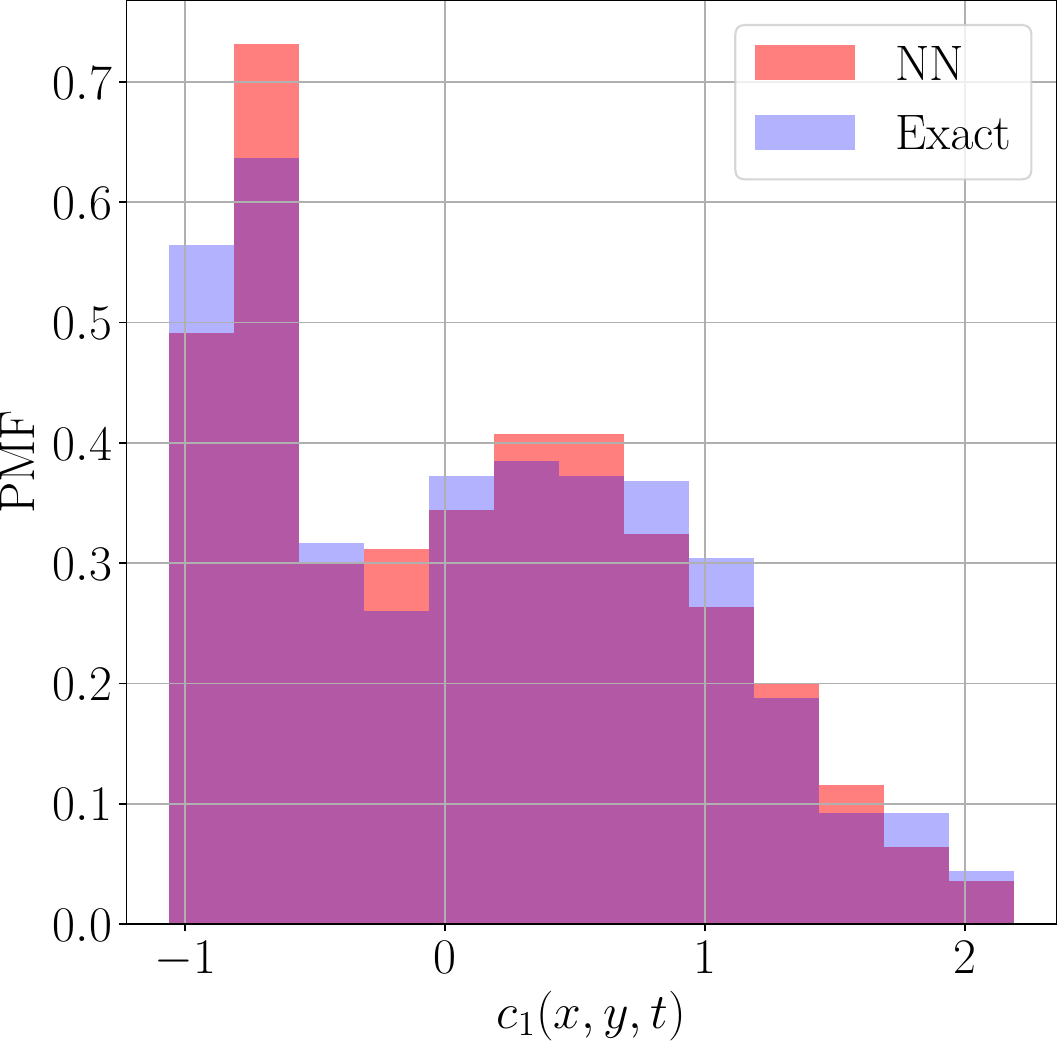}
    \caption{$c_1(x,y,t)$-histogram.}
     \end{subfigure}
     \hfill
     \begin{subfigure}[b]{0.32\textwidth}
         \centering
         \includegraphics[scale=0.21]{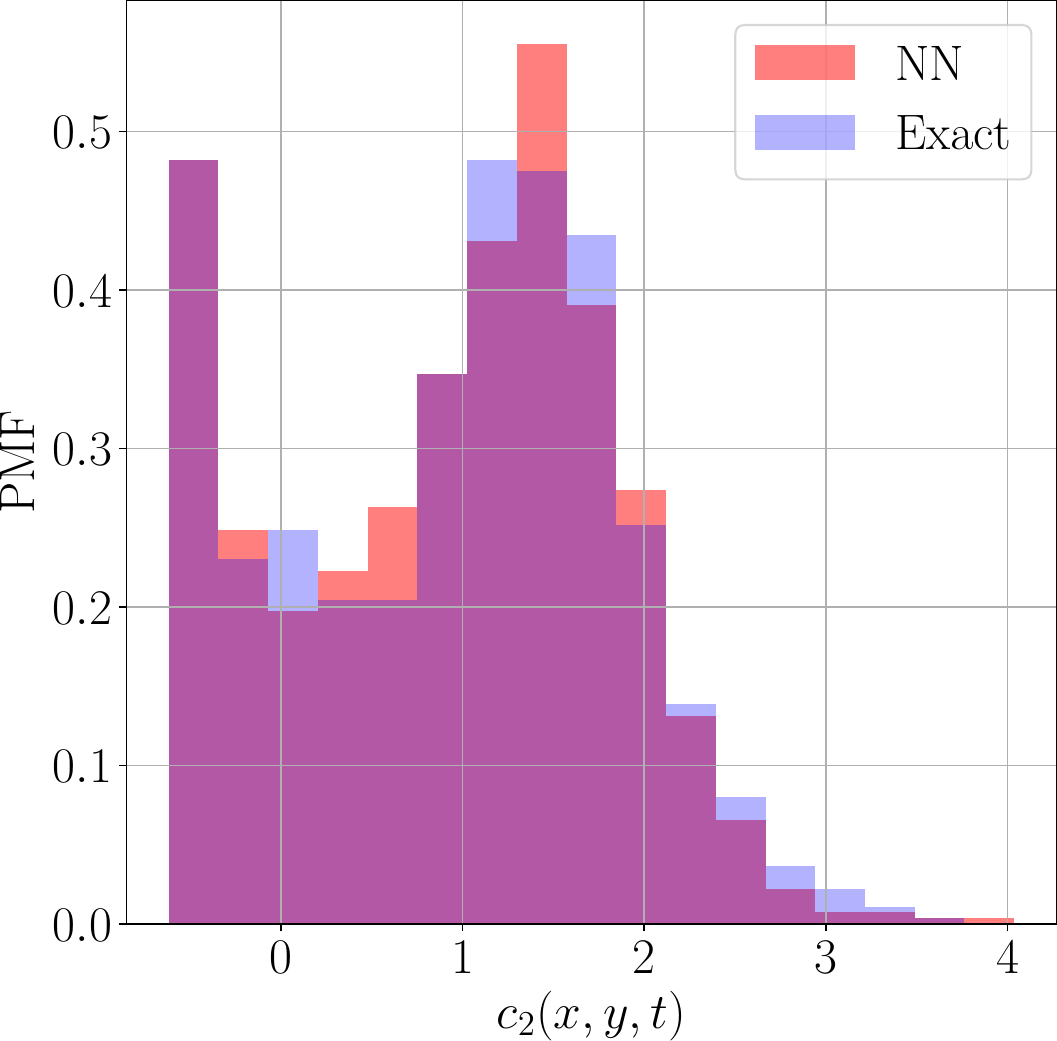}
    \caption{$c_2(x,y,t)$-histogram.}
     \end{subfigure}
    \hfill
     \begin{subfigure}[b]{0.32\textwidth}
         \centering
         \includegraphics[scale=0.205]{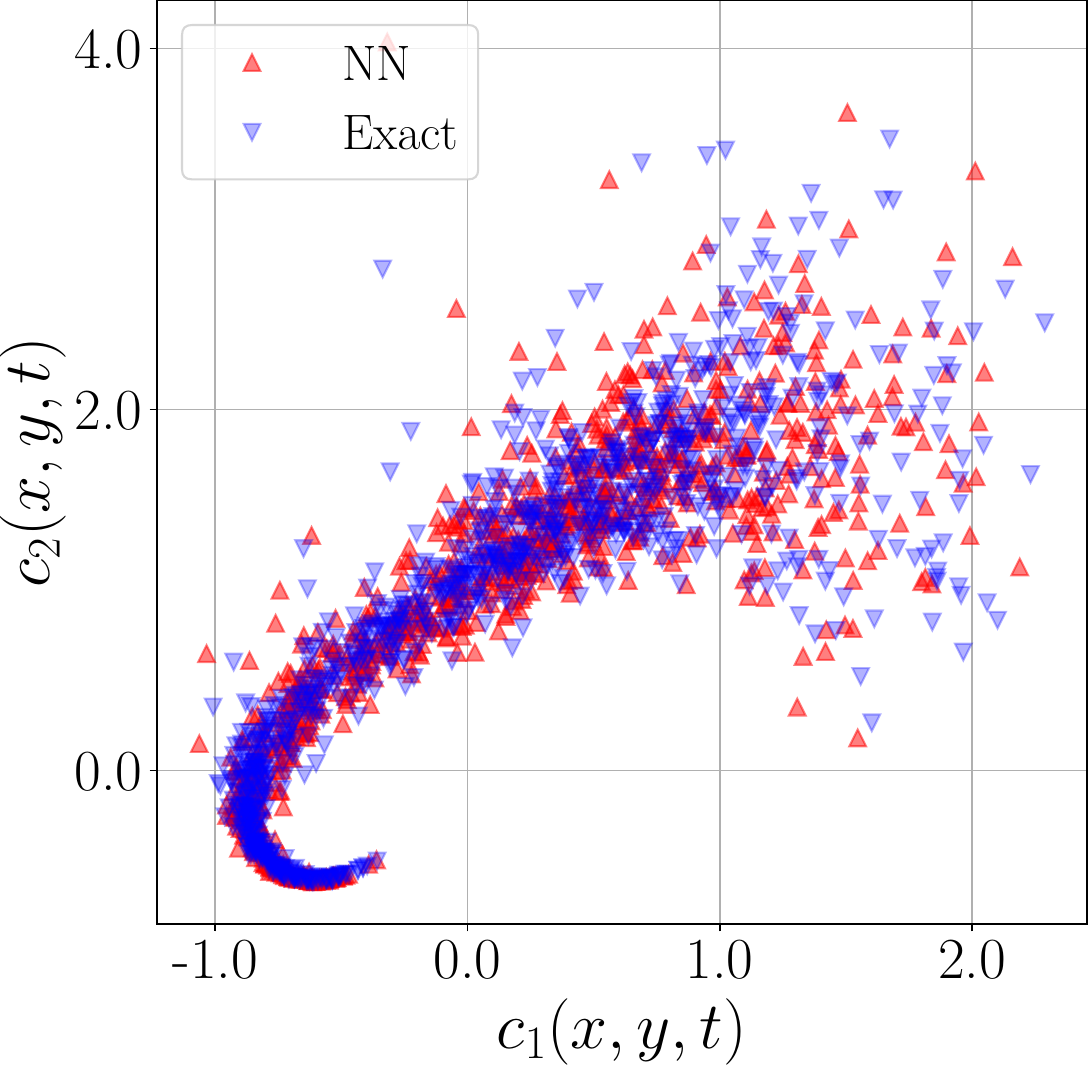}
         \caption{$c_1$-$c_2$ correlation.}
         \label{fig: rd-nf-correlation}
     \end{subfigure}
    \caption{Generative model evaluation of the reaction-diffusion system outputs.}
    \label{fig: RD-NF}
\end{figure}

% Output distributions
Next, Figure~\ref{fig: RD-NF} illustrates the ability of the Real-NVP flow $\mathscr{N}_f$ to learn the joint distribution of two system outputs $c_1(x, y, t)$ and $c_2(x, y, t)$.
High-density areas are well captured by $\mathscr{N}_f$ whereas some approximation is introduced for rare states in the tails, as shown in Figure~\ref{fig: rd-nf-correlation}. 
During our experiments, we notice a strong correlation between those high-density regions and the close-to-steady-state behavior of the reaction-diffusion system. 
% Also, due to their high concentrations in the training dataset, our inVAErt network can easily invert these states.  
%
This brings our first inversion task: Find all combinations of parameters $D_1, D_2, \kappa$, spatial locations $x, y$ and time $t$ where the parametric reaction-diffusion system reaches the selected state $\boldsymbol{c}^* = [c_1^*, c_2^*]^T$. 

We select the state $\boldsymbol{c}^* = [-0.6616, -0.5964]^T$ from the trained $\mathscr{N}_f$, which belongs to the high-density regions as illustrated in Figure~\ref{fig: rd-nf-correlation}. To enrich the latent space representation, we set $\dim(\boldsymbol{\mathcal{W}}) = 8$, providing 4 additional dimensions compared to the difference between input and output dimensionality.

\begin{figure}[ht!]
     \centering
     \begin{subfigure}[b]{0.2455\textwidth}
         \centering
         \includegraphics[scale=0.165]{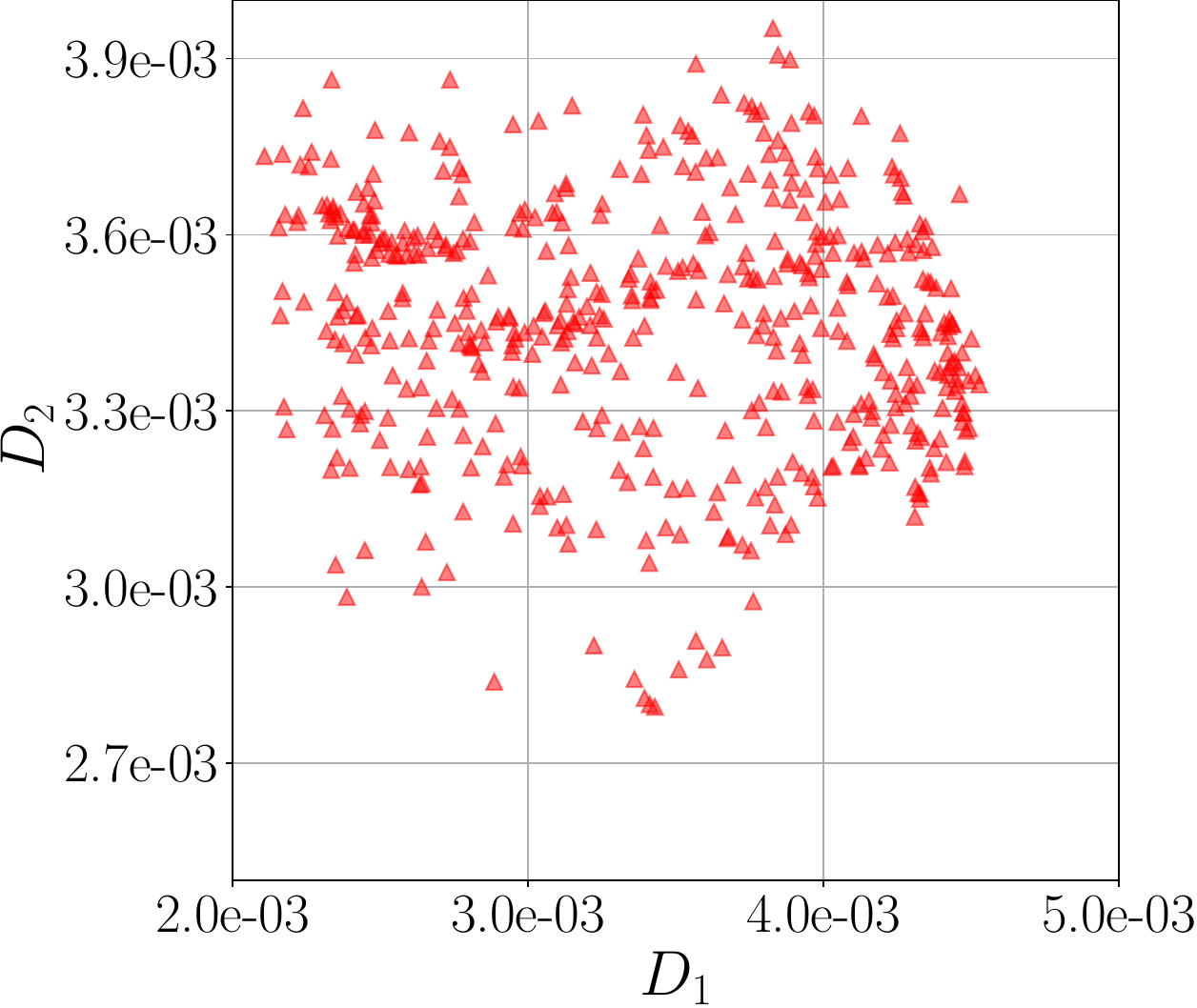}\caption{$\mathcal{N}(\boldsymbol{0}, \mathbf{I})$ sampling.}
         \label{fig: DR-d1d2-n01}
     \end{subfigure}
     \hfill
      \begin{subfigure}[b]{0.2455\textwidth}
         \centering
         \includegraphics[scale=0.165]{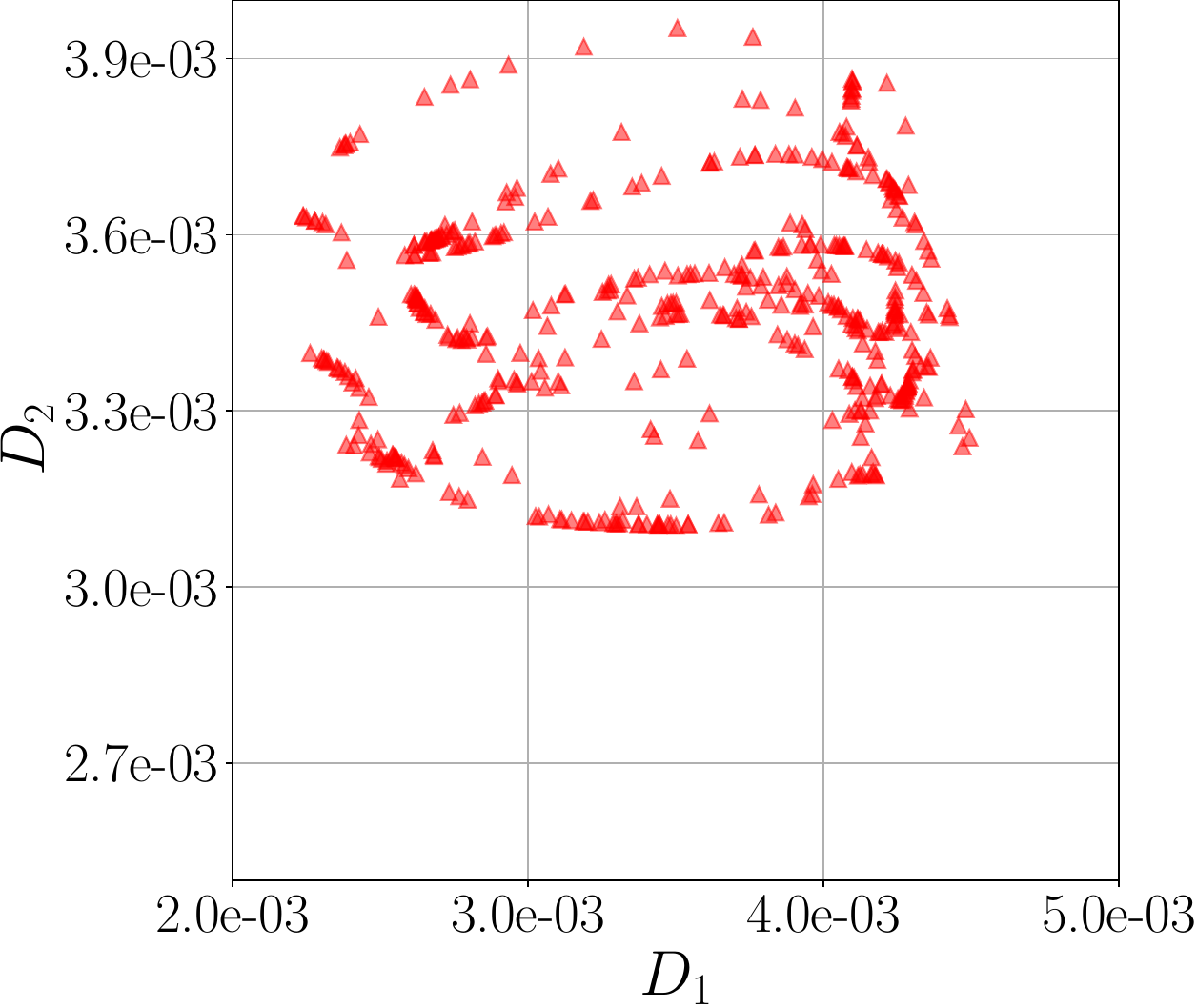}\caption{PC sampling.}
         \label{fig: DR-d1d2-pc}
     \end{subfigure}
     \hfill
      \begin{subfigure}[b]{0.2455\textwidth}
         \centering
         \includegraphics[scale=0.165]{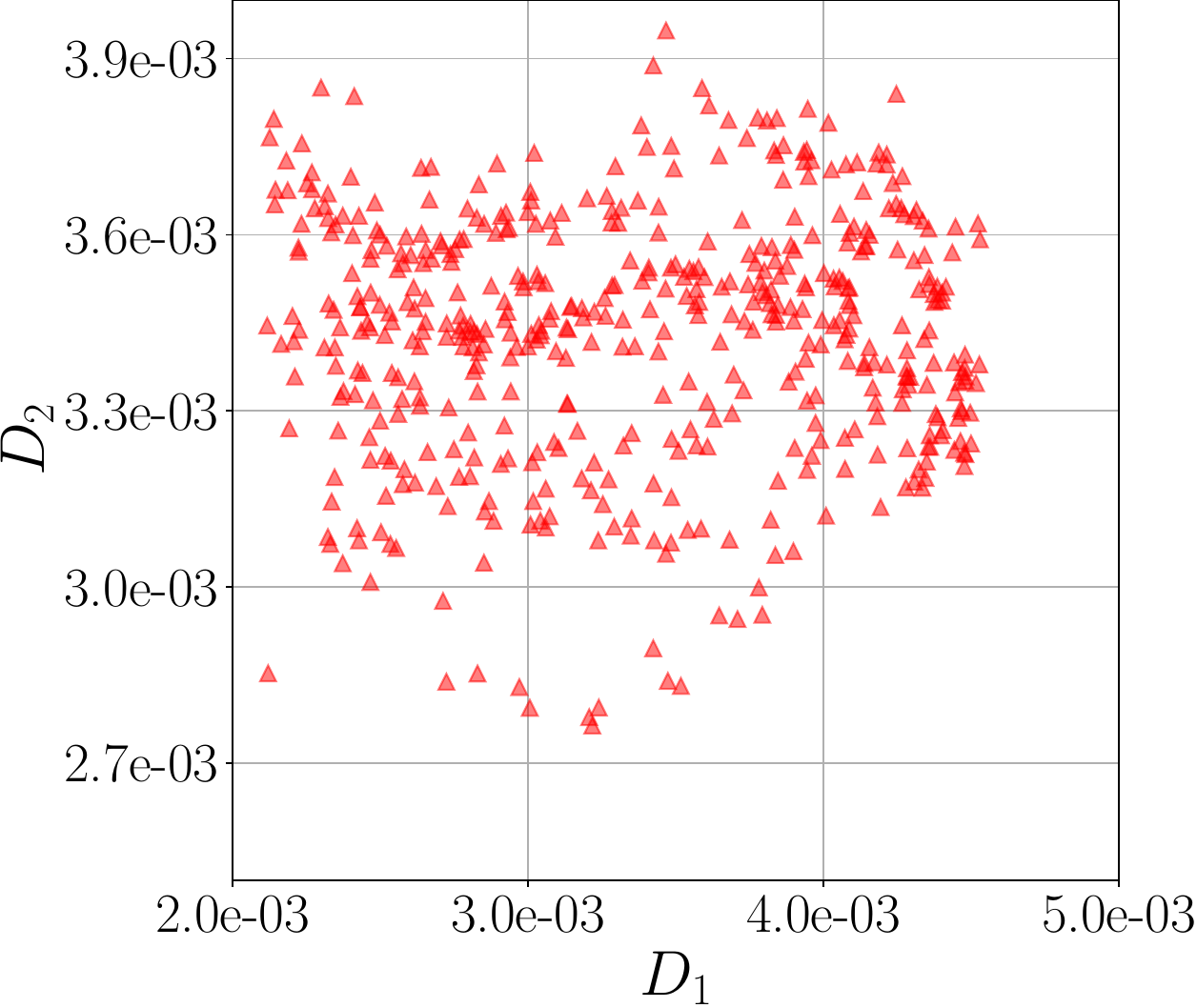}\caption{NF sampling.}
         \label{fig: DR-d1d2-NF}
     \end{subfigure}
     \hfill
     \begin{subfigure}[b]{0.2455\textwidth}
         \centering
         \includegraphics[scale=0.165]{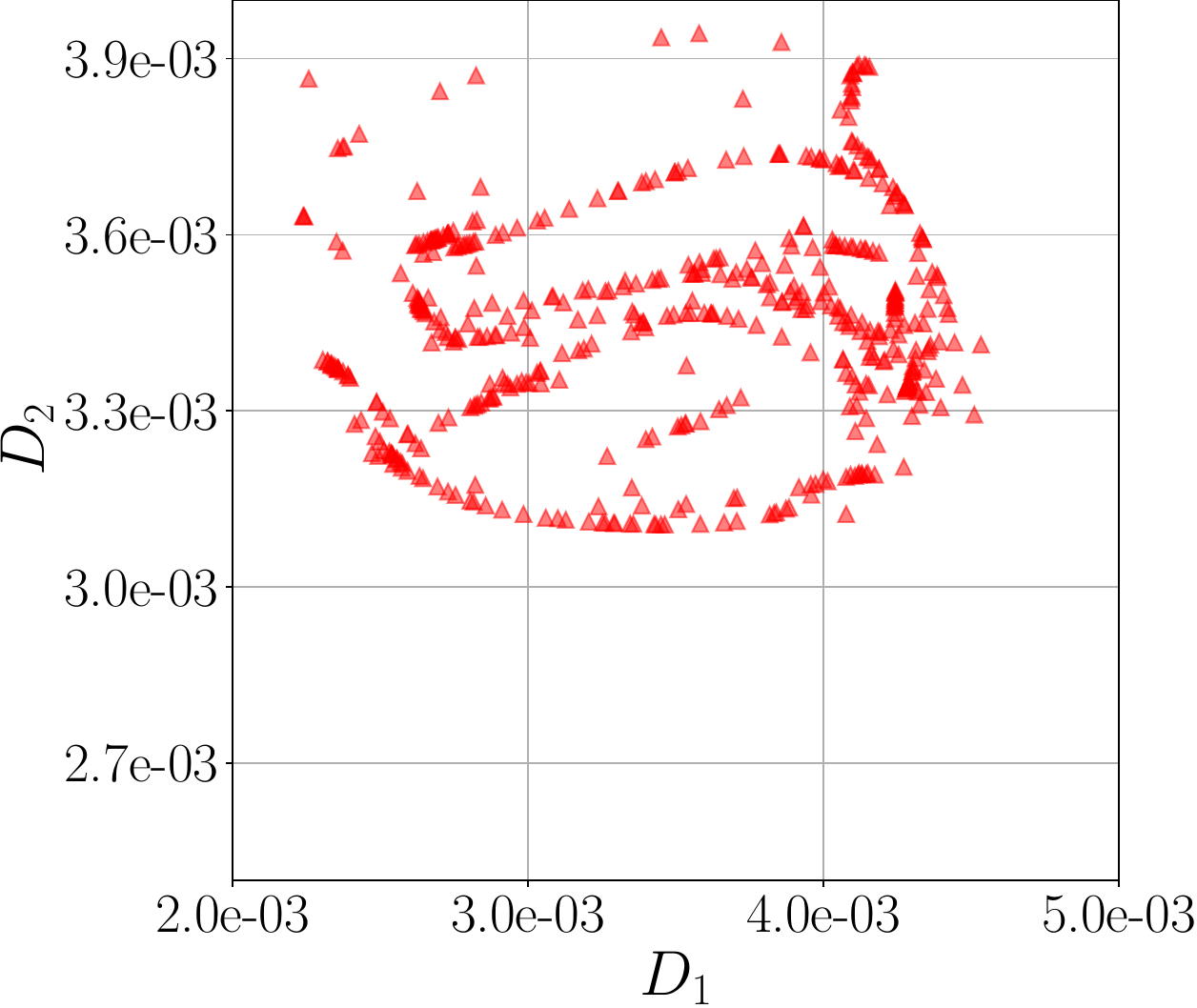}\caption{NF+PC.}
         \label{fig: DR-d1d2-NFPC}
     \end{subfigure}
     \\
     \vspace{0.2cm}
     \begin{subfigure}[b]{0.2455\textwidth}
         \centering
         \includegraphics[scale=0.17]{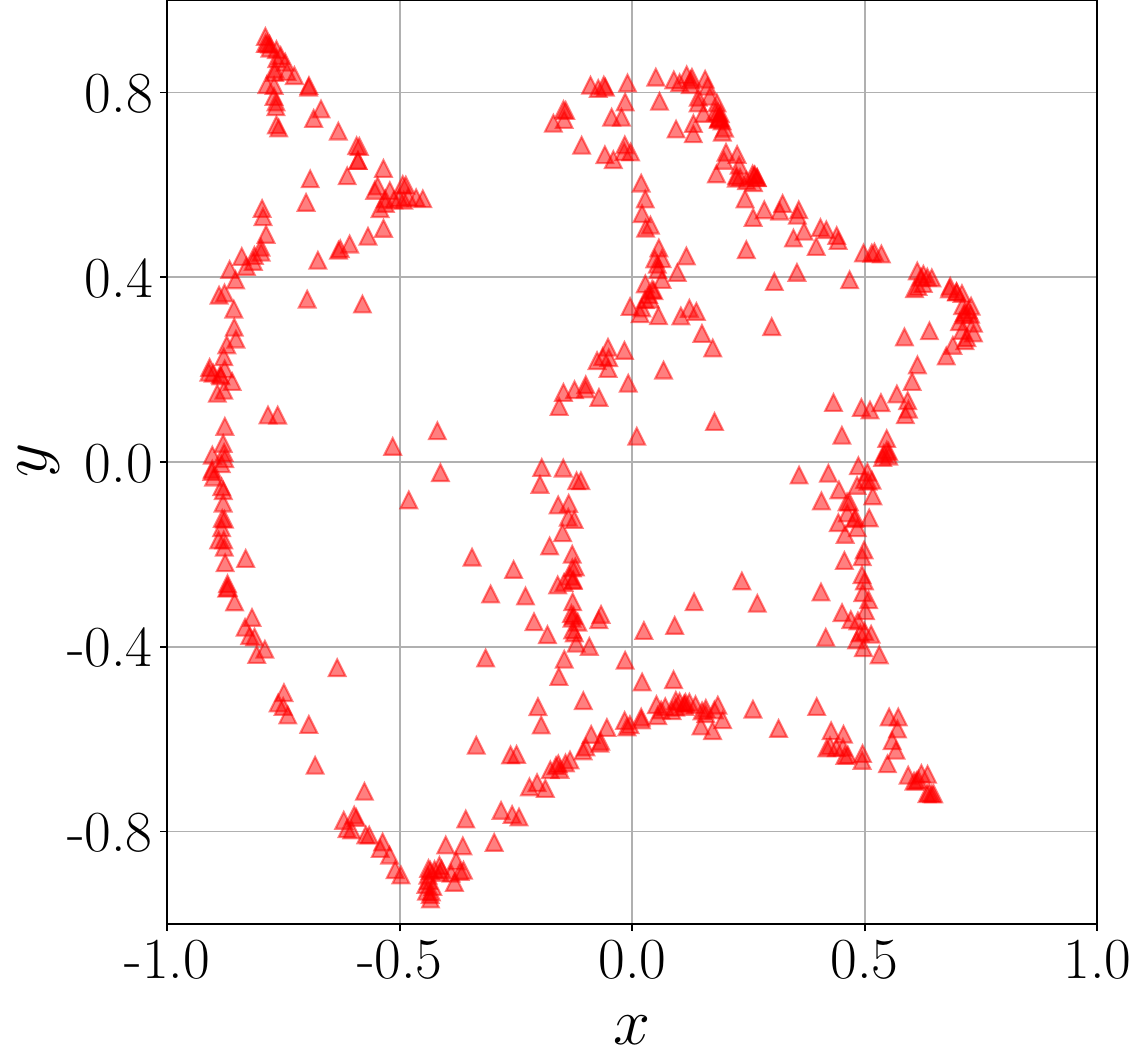}\caption{$\mathcal{N}(\boldsymbol{0}, \mathbf{I})$ sampling.}
         \label{fig: DR-xy-n01}
     \end{subfigure}
     \hfill
     \begin{subfigure}[b]{0.2455\textwidth}
         \centering
         \includegraphics[scale=0.17]{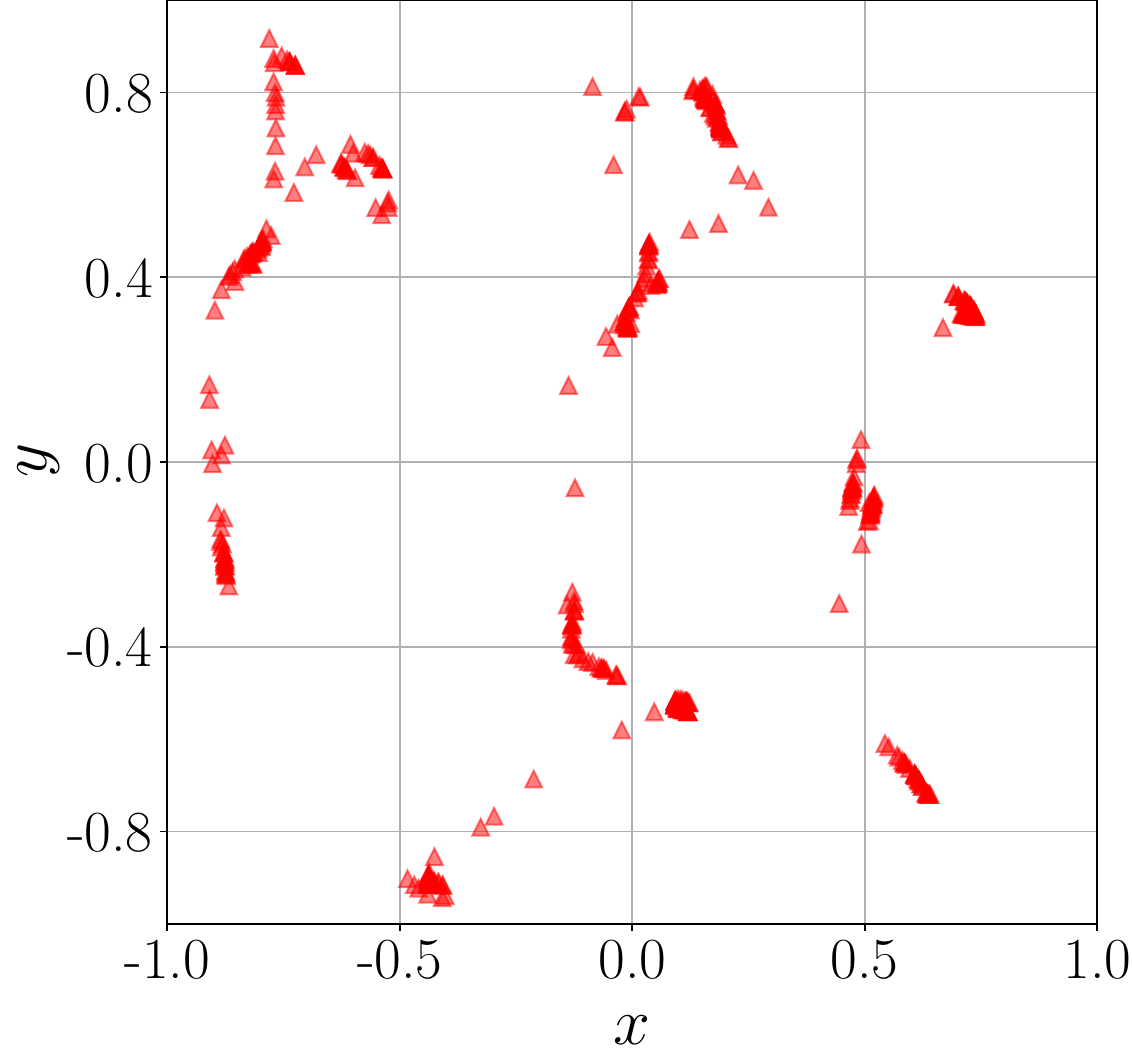}\caption{PC sampling.}
         \label{fig: DR-xy-pc}
     \end{subfigure}
     \hfill
     \begin{subfigure}[b]{0.2455\textwidth}
         \centering
         \includegraphics[scale=0.17]{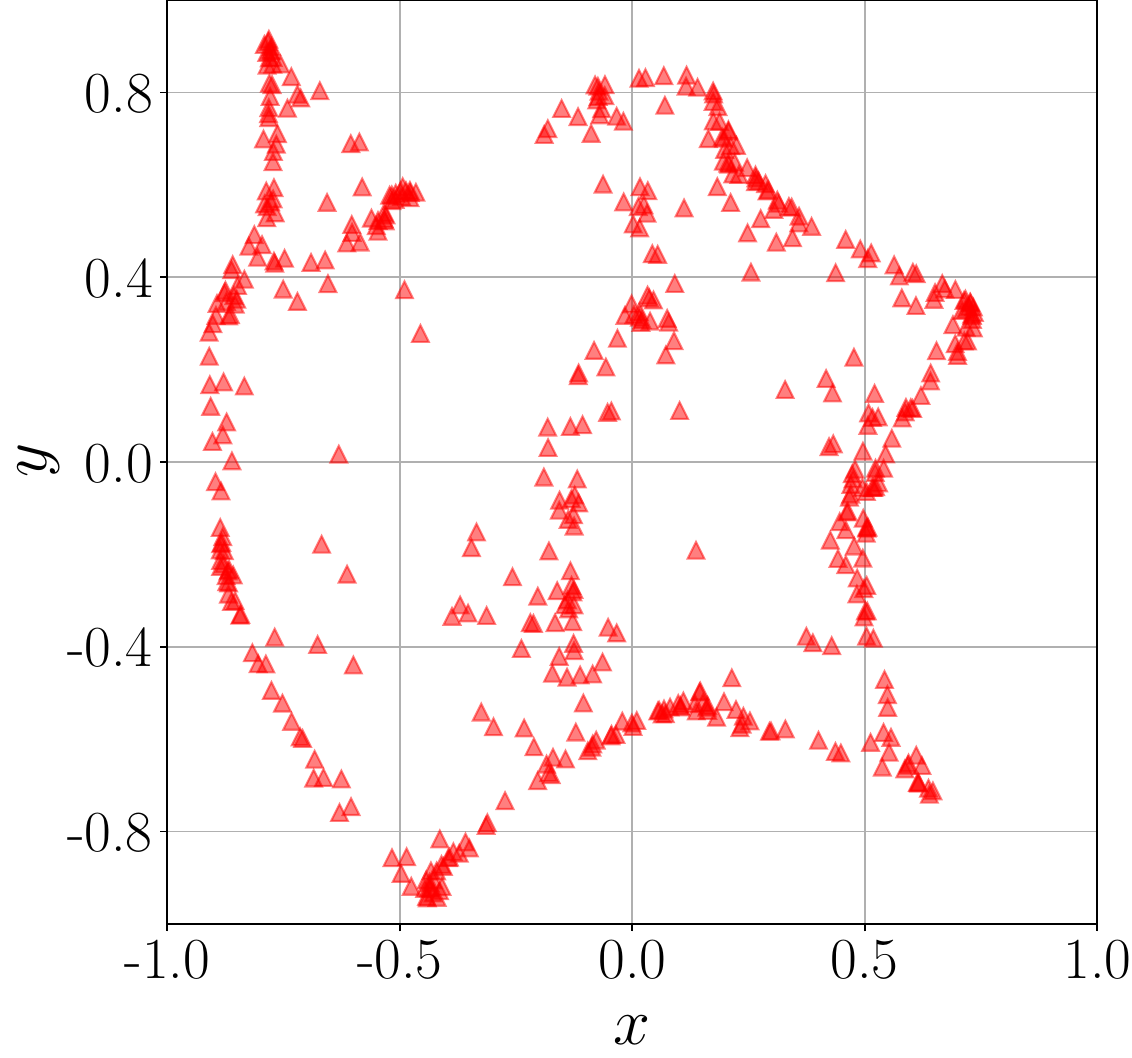}\caption{NF sampling.}
         \label{fig: DR-xy-NF}
     \end{subfigure}
     \hfill
     \begin{subfigure}[b]{0.2455\textwidth}
         \centering
         \includegraphics[scale=0.17]{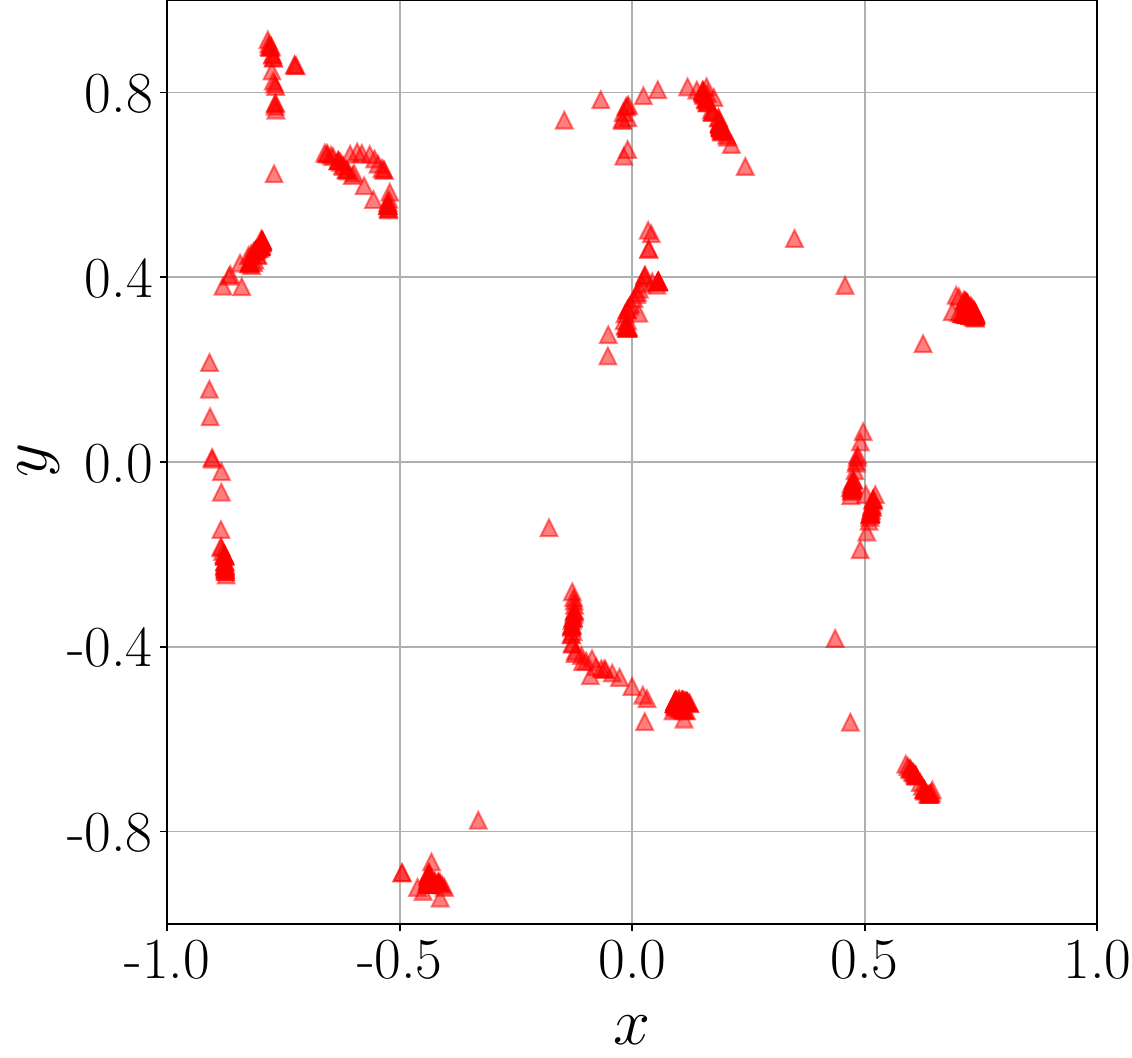}\caption{NF+PC.}
         \label{fig: DR-xy-NFPC}
     \end{subfigure}
     \caption{Model inversion results for the parametric reaction-diffusion system: fix $\boldsymbol{c}^* = [-0.6616, -0.5964]^T$. Comparing $D_1$-$D_2$, $x$-$y$ correlations using different latent variable sampling schemes. Sample size: 500, 8D latent space. Methods: sampling from $\mathcal{N}(\boldsymbol{0}, \mathbf{I})$, PC sampling ($R=6$), NF sampling (see Table~\ref{table: rd-nf}), combined NF+PC sampling (apply PC sampling to inverse predictions $\widehat{\boldsymbol{v}}$ associated with~\Cref{fig: DR-d1d2-NF,fig: DR-xy-NF} with $R=6$).}
     \label{fig: rd-task1-correlations}
\end{figure}

First, we utilize various sampling schemes discussed in Section~\ref{sec:analysis_sampling} to test the trained decoder $\mathscr{N}_d$. For conciseness, we only plot the learned correlations between the two diffusivity coefficients $D_1$, $D_2$ and the spatial coordinates $x, y$ in Figure~\ref{fig: rd-task1-correlations}. 
Like the Lorenz system, we found the posterior distribution $q(\boldsymbol{w}|\boldsymbol{v})$ closely resembles a standard Gaussian density, which explains the resulting similarity of $\mathcal{N}(\boldsymbol{0},\mathbf{I})$ sampling and NF sampling.
The PC sampling method, either applied to the $\mathcal{N}(\boldsymbol{0},\mathbf{I})$ samples or those drawn from the trained $\mathscr{N}_{f,\boldsymbol{w}}$, exhibits promising potential in revealing unrecognizable latent structures polluted by spurious outliers (e.g. compare Figure~\ref{fig: DR-d1d2-n01} to Figure~\ref{fig: DR-d1d2-pc}).

\begin{figure}[ht!]
    \begin{subfigure}[b]{0.32\textwidth}
         \centering
         \includegraphics[scale=0.22]{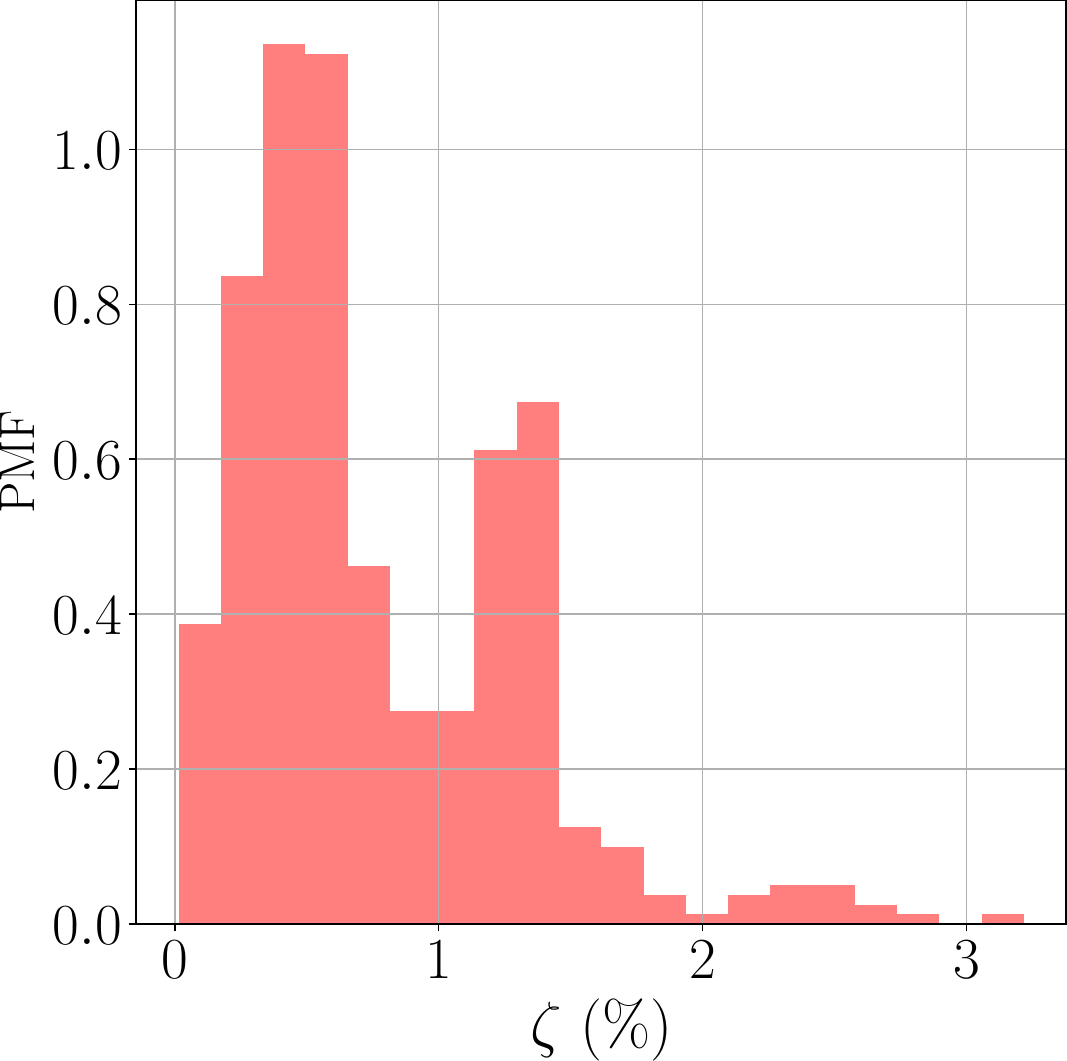}
    \caption{Histogram of $\zeta$.}
    \label{fig: dr-hist-zeta}
     \end{subfigure}
     \hfill
     \begin{subfigure}[b]{0.32\textwidth}
         \centering
         \includegraphics[scale=0.22]{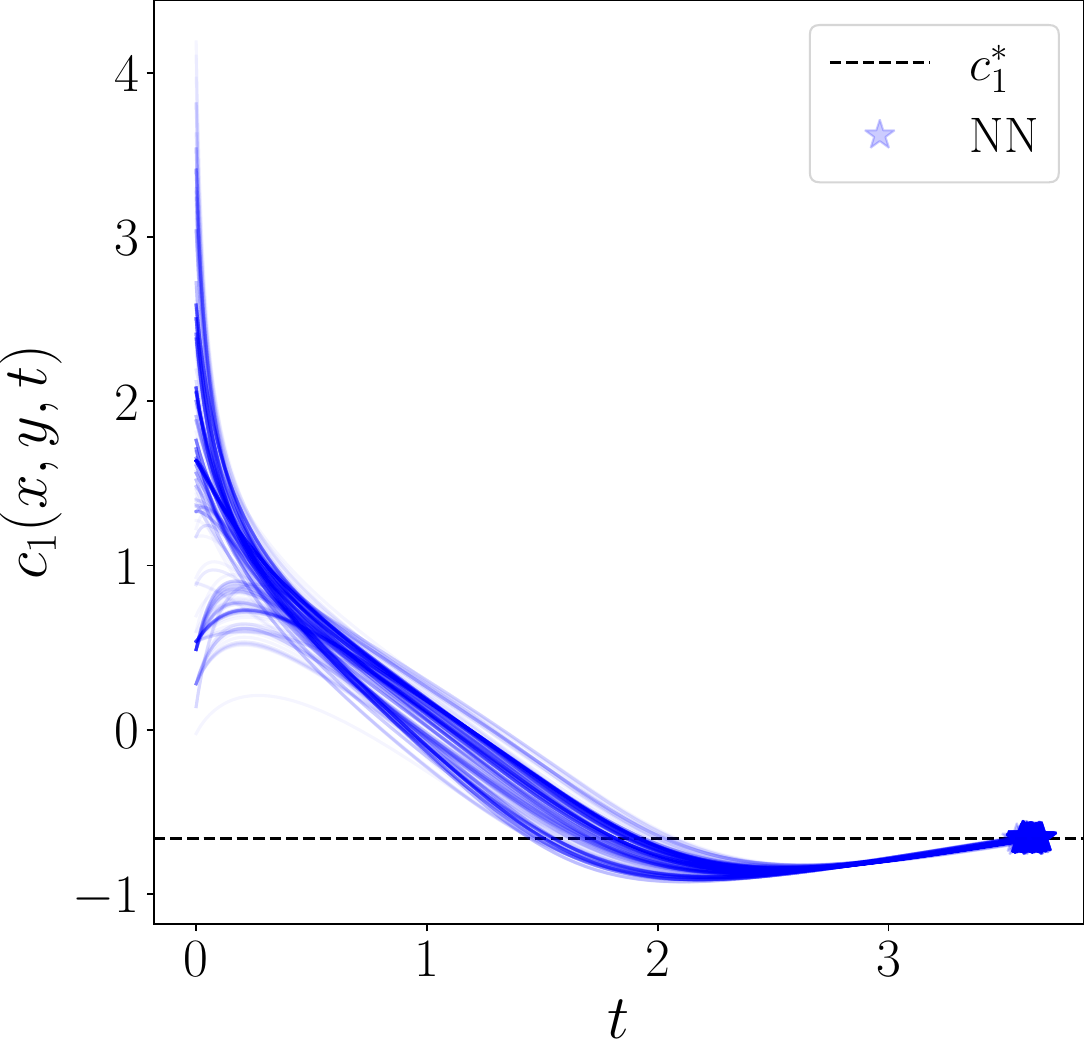}
    \caption{Verification of $c_1^*$.}
    \label{fig: dr-task1-c1}
     \end{subfigure}
     \hfill
     \begin{subfigure}[b]{0.32\textwidth}
         \centering
         \includegraphics[scale=0.22]{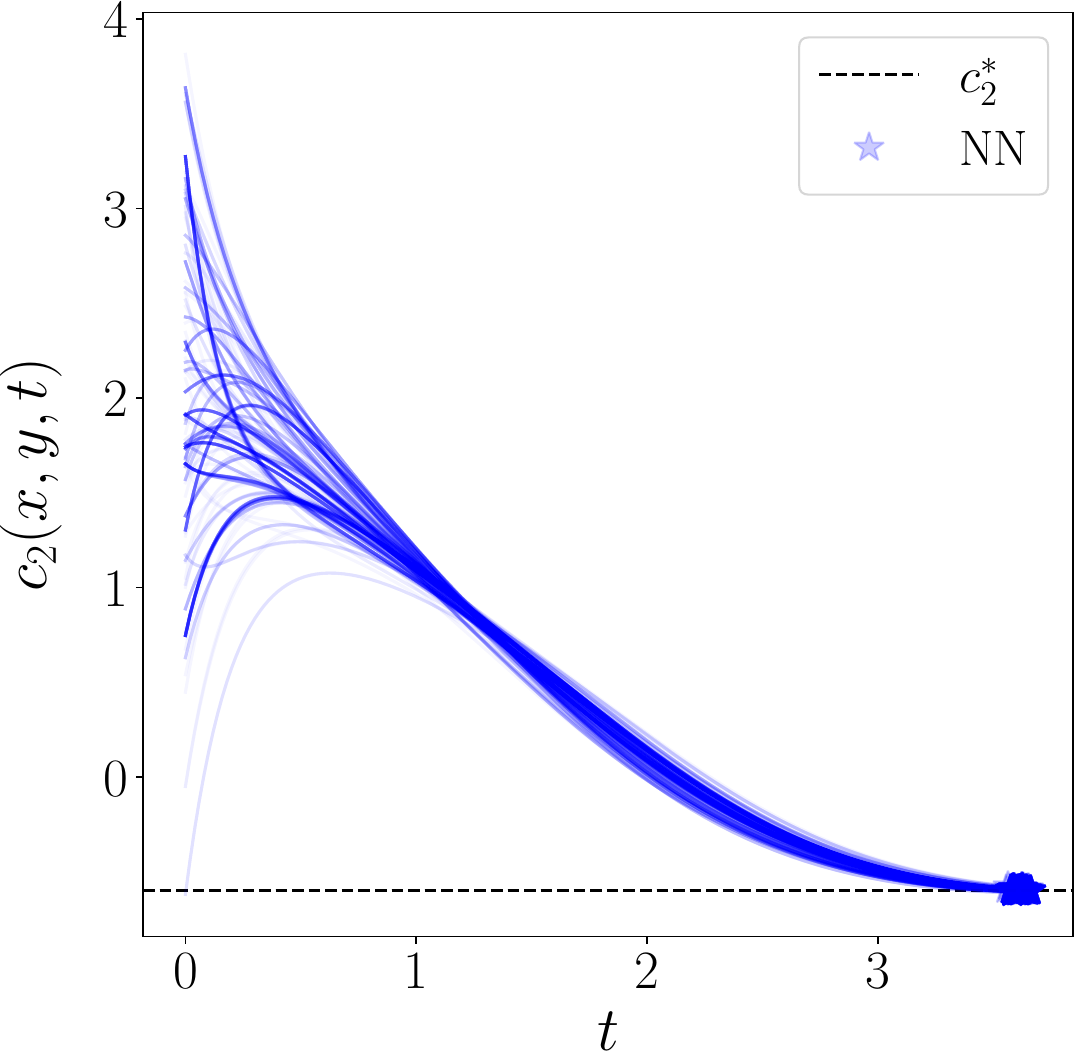}
    \caption{Verification of $c_2^*$.}
    \label{fig: dr-task1-c2}
     \end{subfigure}
      \caption{Model inversion results for the parametric reaction-diffusion system: fix $\boldsymbol{c}^* = [-0.6616, -0.5964]^T$. PC sampling ($R=6$) is applied to generate 500 samples of $\boldsymbol{w}$. Results corresponding to inversion samples of $\widehat{\boldsymbol{v}}$ associated with~\Cref{fig: DR-d1d2-pc,fig: DR-xy-pc}.  
      Each FV-RK4 solution trajectory from $\widehat{\boldsymbol{v}}$ is plotted (NN) and a marker added at the concentration corresponding to $t = \widehat{t}$.}
     \label{fig: DR-TASK1-verify}
\end{figure}

Next, in Figure~\ref{fig: DR-TASK1-verify}, we verify that our parametric reaction-diffusion system is indeed non-identifiable under these inverse predictions.
To realize this, we forward the FV-RK4 solver of PDEB{\scriptsize{ENCH}}~\cite{PDEBench2022} with all 500 $\widehat{\boldsymbol{v}}$'s produced by the trained decoder $\mathscr{N}_d$, plus the PC sampling approach (i.e. inversion samples associated with~\Cref{fig: DR-d1d2-pc,fig: DR-xy-pc}). 
The simulation uses $\widehat{D}_1,\widehat{D}_2, \widehat{\kappa} \in \widehat{\boldsymbol{v}}$ as inputs and finishes at $t = \widehat{t} \in \widehat{\boldsymbol{v}}$. From~\Cref{fig: dr-task1-c1,fig: dr-task1-c2}, we see all solution trajectories of $c_1$, $c_2$, originated from various $(\widehat{x}, \widehat{y}) \in \widehat{\boldsymbol{v}}$, converge towards our prescribed values with a maximum relative error around 3\% (i.e. see Figure~\ref{fig: dr-hist-zeta}). 

However, the accuracy of the proposed approach deteriorates when inverting an output $\by^*$ that is rarely observed during training. This also leads to an insufficient characterization of the fiber $\boldsymbol{\mathcal{M}}_{\bv}$ over $\by^*$.
Besides increasing the training set, one could also use the exact forward model or the proposed emulator $\mathscr{N}_e$ to reject decoded inputs, as discussed in Section~\ref{example: lorenz}.

In the end, we would like to leverage the inVAErt network to quickly answer relevant question on the inverse dynamics of the parametric reaction-diffusion system~\eqref{equ: pde-DR}.
Specifically, we would like to identify the locations in space where the chemical concentrations $c_1, c_2$ may fall within some prescribed ranges.
Answering this type of question represents a challenging inverse problem, but is well within reach for the proposed architecture. 

First, suppose the system output state $\boldsymbol{c}^*$ is of particular interests if it belongs to an \emph{active region} defined as: 
$$\mathcal{A} = \{\boldsymbol{c}^*| -1\leq c_1^*\leq -0.8, 0.0\leq c_2^*\leq 0.2 \} \ .$$ 

\begin{figure}[ht!]
    \begin{subfigure}[b]{0.45\textwidth}
         \centering
         \includegraphics[scale=0.22]{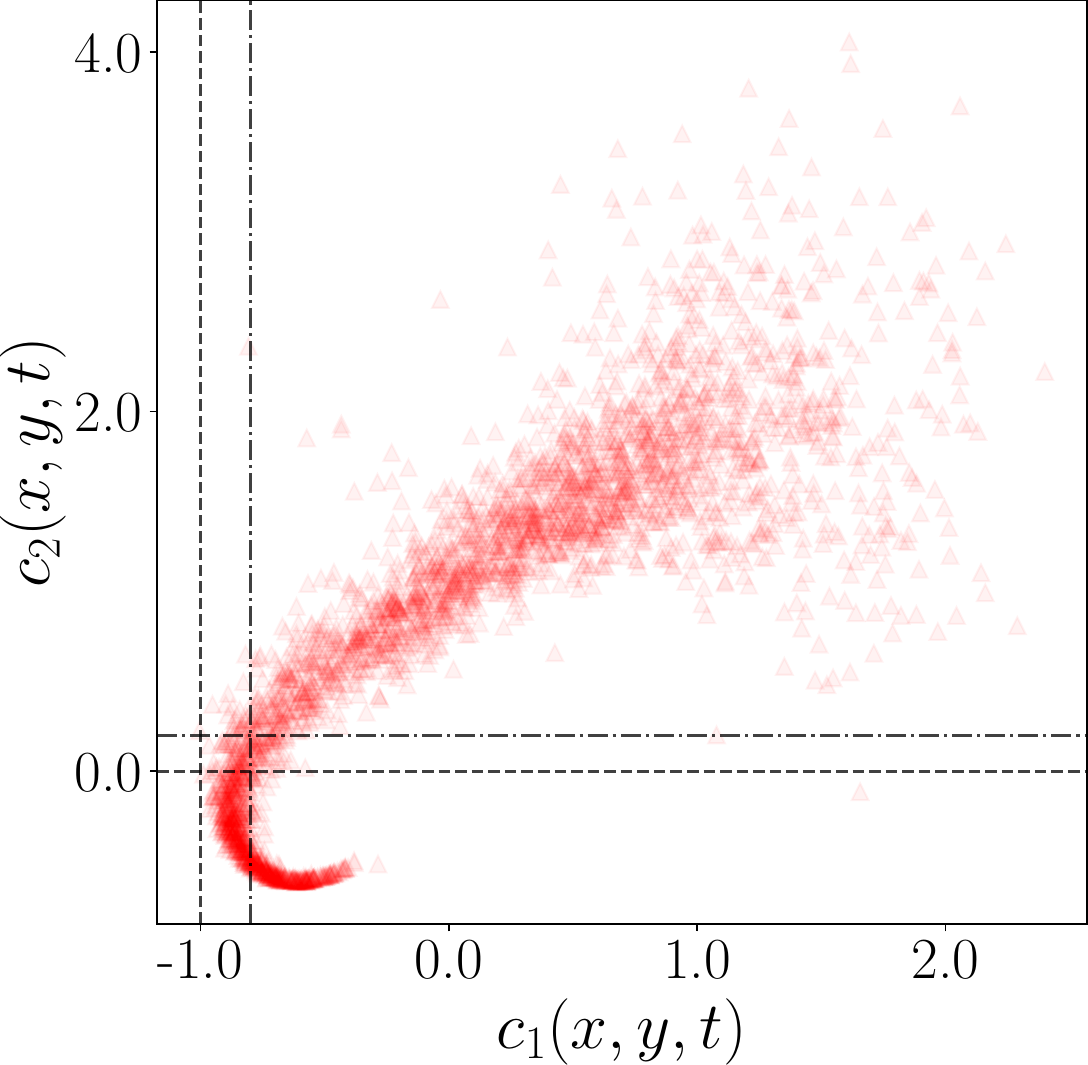}
    \caption{The region $\mathcal{A}$ superimposed to the model outputs sampled from $\mathscr{N}_f$. We utilize a reduced sample size to enhance the clarity of the visualization.}
    \label{fig: mark A}
    \end{subfigure}
     \hfill
    \begin{subfigure}[b]{0.5\textwidth}
         \centering
         \includegraphics[scale=0.22]{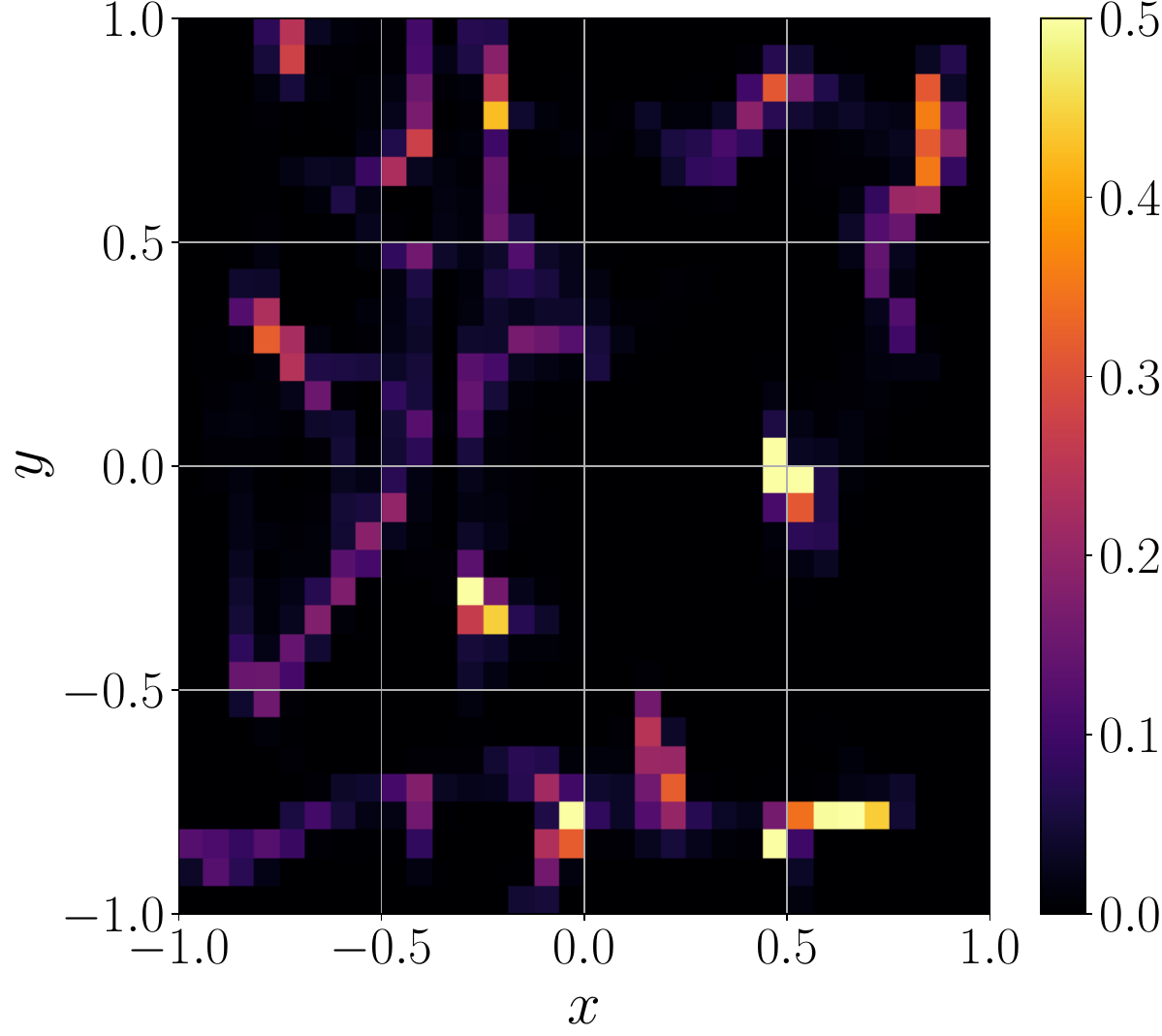}
    \caption{Spatial locations where the output concentrations are compatible with the region $\mathcal{A}$. The maximum value is limited to 0.5 (instead of 1.0) for improved visualization.}
    \label{fig: space id}
    \end{subfigure}
    \caption{Model inversion results for the parametric reaction-diffusion system: concentrations $\boldsymbol{c}^*$ from a selected region $\mathcal{A}$ (left) and corresponding spatial locations (right). Color scale: normalized occurrence of a spatial cell $(\widehat{x},\widehat{y})$ in the decoded samples. Results are generated using PC sampling with $R=6$, an eight-dimensional latent space, 1188/50000 concentrations selected from $\mathcal{A}$, and 500 samples for each application of the decoder.}
    \label{fig: act region}
\end{figure}

To collect states within $\mathcal{A}$, we sample from the trained normalizing flow model $\mathscr{N}_f$ and reject all samples outside of $\mathcal{A}$ (see Figure~\ref{fig: mark A}).
Then, using the trained decoder $\mathscr{N}_d$ to invert each of the selected state, we obtain a sequence of inverse predictions $\widehat{\boldsymbol{v}}$.
Next, we track each ($\widehat{x}, \widehat{y}) \in \widehat{\bv}$ and accumulate the occurrence of its corresponding cell in space. This results in Figure~\ref{fig: space id}, where each cell's occurrence count is normalized by the maximum occurrence across all cells. 

\begin{figure}[ht!]
     \begin{subfigure}[b]{0.33\textwidth}
         \centering
         \includegraphics[scale=0.22]{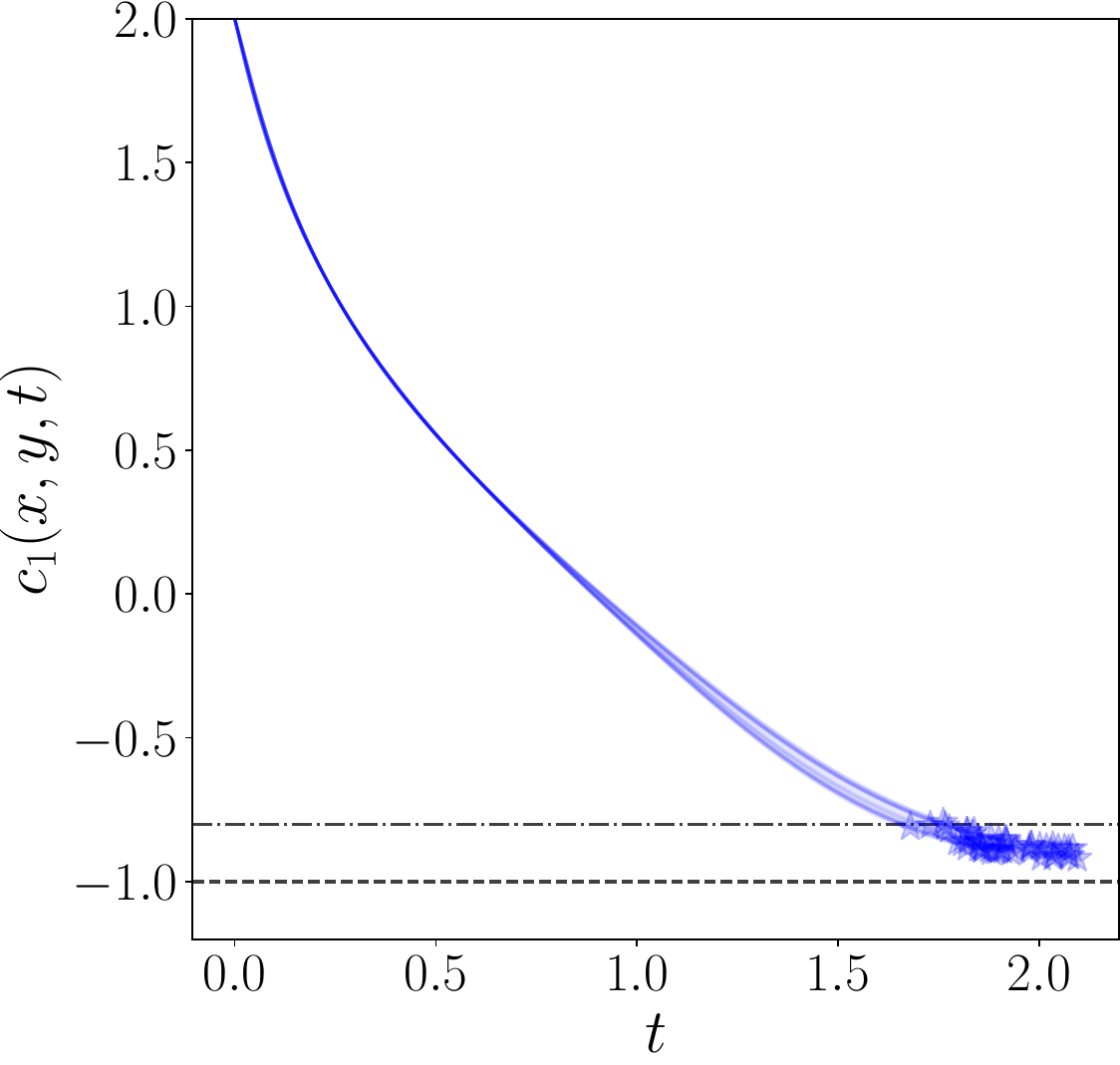}
    \caption{$c_1$: 1st location.}
     \end{subfigure}
      \begin{subfigure}[b]{0.33\textwidth}
         \centering
         \includegraphics[scale=0.22]{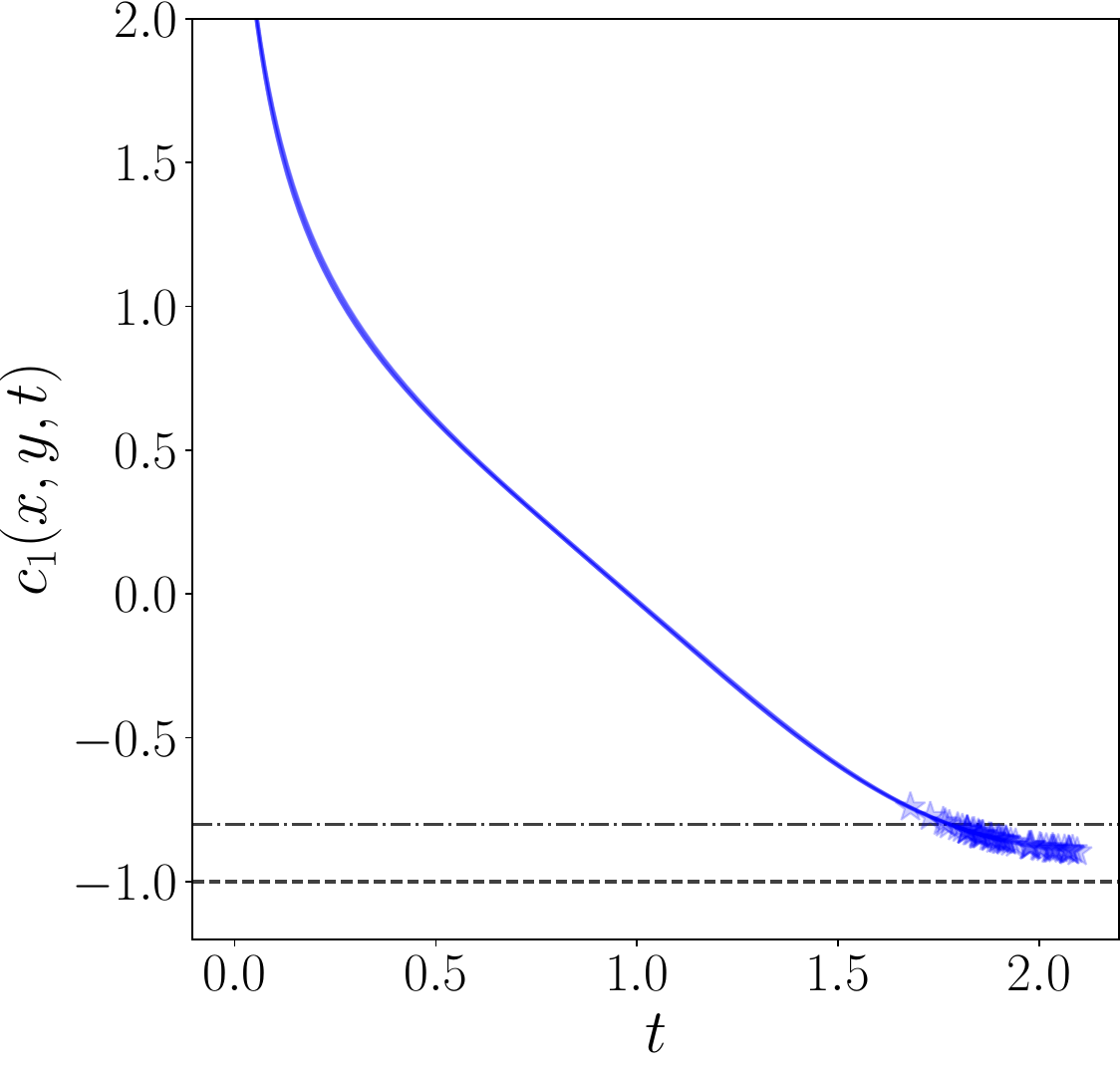}
    \caption{$c_1$: 4th location.}
     \end{subfigure}
      \begin{subfigure}[b]{0.33\textwidth}
         \centering
         \includegraphics[scale=0.22]{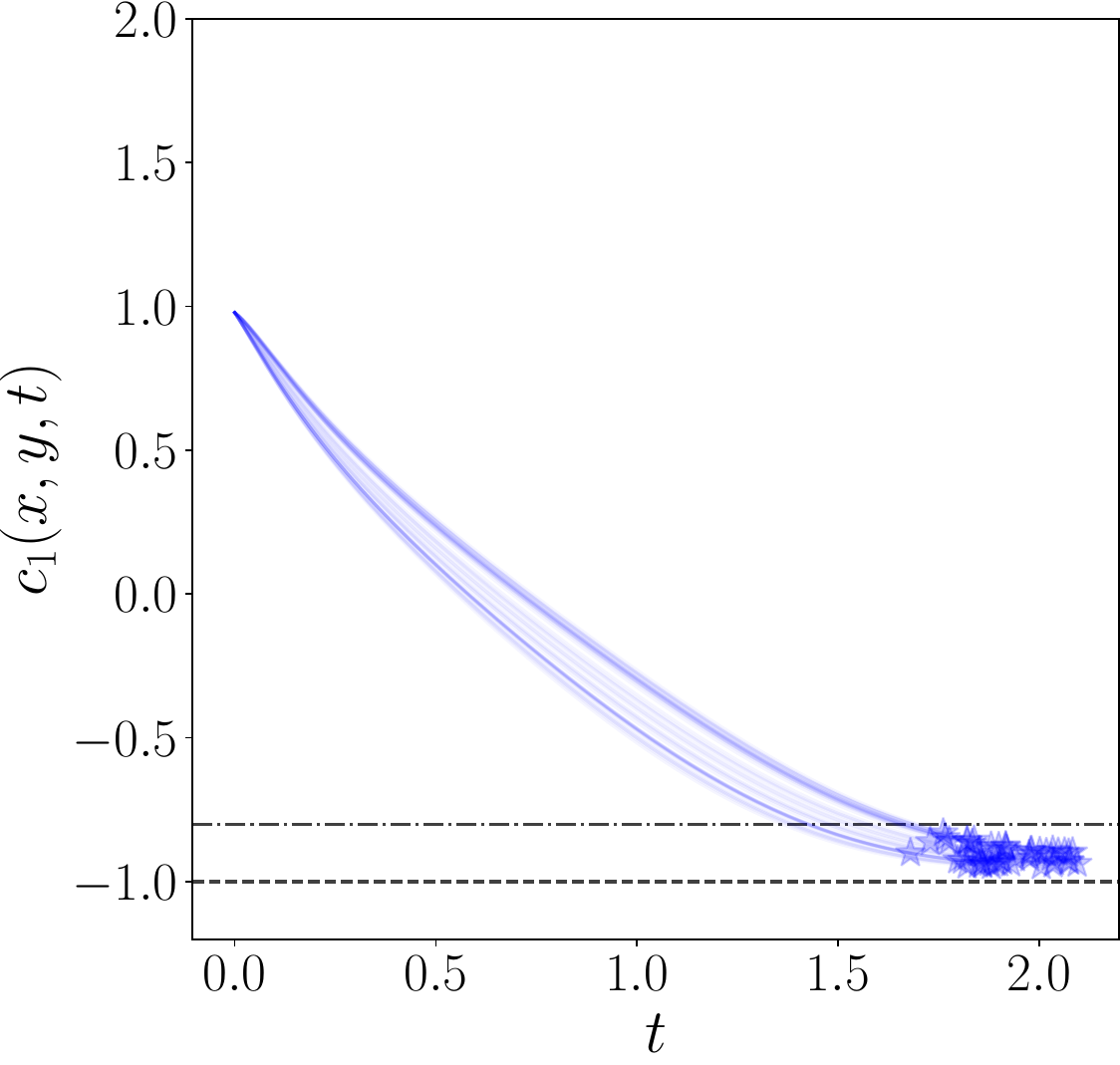}
    \caption{$c_1$: 9th location.}
     \end{subfigure} \\

    \begin{subfigure}[b]{0.33\textwidth}
         \centering
         \includegraphics[scale=0.22]{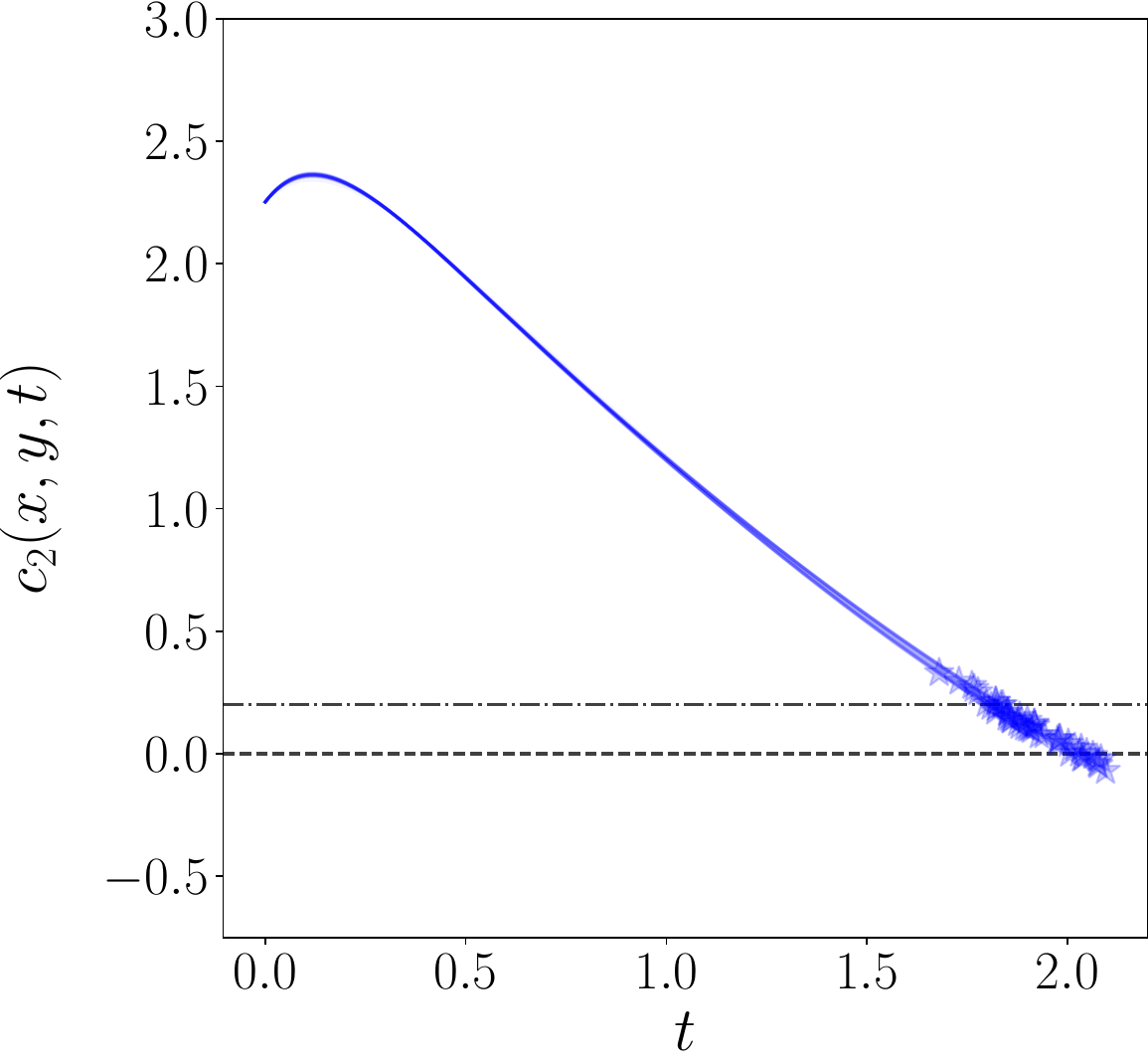}
    \caption{$c_2$: 1st location.}
     \end{subfigure}
      \begin{subfigure}[b]{0.33\textwidth}
         \centering\includegraphics[scale=0.22]{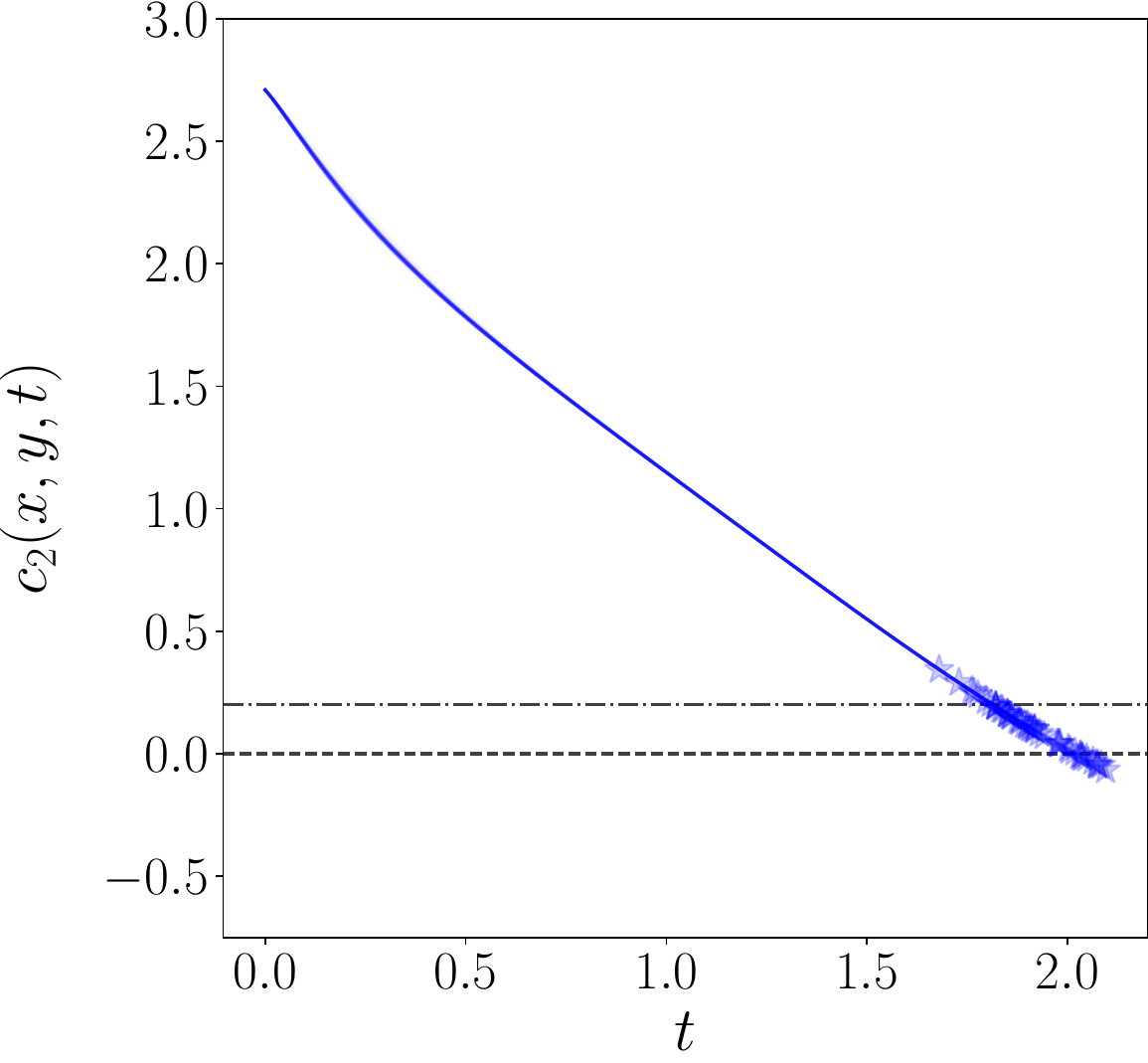}
    \caption{$c_2$: 4th location.}
     \end{subfigure}
      \begin{subfigure}[b]{0.33\textwidth}
         \centering
         \includegraphics[scale=0.22]{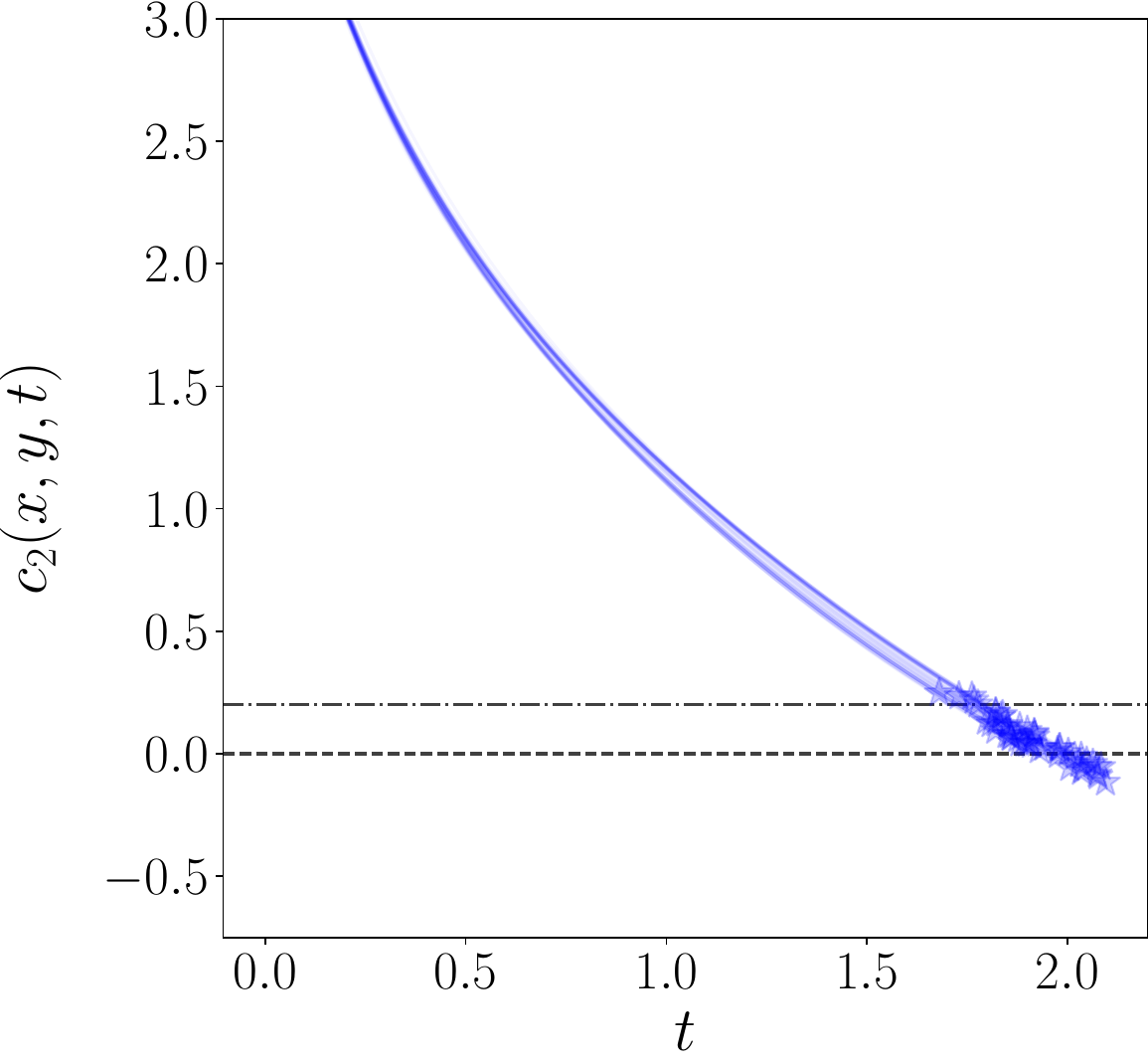}
    \caption{$c_2$: 9th location.}
     \end{subfigure} 
     \caption{Model inversion results for the parametric reaction-diffusion system. Verification of three possible locations $(\widehat{x}, \widehat{y})$ associated with the 1st, 4th and 9th relative frequency among all inverse predictions $\widehat{\boldsymbol{v}}$. Each FV-RK4 solution trajectory from a small subset of all $\widehat{\boldsymbol{v}}$ is plotted and the predicted time point $t = \widehat{t}$ is marked.}
     \label{fig: dr task2-check}
\end{figure}

Finally, for verification, we take a small subset of all inverse predictions and forward the systems with the FV-RK4 solver~\cite{PDEBench2022}. Then, we check the resulting time series solution of $c_1$ and $c_2$ at locations ranked 1st, 4th, and 9th with respect to the normalized occurrence count shown in Figure~\ref{fig: space id}. The results are reported in Figure~\ref{fig: dr task2-check}, showing that most of the solution trajectories converge inside the region $\mathcal{A}$.

%*************NOT DISCUSSED YET****************************% 
%We then concatenate these negative values ({\bf\color{red}WE CAN ALSO GENERATE ADDITIONAL VALUES NOT RELATED TO NF}) with samples from the latent space {\bf\color{red} DIMENSIONALITY? SAMPLING APPROACH?} and decode to obtain corresponding realizations $\widehat{\boldsymbol{v}}$. 
% % We can further filter in space and time to zoom in 
% Note that such realizations can be further filtered based on space and time to focus on a specific spatial or temporal range within the model.
%*********************************************************% 

%==========================================
\section{Conclusion}\label{sec: conclusion}
%==========================================
% 
In this paper we introduce inVAErt networks, a comprehensive framework for data-driven analysis and synthesis of parametric physical systems. 
An inVAErt network is designed to learn many different aspects of a map or differential model including an emulator for the forward deterministic response, a decoder to quickly provide an approximation of the inverse output-to-input map, and a variational auto-encoder which learns a compact latent space responsible for the lack of bijectivity between the input and output spaces. 

We describe each component in detail and provide extensive numerical evidence on the performance of inVAErt networks for the solution of inverse problems for non-identifiable systems of varying complexity, including linear/nonlinear maps, dynamical systems, and spatio-temporal PDE systems.
InVAErt networks can accommodate a wide range of data-driven emulators including dense, convolutional, recurrent, graph neural networks, and others. 
Compared to previous systems designed to simultaneously learn the forward and inverse maps for physical systems, inVAErt networks have the key property of providing a partition of the input space into non-identifiable manifolds and their identifiable complements.

We find that, in practice, the accuracy in capturing the inverse system response is affected by the selection of appropriate penalty coefficients in the loss function and by the ability to sample from the posterior density of the latent variables. 
We have formally analyzed these two aspects, providing a connection between penalty selection, emulator bias, decoder bias and the property of the forward map Jacobian.

Once an inVAErt network is optimally trained from instances of the forward input-to-output map, inverse problems can be solved instantly, and multiple solutions can be determined at once, providing a much broader understanding of the physical response than unique solutions obtained by regularization, particularly when regularization is not strongly motivated in light of the underlying physical phenomena. 

This work represents a demonstration of the capabilities of inVAErt networks and a first step to improve current understanding on the use of data-driven architectures for the analysis and synthesis of physical systems. A number of extensions will be the objective of future research, from applied studies to combination with PINNs or multifidelity approaches in order to minimize the number of examples needed during training. 

%=========================================
\section*{Acknowledgements}\label{sec:ack}
%=========================================

GGT and DES were supported by a NSF CAREER award \#1942662 (PI DES), a NSF CDS\&E award \#2104831 (University of Notre Dame PI DES) and used computational resources provided through the Center for Research Computing at the University of Notre Dame. CSL was partially supported by an Open Seed Fund between Pontificia Universidad Cat\'olica de Chile and the University of Notre Dame, by the grant Fondecyt 1211643, and by Centro Nacional de Inteligencia Artificial CENIA, FB210017, BASAL, ANID. DES would like to thank Marco Radeschi for the interesting discussions.

% BIBLIO
\bibliographystyle{unsrt}
\bibliography{invaert}{}

\appendix
%====================================================
\section{Hyperparameter selection} \label{chp: Hyper}
%====================================================
%
This section outlines the optimal set of hyperparameters employed in each experiment. Unless otherwise specified, our training approach consists of mini-batch gradient descent with the Adam optimizer~\cite{kingma2014adam}. Besides, our learning rate $\eta$ is set through the exponential scheduler $\eta(z)=\eta_0\cdot\gamma^z$. For a given epoch $z$, the initial learning rate $\eta_0$ and the decay rate $\gamma$ are treated as hyperparameters. 

We also found that dataset normalization is particularly important to improve training accuracy. It  helps us control the amount of spurious outliers during inference and significantly helps with inputs having different magnitudes. 
If not stated otherwise, we apply component-wise standardization to both the network inputs and outputs, i.e.
\begin{equation*}
    \bar{v}_i = \frac{v_i - \mu_{v_i}}{\sigma_{v_i}} \ , \ i = 1:\dim(\boldsymbol{v}) \ , 
    \label{equ: z-std}
\end{equation*}
where $\mu_{v_i}$ and $\sigma_{v_i}$ stand for the calculated mean and standard deviation of component $v_i$, from the entire dataset.

The MLP (Multi-Layer Perceptron) serves as the fundamental building block of our neural network. We use the notation: $[a,b,c]$ to represent a MLP, where $a,b,c$ denote the number of neurons per layer, the number of hidden layers and the type of the activation function, respectively. 
We also highlight the use of Swish activation function \texttt{SiLU}~\cite{ramachandran2017searching} in a few numerical examples, as it has been shown to outperform other activation functions in our experiments.

For the Real-NVP sampler $\mathscr{N}_f$, we use the notation $[a,b,d,e]$ to specify the number of neurons per layer ($a$), the number of hidden layers ($b$), and the number of alternative coupling blocks ($c$). The Boolean variable $e$ indicates whether batch normalization is used during training and our choice of activation functions aligns with the original Real-NVP literature~\cite{dinh2016density}.

Note that the three components of our inVAErt network, i.e., emulator $\mathscr{N}_e$, output density estimator $\mathscr{N}_f$, inference engine ($\mathscr{N}_v+\mathscr{N}_d$) can be trained independently, which enables us to apply distinct hyperparameters for achieving optimal performance. 
To make this more clear , we use the notation $\{A,B,C\}$ for each of the aforementioned components, respectively.

Finally, we also include hyperparameter choices of the additional NF sampler $\mathscr{N}_{f, \boldsymbol{w}}$ (see Section~\ref{sec: nf sampling}), if it helps a certain numerical experiment by generating informative samples of latent variable $\boldsymbol{w}$.
%
%=================================================%
% \subsection{Simple linear system} \label{apd: Hyper-linear}
%
\begin{table}[H]
{\footnotesize
\begin{center}
\begin{tabular}{@{} l l @{}}
\toprule
Data scaling & \texttt{True} \\ 
Initial learning rate $\eta_0$ & $\{1\times 10^{-2}, 1\times 10^{-2}, 1\times 10^{-2}\}$ \\
Decay rate $\gamma$ & $\{0.98, 0.98, 0.999\}$ \\
Mini-Batch size & $\{128, 128, 64\}$ \\
Parameters of $\mathscr{N}_e$ & $[3, 2, \texttt{identity}]$ \\
Parameters of $\mathscr{N}_f$ & $[6, 4, 4,  \texttt{False}]$ \\
Parameters of $\mathscr{N}_v$ & $[8, 4, \texttt{ReLU}]$ \\
Parameters of $\mathscr{N}_d$ & $[10, 10, \texttt{SiLU}]$ \\
$\ell^2$-weight decay rate & $\{0,0,1\times 10^{-3}\}$\\
Loss penalty $\lambda_v$, $\lambda_d$ & $1, 40$ \\  
\bottomrule
\end{tabular}
\end{center}}
\caption{Hyperparameter choices of the simple linear system.}
\label{table:linear hyper}
\end{table}
%
%=================================================%

%=================================================%
% \subsection{Simple nonlinear system without periodicity} \label{apd: Hyper-nonlinear}
%
\begin{table}[H]
{\footnotesize
\begin{center}
\begin{tabular}{@{} l l @{}}
\toprule
Data scaling & \texttt{True} \\ 
Initial learning rate $\eta_0$ & $\{1\times 10^{-2}, 1\times 10^{-2}, 1\times 10^{-2}\}$ \\
Decay rate $\gamma$ & $\{0.99, 0.99, 0.998\}$ \\
Mini-Batch size & $\{128, 128, 128\}$ \\
Parameters of $\mathscr{N}_e$ & $[10, 4, \texttt{ReLU}]$ \\
Parameters of $\mathscr{N}_f$ & $[10, 4, 4,  \texttt{False}]$ \\
Parameters of $\mathscr{N}_v$ & $[10, 4, \texttt{ReLU}]$ \\
Parameters of $\mathscr{N}_d$ & $[10, 10, \texttt{SiLU}]$ \\
$\ell^2$-weight decay rate & $\{0,0,1\times 10^{-3}\}$\\
Loss penalty $\lambda_v$, $\lambda_d$ & $1, 200$ \\  
\bottomrule
\end{tabular}
\end{center}}
\caption{Hyperparameter choices of the simple nonlinear system \textbf{without} periodicity.}
\label{table:nonlinear hyper}
\end{table}
%=================================================%

%=======================================================================%
% \subsection{Simple nonlinear system with periodicity} \label{apd: Hyper-nonlinear-periodicity}
%
\begin{table}[H]
{\footnotesize
\begin{center}
\begin{tabular}{@{} l l @{}}
\toprule
Data scaling & \texttt{True} \\ 
Initial learning rate $\eta_0$ & $\{1\times 10^{-3}, 1\times 10^{-3}, 1\times 10^{-3}\}$ \\
Decay rate $\gamma$ & $\{0.998, 0.998, 0.999\}$ \\
Mini-Batch size & $\{64, 64, 32\}$ \\
Parameters of $\mathscr{N}_e$ & $[16, 8, \texttt{SiLU}]$ \\
Parameters of $\mathscr{N}_f$ & $[10, 4, 4,  \texttt{False}]$ \\
Parameters of $\mathscr{N}_v$ & $[24, 8, \texttt{SiLU}]$ \\
Parameters of $\mathscr{N}_d$ & $[48, 8, \texttt{SiLU}]$ \\
$\ell^2$-weight decay rate & $\{0,0,0\}$\\
Loss penalty $\lambda_v$, $\lambda_d$, $\lambda_r$ & $1, 200, 5$ \\  
\bottomrule
\end{tabular}
\end{center}}
\caption{Hyperparameter choices of the simple nonlinear system \textbf{with} periodicity.}
\label{table:nonlinear hyper-periodicity}
\end{table}
%=======================================================================%

%=======================================================================%
% \subsection{Simple nonlinear system with periodicity-additional NF sampler} \label{apd: Hyper-nonlinear-periodicity-nf}
\begin{table}[H]
{\footnotesize
\begin{center}
\begin{tabular}{@{} l l @{}}
\toprule
Data scaling & \texttt{True} \\ 
Subset size $S$ & 10000\\
Sub-sampling size $Q$ & 20 \\
Mini-Batch size & 1024 \\
Initial learning rate $\eta_0$  & $5\times 10^{-3}$\\
Decay rate $\gamma$ & 0.995 \\
Parameters of $\mathscr{N}_{f,\boldsymbol{w}}$ & $[10, 4, 6,  \texttt{False}]$ \\
$\ell^2$-weight decay rate & $0$\\
\bottomrule
\end{tabular}
\end{center}}
\caption{Hyperparameter choices of the additional NF sampler of the simple nonlinear system \textbf{with} periodicity.}
\label{table: nonlinear-nf}
\end{table}
%=======================================================================%

%=================================================%
% \subsection{The RCR system} \label{apd: Hyper-RCR}
%
\begin{table}[H]
{\footnotesize
\begin{center}
\begin{tabular}{@{} l l @{}}
\toprule
Data scaling & \texttt{True} \\ 
Initial learning rate $\eta_0$ & $\{1\times 10^{-2}, 1\times 10^{-2}, 1\times 10^{-2}\}$ \\
Decay rate $\gamma$ & $\{0.995, 0.995, 0.9992\}$ \\
Mini-Batch size & $\{128, 128, 128\}$ \\
Parameters of $\mathscr{N}_e$ & $[10, 6, \texttt{ReLU}]$ \\
Parameters of $\mathscr{N}_f$ & $[10, 4, 6,  \texttt{False}]$ \\
Parameters of $\mathscr{N}_v$ & $[10, 4, \texttt{ReLU}]$ \\
Parameters of $\mathscr{N}_d$ & $[10, 10, \texttt{SiLU}]$ \\
$\ell^2$-weight decay rate & $\{0,0,1\times 10^{-3}\}$\\
Loss penalty $\lambda_v$, $\lambda_d$ & $1, 400$ \\  
\bottomrule
\end{tabular}
\end{center}}
\caption{Hyperparameter choices of the RCR system.}
\label{table:RCR hyper}
\end{table}
%=================================================%

%=================================================%
% \subsection{The Lorenz system} \label{apd: Hyper-Lorenz}
%
\begin{table}[H]
{\footnotesize
\begin{center}
\begin{tabular}{@{} l l @{}}
\toprule
Data scaling & \texttt{False} \\ 
Initial learning rate $\eta_0$ & $\{1\times 10^{-3}, 1\times 10^{-3}, 1\times 10^{-3}\}$ \\
Decay rate $\gamma$ & $\{0.999, 0.998, 0.999\}$ \\
Mini-Batch size & $\{512, 512, 512\}$ \\
Parameters of $\mathscr{N}_e$ & $[64, 15, \texttt{SiLU}]$ \\
Parameters of $\mathscr{N}_f$ & $[12, 4, 8,  \texttt{False}]$ \\
Parameters of $\mathscr{N}_v$ & $[24, 8, \texttt{SiLU}]$ \\
Parameters of $\mathscr{N}_d$ & $[80, 15, \texttt{SiLU}]$ \\
$\ell^2$-weight decay rate & $\{0,0,0\}$\\
Loss penalty $\lambda_v$, $\lambda_d$, $\lambda_r$ & $1, 200, 7$ \\  
\bottomrule
\end{tabular}
\end{center}}
\caption{Hyperparameter choices of the Lorenz system.}
\label{table:Lorenz hyper}
\end{table}
%=================================================%

%=======================================================================%
% \subsection{The Lorenz system-additional NF sampler} \label{apd: Hyper-lorenz-nf}
\begin{table}[H]
{\footnotesize
\begin{center}
\begin{tabular}{@{} l l @{}}
\toprule
Data scaling & \texttt{False} \\ 
Subset size $S$ & 30000\\
Sub-sampling size $Q$ & 15 \\
Mini-Batch size & 2048 \\
Initial learning rate $\eta_0$  & $5\times 10^{-3}$\\
Decay rate $\gamma$ & 0.995 \\
Parameters of $\mathscr{N}_{f,\boldsymbol{w}}$ & $[8, 4, 6,  \texttt{False}]$ \\
$\ell^2$-weight decay rate & $0$\\
\bottomrule
\end{tabular}
\end{center}}
\caption{Hyperparameter choices of the additional NF sampler of the Lorenz system.}
\label{table: lorenz-nf}
\end{table}

% \subsection{The Reaction-diffusion system} \label{apd: Hyper-RD}
%
\begin{table}[H]
{\footnotesize
\begin{center}
\begin{tabular}{@{} l l @{}}
\toprule
Data scaling & \texttt{False} \\ 
Initial learning rate $\eta_0$ & $\{1\times 10^{-3}, 1\times 10^{-3}, 1\times 10^{-3}\}$ \\
Decay rate $\gamma$ & $\{0.999, 0.998, 0.998\}$ \\
Mini-Batch size & $\{1024, 1024, 1024\}$ \\
Parameters of $\mathscr{N}_e$ & $[96, 12, \texttt{SiLU}]$ \\
Parameters of $\mathscr{N}_f$ & $[12, 4, 8,  \texttt{False}]$ \\
Parameters of $\mathscr{N}_v$ & $[16, 8, \texttt{SiLU}]$ \\
Parameters of $\mathscr{N}_d$ & $[64, 8, \texttt{SiLU}]$ \\
$\ell^2$-weight decay rate & $\{0,0,0\}$\\
Loss penalty $\lambda_v$, $\lambda_d$, $\lambda_r$ & $1, 200, 5$ \\  
\bottomrule
\end{tabular}
\end{center}}
\caption{Hyperparameter choices of the reaction-diffusion system.}
\label{table:RD hyper}
\end{table}

% \subsection{The Reaction-diffusion system-additional NF sampler} \label{apd: Hyper-dr-nf}
\begin{table}[H]
{\footnotesize
\begin{center}
\begin{tabular}{@{} l l @{}}
\toprule
Data scaling & \texttt{False} \\ 
Subset size $S$ & 50000\\
Sub-sampling size $Q$ & 10 \\
Mini-Batch size & 2048 \\
Initial learning rate $\eta_0$  & $1\times 10^{-2}$\\
Decay rate $\gamma$ & 0.995 \\
Parameters of $\mathscr{N}_{f,\boldsymbol{w}}$ & $[10, 4, 6,  \texttt{False}]$ \\
$\ell^2$-weight decay rate & $0$\\
\bottomrule
\end{tabular}
\end{center}}
\caption{Hyperparameter choices of the additional NF sampler of the reaction-diffusion system.}
\label{table: rd-nf}
\end{table}
\end{document}